\documentclass[10pt,twocolumn,letterpaper]{article}

\usepackage{times}
\usepackage{epsfig}
\usepackage{graphicx}
\usepackage{amsmath}
\usepackage{amssymb}
\usepackage{array}
\usepackage{adjustbox}
\usepackage{array}
\usepackage{cellspace}
\usepackage{hhline}
\usepackage{subfig}
\usepackage{times}
\usepackage{epsfig}
\usepackage{graphicx}
\usepackage{amsmath}
\usepackage{amssymb}

\usepackage[utf8]{inputenc} 
\usepackage{adjustbox}
\usepackage{array}
\usepackage{cellspace}
\usepackage{hhline}
\usepackage{subfig}
\usepackage{mathtools}
\usepackage{cellspace}
\usepackage{comment}
\usepackage{makecell}
\setcellgapes{4pt}
\DeclarePairedDelimiter{\ceil}{\lceil}{\rceil}
\DeclarePairedDelimiter{\floor}{\lfloor}{\rfloor}
\usepackage[usenames]{color}
\definecolor{Red}{rgb}{1,0,0}
\definecolor{blue}{rgb}{1,0,1}
\graphicspath{{images/}}

\usepackage{multirow}
\usepackage{booktabs}
\usepackage[table,xcdraw]{xcolor}

\DeclareMathOperator*{\argmin}{arg\,min}
\pdfoutput=1

\usepackage[usenames]{color}
\definecolor{Red}{rgb}{1,0,0}
\graphicspath{{images/}}

\usepackage[pagebackref=true,breaklinks=true,letterpaper=true,colorlinks,bookmarks=false]{hyperref}

\def\cvprPaperID{
	2750} 

\begin{document}
\title{Deep Shape-from-Template: Wide-Baseline, Dense and Fast\\Registration and Deformable Reconstruction from a Single Image}

\author{David Fuentes-Jimenez, Daniel Pizarro\\
Universidad de Alcal\'a\\
{\tt\small d.fuentes@edu.uah.es,daniel.pizarro@uah.com}
\and
David Casillas-Perez\\
Department of Signal Processing and Communications, Universidad Rey Juan Carlos\\
{\tt\small david.casillas@urjc.es}
\and
Toby Collins\\
Institut de Recherche contre les Cancers de l’Appareil Digestif\\
{\tt\small toby.collins@gmail.com}
\and
Adrien Bartoli\\
Universit\'e Clermont-Auvergne\\
{\tt\small adrien.bartoli@gmail.com}
}

\maketitle

\begin{abstract}
Shape-from-Template (SfT) solves 3D vision from a single image and a deformable 3D object model, called a template.
Concretely, SfT computes registration (the correspondence between the template and the image) and reconstruction (the depth in camera frame).
It constrains the object deformation to quasi-isometry.
Real-time and automatic SfT represents an open problem for complex objects and imaging conditions.
We present four contributions to address core unmet challenges to realise SfT with a Deep Neural Network (DNN).
First, we propose a novel DNN called DeepSfT, which encodes the template in its weights and hence copes with highly complex templates.
Second, we propose a semi-supervised training procedure to exploit real data. This is a practical solution to overcome the render gap that occurs when training only with simulated data.
Third, we propose a geometry adaptation module to deal with different cameras at training and inference.
Fourth, we combine statistical learning with physics-based reasoning.
DeepSfT runs automatically and in real-time and we show with numerous experiments and an ablation study that it consistently achieves a lower 3D error than previous work.
It outperforms in generalisation and achieves an unprecedented performance with wide-baseline, occlusions, illumination changes, weak texture and blur.
\end{abstract}

\section{Introduction}

\subsection{Context and the SfT problem}
\begin{figure}[!htbp]
	\centering
	\includegraphics[width=\linewidth]{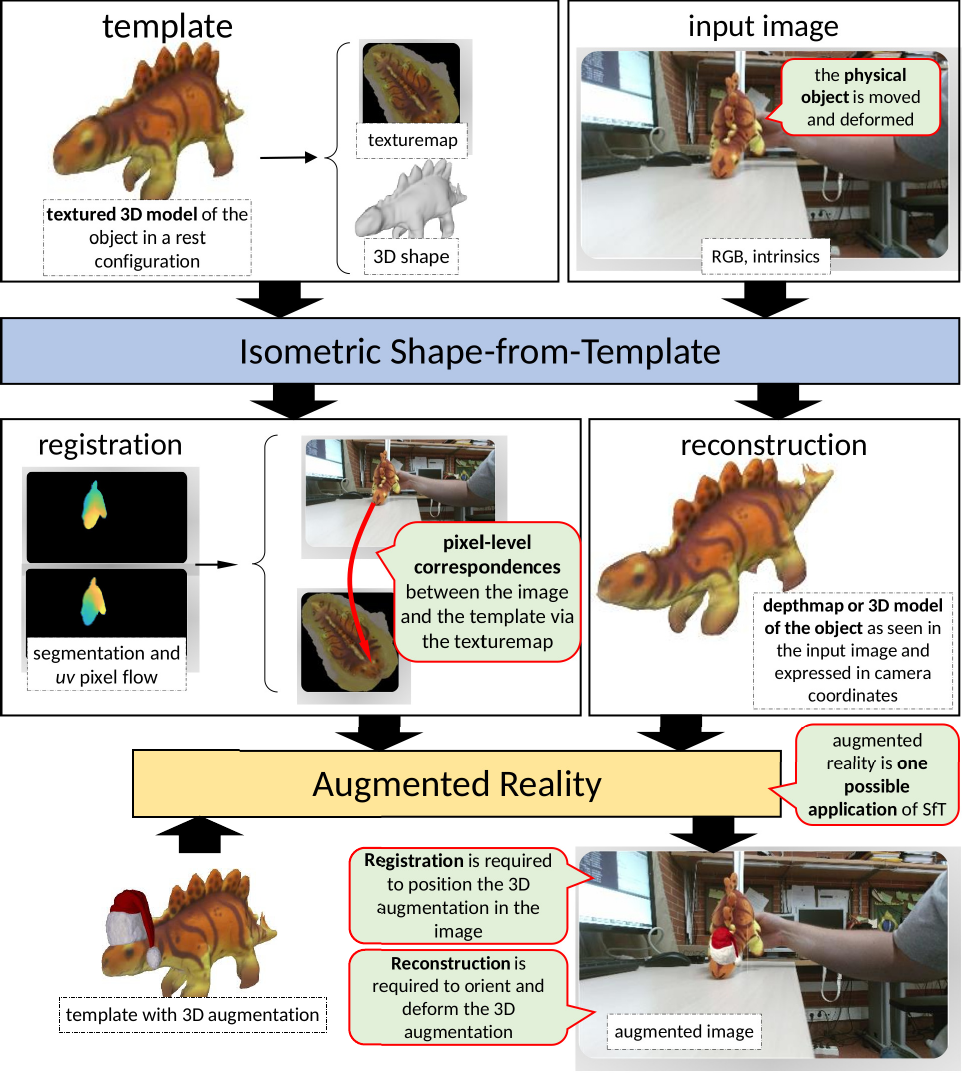} 
	\caption{The principle of SfT and its application to augmented reality. Results obtained with DeepSfT}
	\label{fig:arexample}
\end{figure}

The tasks of image registration (i.e., the computation of correspondences) and image-based reconstruction (i.e., the computation of depth) are fundamental in computer vision\footnote{By `image', we always mean an RGB image taken by a regular camera.}.
Solving both tasks is required in applications such as 3D object tracking and augmented reality.
To date, there exist mature techniques for rigid objects, such as Structure-from-Motion (SfM)~\cite{Hartley2003}.
The case of deformable objects is however largely unresolved.
The existing work has considered two main scenarios.
In Non-Rigid SfM (NRSfM)~\cite{Bregler2000,Torresani2008,Dai2012,Chhatkuli2017a}, the inputs are a set of images and the problem is to find correspondences across images (registration) and depth (reconstruction).
In Shape-from-Template (SfT)~\cite{Salzmann2008,Bartoli2015,Ngo2016,Chhatkuli2017}, the inputs are a single image, a 3D object model (template) is known, and the problem is to find correspondences between the model and the image (registration) and depth (reconstruction).
Obviously, as the object is deformable, the image is not a photo of the model under some unknown pose: rather, it is a photo of the model taken after some unknown deformation.
The most common type of deformation prior used in NRSfM and SfT is the widely applicable {\em quasi-isometry}, which prevents significant stretching or shrinking of the object.
An illustration of SfT is shown in figure~\ref{fig:arexample}.
A very important concept in SfT is the {\em template}, which is the known textured 3D object model.
Concretely, the template is a 3D shape (e.g., a triangulated 3D mesh) and a texture map (e.g. an image giving colours for the mesh's facets), which is acquired straightforwardly using a 3D scanner, an RGB-D sensor or SfM.

SfT is a difficult and unresolved problem.
The core challenges are related to the object (typically, a rich texture and a flat template shape are easier to deal with), to the imaging conditions (typically, a sharp and well-lit image with strong visibility are easier to deal with) and to the availability of an initial solution guess.
The latter is generally available when the input image is extracted from a continuous video, where the solution to the past frame forms a guess for the current frame, and forms the so-called {\em short-baseline} case.
In contrast, the {\em wide-baseline} case occurs when the input image is processed individually, without having a solution guess.
The short-baseline condition (that a solution guess is available) is obviously a strong weakness, as it assumes that camera motion and object deformation are small between frames, and fails if, for instance, the object goes outside the field of view.
The wide-baseline case, despite its increased difficulty, is thus very important to achieve highly robust deformable object registration and reconstruction.

SfT has been widely investigated with non-DNN approaches for almost two decades and only recently within the DNN framework.
Non-DNN SfT methods fall into two broad categories.
Methods in the first category compute registration before reconstruction with existing keypoint-based or dense matching methods~\cite{Pizarro2012,Gay-Bellile2010,Collins2014}.
They thus deal with the wide-baseline case but are tremendously limited by the catastrophic failure of registration, for many of the challenging object or imaging conditions (e.g., blur will typically defeat the extraction of keypoints).
Methods in the second category compute registration and reconstruction simultaneously~\cite{Ngo2016,Collins2016,Agudo2016}.
They proceed by numerical optimisation from an initial guess and hence only work in the short-baseline case.
They may catastrophically fail for many of the challenging object or imaging conditions.
Using the DNN framework to solve SfT is an attractive idea.
The general concept is to learn a function that maps the input image to 3D deformation parameters~\cite{Pumarola2018,hdm_net,Shimada2019}.
This solves registration and reconstruction jointly, without iterative optimisation at run-time, and copes with the wide-baseline case.
The attempts to develop DNN SfT methods are promising but also bear three important limitations.
First, they are very restrictive with the object template, requiring regular rectangular meshes with a relatively small number of vertices (namely, $73\times73$ in \cite{hdm_net,Shimada2019} and $10\times10$ in \cite{Pumarola2018}).
Second, they require labelled registration and reconstruction data for training.
This relegates their training to only use synthetic data, affecting their accuracy in real conditions.
Third, they require that training and inference are done with images coming from the same camera, which is a strong practical limitation.
In spite of the progress brought by these works, there does not currently exist an SfT method capable of handling the wide-baseline case robustly for the challenging object and imaging conditions, densely and in real-time.
\subsection{Related vision problems}
SfT is closely related to some other vision problems, namely optical flow, scene flow, monocular depth reconstruction, pose estimation and Shape-from-Shading. However, SfT is unique in its own right as it has specific inputs-outputs and challenges. As a consequence, existing methods to these related problems either do not apply or cannot compete with specific SfT solutions. 
\\ 
\textbf{Optical flow}. Optical flow~\cite{Collins2016,Liu-Yin2017,FlowNet20,FlowNet,optical_flow} solves registration between two consecutive images in a video. It differs from SfT in terms of its inputs (which are two images) and because it is only solved in the short-baseline configuration. Additionally, SfT involves reconstruction, while optical flow does not.
Applications in AR, for instance, cannot be realised from optical flow only.
\\
\textbf{Scene flow}. Scene flow solves registration between consecutive depth maps in an RGB-D video, obtained from an active sensor~\cite{lv2018learning,hornacek2014sphereflow} or stereo~\cite{schuster2018sceneflowfields,wedel2011stereoscopic}. It differs from SfT in terms of its inputs and because it is only solved in the short-baseline configuration. In SfT, the sensor at run-time is a regular camera, whose image cannot be fed to scene flow because of the missing depth channel.
Additionally, DNN-based scene-flow methods~\cite{hur2020,zou2018df,brickwedde2019monosf} have shown limited success using piecewise rigid motion and self-supervised approaches, with very similar limitations to optical flow methods.
\\
\textbf{Monocular depth reconstruction}. Monocular depth reconstruction~\cite{Eigen2015,Garg2016,Liu2016,densedepth,BTS} infers depth from a single image for a scene category, such as rooms and road scenes. It is a hard problem with ambiguities due to the wide variability of objects, textures and shapes inside the scene.
It is a reconstruction method, not involving registration, in contrast with SfT which involves both.
Applications in augmented reality, for instance, cannot be realised from monocular depth reconstruction only.
\\
\textbf{Pose estimation}. Pose estimation~\cite{Martinez2017,alp2018densepose,posenet} computes the articulated pose of a person, typically defined by a 3D skeleton model, from a single image. In this sense it computes both registration and reconstruction, as the skeleton model is recovered in 3D. In a way, the skeleton model represents a category-level template. SfT differs from human pose estimation because its template is object-specific and deformation is of a much broader dimensionality.
Specifically, a skeleton model typically has about 16 vertices, while an SfT template typically has several thousand vertices (e.g., 36256 vertices for the dinosaur template shown in figure~\ref{fig:arexample}).
\\
\textbf{Shape-from-Shading}.
Shape-from-Shading (SfS) is a reconstruction method which estimates depth and normal maps from an image of a textureless object.
SfS does not generally consider a 3D object model and does not solve registration.
The recent DNN methods~\cite{bednarik,patchbased} have however solved SfS for object categories, such as pieces of cloth or paper.
While \cite{patchbased} does not use an explicit object model, \cite{bednarik} fits a $31\times 31$ regular rectangular mesh to guide reconstruction. It uses a depth sensor for labelling real training data. The experimental setup ensures that the object is easily segmented from a dark background and the illumination is controlled with at least three light sources.
Both approaches stick to the classical SfS setting where the object must be mainly textureless, the scene illumination must produce significant shading, and the object must be segmented from the background.
They are thus not applicable in the general AR context.

\subsection{Summary of contributions}
We present four contributions to advance the state-of-the-art in SfT within the DNN framework.

First, we propose DeepSfT, a novel DNN specifically tailored to SfT.
Technically, DeepSfT is fully-convolutional and based on residual encoder-decoder structures with refining blocks.
DeepSfT has an original architecture compared to previous DNN SfT methods~\cite{Pumarola2018,hdm_net,Shimada2019}.
First, in terms of its inputs: DeepSfT only takes the image as input, but not the template.
This means that DeepSfT is {\em object-specific}, as the template is encoded in its weights at training time.
Second, in terms of its outputs: while previous methods output 3D vertices, DeepSfT produces a dense optical flow to represent registration and a dense depth map to represent reconstruction.
If required, the full object shape is then obtained from a physics-based model a posteriori. 
These choices have important practical consequences.
First, DeepSfT is an efficient network with real-time inference capability.
Second, it is independent of the 3D object model representation, hence capable to exploit models with fine geometric details, complex topology and advanced material and illumination parameters. The computational cost of inference is independent of the number of parameters used to represent the object, such as the number of vertices with a mesh model. It thus solves the problem of limited template complexity of previous DNN SfT approaches~\cite{Pumarola2018,hdm_net,Shimada2019}.

Second, we propose a semi-supervised end-to-end training procedure, capable of training from synthetic and real data.
Training from synthetic data is simple, by synthesising images from random quasi-isometric deformations of the template, whose registration and reconstruction parameters are readily available.
Training from real data is however a difficult issue in SfT, yet is required to achieve good generalisation and to overcome the so-called \emph{render gap}.
Indeed, while the depth label can be obtained by acquiring data with a standard RGB-D sensor, the registration label cannot be obtained.
Our procedure first trains DeepSfT from synthetic data with a combination of supervised loss functions that measures the error of the predicted registration and depth in different points of the network, forcing it to a coarse initial output.
The refining blocks of the network are then trained from real data, with a combination of a supervised reconstruction loss function and a self-supervised registration loss function, based on image colour photo-consistency. 
Importantly, the quasi-isometric deformation of the object is learnt by DeepSfT because the training data, whether synthetic or real, exhibit quasi-isometric deformations of the object.
Our training procedure thus strongly reduces the requirement for fully-labelled data of previous DNN SfT approaches~\cite{Pumarola2018,hdm_net,Shimada2019}.

Third, we propose a solution to cope with multiple imaging geometries (caused by changing the intrinsics of the physical camera, typically by zooming in or out, or by using a different camera), at training and inference.
A natural idea is to train the network with a variety of imaging geometries and possibly to also have it to output the calibration parameters.
This is risky in at least two respects.
First, quasi-isometric SfT has been shown to have a unique and well-posed solution in the general case for a calibrated camera, but not for an uncalibrated camera.
Second, this will increase the size of the network and decrease its generalisability.
Our proposal simply exploits the known camera geometry (intrinsics and distorsion parameters) to warp the input image to a standard configuration.
With this standardisation, DeepSfT can be trained to a single imaging geometry, and yet, handles any camera in any configuration.
Our solution thus resolves the need of using the same camera to acquire or simulate training data and at inference time of previous DNN SfT approaches~\cite{Pumarola2018,hdm_net,Shimada2019}.

Fourth, we propose to combine DeepSfT with a physics-based estimation procedure.
This combination is intrinsically related to the choice of outputs we made for DeepSfT which, recall, is the optic flow field (registration) and the depth map (reconstruction).
These outputs only solve SfT on the part of the object visible in the input image.
For a volumic object such as a shoe, there are always occluded parts, which, for some applications, may be important to register and reconstruct too.
We propose to use the result of SfT for the visible part to solve for the occluded part based on the physics-based deformation model induced by the template and quasi-isometry.
This problem has received a stable solution in computer graphics, implementing quasi-isometry by the so-called As-Rigid-As-Possible (ARAP) prior.
We show how DeepSfT can be directly connected to this solution to recover the full object solution.
Importantly, some applications such as AR do not require one to resolve the occluded object part, in which case this last step can be left aside and computation time saved.
Our solution guarantees that the recovered occluded part of the object fulfills the physics-based constraints pertaining to the real world.
In contrast, in previous DNN SfT approaches~\cite{Pumarola2018,hdm_net,Shimada2019}, these constraints are learnt by the network and thus only hold approximately on the solution. 

Table~\ref{tb:methods} summarises the comparison between DeepSfT, other SfT methods, and related vision methods.
We present quantitative and qualitative experimental results showing that DeepSfT outperforms the state-of-the-art in accuracy, robustness and computation time.
These results include the wide-baseline case and severe  imaging conditions, with strong occlusions, illumination changes, weak texture and blur.
Importantly, we intend to {\em release all the data created or used for this work for public use}.

\begin{table*}
	\centering
	\arrayrulecolor{black}
	\resizebox{\textwidth}{!}{
	\begin{tabular}{|l|l|c|c|c|c|c|c|c|c|c|} 
		\hline
		\rowcolor[rgb]{0.608,0.608,0.608} \multicolumn{1}{|c|}{}                                                                                                            & Methods/problems         & Baseline                 & Accuracy                         & \begin{tabular}[c]{@{}>{\cellcolor[rgb]{0.608,0.608,0.608}}c@{}}Number\\ of\\ vertices \end{tabular} & \begin{tabular}[c]{@{}>{\cellcolor[rgb]{0.608,0.608,0.608}}c@{}}Needs\\ rectangular\\ template \end{tabular} & \multicolumn{1}{l|}{Volumic} & \begin{tabular}[c]{@{}>{\cellcolor[rgb]{0.608,0.608,0.608}}c@{}}Solves\\ dense\\ registration \end{tabular} & \begin{tabular}[c]{@{}>{\cellcolor[rgb]{0.608,0.608,0.608}}c@{}}Training\\ needs full\\ supervision \end{tabular} & \begin{tabular}[c]{@{}>{\cellcolor[rgb]{0.608,0.608,0.608}}c@{}}Real\\ time \end{tabular} & References                                      \\ 
		\hline
		{\cellcolor[rgb]{0.753,0.753,0.753}}                                                                                                                 & Decoupled methods        & \textcolor{green}{Wide } & \textcolor[rgb]{1,0.647,0}{Med } & \textcolor{red}{Low }                                                                             & \textcolor{green}{No}                                                                                        & \textcolor{green}{Yes }         & \textcolor{red}{No }                                                                                        & N/A                                                                                                               & \textcolor{green}{Yes }                                                                      & \cite{Ngo2016,collins14b}                      \\ 
		\hhline{|>{\arrayrulecolor[rgb]{0.753,0.753,0.753}}->{\arrayrulecolor{black}}----------|}
		\multirow{-2}{*}{{\cellcolor[rgb]{0.753,0.753,0.753}}\begin{tabular}[c]{@{}>{\cellcolor[rgb]{0.753,0.753,0.753}}l@{}} Classical \\SfT\end{tabular}}  & Integrated methods       & \textcolor{red}{Short }  & \textcolor{green}{High }         & \textcolor{green}{High }                                                                             & \textcolor{green}{No}                                                                                        & \textcolor{green}{~Yes }        & \textcolor{green}{Yes }                                                                                     & N/A                                                                                                               & \textcolor{green}{Yes }                                                                   & \cite{Salzmann2009,Perriollat2011,Brunet2014}  \\ 
		\hline
		{\cellcolor[rgb]{0.753,0.753,0.753}}                                                                                                                                & Previous methods                  & \textcolor{green}{Wide}  & \textcolor{green}{High}          & \textcolor{red}{Low}                                                                                 & \textcolor{green}{No}                                                                                        & \textcolor{red}{No}             & \textcolor{green}{Yes}                                                                                      & \textcolor{red}{Yes}                                                                                              & \textcolor{green}{Yes}                                                                    & \cite{hdm_net,Pumarola2018,Shimada2019}       \\ 
		\hhline{|>{\arrayrulecolor[rgb]{0.753,0.753,0.753}}->{\arrayrulecolor{black}}----------|}
		\multirow{-2}{*}{{\cellcolor[rgb]{0.753,0.753,0.753}}\begin{tabular}[c]{@{}>{\cellcolor[rgb]{0.753,0.753,0.753}}l@{}} DNN \\SfT\end{tabular}}                                                                                                       & \textbf{DeepSfT}         & \textcolor{green}{Wide}  & \textcolor{green}{High}          & \textcolor{green}{High}                                                                              & \textcolor{green}{No}                                                                                        & \textcolor{green}{Yes}          & \textcolor{green}{Yes}                                                                                      & \textcolor{green}{No}                                                                                             & \textcolor{green}{Yes}                                                                    & Proposed                                        \\ 
		\hline

		{\cellcolor[rgb]{0.753,0.753,0.753}}                                                                                                                                & Optical flow & \textcolor{red}{Short}                       & \textcolor{green}{High}             & \textcolor{green}{High}                                                                                 & \textcolor{red}{Yes}                                                                                        & \textcolor{green}{Yes}             & \textcolor{green}{Yes}                                                                                      & \textcolor{red}{Yes}                                                                                             & \textcolor{green}{Yes}                                                                    &  \makecell{\cite{Collins2016,Liu-Yin2017,FlowNet20}\\ \cite{FlowNet,optical_flow}}                         \\ 
		\hhline{|>{\arrayrulecolor[rgb]{0.753,0.753,0.753}}->{\arrayrulecolor{black}}----------|}
		{\cellcolor[rgb]{0.753,0.753,0.753}}                                                                                                                                & Scene flow            & \textcolor{red}{Short }  & \textcolor{green}{High }         & \textcolor{green}{High }                                                                             & \textcolor{green}{No }                                                                                        & \textcolor{green}{Yes }         & \textcolor{green}{Yes }                                                                                     & \textcolor{red}{Yes }    & \textcolor{green}{Yes}  &                                                                                           
		\makecell{\cite{lv2018learning,hornacek2014sphereflow,schuster2018sceneflowfields}\\ \cite{hur2020,zou2018df,brickwedde2019monosf} \\ \cite{wedel2011stereoscopic}}                 \\ 
		\hhline{|>{\arrayrulecolor[rgb]{0.753,0.753,0.753}}->{\arrayrulecolor{black}}----------|}
		{\cellcolor[rgb]{0.753,0.753,0.753}}                                                                                                                                & \makecell[l]{Monocular depth\\ reconstruction}          & N/A & \textcolor{red}{Low }        & \textcolor{red}{Low }                                                                                & \textcolor{green}{No }                                                                                       & \textcolor{red}{No }            & \textcolor{green}{Yes }                                                                                        & \textcolor{green}{No }                                                                                             & \textcolor{green}{Yes }                                                                   & \makecell{\cite{densedepth,BTS,Eigen2015}\\ \cite{Garg2016,Liu2016}}                    \\ 
		\hhline{|>{\arrayrulecolor[rgb]{0.753,0.753,0.753}}->{\arrayrulecolor{black}}----------|}
		\multirow{-4}{*}{{\cellcolor[rgb]{0.753,0.753,0.753}}\begin{tabular}[c]{@{}>{\cellcolor[rgb]{0.753,0.753,0.753}}l@{}}Related \\problems\end{tabular}} & Pose estimation    & \textcolor{green}{Wide } & \textcolor[rgb]{1,0.647,0}{Med } & \textcolor{red}{Low }                                                                                & \textcolor{green}{No }                                                                                       & \textcolor{red}{No }            & \textcolor{green}{Yes }                                                                                     & \textcolor{red}{Yes }                                                                                             & \textcolor{green}{Yes }                                                                   & \cite{alp2018densepose,posenet,Martinez2017}                        \\
		\hhline{|>{\arrayrulecolor[rgb]{0.753,0.753,0.753}}->{\arrayrulecolor{black}}----------|}
		\multirow{-4}{*}{{\cellcolor[rgb]{0.753,0.753,0.753}}\begin{tabular}[c]{@{}>{\cellcolor[rgb]{0.753,0.753,0.753}}l@{}}\end{tabular}} & Shape-from-Shading    & N/A& \textcolor[rgb]{1,0.647,0}{Med} & \textcolor{green}{High }                                                                                & \textcolor{green}{No }                                                                                       & \textcolor{red}{No }            & \textcolor{green}{Yes }                                                                                     & \textcolor{red}{Yes }                                                                                             & \textcolor{green}{Yes }                                                                   & \cite{bednarik,patchbased}                        \\
		\hline
	\end{tabular}
}
	\caption{Characteristics of existing SfT methods and other related problems with state-of-the-art solutions provided by DNNs. Existing SfT methods are divided into classical (non-DNN) and DNN methods.}
	\label{tb:methods}
\end{table*}

\section{Previous Work}
We first review the non-DNN SfT methods, which we call \emph{classical SfT} methods, forming the vast majority of existing work.
We start with the \emph{decoupled methods}, which solve registration and reconstruction as independent problems, and then we discuss the \emph{integrated methods} that solve registration and reconstruction jointly.
We finally review the DNN SfT methods. We have categorised state-of-the-art methods, their properties, and problems related to SfT in table \ref{tb:methods}.

\subsection{Classical SfT decoupled methods}
Decoupled methods first compute registration and then reconstruction as two independent and sequential stages \cite{Salzmann2009,Perriollat2011,Brunet2014}. Their main advantages are simplicity, problem decomposition, and to leverage existing mature registration approaches. However, they tend to produce sub-optimal solutions because they do not consider all physical constraints that connect reconstruction and registration. Decoupled methods typically solve wide-baseline registration with an existing method that is not specific to SfT, using feature-based matching with keypoints such as SIFT~\cite{sift}, with filtering to reduce the mismatches~\cite{Pizarro2012,pilet08a}. These approaches inherit the advantages of wide-baseline registration: they can deal with individual images and strong deformation without requiring temporal consistency. However, they are fundamentally limited by feature-based registration, which fails when the object has a weak or repetitive texture, or when the imaging conditions are challenging (low image resolution, blur or strong viewpoint distortion).
Furthermore, accurate results demand an expensive optimisation process at run-time.
Because of these limitations, the existing real-time wide-baseline decoupled methods require simple objects with simple deformations, such as bending sheets of paper. 

Various reconstruction methods have been considered in decoupled methods, and they can be classified according to the deformation model. The most popular deformation model is isometry, which approximately preserves geodesic distances. These methods follow one of three main strategies: \textit{i)} using a convex relaxation of isometry called inextensibility~\cite{Salzmann2009,Salzmann2008,Perriollat2011,Brunet2014}, \textit{ii)} using local differential geometry~\cite{Bartoli2015,Chhatkuli2017} and \textit{iii)} minimising a global non-convex cost~\cite{Brunet2014,Oezguer2017}. Methods in \textit{iii)} are the most accurate but also the most computationally expensive.
They require an initial solution found using a method from \textit{i)} or \textit{ii)}.  
There also exist methods that relax isometry in an attempt to handle elastic deformations. These include the angle-preserving conformal model~\cite{Bartoli2015}, or simple mechanical models with linear~\cite{Malti2013,Malti2015} or non-linear elasticity~\cite{Haouchine2017,Haouchine2014,Agudo2015,equiareal}. These models all require boundary conditions in the form of known 3D points, which is a fundamental limitation. 
The well-posedness of non-isometric methods remains an open research question. 

\subsection{Classical SfT integrated methods}
Integrated methods compute both registration and reconstruction jointly. All existing methods are short-baseline, restricted to video data, and may work in real time~\cite{Ngo2016,Collins2016,Liu-Yin2017}.
They are based on the iterative minimisation of a non-convex cost that deforms the template in 3D so that its projection agrees with the image data. Some methods use keypoint correspondences that can be re-estimated during optimisation~\cite{Ngo2016}, and others use pixel-level information~\cite{Collins2016,Liu-Yin2017} and a data cost based on template/image photo-consistency. These latter methods support dense solutions and resolve complex, high-frequency deformations. 
Their main limitations are two-fold. First, they break down with fast deformation or camera motion. 
Second, at run-time, they must solve an optimisation process that is highly non-convex and computationally demanding, requiring careful hand-crafted design and a correct balance of data and deformation constraints.

\subsection{DNN SfT methods}
Several DNN-based methods have been recently proposed~\cite{Pumarola2018, hdm_net, Shimada2019}. These methods assume a flat template, described with a regular mesh. We refer to this special type of template as a \emph{rectangular template}. They all use encoder-decoder neural architectures, and differ in the way the mesh vertex coordinates are parameterised and the learning strategy.~\cite{Pumarola2018} first solves registration by regressing many 2D belief maps (three per vertex), giving their likely 2D coordinates in the image. A depth estimation network is then used to reconstruct vertex depth coordinates. This strategy does not scale well to many vertices, limiting its applicability, as shown by the reported experiments with $10\times10$ vertices or fewer. ~\cite{hdm_net,Shimada2019} use three-channel 2D outputs to parameterise the 3D coordinates of the mesh vertices. This strategy allows~\cite{hdm_net} and~\cite{Shimada2019} to use a rectangular template with a greater number of vertices than~\cite{Pumarola2018}, showing results with $73\times73$ vertices in both cases. ~\cite{hdm_net} use supervised learning that minimises the mean squared error between the network outputs and reconstruction labels with a synthetic training data base. \cite{Shimada2019} uses an adversarial learning approach, introducing a discriminator network. 
The methods of ~\cite{Pumarola2018,hdm_net,Shimada2019} share four important common weaknesses. First, they only work with rectangular templates, limiting their application to e.g. paper sheets or rectangular cloth sections. They cannot be used with non-rectangular templates, such as volumic templates or objects with complex geometries like the shoe of figure \ref{fig:scheme}.  Second, they do not scale well for larger meshes, as it increases the network size. Third, the camera used for training and run-time must be the same. Fourth, they are \emph{fully-supervised} methods, requiring fully labeled data. Due to the difficulty of obtaining labels with real data, they rely on simulated data. This strategy limits prediction accuracy in real images due to the render gap between simulated and real data \cite{rendergap}. For instance,~\cite{hdm_net,Shimada2019} use Blender~\cite{Blender} to create synthetic images of a deforming paper sheet or clothing. In all reported experiments the simulated images have controlled background and lighting conditions. In all these previous DNN methods, the experimental results with real data are mostly qualitative and with a controlled environment, to mitigate the render gap between the synthetic and real data.

In summary, the previous DNN SfT methods have shown that SfT can be learnt by a DNN. However, they have not been shown to work in real-world challenging conditions, and suffer four main limitations discussed above. Our proposed approach DeepSfT does not have these limitations, signifying a considerable step forward in SfT research and real-world application.

\section{Methodology}
\label{sec:problemForm}
\subsection{Scene geometry}
\begin{figure}[!htbp]
	\centering
	\includegraphics[width=\linewidth]{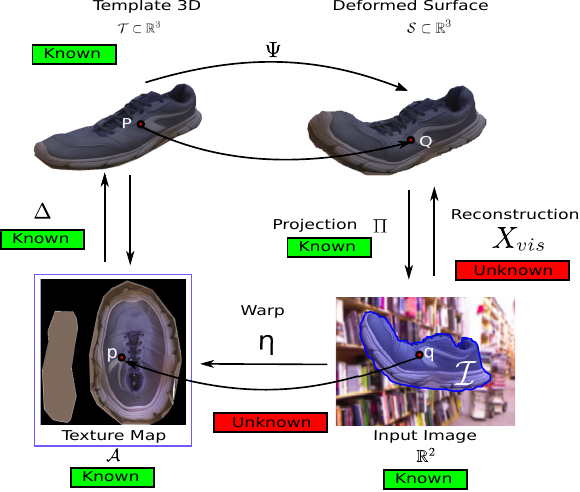}
	\caption{Geometric model of SfT.}
	\label{fig:scheme}
\end{figure}

\noindent \textbf{Template.} Figure~\ref{fig:scheme} shows the geometric model of SfT, including the camera image and template deformation. The template is known and represented by a 3D
surface~$\mathcal{T}\subset\mathbb{R}^3$ jointly with an appearance model, described
as a \emph{texture map}~$\mathcal{A}_\mathcal{T}=(\mathcal{A},A)$.

The texture map consist of an $\mathbb{R}^2$ domain $\mathcal{A}\subset\mathbb{R}^2$ and a function
$A\colon\mathcal{A}\to(r,g,b)$ which maps it to the RGB space.
The texture map domain $\mathcal{A}$
is represented as a collection of flattened \emph{texture charts} $\mathcal{U}_i$ whose union covers the appearance of the whole
template \cite{Hughes}.
We use normalised texture coordinates for
$\mathcal{A}$, drawn from the unit square. In our
approach the template is not restricted to a specific topology, and can
be \textit{thin-shell} or \textit{volumic}, without requiring
modification to our DNN architecture. Our approach is also not restricted to a
specific surface representation. In our experiments section we use mesh
representations because of their generality, but this is not a requirement of
the DNN. The bijective map between $\mathcal{A}$ and $\mathcal{T}$ is known
and denoted by $\Delta\colon\mathcal{A} \to \mathcal{T}$.  

\noindent \textbf{Deformation.} We assume that $\mathcal{T}$ is deformed with an unknown quasi-isometric map~$\Psi\colon\mathcal{T} \to \mathcal{S}$, where $\mathcal{S}\subset\mathbb{R}^3$ denotes the unknown deformed surface. Quasi-isometric maps permit mild extension and compression, common with many real world deformable objects. 

\noindent \textbf{Camera projection.} The input image is modeled as a 3-channel colour intensity function $I\colon\mathbb{R}^2 \to (r,g,b)$, which is discretised into a regular grid of pixels. We model the camera with perspective projection:  

\begin{align}
\label{eq:cameraPerspectiva}
\left(x,y,z\right) &\longmapsto \left(\frac{x}{z},\frac{y}{z}\right)=(u,v).
\end{align}
We assume that the camera is intrinsically calibrated: radial distortion, focal length and aspect ratio are all known parameters. This is a very common assumption in SfT. Hence, $(u,v)$ are retinal coordinates that, without loss of generality, can be readily obtained from the image coordinates. 

\noindent \textbf{Visible surface region and registration map.}
The surface region that is visible in the camera image (unobstructed by self or external occlusion) is unknown and denoted by $\mathcal{S}_{vis}\subset \mathcal{S}$. This region projects onto the image plane to define an unknown 2D region $\mathcal{I} \subset \mathbb{R}^2$. We relate $\mathcal{S}_{vis}$ and $\mathcal{I}$ with a perspective embedding function $X_{vis}\colon\mathcal{I} \to \mathcal{S}_{vis}$ with $X_{vis}(u,v) = \rho(u,v) \left(u,v,1\right)$ and 
where the unknown depth function $\rho\colon\mathcal{I} \to \mathcal{S}_{vis}$ gives the depth of $\mathcal{S}_{vis}$ in camera coordinates at each pixel in $\mathcal{I}$. In the absence of self-occlusions, $\mathcal{S}_{vis}=\mathcal{S}$. Volumic templates always induce self-occlusions.  
The unknown registration map, $\eta\colon\mathcal{I} \to \mathcal{A}$ is an injective map that relates each point of $\mathcal{I}$ to its corresponding point in $\mathcal{A}$.

\subsection{Object-specific approach}
Our proposed DNN SfT solution DeepSfT estimates $\rho(u,v)$, $\eta(u,v)$ and $\mathcal{I}$ directly from the input image $I$.  DeepSfT is object-specific, as the template information is encoded from the training data into the network weights, as~\cite{Pumarola2018}. In other words, the trained network's weights `memorise' the object shape. This reduces the difficulty of the learning problem, requiring a considerably lower amount of training data, and allows us to propose a compact architecture that runs in real time. DeepSfT is much more accurate than object-generic methods~\cite{hdm_net, Shimada2019, densedepth, BTS}, which are not mature enough to solve SfT in challenging conditions, as we show in the experiments section. Importantly, we also provide a semi-supervised method to train DeepSfT without the need of manual labelling, which is a main limitation of the state-of-the-art. It combines synthetic data generated with Blender, with real data captured with a low-cost commercial RGB-D sensor. Generating data for a new template is thus done easily and can be implemented as a highly automatized process. 

\subsection{DNN architecture}
\label{sec:arch}
\begin{figure*}[!htbp]
	\centering
	\includegraphics[width=\linewidth]{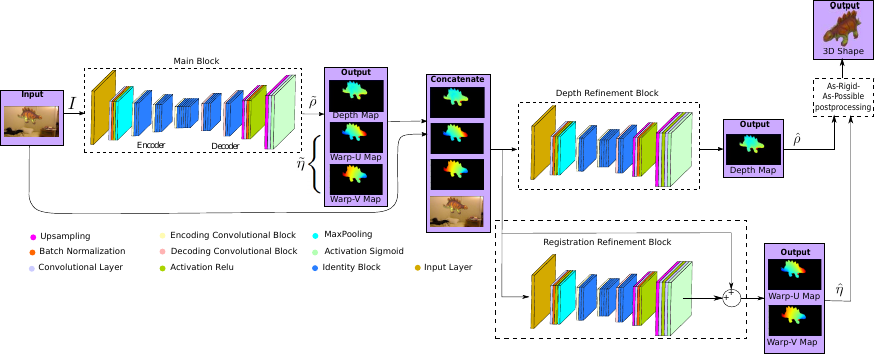}
	\caption{DeepSfT architecture. The proposed network architecture is composed of three principal blocks: the Main Block, the Depth Refinement Block and the Registration Refinement Block. Each block is an encoder-decoder designed for SfT. The Main Block receives an RGB input image $I$ and outputs a first estimate of the registration and depth maps. The Depth and Registration Refinement Blocks improve the initial estimates, taking as input $I$ and the Main Block outputs, and producing the final depth and registration maps.}
	\label{fig:arquitectura}
\end{figure*}

\begin{figure}[!htbp]
	\centering
	\includegraphics[width=0.9\linewidth]{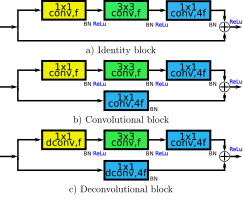}
	\caption{Identity, convolutional and deconvolutional residual blocks.}
	\label{fig:blocks}
\end{figure}

We encode DeepSfT outputs as DNN functions taking $I$ as input, which is resized to a canonical resolution of $270\times 480$ px:
\begin{equation} 
(\hat{\rho},\hat{\eta}) = \mathcal{D}(I,\theta_\mathcal{W}),
\label{eq:dnnFunc}
\end{equation}
where $\theta_\mathcal{W}$ are the network weights. We encode $\mathcal{I}$ in the network outputs $\hat{\rho}$ and $\hat{\eta}$ as follows:
\begin{eqnarray}
\label{eq:rhothau}
\hat{\rho}(u,v) & \approx& \begin{cases}\rho(u,v) &\quad(u,v)\in \mathcal{I}\\ -1 &\quad\textrm{otherwise}\end{cases}\\\nonumber
\hat{\eta}(u,v) &\approx& \begin{cases}\eta(u,v) &(u,v)\in \mathcal{I}\\ (-1,-1) &\textrm{otherwise}.\end{cases}
\end{eqnarray}
Figure~\ref{fig:arquitectura} shows the proposed network architecture. 
It uses a cascaded structure divided into three principal blocks shown in figure~\ref{fig:blocks}. The Main Block is denoted as $\mathcal{D}_{\mathcal{M}}$:
\begin{equation} 
(\tilde{\rho},\tilde{\eta}) = \mathcal{D}_{\mathcal{M}}(I,\theta_{\mathcal{M}}),
\label{eq:dnnFuncM}
\end{equation}
where $\tilde{\rho}$ and $\tilde{\eta}$ are estimates of the depth and registration maps and $\theta_{\mathcal{M}}$ contains the Main Block network weights. The Depth Refinement Block $\mathcal{D}_{\mathcal{D}}$ inputs $I$, $\tilde{\rho}$ and $\tilde{\eta}$ and outputs a refined depth map $\hat{\rho}$:
\begin{equation} 
\hat{\rho} = \mathcal{D}_{\mathcal{D}}(I,\tilde{\rho},\tilde{\eta},\theta_{\mathcal{D}}),
\label{eq:dnnFuncD}
\end{equation}
where $\theta_{\mathcal{D}}$ are the Depth Refinement Block network weights. The Registration Refinement Block $\mathcal{D}_R$ inputs $I$, $\tilde{\rho}$ and $\tilde{\eta}$ and outputs a refined registration map $\hat{\eta}$:
\begin{equation} 
\hat{\eta} = \mathcal{D}_{\mathcal{R}}(I,\tilde{\rho},\tilde{\eta},\theta_{\mathcal{R}}),
\label{eq:dnnFuncD}
\end{equation}
where $\theta_{\mathcal{R}}$ are the Registration Refinement Block network weights. The weights of the three blocks define the network's total weights $\theta_\mathcal{W}=(\theta_{\mathcal{M}},\theta_{\mathcal{D}},\theta_{\mathcal{R}})$.

The refinement blocks play an important role to adapt the network to real data, as described in \S\ref{sec:training}. The three blocks use identity, convolutional and deconvolutional residual feed-forwarding structures based on ResNet50~\cite{he2016deep}. They use encoder-decoder architectures, very similar to those used in semantic segmentation \cite{segnet}. 
Each block is composed of two unbalanced parallel branches with convolutional layers that propagate information to deeper layers, preserving high spatial frequencies.

Table~\ref{tab:capas_main_block} shows the layered decomposition of the Main Block. It first receives $I$ and performs a first spatial reduction using a 2D convolutional layer with ReLu activation and Max Pooling. Then, a sequence of three convolutional and identity blocks is used to encode texture and depth information as deep features (figure~\ref{fig:blocks}).
\begin{table}[!htbp]
	\setcellgapes{3pt}\makegapedcells
	\centering
	\begin{adjustbox}{max width=\linewidth}
		\begin{tabular}{|c|c|c|c|}
			\hline
			Layer num & Type  & Output size & Kernels/Activation    	\\ \hline
			1 & Input & (270,480,3) & -- \\ \hline
			2 & Convolution 2D  & (135,240,64) & (7,7)   \\ \hline
			3 & Batch Normalisation & (135,240,64) &  --  \\	\hline	
			4 & Activation & (135,240,64) & Relu \\		\hline
			5 & Max Pooling 2D & (45,80,64) & (3,3) \\		\hline
			6 & Encoding Convolutional Block & (45,80,[64, 64, 256]) & (3,3) \\		\hline
			7-8 & Encoding identity Block x 2 & (45,80,[64, 64, 256]) & (3,3) \\		\hline	
			9 & Encoding Convolutional Block & (23,40,[128, 128, 512]) & (3,3) \\		\hline
			10-12 & Encoding identity Block x 3 & (23,40,[128, 128, 512]) & (3,3) \\		\hline
			13 & Encoding Convolutional Block & (12,20,[256, 256, 1024]) & (3,3) \\		\hline
			14-16 & Encoding identity Block x 3 & (12,20,[256, 256, 1024]) & (3,3) \\		\hline
			17-20 & Encoding identity Block x 3 & (12,20,[1024, 1024, 256]) & (3,3) \\		\hline
			21 & Decoding Convolutional Block & (24,40,[512, 512, 128]) & (3,3) \\		\hline
			22 & Cropping 2D & (23,39,128) & (1,1) \\		\hline	
			23-25 & Encoding identity Block x 3 & (23,39,[512, 512, 128]) & (3,3) \\		\hline	
			26 & Decoding Convolutional Block & (46,78,[256, 256, 64]) & (3,3) \\		\hline
			27 & Zero Padding & (46,80,64) & (0,1) \\		\hline	
			28-29 & Encoding identity Block x 2 & (46,80,[256, 256, 64]) & (3,3) \\		\hline
			30 & Upsampling & (138,240,64) & (3,3) \\		\hline	
			31 & Cropping 2D & (136,240,64) & (2,0) \\		\hline
			32 & Convolution 2D &  (136,240,64)  & (7,7) \\		\hline
			33 & Batch Normalisation & (135,240,64) &  --  \\	\hline	
			34 & Activation & (136,240,64) & Relu \\		\hline
			35 & Upsampling & (272,480,64)  & (3,3) \\		\hline	
			36 & Cropping 2D & (270,480,64) & (2,0) \\		\hline	
			
			37 & Convolution 2D & (272,480,3) & (3,3) \\		\hline
			38 & Activation & (270,480,1) & Linear \\		\hline
			\multicolumn{1}{|c|}{\rule{0pt}{2.5ex}Number of parameters}&\multicolumn{3}{c|}{81\space664\space765}
			\\ \hline
		\end{tabular} 
	\end{adjustbox}
	\caption {Main Block architecture.}
	\label{tab:capas_main_block}
\end{table}
Image information from $I$ is reduced to a compressed feature vector in a representation space of dimension $12\times20\times1024$. Decoding is performed with decoding blocks. These require upsampling layers to increase the dimensions of the input features before passing through the convolution layers, as shown in figure~\ref{fig:blocks}. 
Finally, the last layers have convolutional and cropping layers that adapt the output of the decoding block to the size of the output maps ($270\times480\times3$). The first output channel provides the depth estimate $\tilde{\rho}$, and the last two output channels provide the registration estimate $\tilde{\eta}$.

The Depth Refinement and Registration Refinement Blocks share the same structure, shown in table~\ref{tab:capas_refinement_block}, which is a reduced version of the Main Block using only the first two encoder and decoder blocks. The Depth Refinement Block takes as input the concatenation of $I$, $\tilde{\rho}$, and $\tilde{\eta}$ (6 channels) and it outputs $\hat{\rho}$. The Registration Refinement Block takes as input the concatenation of $I$, $\tilde{\rho}$, and $\tilde{\eta}$ (6 channels). Its output is added as an offset to $\tilde{\eta}$ (last two channels) to produce $\hat{\eta}$. 
\begin{table}[!htbp]
	\setcellgapes{3pt}\makegapedcells
	\centering
	\begin{adjustbox}{max width=\linewidth}
		\begin{tabular}{|c|c|c|c|}
			\hline
			Layer num & Type  & Output size & Kernels/Activation    	\\ \hline
			1 & Input & (270,480,6) & -- \\ \hline
			2 & Convolution 2D  & (135,240,64) & (7,7)   \\ \hline
			3 & Batch Normalisation & (135,240,64) &  --  \\	\hline
			4 & Activation & (135,240,64) & Relu \\		\hline
			5 & Max Pooling 2D & (45,80,64) & (3,3) \\		\hline
			6 & Encoding Convolutional Block & (45,80,[64, 64, 256]) & (3,3) \\		\hline
			7-8 & Encoding identity Block x 2 & (45,80,[64, 64, 256]) & (3,3) \\		\hline
			9 & Encoding Convolutional Block & (23,40,[128, 128, 512]) & (3,3) \\		\hline
			10-13 & Encoding identity Block x 4 & (23,40,[128, 128, 512]) & (3,3) \\		\hline
			14 & Decoding Convolutional Block & (46,80,[512, 512, 128]) & (3,3) \\		\hline
			15-16 & Encoding identity Block x 2 & (46,80,[512, 512, 128]) & (3,3) \\		\hline
			17 & Upsampling & (92,160,128) & (2,2) \\		\hline	
			18 & Cropping 2D & (92,160,128) & (2,0) \\		\hline	
			19 & Convolution 2D & (90,160,64)& (3,3) \\		\hline
			20 & Batch Normalisation & (90,160,64) &  --  \\	\hline
			21 & Activation & (90,160,64) & Relu \\		\hline
			22 & Upsampling & (270, 480, 64) & (3,3) \\		\hline           
			23 & Convolution 2D & (270, 480, 32)& (3,3) \\		\hline
			24 & Activation & (270, 480, 32)& Relu \\		\hline
			25 & Convolution 2D & (272,480,1) & (3,3) \\		\hline
			26 & Activation & (270,480,1) & Linear \\		\hline
			\multicolumn{1}{|c|}{\rule{0pt}{2.5ex}Number of parameters}&\multicolumn{3}{c|}{13\space618\space689}
			\\ \hline
		\end{tabular} 
	\end{adjustbox}
	\caption {Depth and Registration Refinement Block architectures.}
	\label{tab:capas_refinement_block}
\end{table}

\begin{table}[!htbp]
	\setcellgapes{3pt}\makegapedcells
	\begin{adjustbox}{max width=\linewidth}
		\begin{tabular}{|c|c|c|c|}
			\hline
			Sequence&Samples&Train&Test \\ \hline
			
			DS1S & 60000 & 47000& 5000 \\ \hline
			
			DS2S & 60000 & 47000& 5000 \\ \hline
			
			DS3S & 60000 & 47000& 5000  \\ \hline
			
			DS4S & 60000 & 47000& 5000  \\ \hline
			
			DS1R & 2116 & 1884& 232 \\ \hline
			
			DS2R & 3100 & 2728&  373 \\ \hline
			
			DS3R & 4800 & 3500& 1300  \\ \hline
			
			DS4R & 4200 & 3650 & 550  \\ \hline
			
			DS5R & 193 & 143 & 50  \\ \hline

		\end{tabular} 
	\end{adjustbox}
	\centering \caption {Train and test split for each image sequence. `S' stands for synthetic generated with Blender and `R' stands for for real generated with Kinect V2.}
	\label{tb:train_test_split}
\end{table}

\subsection{Recovering occluded surface regions}
\label{subsec:arap}
Our DeepSfT network registers and reconstructs $\mathcal{S}_{vis}$ by its outputs $\hat{\rho}$ and $\hat{\eta}$. Due to self or external occlusions, always occurring with volumic templates, the hidden surface part $\mathcal{S}_{h}=\mathcal{S}\setminus\mathcal{S}_{vis}$ can be large and important. However, learning to infer $\mathcal{S}_{h}$ from a single image is a very ill-posed problem due to ambiguities, and can be very difficult to train with real data. We propose a post-processing step to recover $\mathcal{S}_{h}$ based on minimising the As-Rigid-As-Possible (ARAP) cost, widely used in graphics and mesh processing~\cite{ARAP}. ARAP is also the most natural prior for quasi-isometric templates~\cite{asrigidasposible,Collins2015} and it does not require additional learning.

Unlike the DNN, which is independent of surface representation, the shape completion process requires the template to be represented as a triangular mesh. We use $\mathcal{M}_{\mathcal{S}}$ and $\mathcal{M}_{\mathcal{T}}$ to represent the deformed and rest template meshes respectively. These have 3D vertices $\mathcal{V}_{\mathcal{S}}=\{\mathbf{p}_1, \dots, \mathbf{p}_N \}$ and $\mathcal{V}_{\mathcal{T}}=\{\mathbf{q}_1, \dots, \mathbf{q}_N \}$ respectively. The objective of ARAP shape completion is to recover $\mathcal{V}_\mathcal{S}$ (and hence $\mathcal{S}$) from $\mathcal{S}_{vis}$ and $\mathcal{V}_{\mathcal{T}}$ by solving the following optimisation problem:
\begin{equation}
\label{eq:arap}
\mathcal{V}_{\mathcal{S}}= \argmin_{\mathcal{V}_\mathcal{S}}{E}(\mathcal{V}_{\mathcal{S}}),
\end{equation}
where:
\begin{equation}
E=E_{d}(\mathcal{V}_\mathcal{S},\mathcal{S}_{vis}) + \lambda_{a} E_{a}(\mathcal{V}_\mathcal{S},\mathcal{V}_\mathcal{T})+ \lambda_s E_{s}(\mathcal{V}_\mathcal{S},\mathcal{V}_\mathcal{T}).     
\end{equation}
$E_{d}$ is the data term. It uses the Euclidean norm between the set of visible vertices in $\mathcal{V}_\mathcal{S}$ and their corresponding 3D coordinates in $\mathcal{S}_{vis}$, as produced by DeepSfT. $E_{a}$ is the ARAP prior \cite{Collins2015}, that encourages the deformed mesh to be isometric with respect to the rest mesh. Finally, $E_{s}$ is a smoothing term that penalizes large deviations in the local curvature of $\mathcal{S}$ with respect to the template. The hyperparameters $\lambda_{a}=20$ and $\lambda_s=0.005$ control the influence of the ARAP and smoothing terms. We set them to a fixed value selected experimentally. We implement $E_{a}$ and $E_{s}$ following \cite{Collins2015} and optimise $E$ with Gauss-Newton, which typically converges in fewer than 10 iterations. This can be implemented easily on a GPU enabled device for real-time shape completion.

\section{DNN Training}
\label{sec:training}

\subsection{Training process overview}
For a given template, we create a quasi-photorealistic synthetic dataset using rendering software. This process is described in detail in \S\ref{sec:templates} and it is used to train DeepSfT with supervised learning. We also record a much smaller dataset with a real RGB-D camera capturing some representative deformations and poses of the object. We emphasize that the RGB-D camera provides only depth labels and not registration labels, so it cannot be used for supervised learning of the registration.

Using both simulated and real data, we train DeepSfT in three steps. In the first step we use the synthetic data to train the Main, Depth Refinement and Registration Refinement Blocks end-to-end. In the second step we refine the Depth Refinement Block weights using real training data. In the third step we refine the Registration Refinement Block weights using real training data with unsupervised learning, by minimising a loss function that enforces the registered template to be photometrically consistent with the input images. 

DeepSfT has been implemented in Keras/Tensorflow \cite{tensorflow}. We have observed that Stochastic Gradient Descent (SGD) achieves better generalisation results when fine tuning the network with real data while Adaptive Moment Estimation (ADAM)~\cite{adam} performs
better when training from scratch. We thus use ADAM in the first step and SGD in the second and third steps. Mixing ADAM and SGD is common practice \cite{swats,adamsgd}.

\subsection{Training step 1: initial global training}
The Main, Depth Refinement and Registration Refinement Blocks are trained end-to-end with the following supervised loss function: 
\begin{equation}
\begin{split}
& \mathcal{L}_1(\theta_\mathcal{W}) =
\frac{1}{2}\sum_{i=1}^{M} \| \hat{\eta}_i - \eta_i \|^2_F +
\sum_{i=1}^{M} \| \hat{\rho}_i - \rho_i \|^2_F + \\ &
\frac{1}{2}\sum_{i=1}^{M} \| \tilde{\eta}_i - \eta_i \|^2_F +
\sum_{i=1}^{M} \| \tilde{\rho}_i - \rho_i \|^2_F,
\end{split}
\end{equation}
where $\hat{\rho}_i$ and $\hat{\eta}_i$ are the estimated depth and
registration maps, and $\tilde{\rho}_i$ and
$\tilde{\eta}_i$ are the outputs from the Main Block. The
terms $\rho_i$ and $\eta_i$ are the label maps, and
$M$ is the number of synthetic images.
The symbol $\|.\|_F$
is the Frobenius norm. 
We use ADAM optimisation
with a learning rate of $10^{-3}$ and parameters $\beta_1=\beta_2 = 0.9$. Training is fixed to $40$ epochs with a batch size of $7$, taking approximately 12 hours in a single GPU
workstation (Nvidia GTX1080). The weights are initialised with random uniform sampling~\cite{xavier_glorot}.

\subsection{Training step 2: Depth Refinement Block fine tuning}
We fine-tune the Depth Refinement Block weights using real data while freezing the weights $\theta_{\mathcal{M}}$ of the Main Block. This step is crucial to adapt the network to handle the render gap and to cope with real illumination conditions, camera response and color balance.
In this step a different loss function $\mathcal{L}_2$ is used, which combines a supervised loss for the Depth Refinement Block and a spatial regulariser:
\begin{equation}
\mathcal{L}_2(\theta_\mathcal{D}) = \sum_{i=1}^{M'} \|
\hat{\rho}_i - \rho_i \|^2_F +\lambda \sum_{i=1}^{M'}\|{\nabla \hat{\rho}}_i\|^2_1,
\label{eq:depthloss}
\end{equation}
where $M'$ is the number of real training images.
We include total variation regularisation~\cite{rudin} to mitigate the effect of noise in the depth labels while preserving edges and details~\cite{strong}. The hyperparameter $\lambda$ is set to $10^{-9}$ in all experiments and chosen empirically. We train with SGD and a small and fixed learning rate of $10^{-5}$. We train for $10$ epochs with a batch size of $7$. Having both a low learning rate and a reduced number of epochs allows us to adapt our network to real data while avoiding overfitting.

\subsection{Training step 3: Registration Refinement Block fine tuning}
In this step we use a property of SfT, which is that the input image can be synthesised from the registration solution, by warping the template texture map. We propose a self-supervised fine tuning algorithm for the Registration Refinement Block, based on minimising a photo-consistency loss that computes the error between the synthesised image and the input image. 
For each input image $I_i$, the corresponding synthesised image $I'_i$ is computed as follows:
\begin{equation}
I'_{i}(u, v)=\left\lbrace\begin{array}{l}
{A}\left(\hat\eta_{i}(u, v)\right) \\
0
\end{array}\right.\begin{array}{l}
(u,v)\in \hat{\mathcal{I}} \\
\text { otherwise,}
\end{array}
\label{eq:syntheticinput}
\end{equation}
where
$\hat{\mathcal{I}}(u,v) \triangleq \hat\rho(u,v)>-1$
is the object segmentation obtained from the Depth Refinement Block.
The computation of equation~\eqref{eq:syntheticinput} is first-order differentiable in the Registration Refinement Block network weights $\theta_{\mathcal{R}}$, as described in Appendix A\ref{app:appendixA}. 

The unsupervised loss function $\mathcal{L}_{u}$ forces the network to produce synthesised images that are photometrically similar to the input images. The loss involves the registration map
$\hat{\eta}_i$ computed by the Registration Refinement Block, each input image $I_i$ and their corresponding
synthesised image $I'_i$, and is defined as follows:
\begin{equation}
\begin{split}
\mathcal{L}_{u}(\theta_\mathcal{R}) =
\sum_{i=1}^{M'} \sum_{(u,v)\in{\mathcal{I}}} 
\chi \left(  \left(I'_i(u,v) - I_i(u,v)\right)^2\right)+ \\
\mu\sum_{i=1}^{M'} \sum_{(u,v)\in{\mathcal{I}}} 
\chi \left ( \left(I'^{\downarrow}_{i}(u,v) - I^{\downarrow}_{i}(u,v) \right)^2\right )
+\lambda \sum_{i=1}^{M'}\|{\nabla \hat{\eta}}_i\|^2_1.
\end{split}
\label{eq:photoref}
\end{equation}
where $M'$ is the number of training images, $\chi(x)$ is an M-estimator, and images $I'^{\downarrow}_{i}$ and $I^{\downarrow}_{i}$ are downsized versions of
$I'_{i}$ and $I_{i}$ respectively by a factor $2$.
These are used to include losses at two spatial scales, which improves convergence similarly to image pyramids used in unsupervised optical flow \cite{Ren2017UnsupervisedDL}. The loss is controlled by a
hyperparameter weight $\mu$, fixed to $0.5$ in all our
experiments. To handle illumination
changes and shading effects that violate photo-consistency, we use the Cauchy
M-estimator:
\begin{equation}
\chi(x)=\frac{c^{2}}{2} \log \left(1+(x / c)^{2}\right),
\label{eq:cauchy}
\end{equation}
with $c=4$ as default. We also include a total variation regularisation term in the loss that imposes smoothness in the registration output while preserving discontinuities. This term is usually included in optical flow methods~\cite{Ren2017UnsupervisedDL} to improve convergence.  The hyperparameter $\lambda$ is set empirically to $10^{-9}$ in all experiments. 
We optimise $\mathcal{L}_{u}$ using SGD with momentum. We found that optimisers with an adaptive step, such as ADAM, or large learning rates cause convergence problems when minimising $\mathcal{L}_{u}$. We use an initial learning rate of $10^{-5}$ and a decay of $10^{-9}$. The Registration Refinement Block is trained for $10$ epochs with a batch size of $6$.

\subsection{Handling different camera intrinsics}
\label{sec:otherCams}
We have been able to generalise DeepSfT to work with a different camera at test time without any need to retrain the network weights. This has not been achieved with other DNN-based SfT methods and it significantly broadens our applicability for real-world use. Recall that we train DeepSfT with images generated by a camera with fixed intrinsics (called the training camera), which may potentially have different intrinsics to the test camera. Once the network is trained, we cannot immediately input images from the test camera into the network because its weights are trained specifically for the intrinsics of the training camera. We propose to handle this by adapting the test camera's effective intrinsics to match the training camera. Because the object's depth within the training set varies (and so do the perspective effects), we can emulate testing on the training camera by applying a known affine transform to images from the test camera. The affine transform is the matrix $A=K_{train} K_{test}^{-1}$, where $K_{train}$ and $K_{test}$ are the intrinsic matrices of the training and test cameras respectively.

The transformed test image is then clipped about its principal point and zero padded, if necessary, to obtain the canonical resolution of $270\times480$ (the input image size of DeepSfT).

\section{Experimental Results}
\label{sec:Experiments}
\subsection{Datasets}
\subsubsection{Templates}
\label{sec:templates}
We have tested DeepSfT with 5 objects represented by 3 thin-shell and 2 volumic templates shown in table~\ref{tb:synDeformations}. We refer to these as DS1 to DS5. DS1 models an A4 paper sheet with very poor texture. DS2 models an A4 paper sheet with a richer texture and DS5 models an A4 paper sheet from a well-known public dataset~\cite{cvlab}. DS3 is a volumic model of a soft toy and DS4 is a volumic model of an adult sneaker. DS1, DS2 and DS5 can be modelled with a rectangular template, however DS3 and DS4 cannot. They were built with triangular meshes using dense SfM (Agisoft Photoscan~\cite{photoscan}). We emphasize that no previous work has been able to solve SfT for volumic templates like DS3 and DS4 in the wide-baseline setting. 

\subsubsection{Synthetic datasets}
\label{sec:synDB}
For each template a synthetic dataset was constructed by deforming the template with random quasi-isometric deformations and rendering the deformed template with fixed camera intrinsics and random viewpoints. 
\begin{table*}[!htbp]
	
	\centering
	\begin{adjustbox}{max width=\linewidth}
		\begin{tabular}{m{3cm}m{3cm}m{3cm}m{3cm}m{3cm}}
			\multicolumn{5}{c}{\rule{0pt}{3ex} \large{Template 3D shapes}}\\\hline 
			
			\vspace{1.12mm}{\normalsize DS1}&\vspace{1.12mm}{\normalsize DS2}&\vspace{1.12mm}{\normalsize DS3}&\vspace{1.12mm}{\normalsize DS4}&\vspace{1.12mm}{\normalsize DS5}\\
			\vspace{1.52mm}\includegraphics[width=30mm]{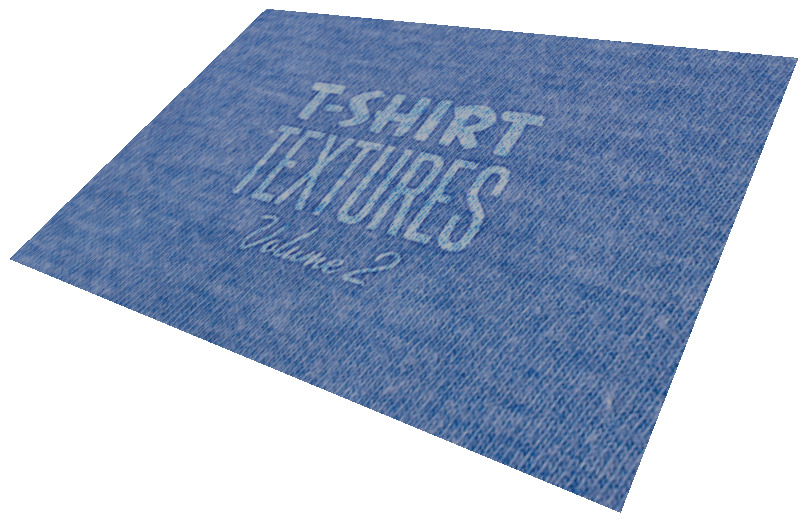} & \vspace{1.52mm}\includegraphics[width=30mm]{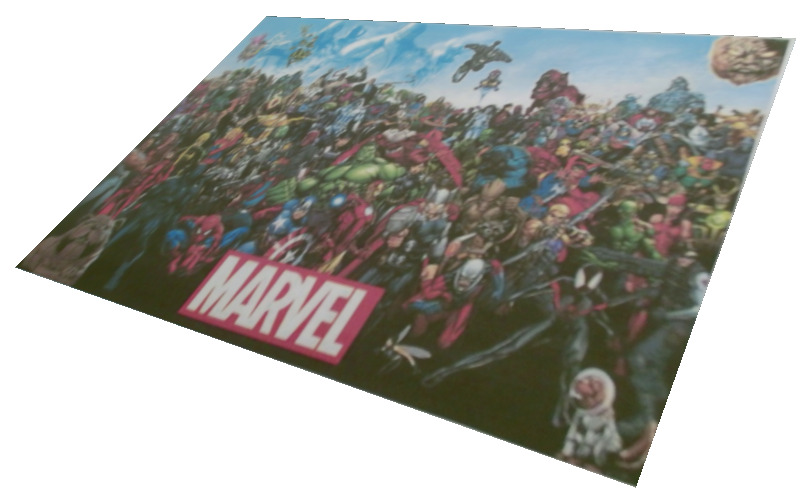}
			& \vspace{1.52mm}\includegraphics[width=30mm]{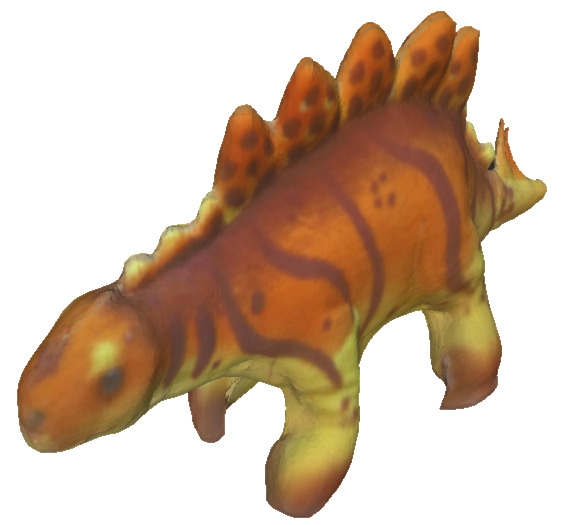}  & 
			\vspace{1.52mm}\includegraphics[width=30mm]{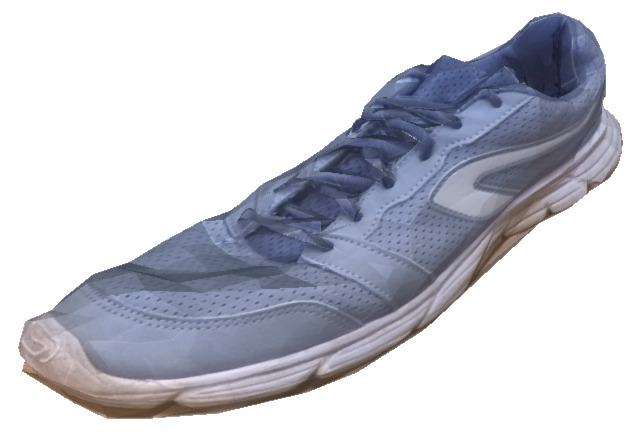} & 
			\vspace{1.52mm}\includegraphics[width=30mm]{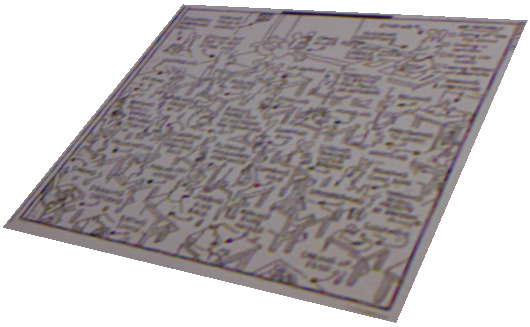}  \\
			Mesh Faces=1521& Mesh Faces=1521&Mesh Faces=36256&Mesh Faces=5212&Mesh Faces=1521\\      
			\multicolumn{5}{c}{\rule{0pt}{3ex} \large{Template texture maps}}\\\hline          
			\vspace{1.52mm}\includegraphics[width=30mm]{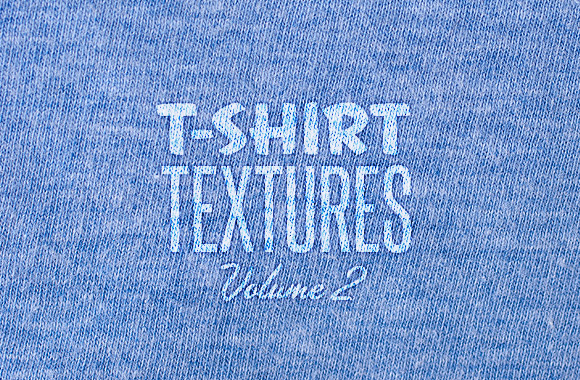} & \vspace{1.52mm}\includegraphics[width=30mm]{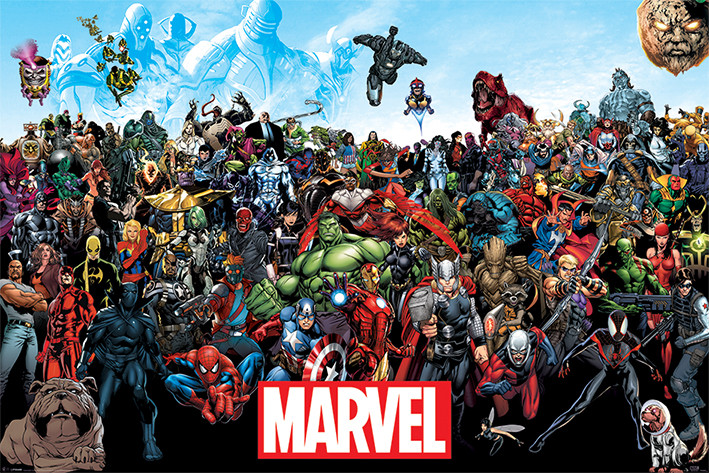}
			& \vspace{1.52mm}\includegraphics[width=30mm]{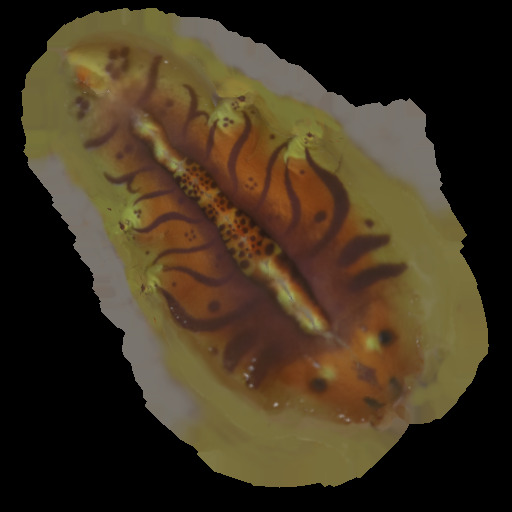}  & 
			\vspace{1.52mm}\includegraphics[width=30mm]{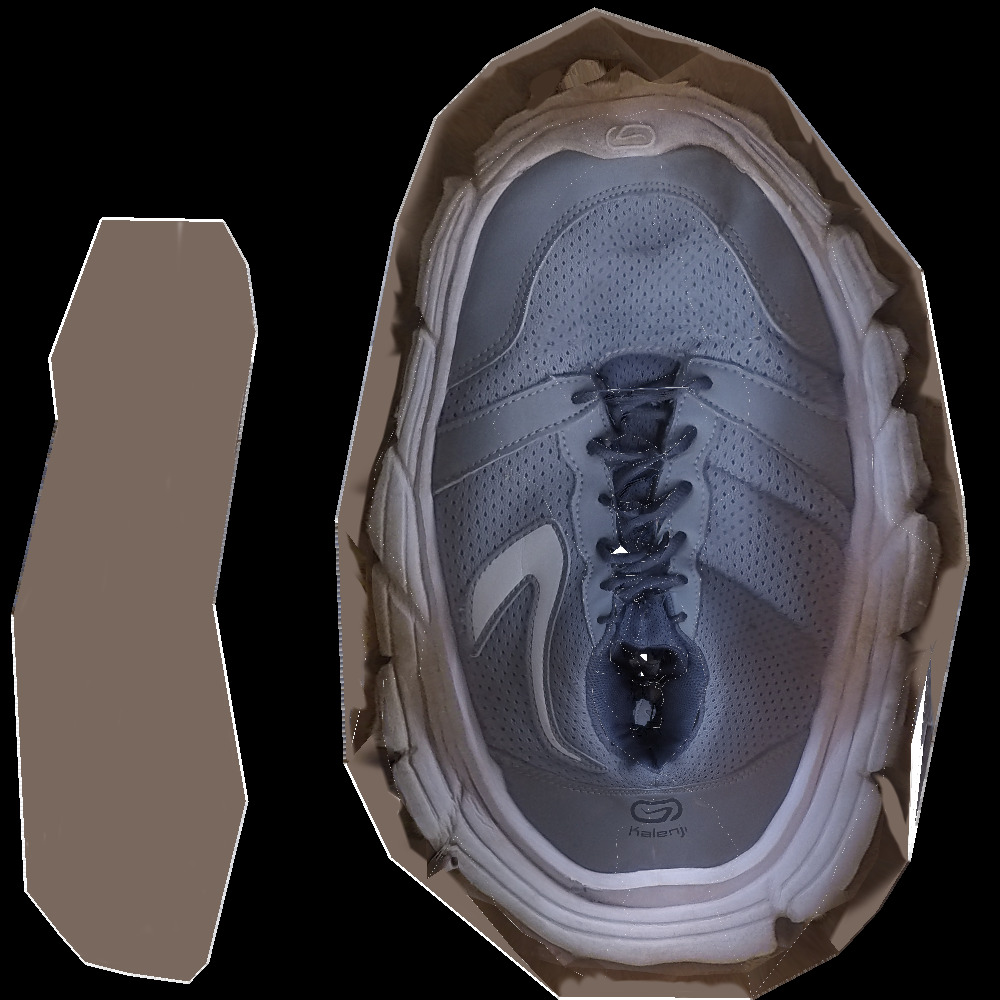} 
			& 
			\vspace{1.52mm}\includegraphics[width=30mm]{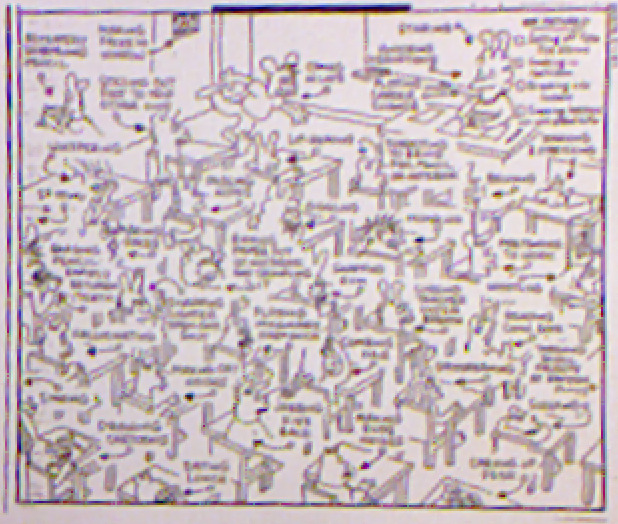}  \\      
			
			\multicolumn{5}{c}{\rule{0pt}{3ex} \large{Synthetic images}}\\\hline
			\vspace{1.52mm}\includegraphics[width=30mm]{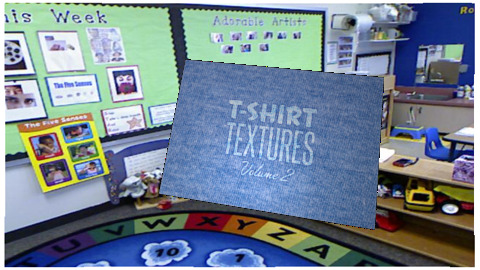} & \vspace{1.52mm}\includegraphics[width=30mm]{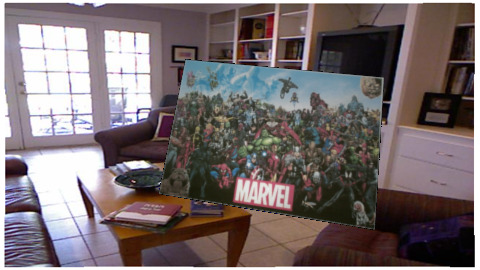}
			& \vspace{1.52mm}\includegraphics[width=30mm]{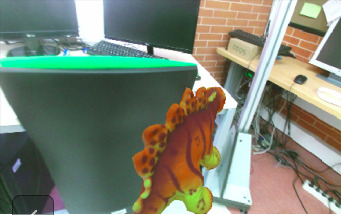}  & \vspace{1.52mm}\includegraphics[width=30mm]{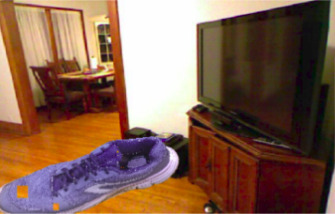}& \vspace{1.52mm}\includegraphics[width=30mm]{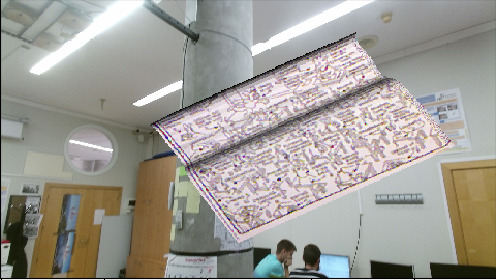}  \\ 

			\vspace{1.52mm}\includegraphics[width=30mm]{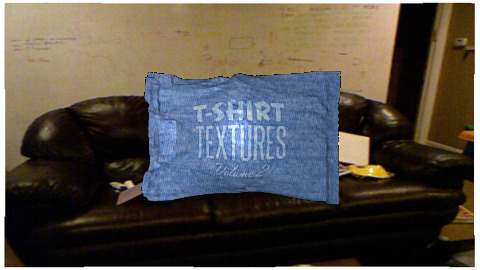}  &
			\vspace{1.52mm}\includegraphics[width=30mm]{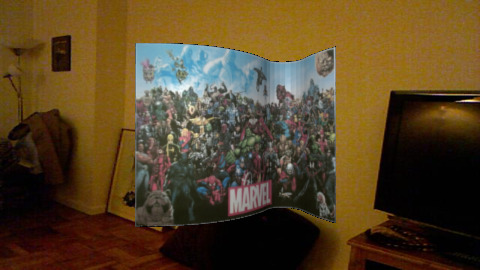} & \vspace{1.52mm}\includegraphics[width=30mm]{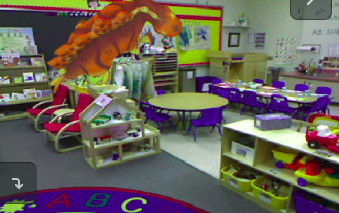} &\vspace{1.52mm}\includegraphics[width=30mm]{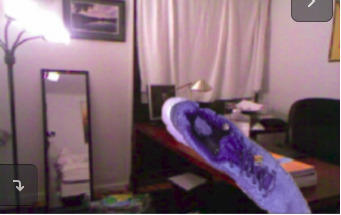}& \vspace{1.52mm}\includegraphics[width=30mm]{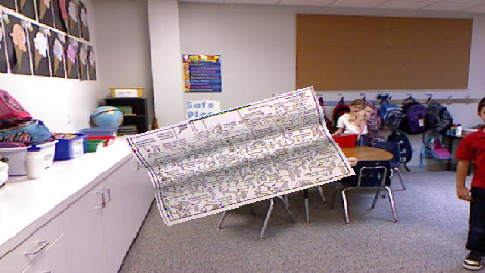}  \\ 
			\multicolumn{5}{c}{\rule{0pt}{3ex} \large{Real images}}\\\hline
			\vspace{1.52mm}\includegraphics[width=30mm]{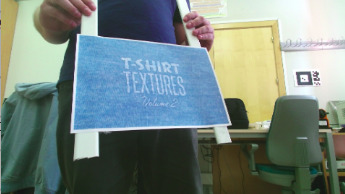} & \vspace{1.52mm}\includegraphics[width=30mm]{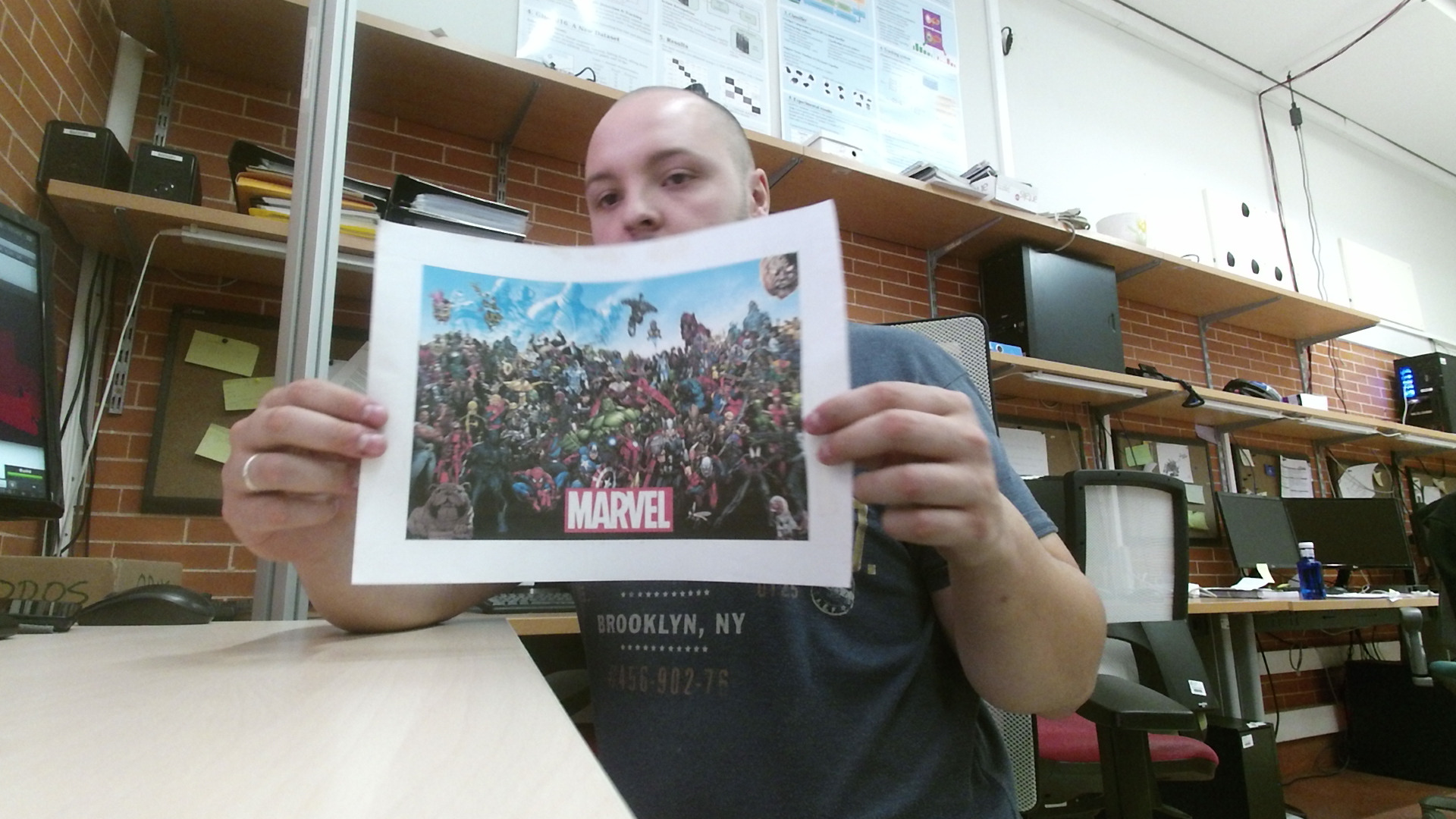}
			& \vspace{1.52mm}\includegraphics[width=30mm]{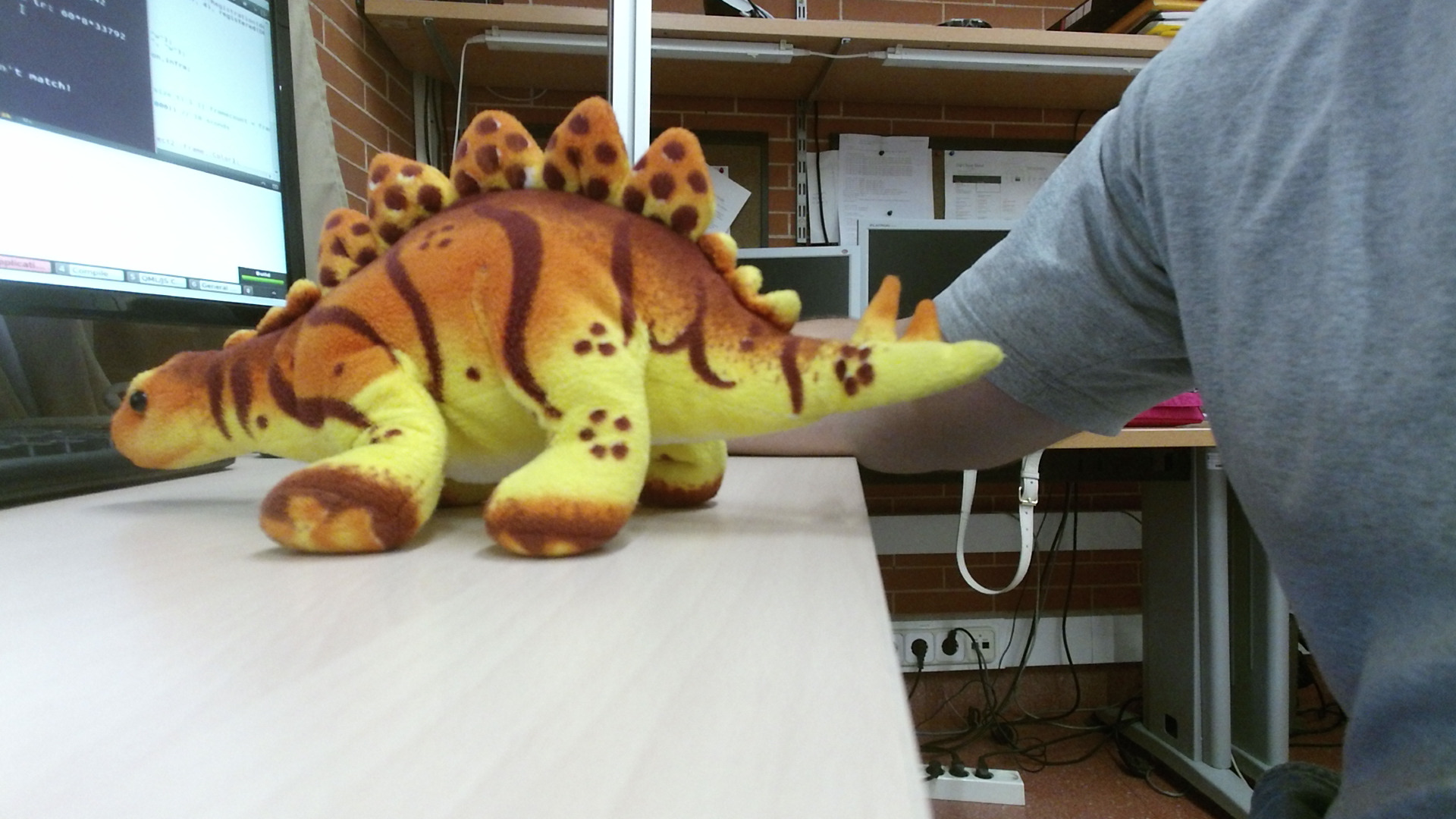}  & \vspace{1.52mm}\includegraphics[width=30mm]{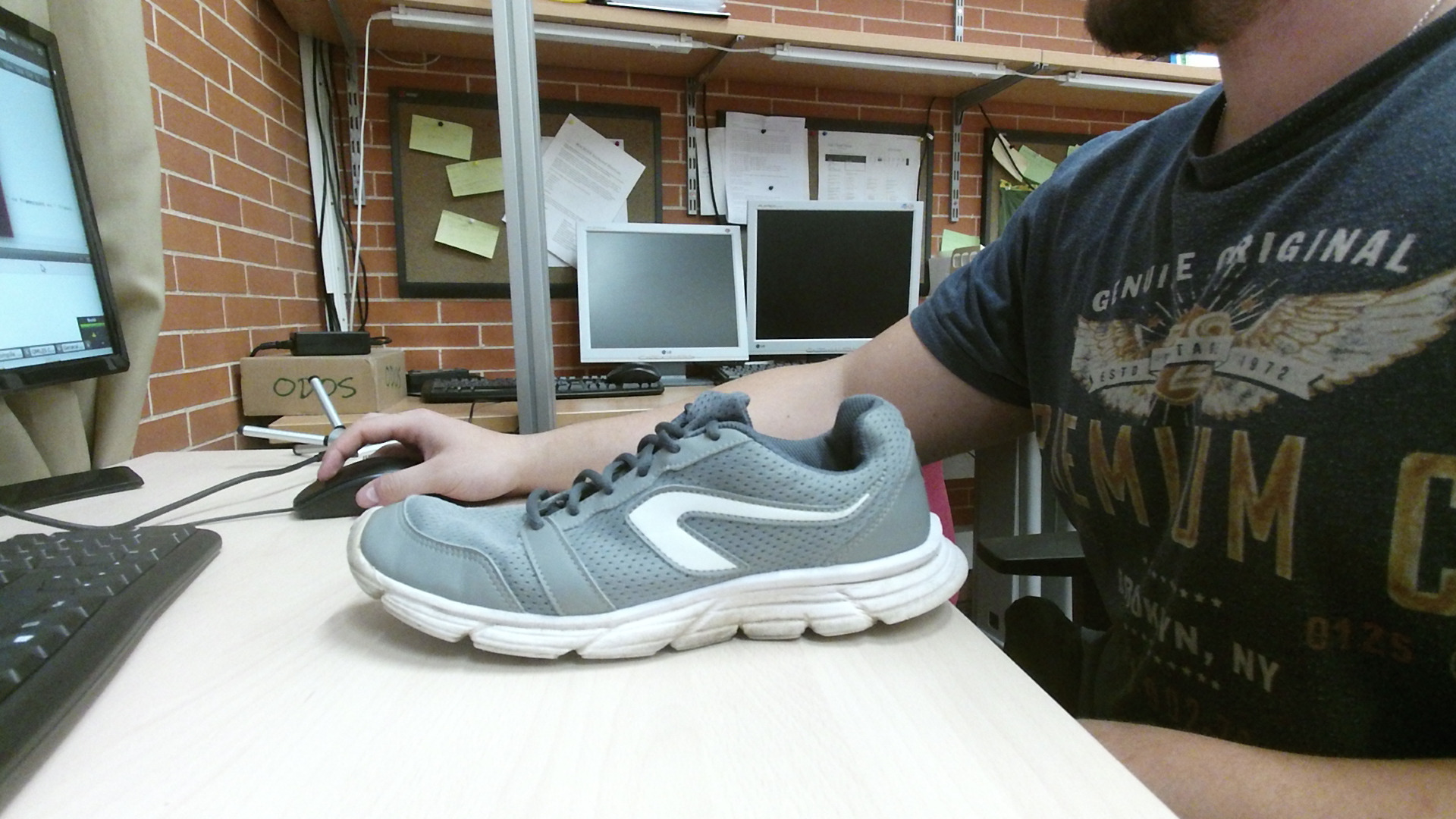} & \vspace{1.52mm}\includegraphics[width=30mm]{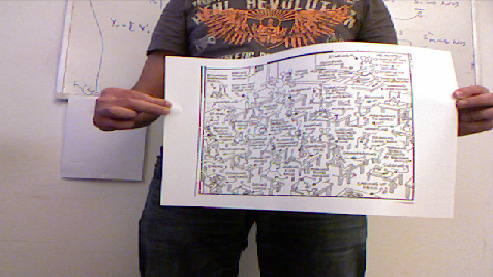}  \\ 
			
			\vspace{1.52mm}\includegraphics[width=30mm]{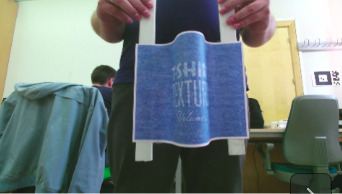} & \vspace{1.52mm}\includegraphics[width=30mm]{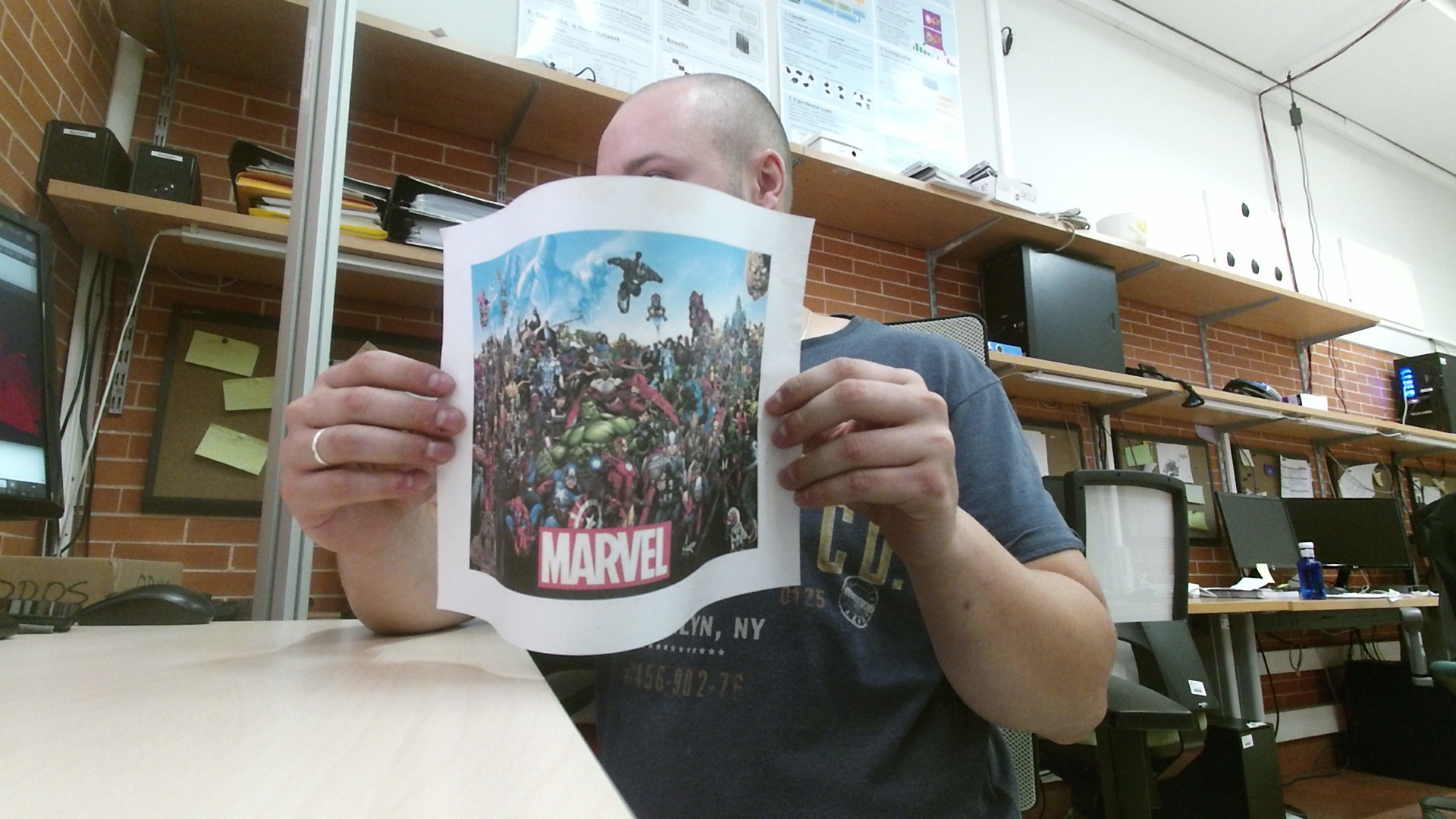}
			& \vspace{1.52mm}\includegraphics[width=30mm]{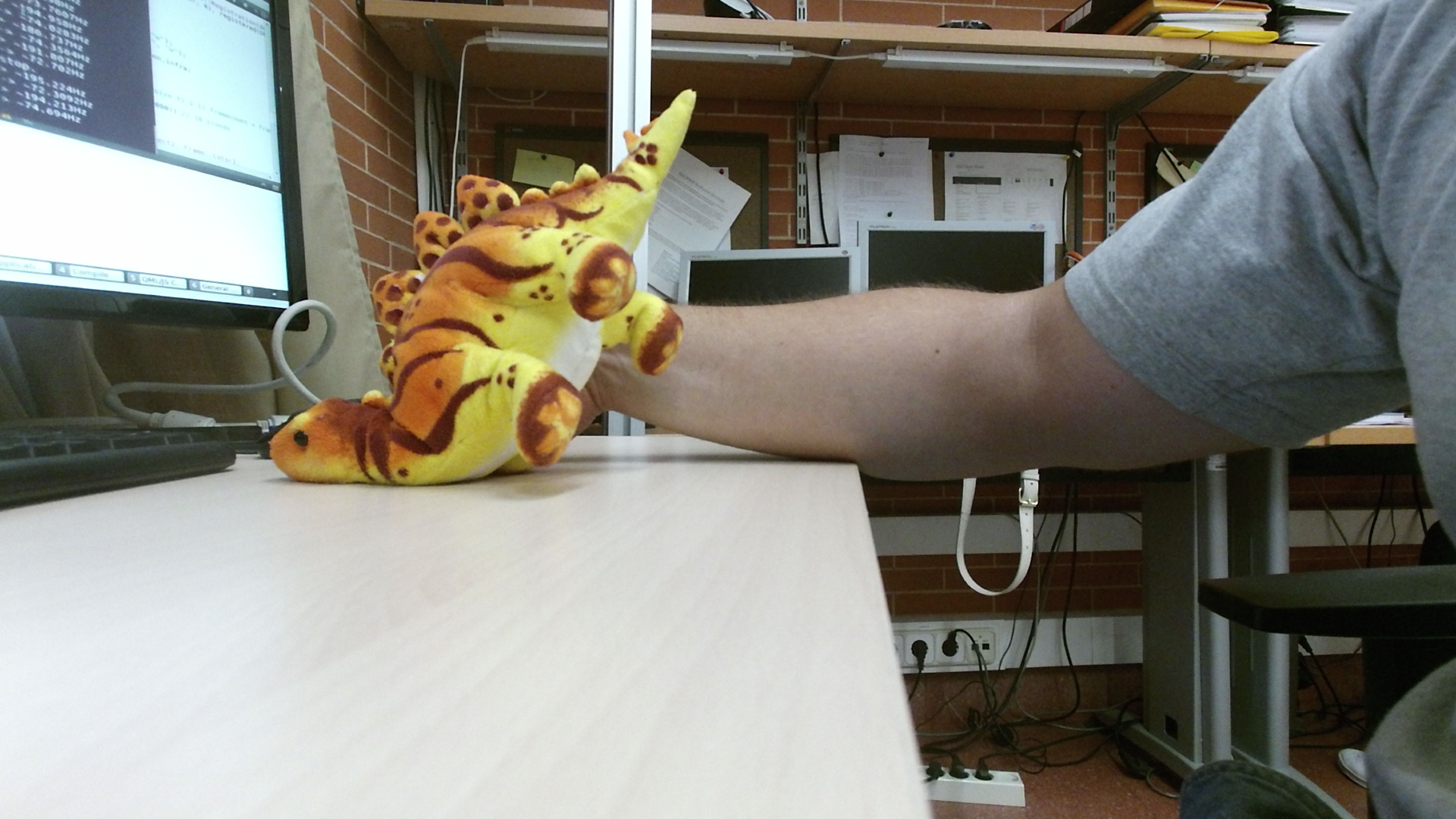} &  \vspace{1.52mm}\includegraphics[width=30mm]{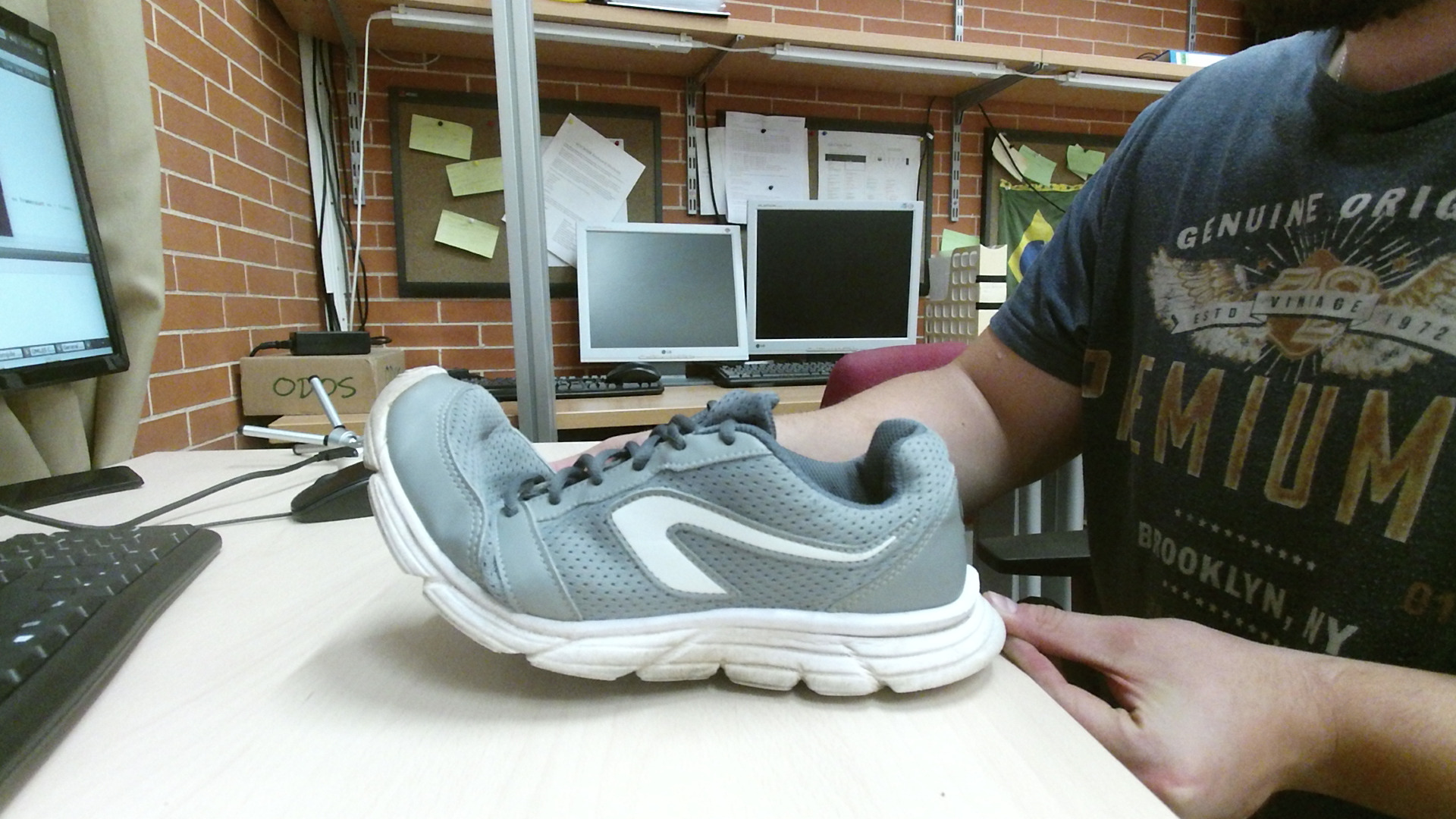}&  \vspace{1.52mm}\includegraphics[width=30mm]{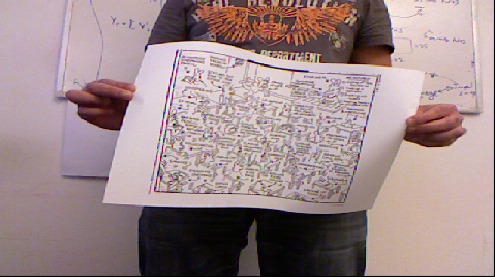} \\ 

		\end{tabular} 
	\end{adjustbox}
	\caption {Visualization of templates and input images. Rows 1 and 2 show the five templates DS1, DS2, DS3, DS4 and DS5. Rows 3 and 4 show example renders with simulated deformations. Rows 5 and 6 show real deformations of the physical objects.}
	\label{tb:synDeformations}
\end{table*}
We used \emph{Blender}~\cite{Blender}, which includes a physics-based
simulation engine to simulate deformations with different degrees of stiffness
using position-based dynamics. For DS1, DS2 and DS5 (rectangular templates) we simulated continuous videos with a high stiffness term and randomly located 3D anchor points. We applied tensile and compressive forces in randomised 3D directions. The simulation parameters are given in the supplementary material. For DS3 and DS4 (volumic
templates) we used rig-based deformations with hand-crafted rigs. We generated
independent deformations for each image using random joint angles. 

For each deformation we rendered an image with a random camera pose (random 
rotation around the camera's optical axes with angle variations in the interval
$[-\frac{\pi}{4},\frac{\pi}{4}]$ radians and random translations in the intervals $t_x\in [-150,150]$
mm, $t_y\in [-150,150]$ mm and $ t_z \in [100,600] $ mm). A distant light model was used with illumination angles parameterised by spherical coordinates that was drawn randomly in the interval $[-\frac{\pi}{18},\frac{\pi}{18}]$ radians
around the camera's optical axis. The diffuse surface reflectance component was modelled as Lambertian and the specular component was modelled with Blender's Cook-Torrence model.  We generated
brightness variations by a random gain in the range $[0.9,1.1]$. We randomly changed the image background with
images from \cite{nyudepth}. To simulate occlusions, we randomly
introduced a maximum of $4$ synthetically generated circles of constant random color
in each image with variable diameter in the range $[1,10]$ px at random locations. In total, each dataset consists of $60\space000$ RGB images with labelled depth and
registration maps. These were standardised to a canonical resolution of $270\times480$ px.

\subsubsection{Real datasets}
\label{sec:realDB}
Real datasets of each object were recorded with Microsoft Kinect v2 with deformations caused by hand manipulation, as shown in table~\ref{tb:synDeformations}. Videos for DS1, DS2, DS3 and DS4 were recorded by us and the video for DS5 was provided in the public dataset (192 frames). The recorded depth maps were aligned with the RGB images using the extrinsic parameters and downsized to $270\times480$ px. Note that these RGB-D videos do not provide labelled registration data.

\subsubsection{Training/testing data splits}
We evaluate DeepSfT in terms of reconstruction and registration errors with synthetic and real test data. Synthetic test data were generated using the same process as the synthetic training data (\S\ref{sec:synDB}), using random configurations not present in the training data. Real test data were generated using the same process as the real training data, using new videos, consisting of new viewpoints and object manipulations not present in the training data. We also generated test data using two new real cameras: an Intel Realsense D435\cite{realsense} (an RGB-D camera for quantitative reconstruction evaluation) and a Gopro Hero V3\cite{gopro} (an RGB camera for qualitative evaluation). Table \ref{tb:qualitative_cameras} shows their respective camera intrinsics. 
\begin{table}[!htbp]
	\setcellgapes{3pt}\makegapedcells
	\begin{adjustbox}{max width=\linewidth}
		\begin{tabular}{|c|c|c|c|c|c|}
			\hline
			Camera & Resolution & $f_u$ & $f_v$ & $c_u$ & $c_v$ \\ \hline			
			Kinect V2 & $1920\times1080$ & $1057.8$ & $1064.0$ & $947.6$ & $530.4$ \\ \hline
			Intel Realsense D435 & $1270\times720$& $915.5$ & $915.5$ & $645.5$& $366.3$ \\ \hline
			Gopro Hero V3 &$1920\times1080$& $1686.8$ & $1694.2$ & $952.8$ & $563.5$ \\ \hline			
		\end{tabular} 
	\end{adjustbox}
	\centering \caption {Camera intrinsics of the different real cameras used in our experiments. We use Kinect V2 for training and all three cameras for testing.}
	\label{tb:qualitative_cameras}
\end{table}

Table \ref{tb:train_test_split} shows the train and test split for all real datasets. When testing DeepSfT with synthetic data, results from the Main Block are evaluated. When testing with real data, results from the Depth Refinement Block and Registration Refinement Block are evaluated. 

\subsection{Compared methods and evaluation metrics}
We compare DeepSfT with two classical state-of-the-art SfT methods. The first is an isometric SfT method~\cite{Chhatkuli2017} with public code, referred to as CH17. We provide this method with two types of registration: CH17+GTR uses Ground-truth Registration (indicating its best possible performance independent of the registration method) and CH17+DOF uses a state-of-the-art Dense Optical Flow registration method~\cite{optical_flow}. In the latter case we generate registration only for image sequences using frame-to-frame tracking. We also add to these two methods a final refinement step based on minimising a statistically optimal non-convex cost function with Levenberg-Marquardt~\cite{Bartoli2015}. We refer to the refined solutions as CH17R+GTR and CH17R+DOF. The second classical SfT method we test is~\cite{7410619} with public code, referred to as NGO15.

We compare DeepSfT with three DNN-based methods. The first is a na{\"i}ve application of the popular ResNet architecture~\cite{he2016deep} to solve SfT, referred to as R50F. The reason to include the ResNet model was to compare our fully convolutional encoder-decoder architecture against a combination of an encoder and a fully connected model. This comparison demonstrates that the proposed architecture outperforms the classic encoder-fully connected architectures such as ResNet. We adapt ResNet by removing the final two layers and introduce a dense layer with 200 neurons and a final dense layer with a 3-channel output (for depth and registration maps) of the same size as the input image. We trained ~R50F with exactly the same training data as DeepSfT and with real-data fine tuning. Fine-tuning was implemented by optimising the depth loss, using the same optimiser and learning rate as we used for DeepSfT. The second DNN method is \cite{hdm_net}, which we refer to as HDM-net. This is tested only with rectangular templates (DS1 and DS2) because it only handles textureless or weakly textured rectangular templates.

We carefully re-implemented \cite{hdm_net}, requiring an adaptation of the image input size and the mesh size so that it matched the size of the template meshes. The third DNN method is \cite{Shimada2019} using the authors' code, which we refer to as IsMo-GAN, that is also applied only to DS1 and DS2 as it requires a rectangular template. 

We evaluate reconstruction error using the Root Mean Square Error (RMSE) in millimeters. We also use RMSE to evaluate the registration accuracy in pixels. The evaluation of registration accuracy is notoriously difficult with real data because there is no way to obtain reliable ground-truth. We propose to use as a proxy for the ground-truth the output from a state-of-the-art dense trajectory optical flow method DOF \cite{optical_flow}. We only make this quantitative evaluation for videos, for which DOF can reliably compute registration. We manually selected sequences where DOF produces stable tracks. The use of DOF or any other optical flow method as a registration baseline can introduce bias. However, obtaining registration results with a wide-baseline method such as DeepSfT that are comparable with DOF is considered a very strong result for a wide-baseline method.

\begin{table*}[!htbp]
	\setcellgapes{3pt}\makegapedcells
	\begin{adjustbox}{max width=0.8\linewidth}
		\begin{tabular}{|c|c|c|c|c|c|c|c|c|c|c|c|c|c|}
			\hline
			\multicolumn{2}{|c}{}&\multicolumn{3}{|c|}{Registration RMSE (px) }&\multicolumn{9}{c|}{Reconstruction RMSE (mm)} \\ \hline
			Sequence & Samples & DOF &  R50F & DeepSfT & CH17+GTR & CH17+DOF & CH17R+GTR & CH17R+DOF& NGO15 & HDM-net& IsMo-GAN & R50F & DeepSfT   \\ \hline
			
			\large{DS1S} & \large{5000} & \large{4.63}  & \large{6.69}  & \large{\textbf{1.87}} & \large{6.89} & \large{15.60} & \large{8.27} & \large{15.41}& \large{18.77} & \large{10.80}&  \large{7.32}& \large{7.99} & \large{\textbf{1.68}}   \\ \hline
			
			\large{DS2S} & \large{5000} & \large{5.91}  & \large{6.13}  & \large{\textbf{1.34}} & \large{6.89} & \large{28.26} & \large{8.27} & \large{28.04}& \large{21.32} & \large{9.92}&  \large{6.94} & \large{7.75} & \large{\textbf{1.63}}   \\ \hline
			
			\large{DS1R} & \large{232} & -  & \large{5.02}  & \large{\textbf{2.32}} & \large{-} & \large{38.12} & \large{-} & \large{34.24}& - & \large{-}&  \large{-} & \large{17.53} & \large{\textbf{9.51}}   \\ \hline
			
			\large{DS2R} & \large{373} & \large{-}  & \large{4.13}  & \large{\textbf{1.53}} & \large{-} & \large{27.31} & \large{-} & \large{25.24}& - & \large{-}&  \large{-} & \large{14.45} & \large{\textbf{7.37}}   \\ \hline
			
			\large{DS5R} & \large{50} & \large{-}  & \large{6.33}  & \large{\textbf{2.74}} & \large{-} & \large{22.57} & \large{-} & \large{19.42}& \large{32.3} & \large{-}&  \large{-} & \large{16.30} & \large{\textbf{6.97}}   \\ \hline				
		\end{tabular}
	\end{adjustbox}
	\centering \caption {Quantitative evaluation on synthetic and real test data with rectangular templates (DS1S, DS2S, DS1R, DS2R and DS5R).}
	\label{tb:exp_sintetic}
\end{table*}

\begin{table}[!htbp]
	\begin{adjustbox}{max width=\linewidth}
		\begin{tabular}{lm{2cm}m{2cm}m{2cm}m{2cm}}

			\rule{0pt}{2ex} \centering  &\multicolumn{1}{r}{DS1}&\multicolumn{1}{r}{DS3} &\multicolumn{1}{r}{DS4}&\multicolumn{1}{r}{DS5}\\ 
			
			\rule{0pt}{2ex} \centering   Input Image &\vspace{1.52mm}\includegraphics[width=20mm]{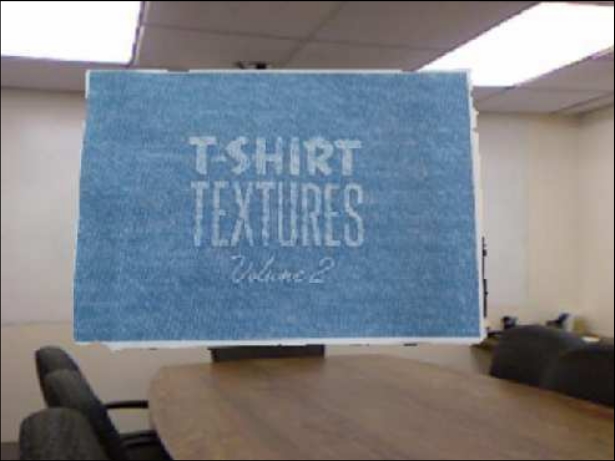}& \vspace{1.52mm}\includegraphics[width=20mm]{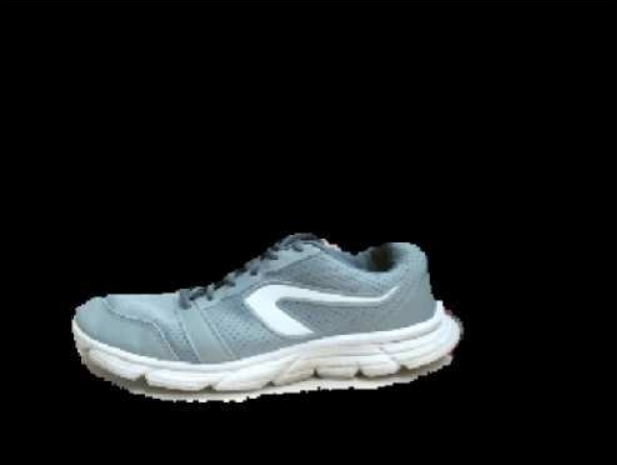}& \vspace{1.52mm}\includegraphics[width=20mm]{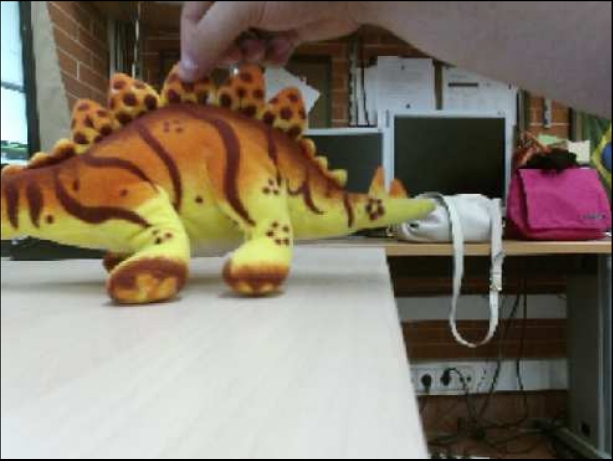}&\vspace{1.52mm}\includegraphics[width=20mm]{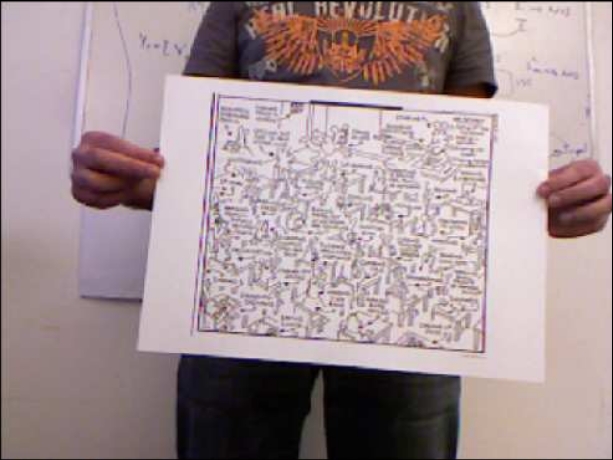}\\
			
			\rule{0pt}{2ex} \centering   Ground-truth &\vspace{1.52mm}\includegraphics[width=20mm]{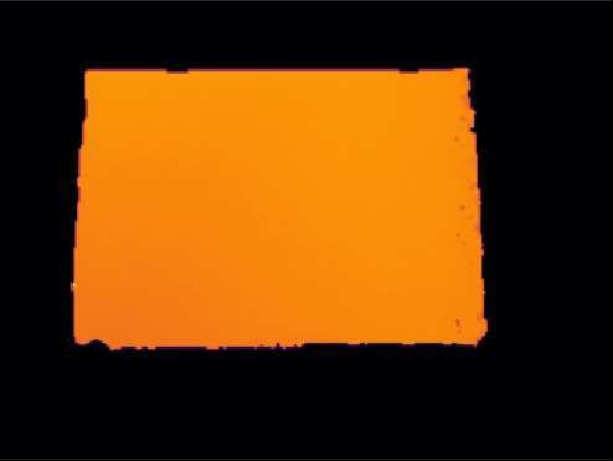}& \vspace{1.52mm}\includegraphics[width=20mm]{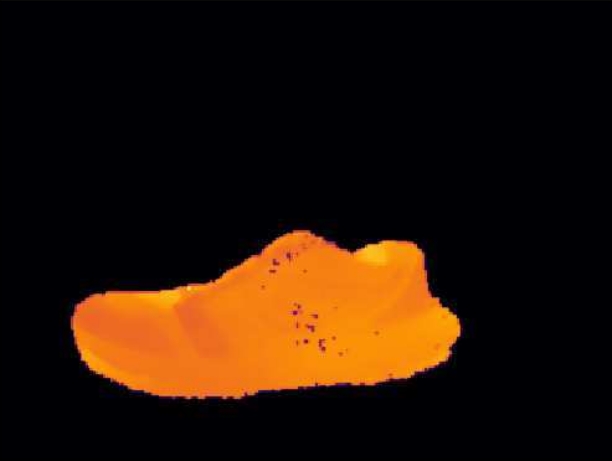}& \vspace{1.52mm}\includegraphics[width=20mm]{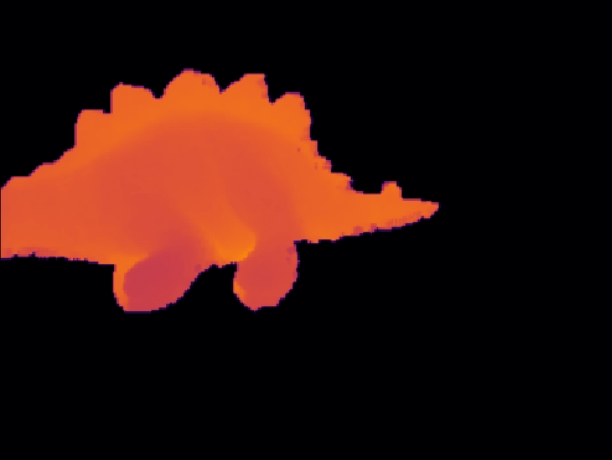}&\vspace{1.52mm}\includegraphics[width=20mm]{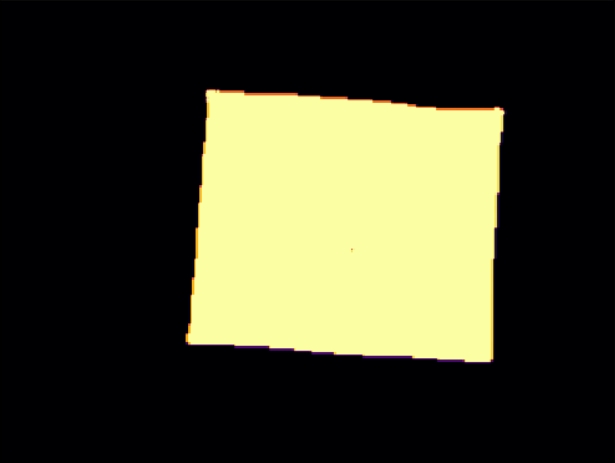}\\ 

			\rule{0pt}{2ex} \centering  DenseDepth+FT  & \vspace{1.52mm}\includegraphics[width=20mm]{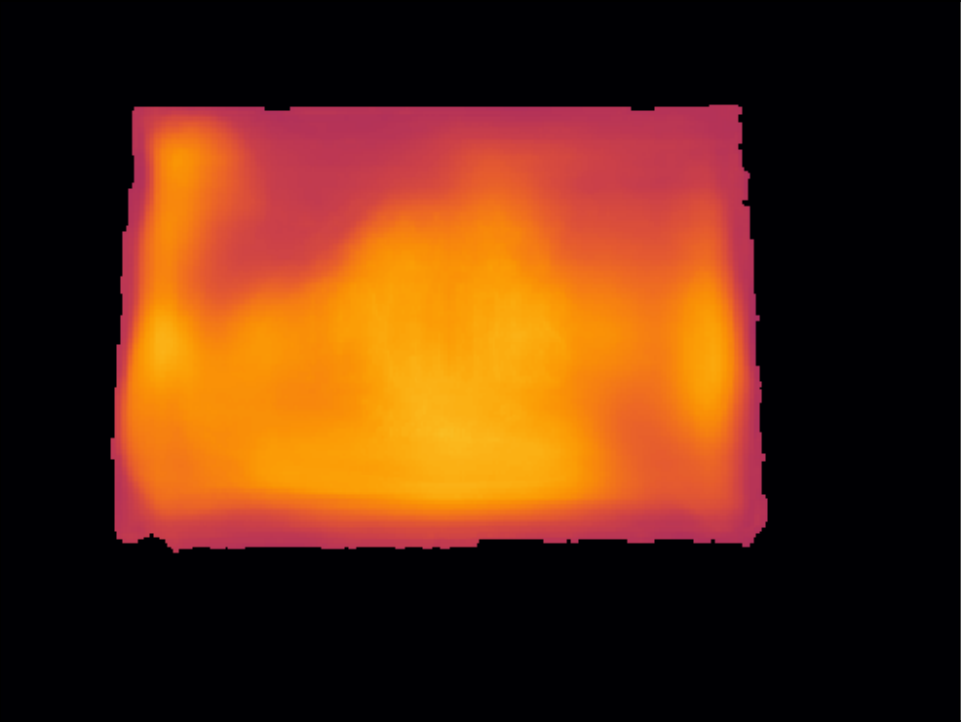}& \vspace{1.52mm}\includegraphics[width=20mm]{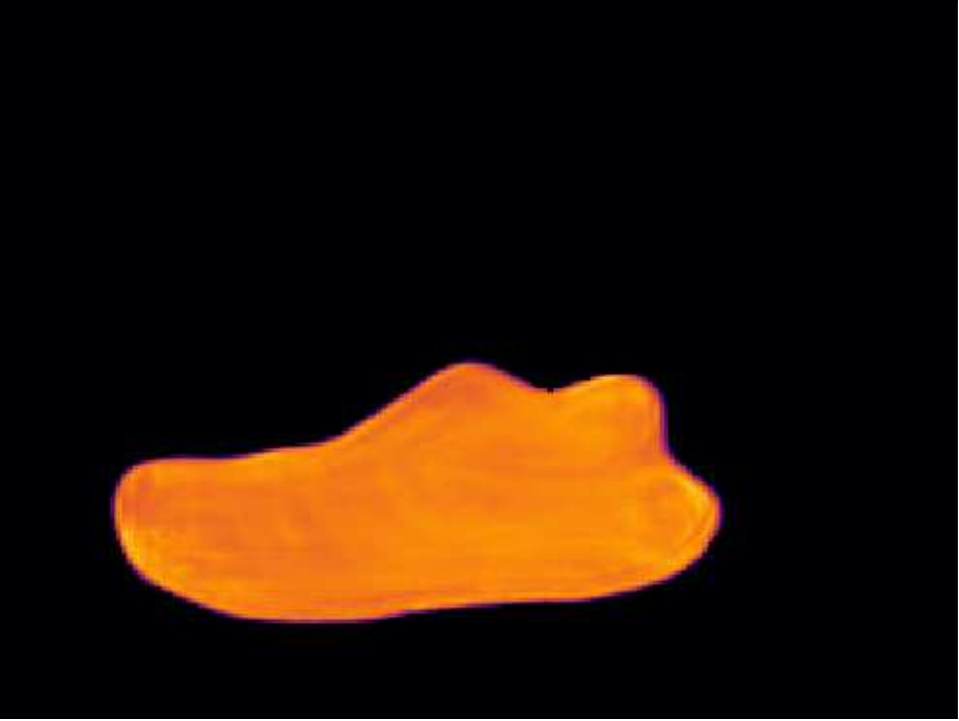}&  \vspace{1.52mm}\includegraphics[width=20mm]{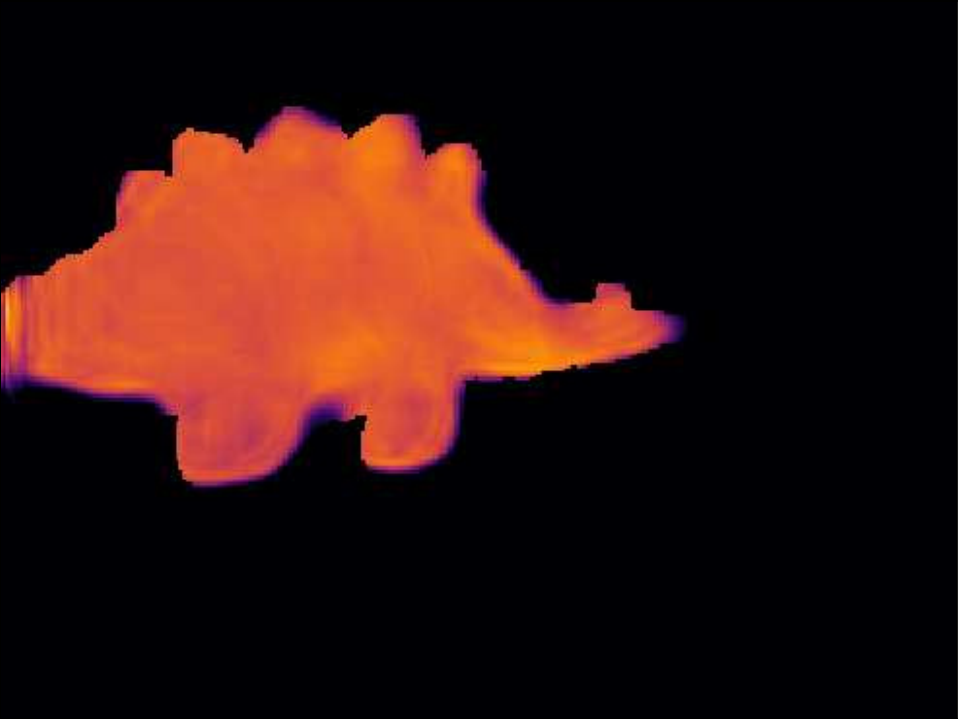}& \vspace{1.52mm}\includegraphics[width=20mm]{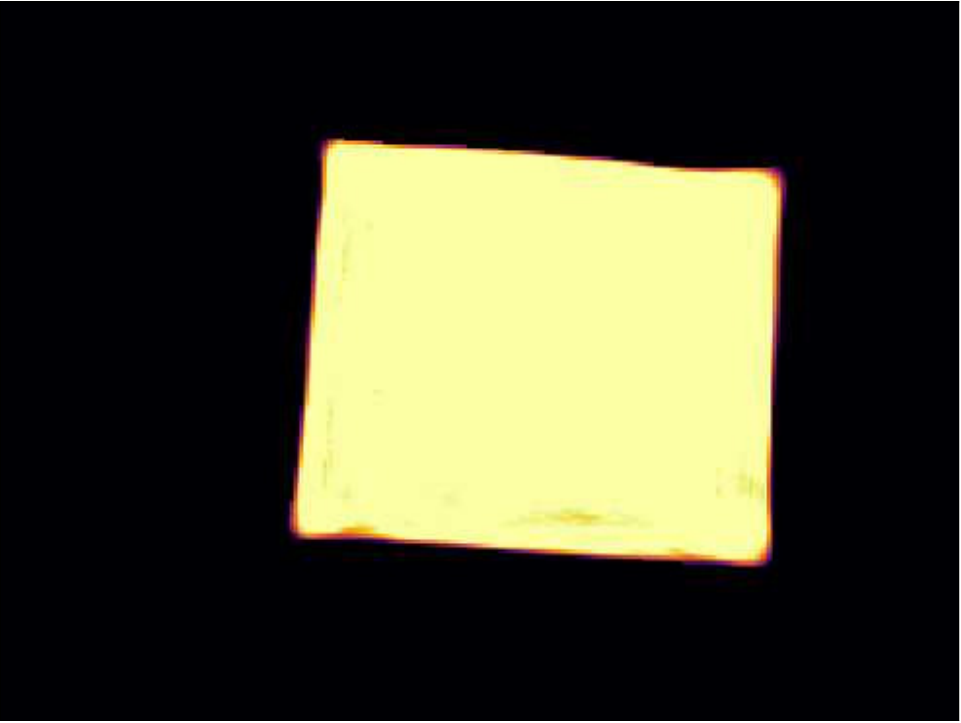}\\
			
			\rule{0pt}{2ex} \centering  &\multicolumn{1}{r}{46.31}&\multicolumn{1}{r}{17.46} &\multicolumn{1}{r}{28.68}&\multicolumn{1}{r}{20.73}\\ 
			
			\rule{0pt}{2ex} \centering  BTS+FT  &\vspace{1.52mm}\includegraphics[width=20mm]{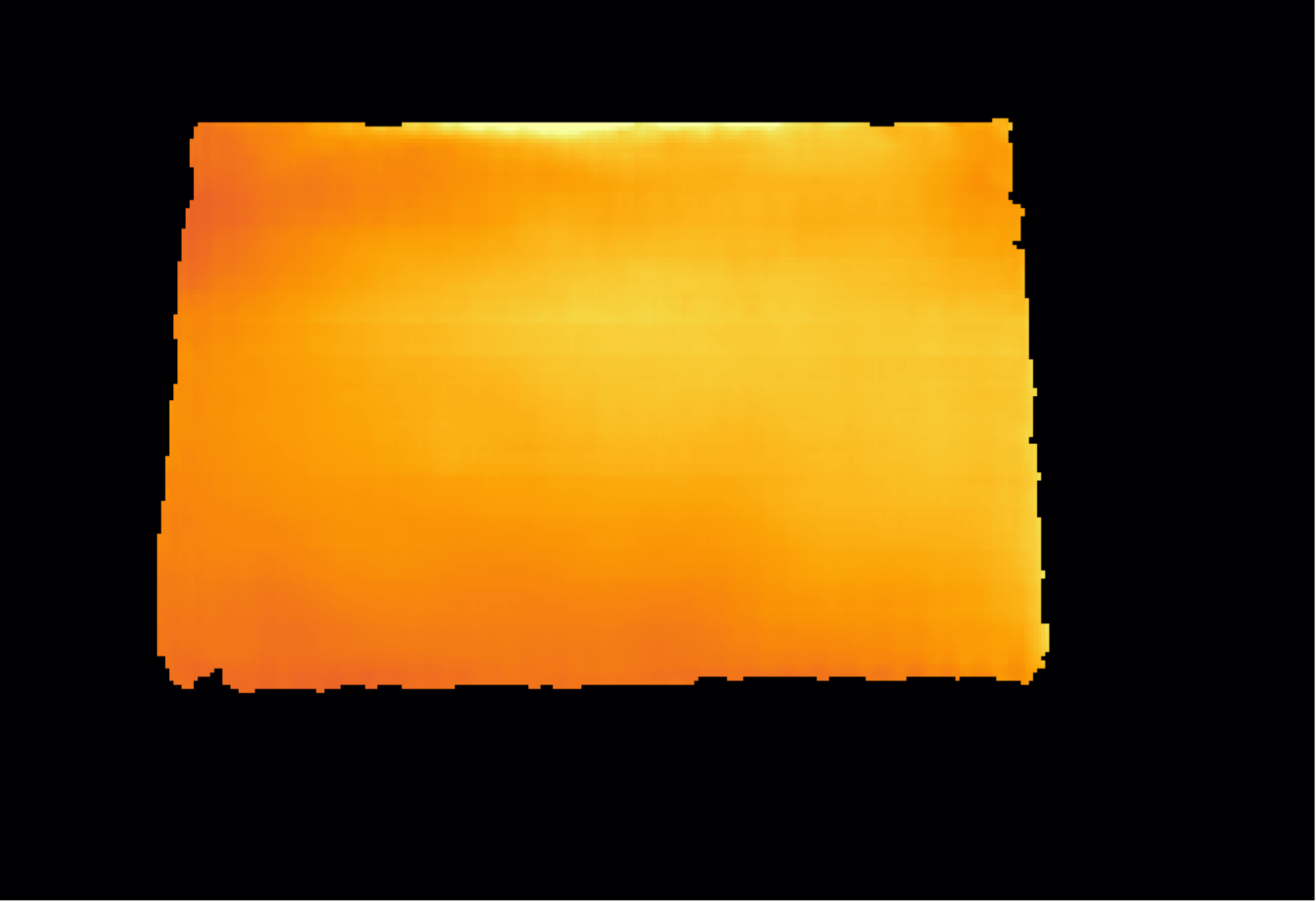}&\vspace{1.52mm}\includegraphics[width=20mm]{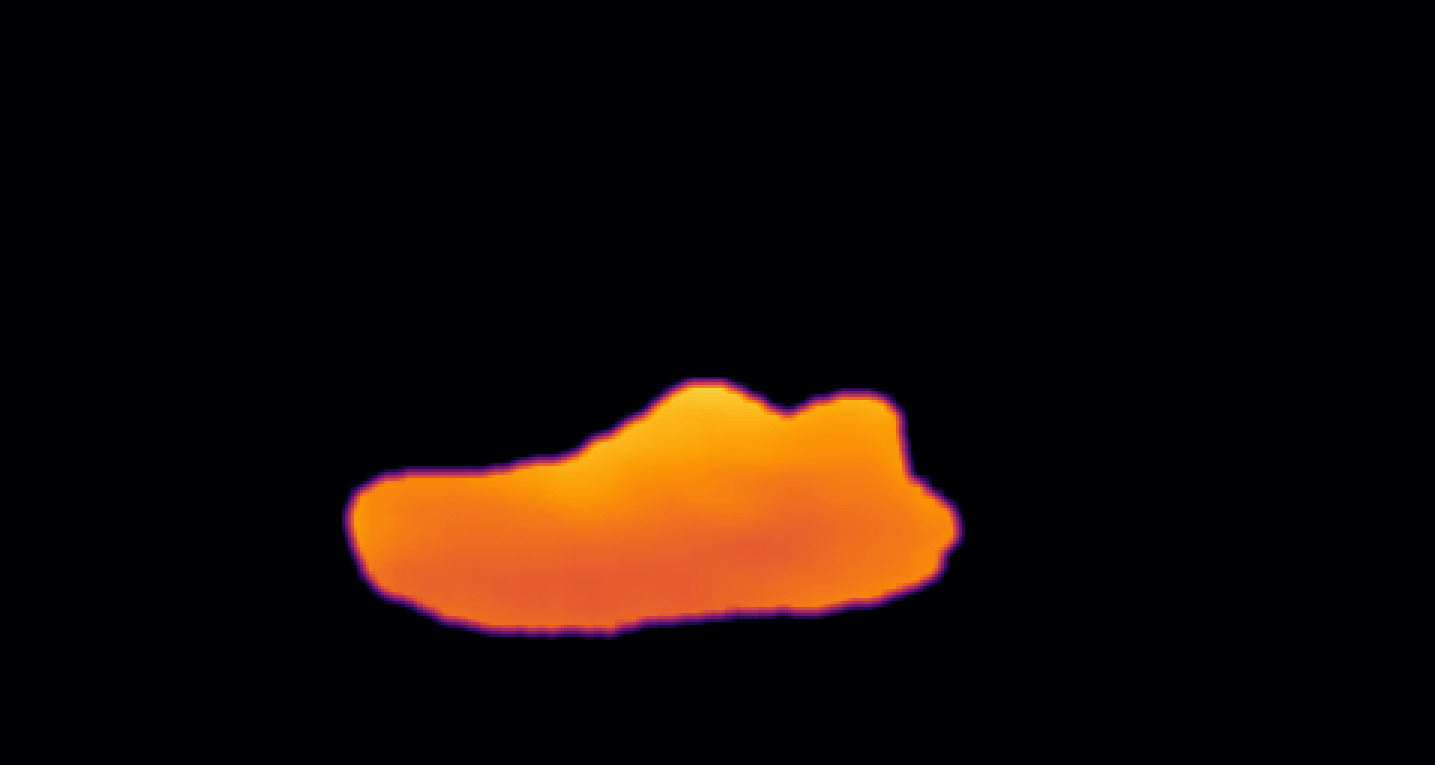}&\vspace{1.52mm}\includegraphics[width=20mm]{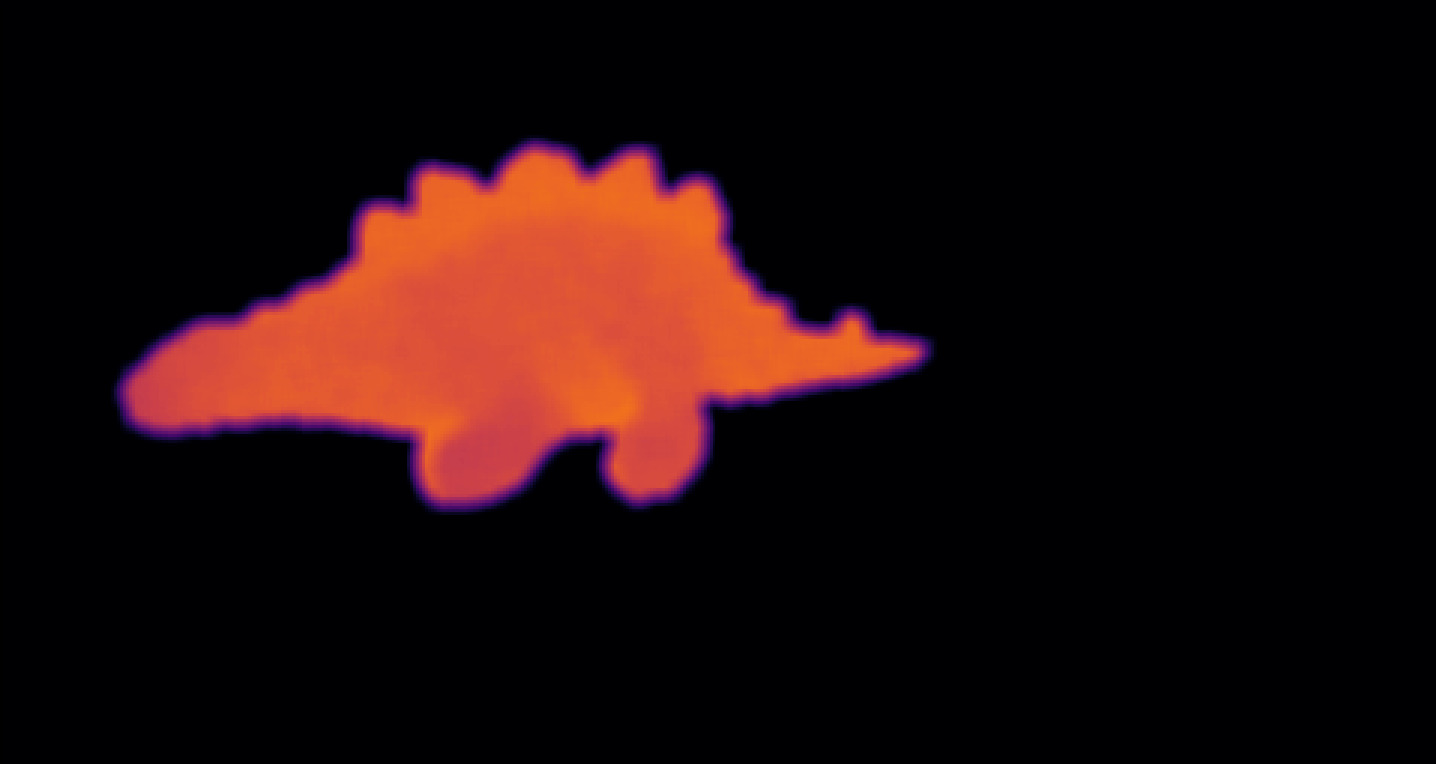}&\vspace{1.52mm}\includegraphics[width=20mm]{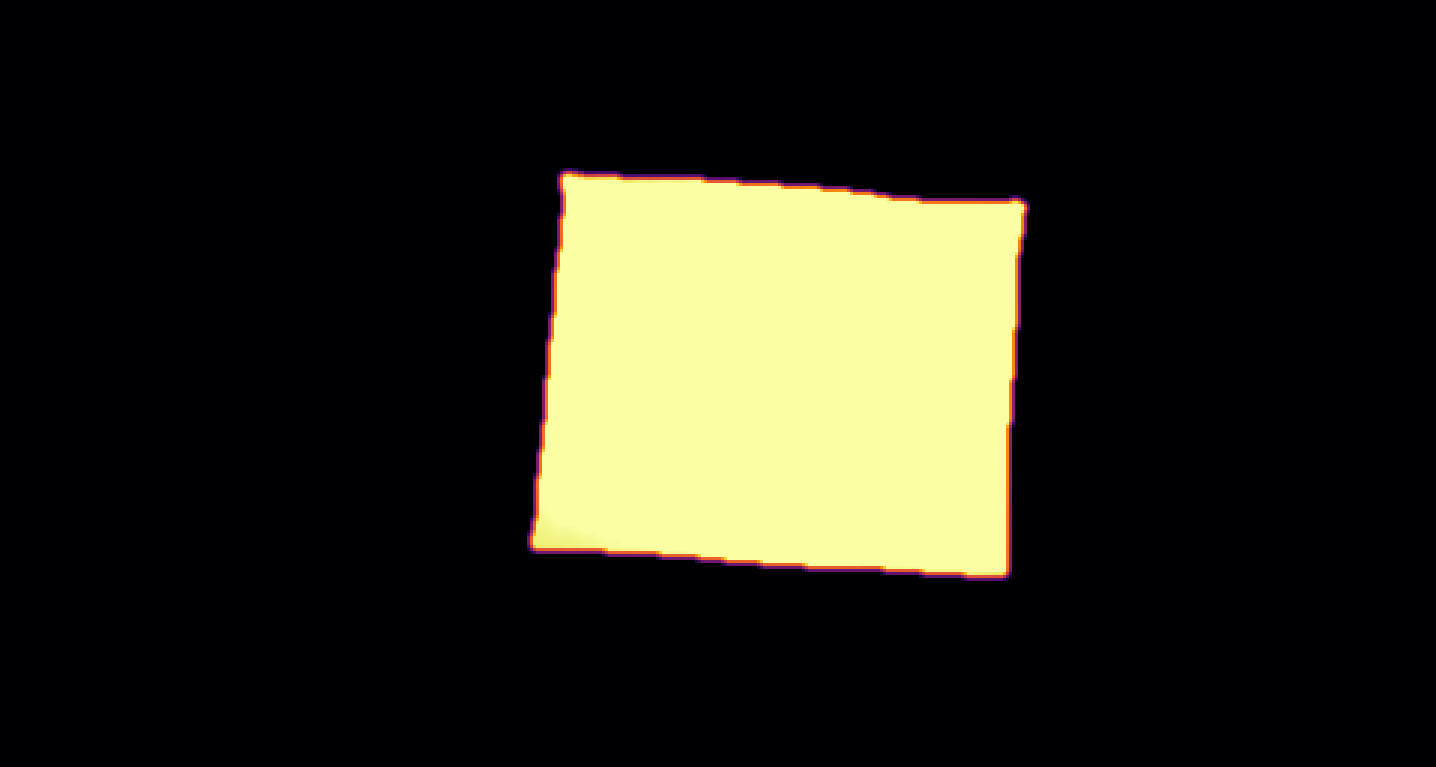}  \\ 
			
			\rule{0pt}{2ex} \centering &\multicolumn{1}{r}{19.84}&\multicolumn{1}{r}{19.35} &\multicolumn{1}{r}{19.22}&\multicolumn{1}{r}{14.61}\\ 
			
			\multicolumn{1}{l}{\rule{0pt}{2ex}}&\multicolumn{4}{c}{\textbf{0} \includegraphics[width=70mm,height=2mm]{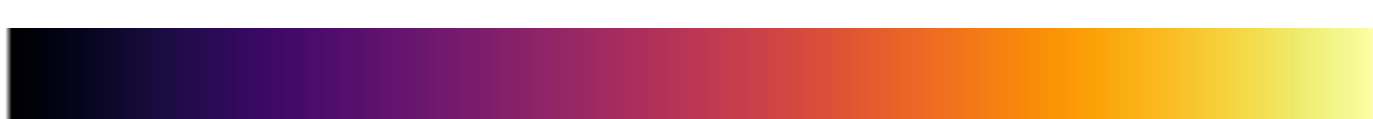} \textbf{500} mm}\\
			\rule{0pt}{2ex} \centering  DenseDepth & \vspace{1.52mm}\includegraphics[width=20mm]{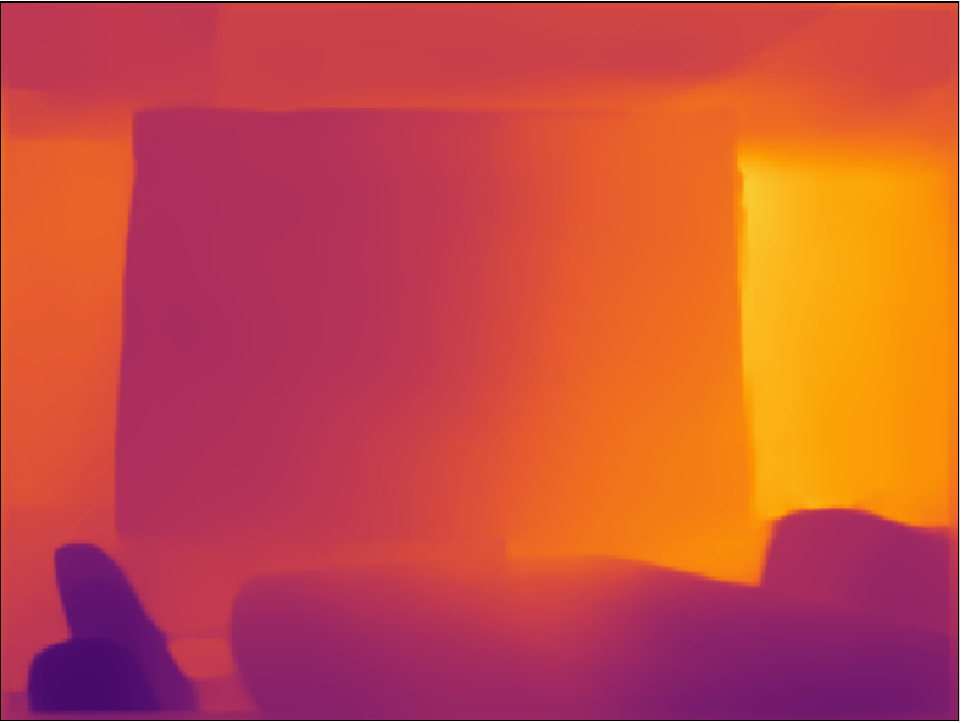}& \vspace{1.52mm}\includegraphics[width=20mm]{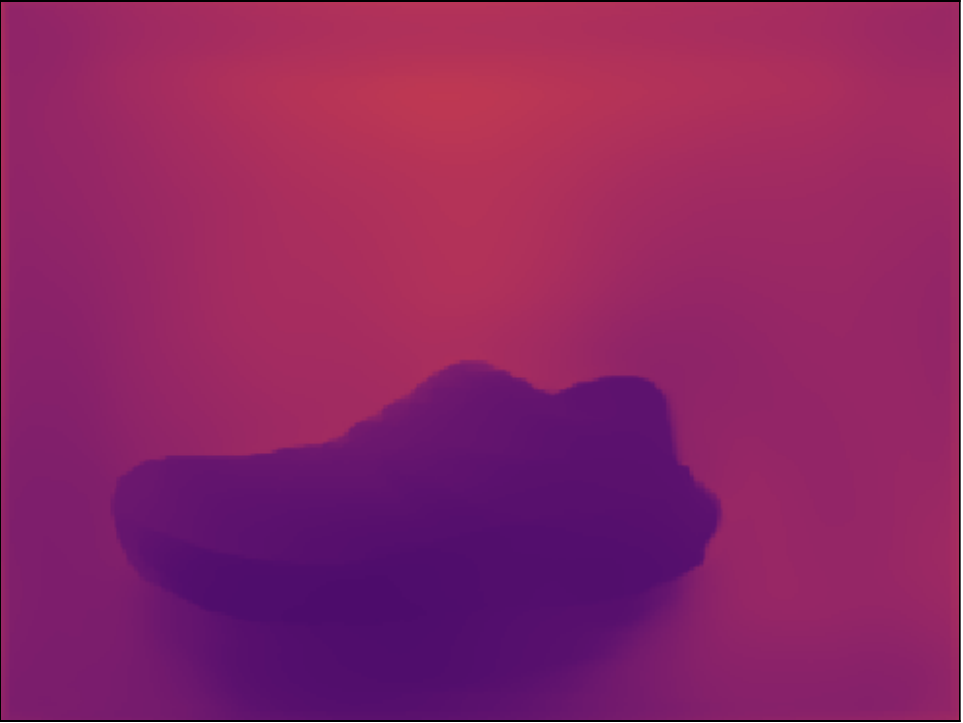}&  \vspace{1.52mm}\includegraphics[width=20mm]{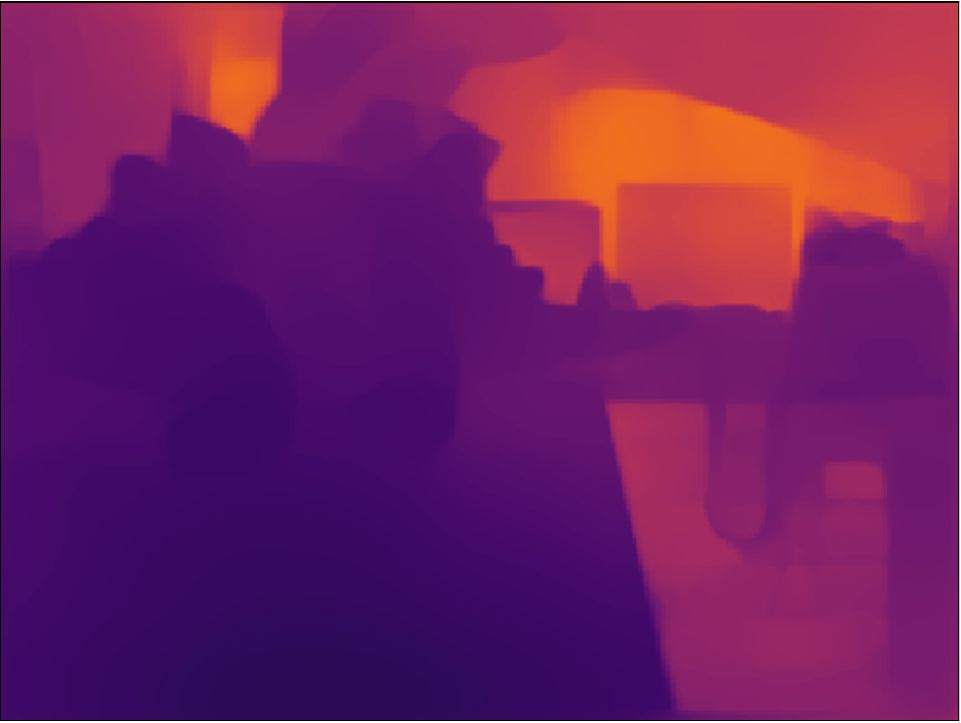}& \vspace{1.52mm}\includegraphics[width=20mm]{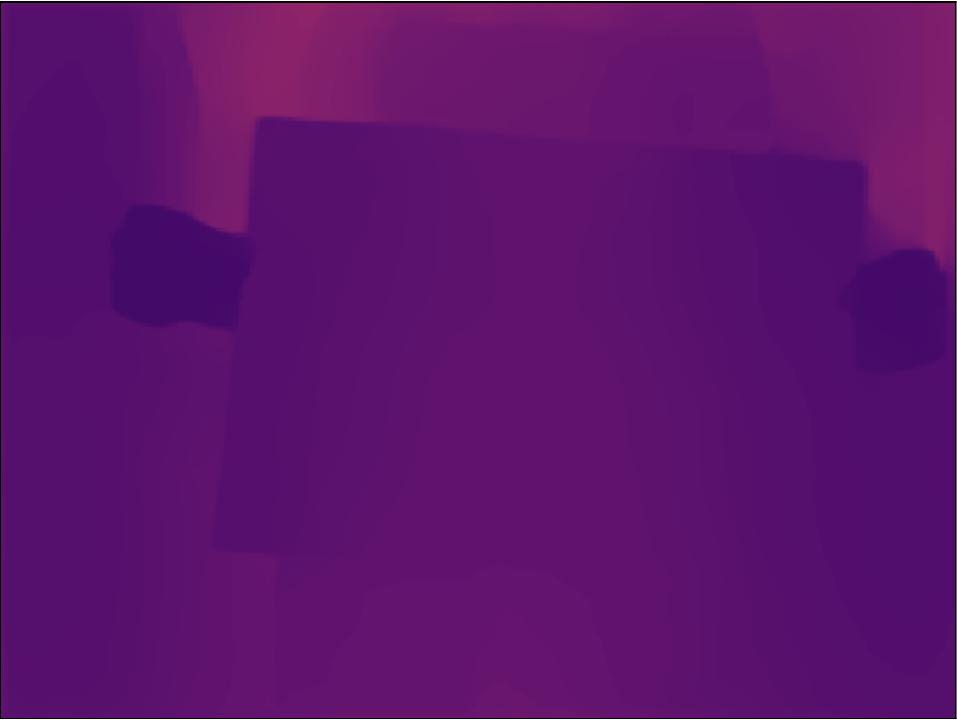}\\
			
			\rule{0pt}{2ex} \centering &\multicolumn{1}{r}{96.42}&\multicolumn{1}{r}{205.23} &\multicolumn{1}{r}{172.32}&\multicolumn{1}{r}{272.70}\\
			
			\rule{0pt}{2ex} \centering  BTS  &\vspace{1.52mm}\includegraphics[width=20mm]{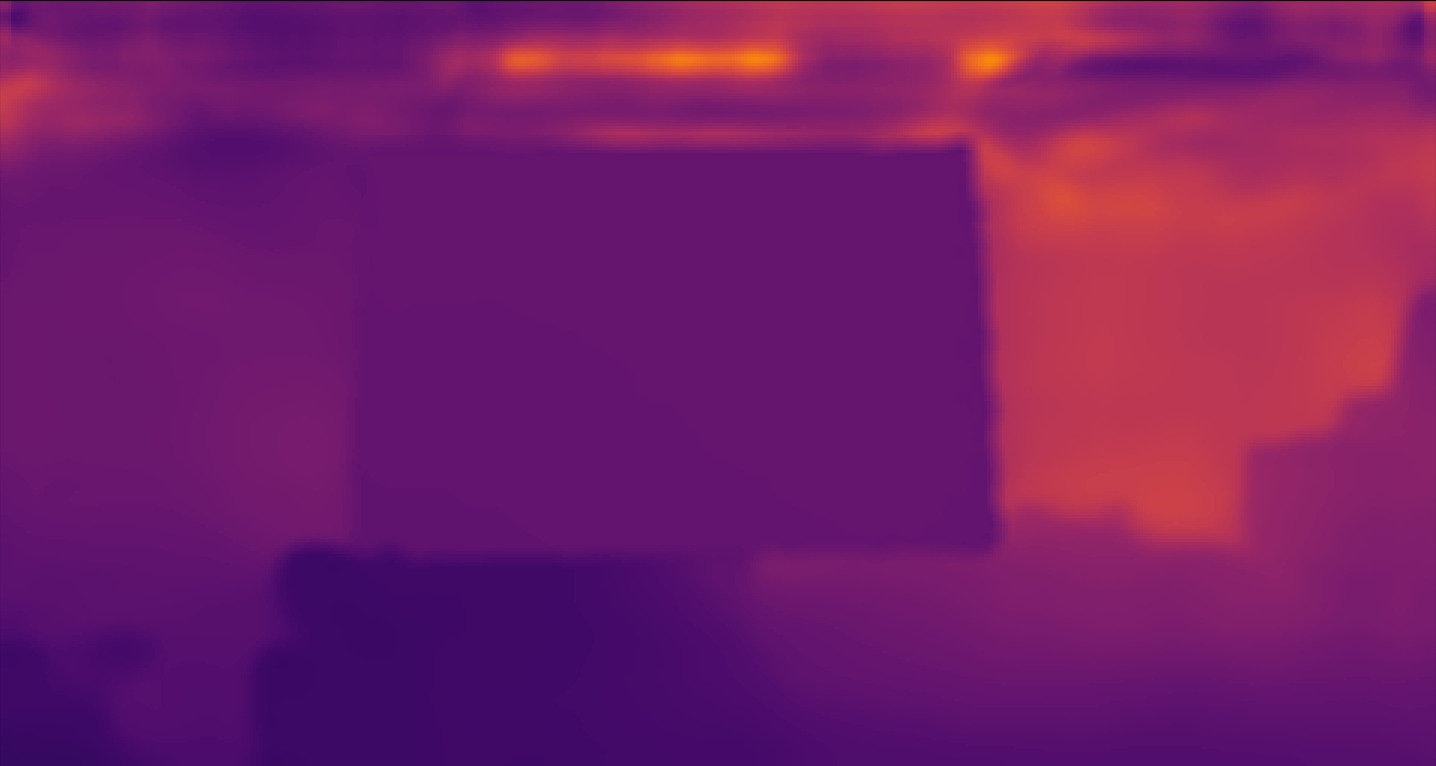}&\vspace{1.52mm}\includegraphics[width=20mm]{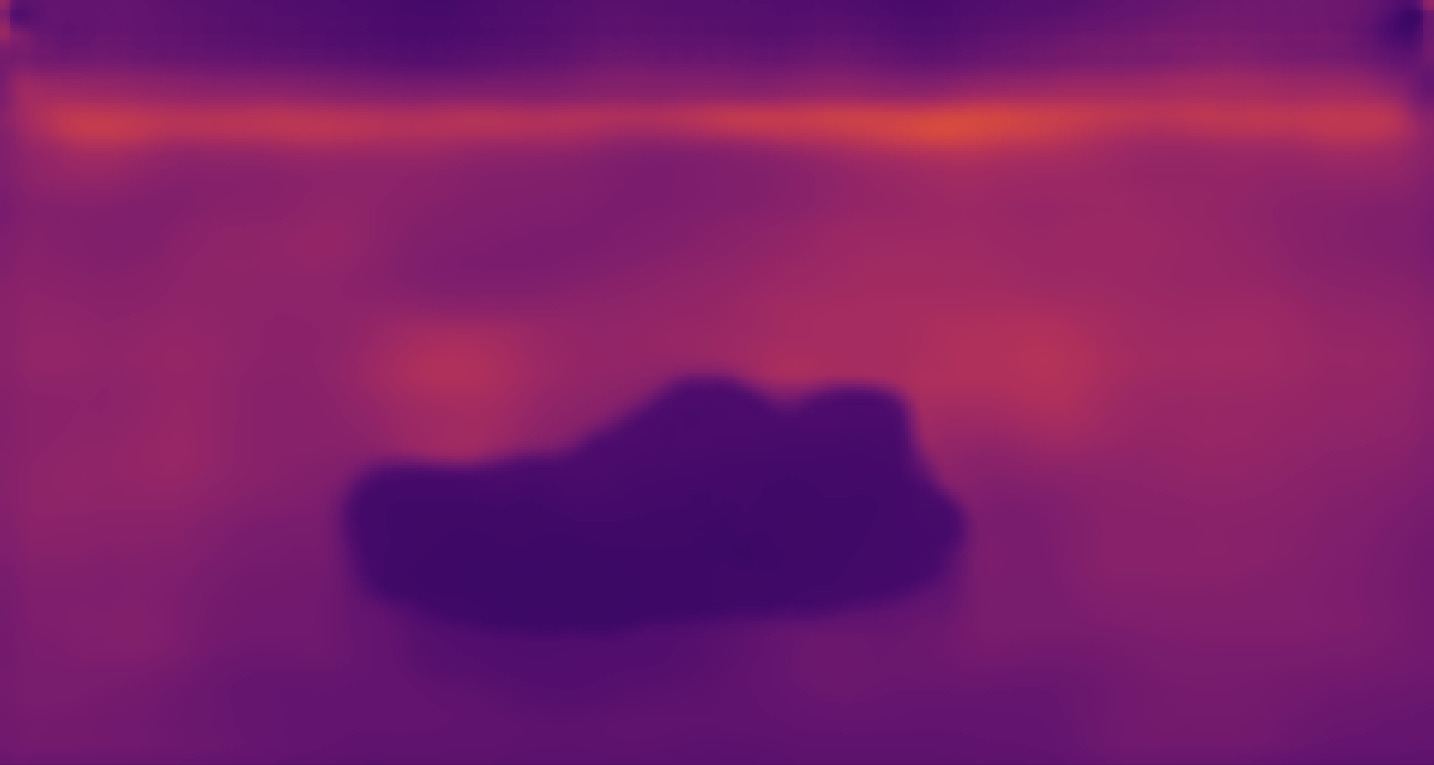}&\vspace{1.52mm}\includegraphics[width=20mm]{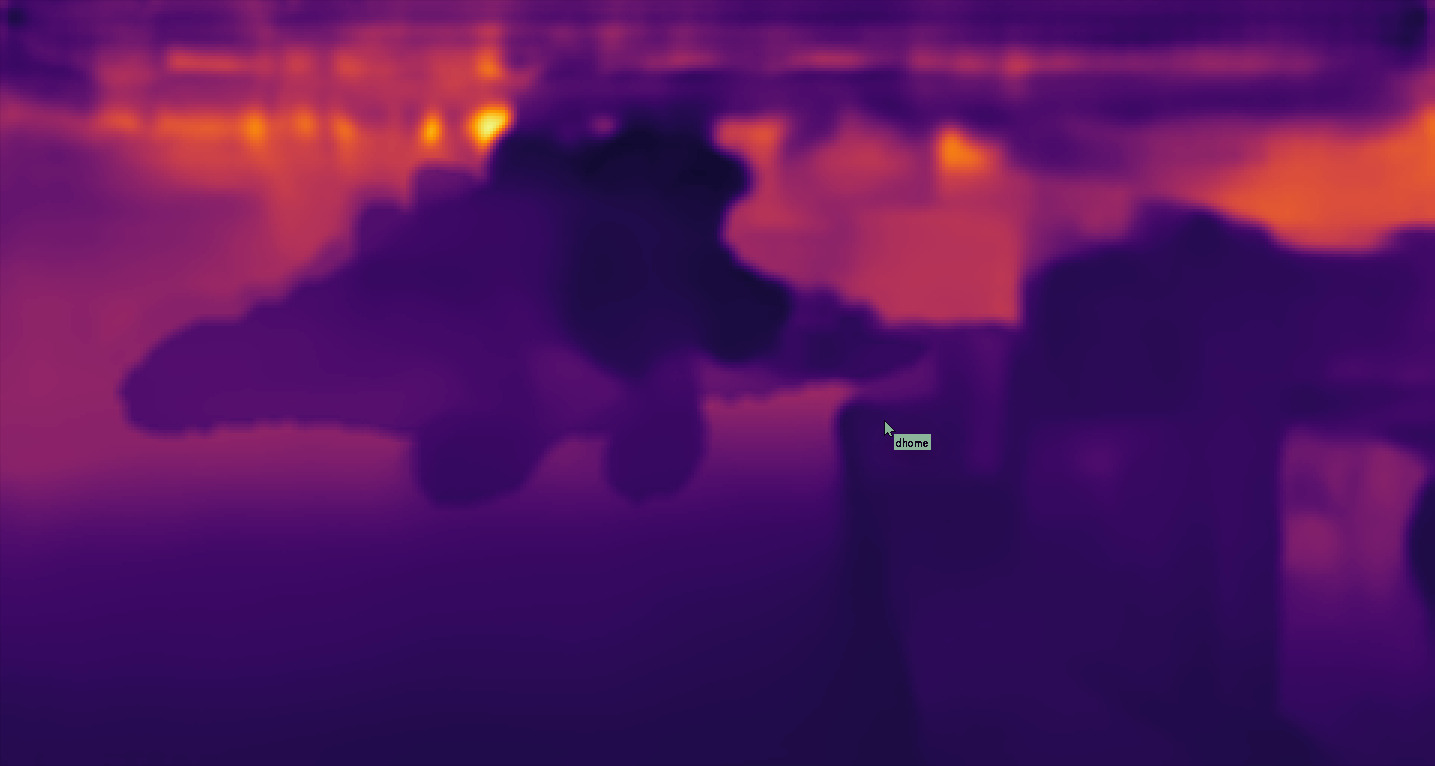} &\vspace{1.52mm}\includegraphics[width=20mm]{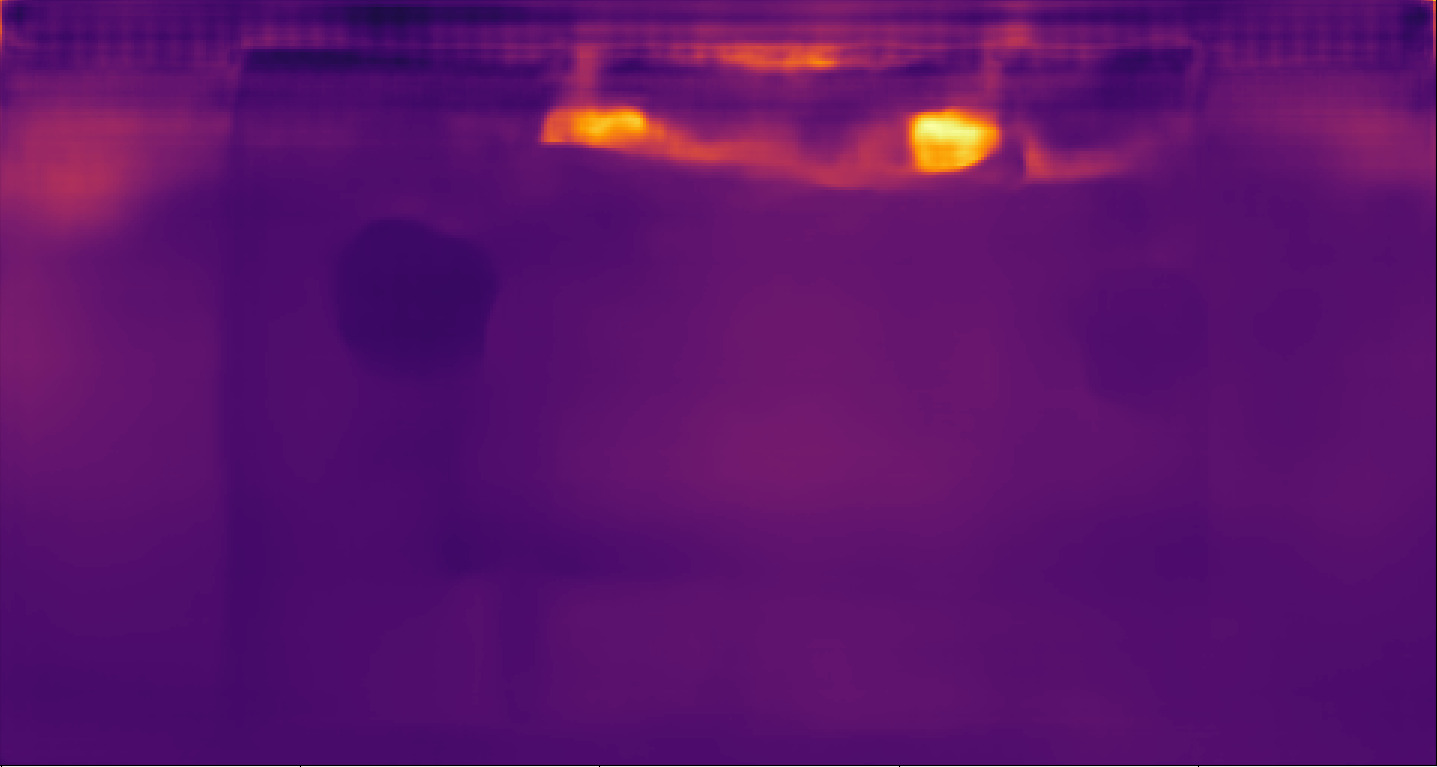}\\ 
			
			\rule{0pt}{2ex} \centering &\multicolumn{1}{r}{72.12}&\multicolumn{1}{r}{132.16} &\multicolumn{1}{r}{134.63}&\multicolumn{1}{r}{126.11}\\ 
			
			\multicolumn{1}{l}{\rule{0pt}{2ex}}&\multicolumn{4}{c}{\textbf{0} \includegraphics[width=70mm,height=2mm]{colormap_inferno2.jpg} \textbf{3000} mm}\\ 

			\rule{0pt}{2ex} \centering   Input Image &\vspace{1.52mm}\includegraphics[width=20mm]{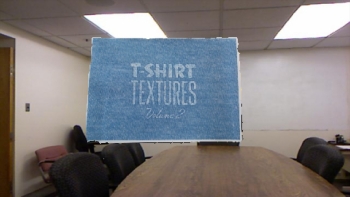}& \vspace{1.52mm}\includegraphics[width=20mm]{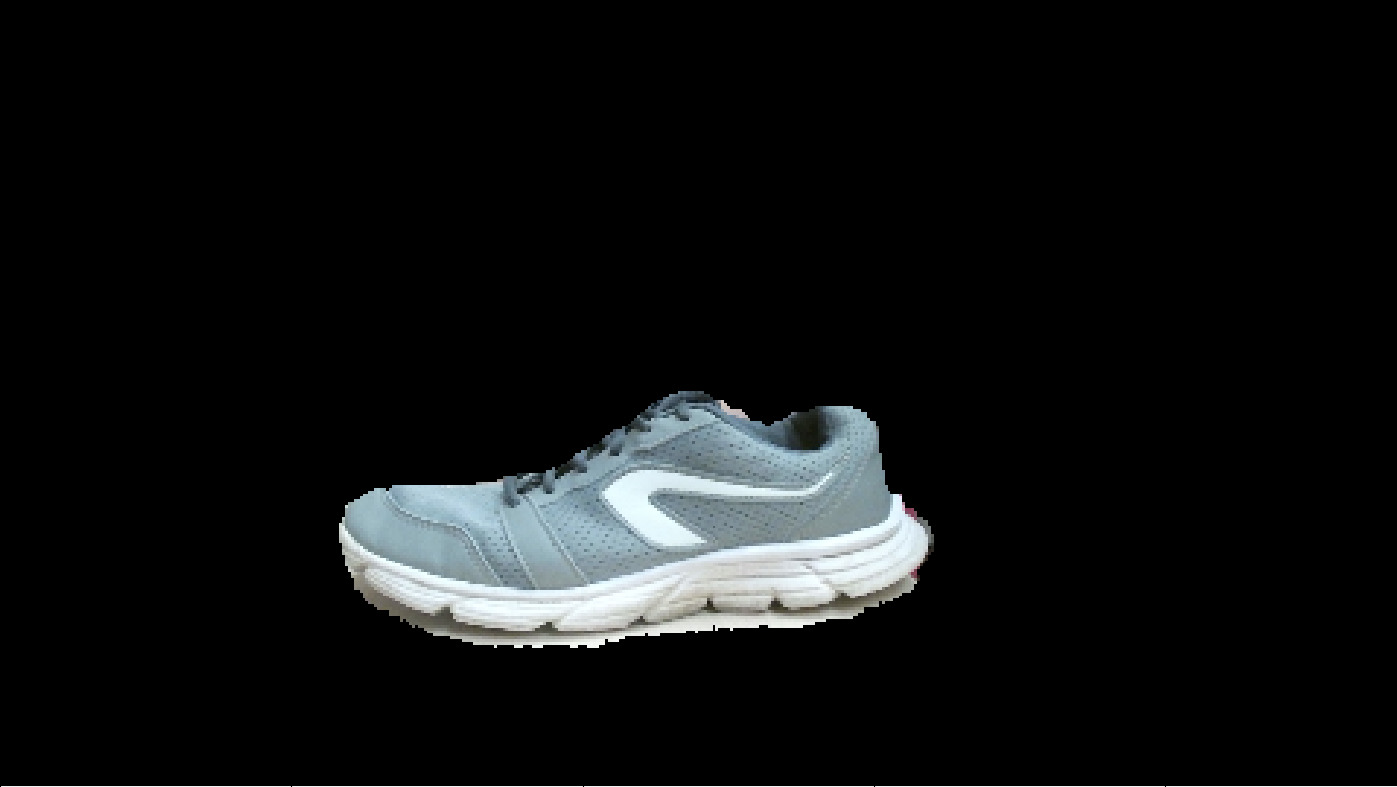}& \vspace{1.52mm}\includegraphics[width=20mm]{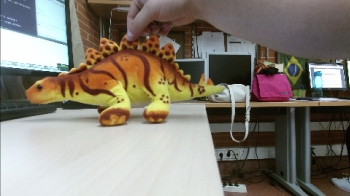}&\vspace{1.52mm}\includegraphics[width=20mm]{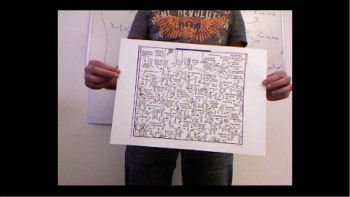}\\ 
			
			\rule{0pt}{2ex} \centering   DeepSfT &\vspace{1.52mm}\includegraphics[width=20mm]{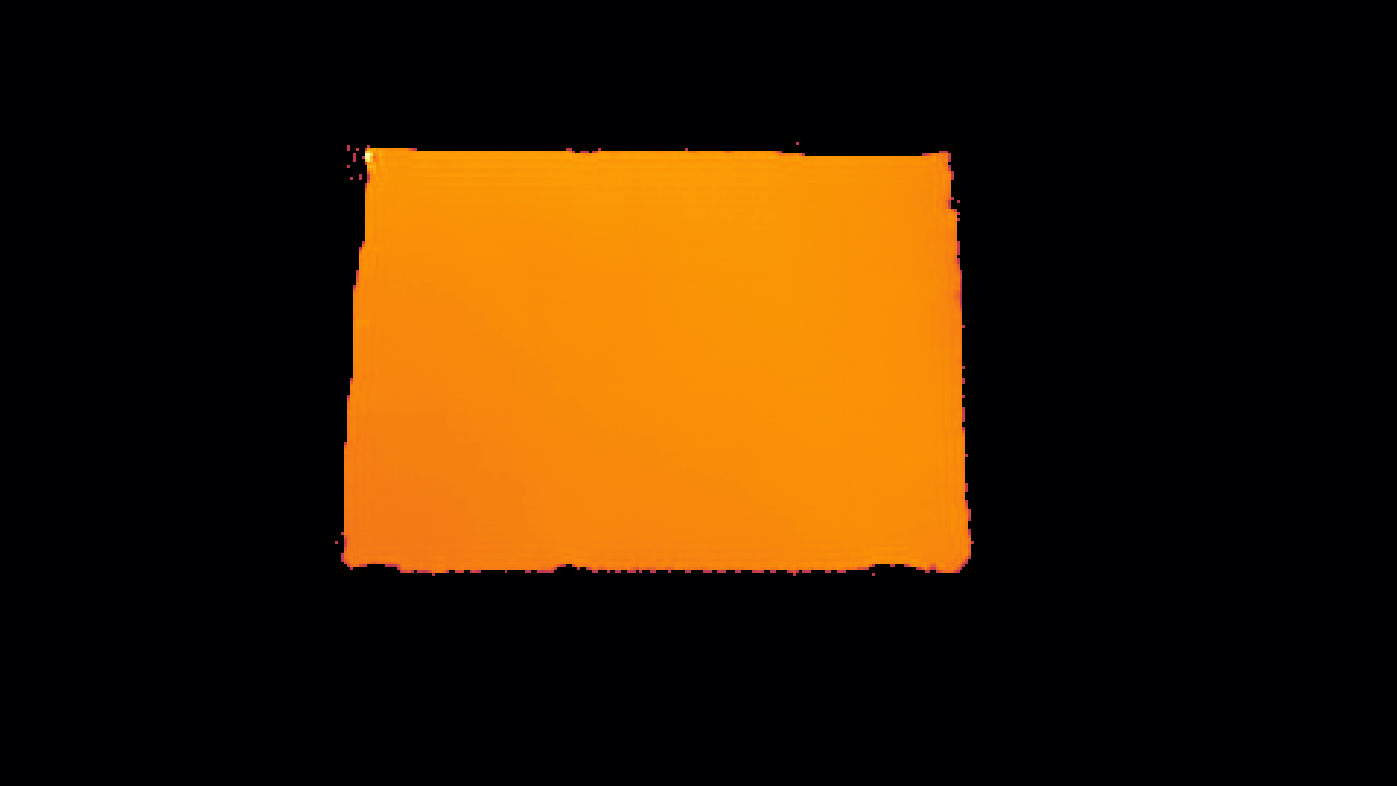}& \vspace{1.52mm}\includegraphics[width=20mm]{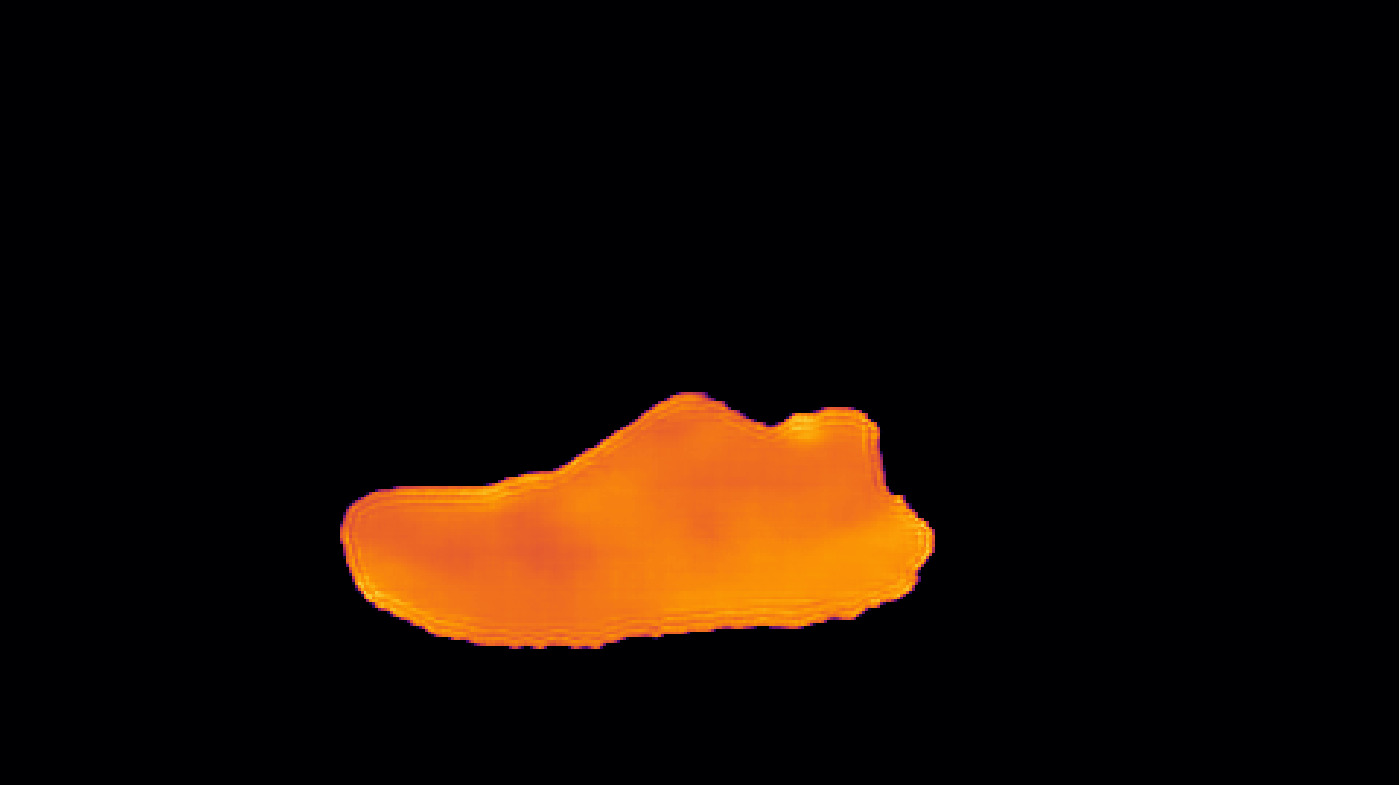}& \vspace{1.52mm}\includegraphics[width=20mm]{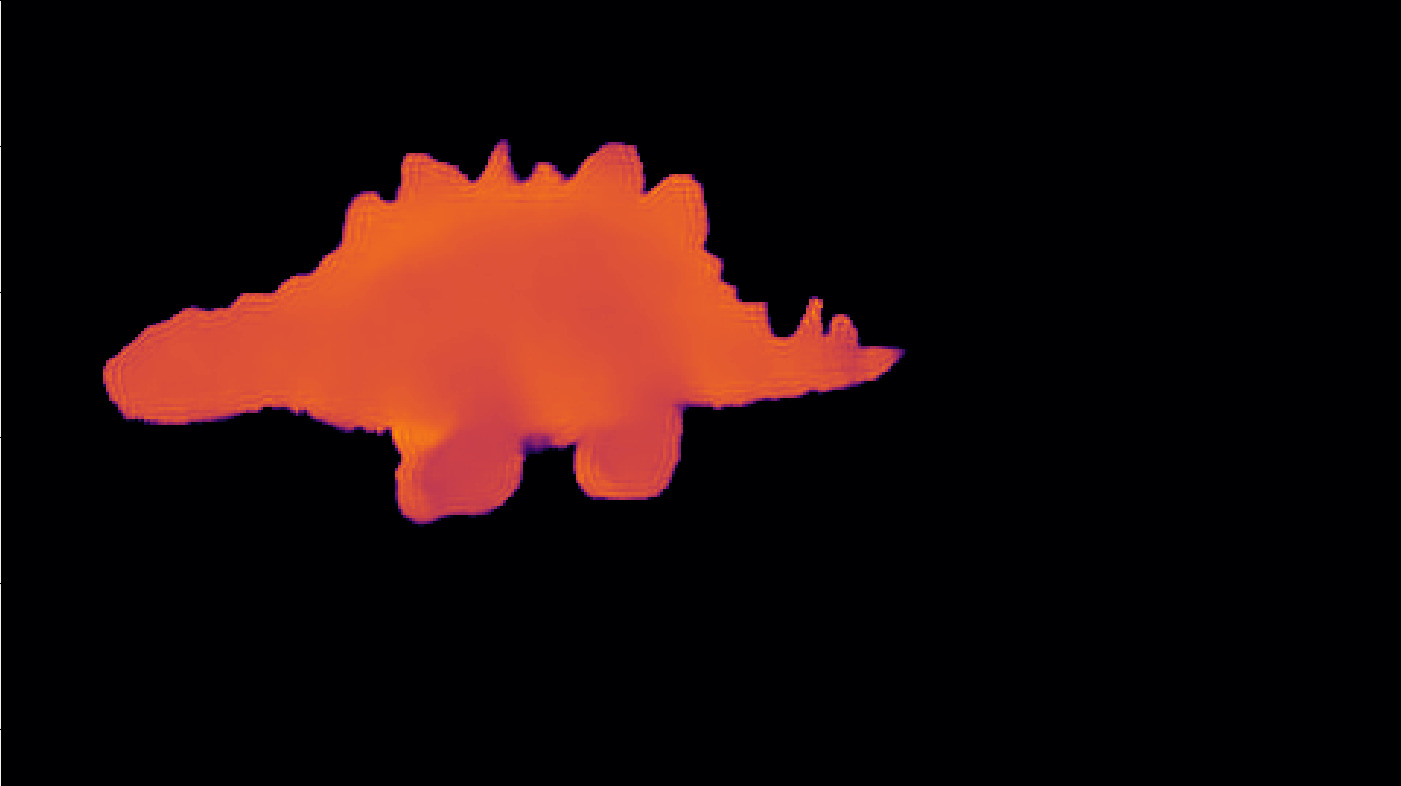}&\vspace{1.52mm}\includegraphics[width=20mm]{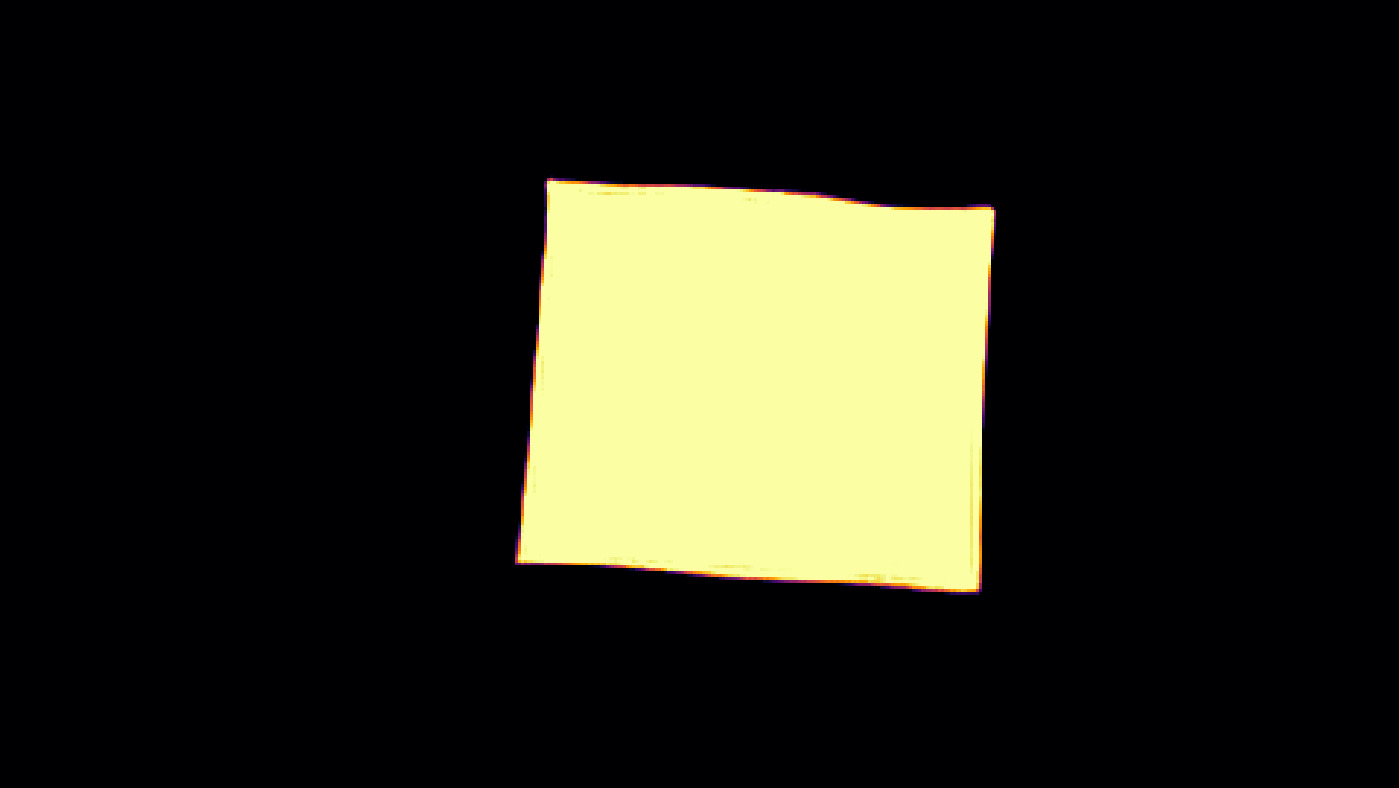}\\ 
			
			\rule{0pt}{2ex} \centering &\multicolumn{1}{r}{\textbf{3.26}}&\multicolumn{1}{r}{\textbf{3.98}} &\multicolumn{1}{r}{\textbf{4.81}}&\multicolumn{1}{r}{\textbf{4.29}}\\ 
			
			\multicolumn{1}{l}{\rule{0pt}{2ex}}&\multicolumn{4}{c}{\textbf{0} \includegraphics[width=70mm,height=2mm]{colormap_inferno2.jpg} \textbf{500} mm}\\ 
			
		\end{tabular} 
	\end{adjustbox}
	\centering \caption {Representative results and comparison of DeepSfT with other monocular depth reconstruction methods. The estimated depth maps and corresponding RMSE error in mm are shown for each method with one example input image from 4 templates (arranged in 4 columns).}
	\label{tb:qualitative_monocular}
\end{table}
\subsection{Evaluation with rectangular templates}
We show in tables \ref{tb:exp_sintetic}, \ref{tb:qualitative} and \ref{tb:examples} quantitative and qualitative results obtained with rectangular templates and synthetic test datasets, denoted by DS1S and DS2S, and real test datasets, denoted by DS1R, DS2R and DS5R. In terms of reconstruction error, DeepSfT is considerably better than the other methods, both in synthetic test data, where the RMSE remains below 2 mm, and for real test data, where the RMSE is below 10 mm. Kinect V2 has an uncertainty of about 10 mm at a distance of one meter, which partially explains the higher error for real data. The second and third best methods are IsMo-GAN and R50F respectively, also DNN-based. However, their errors are far worse compared to DeepSfT. CH17 obtains reasonable results when it is provided with ground-truth registration (CH17-GTR and CH17R-GTR). However, the performance is considerably worse when real registration is provided by DOF (CH17-DOF and CH17R-DOF). NGO15 obtains the worst result on DS1 and the second worst result on DS2. This was expected because we evaluate this algorithm in a wide-baseline setting and, as mentioned by the authors, this method was designed to work only for small deformations (small-baseline).

In terms of registration error, DeepSfT also has the best results both for synthetic test data, where ground-truth registration is available, and real test data, where DOF is used as the ground-truth proxy. In all cases DeepSfT has a mean registration RMSE of approximately 2 px. The performance of R50F is competitive with DOF, with registration RMSE of approximately 5 px.

\subsection{Evaluation with volumic templates}

The quantitative and qualitative results of the experiments for the volumic templates DS3 and DS4 are provided in tables \ref{tb:exp_sintetic2}, \ref{tb:qualitative} and \ref{tb:examples}, with both synthetic test data, denoted by DS3S and DS4S, and real test data, denoted by DS4R and DS4R. Recall that the test datasets consist of unorganised images, unlike DS1, DS2 and DS5, and it is thus impossible to estimate registration reliably with DOF. Therefore we only compute registration error with synthetic data (DS3S and DS4S). CH17+GTR and CH17R+GTR are tested only on DS3S and DS4S, because these are the only datasets they can handle.
\begin{table}[!htbp]
	\setcellgapes{3pt}\makegapedcells
	\begin{adjustbox}{max width=\linewidth}
		\begin{tabular}{|c|c|c|c|c|c|c|c|}
			\hline
			\multicolumn{2}{|c}{}&\multicolumn{2}{|c|}{Registration RMSE (px)}&\multicolumn{4}{c|}{Reconstruction RMSE (mm)} \\ \hline
			Sequence  & Samples &  R50F & DeepSFT & CH17+GTR & CH17R+GTR & R50F & DeepSfT   \\ \hline
			
			\large{DS3S} & \large{5000}  & \large{7.14}  & \large{\textbf{1.05}}& \large{45.21} & \large{43.67}  & \large{6.34} & \large{\textbf{1.16}}  \\ \hline
			
			\large{DS4S} & \large{5000}  & \large{8.93}  & \large{\textbf{3.60}} & \large{73.80} & \large{70.70} & \large{12.62} & \large{\textbf{1.57}}  \\ \hline
			
			\large{DS3R} &\large{1300}  & \large{-}  & \large{-} & \large{-} & \large{-} & \large{12.43} & \large{\textbf{8.12}}  \\ \hline
			
			\large{DS4R} &\large{550}  & \large{-}  & \large{-} & \large{-} & \large{-} & \large{27.31} & \large{\textbf{6.86}}  \\ \hline
		\end{tabular} 
	\end{adjustbox}
	\centering \caption {Quantitative evaluation on synthetic and real data with volumic templates (DS3S, DS4S, DS3R and DS4R).}
	\label{tb:exp_sintetic2}
\end{table}

\begin{table}[!htbp]
	\setcellgapes{3pt}\makegapedcells
	\begin{adjustbox}{max width=\linewidth}
		\begin{tabular}{|c|c|c|c|c|c|}
			\hline
			\multicolumn{2}{|c}{}&\multicolumn{2}{|c|}{Reconstruction RMSE (mm)}&\multicolumn{2}{c|}{Registration (photometric error $\mathcal{E}_{pr}$)} \\ \hline
			Sequence  & Samples & DeepSfT & DeepSfT+TV&DeepSfT&DeepSfT+PR   \\ \hline
			
			\large{DS1R} & \large{5000} & \large{9.51} &  \large{\textbf{4.12}}&\large{0.266}&\large{\textbf{0.211}} \\ \hline
			
			\large{DS2R} & \large{5000} & \large{7.37} & \large{\textbf{3.39}}&\large{0.094}&\large{\textbf{0.015}}  \\ \hline
			
			\large{DS3R} &\large{1300} & \large{8.12} & \large{\textbf{7.40}}&\large{0.141}&\large{\textbf{0.109}}  \\ \hline
			
			\large{DS4R} &\large{550} & \large{6.86} & \large{\textbf{5.80}}&\large{0.196}&\large{\textbf{0.184}}  \\ \hline
			
			\large{DS5R} &\large{50} & \large{6.97} & \large{\textbf{6.89}}&\large{0.388}&\large{\textbf{0.203}}  \\ \hline
		\end{tabular} 
	\end{adjustbox}
	\centering \caption {Quantitative evaluation on real test data and volumic templates (DS1R, DS2R, DS3R, DS4R and DS5R).}
	\label{tb:total_variation}
\end{table}

\begin{table}[!htbp]
	\begin{adjustbox}{max width=0.9\linewidth}
		\begin{tabular}{lm{2cm}m{2cm}}
			
			&\multicolumn{1}{c}{\centering{Ground-truth 3D surface}}&\multicolumn{1}{c}{\centering DS1 Input Image} \\ 
			
			&\includegraphics[width=30mm]{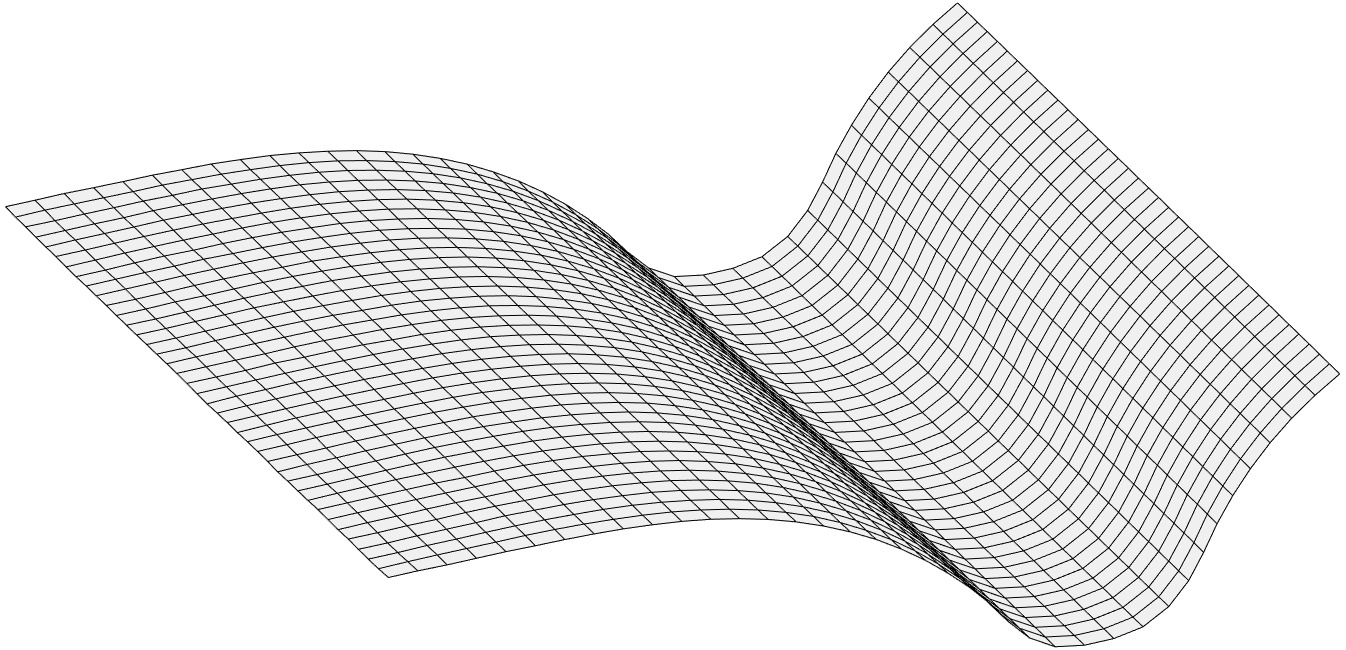}&\includegraphics[width=30mm]{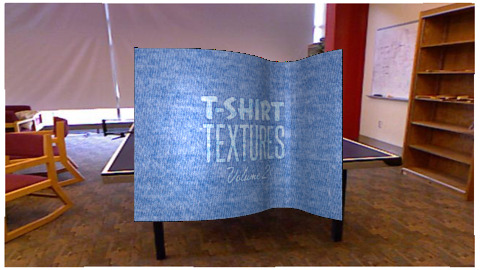}\\ 

			\multicolumn{1}{c}{Method}& \multicolumn{1}{c}{3D Reconstruction \& RMSE colormap}&\multicolumn{1}{c}{Registration ROI \& RMSE colormap} \\\hline

			\multicolumn{1}{c}{\rule{0pt}{2ex} \centering   CH17+DOF}&\includegraphics[width=30mm]{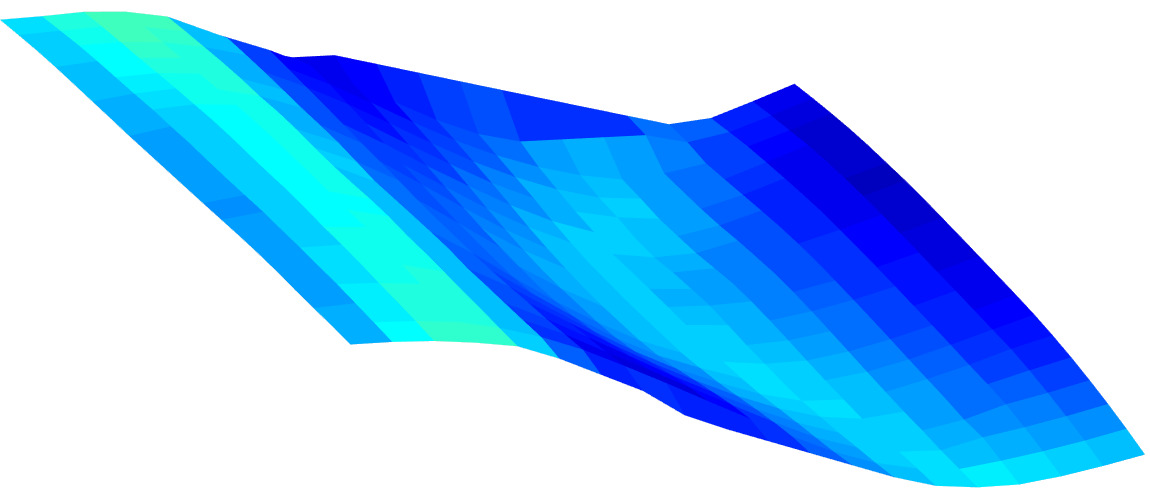}&\includegraphics[width=30mm]{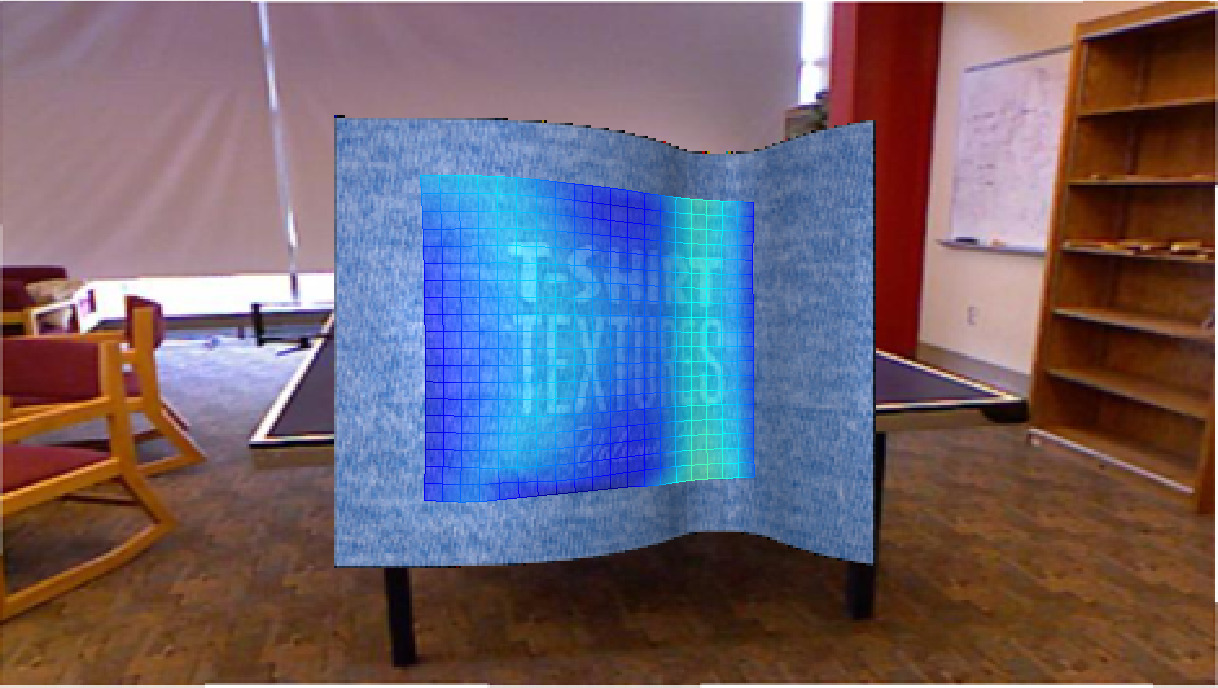}\\ 

			\multicolumn{1}{c}{\rule{0pt}{2ex} \centering   CH17R+DOF}&\includegraphics[width=30mm]{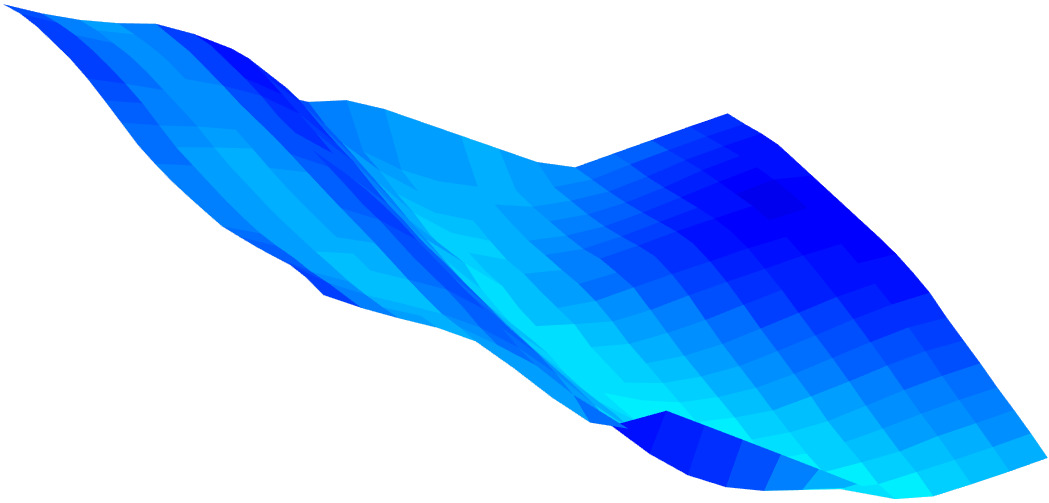}&\includegraphics[width=30mm]{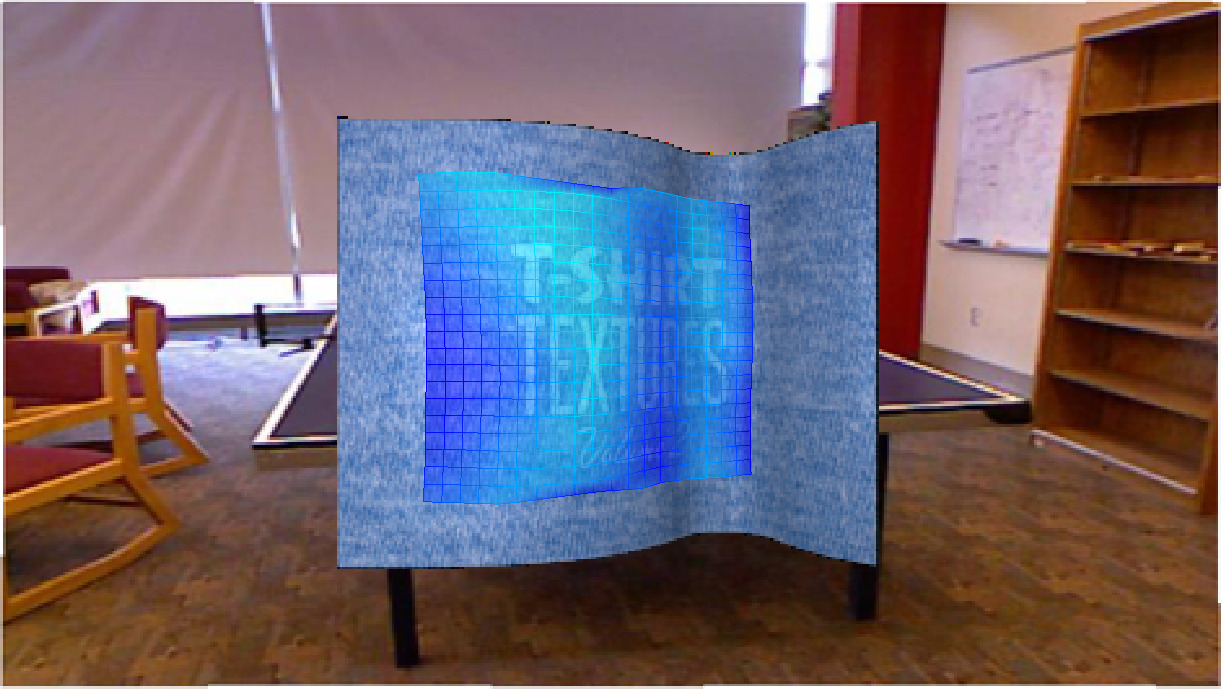}\\

			\multicolumn{1}{c}{\rule{0pt}{2ex} \centering   NGO15}&\includegraphics[width=30mm]{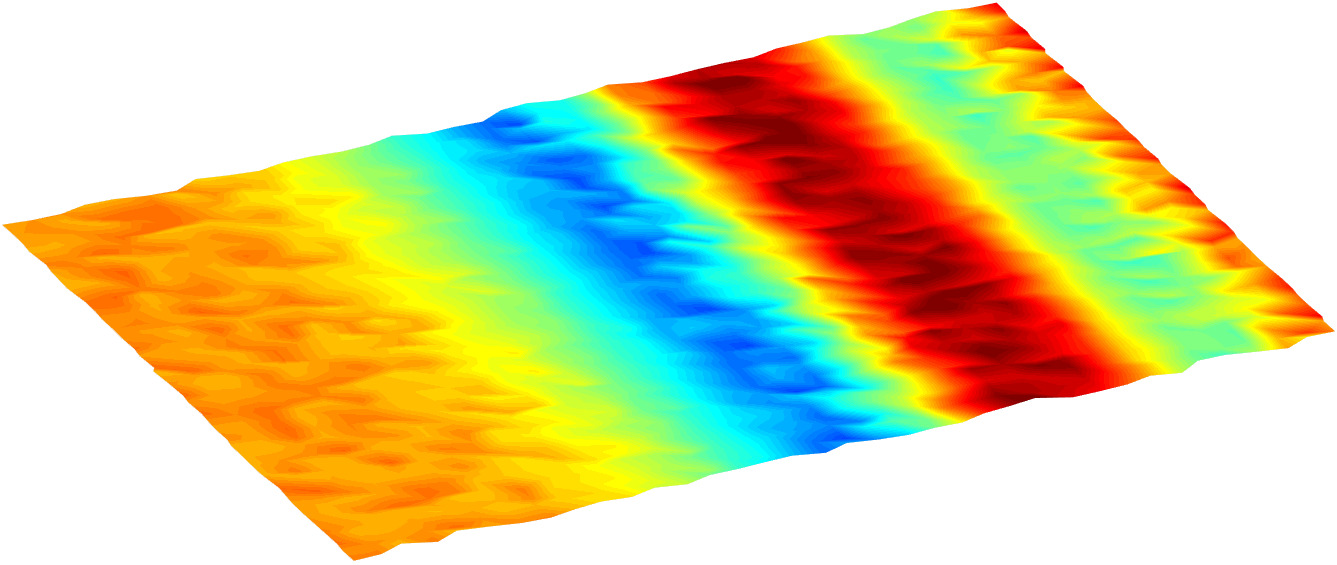}&\includegraphics[width=30mm]{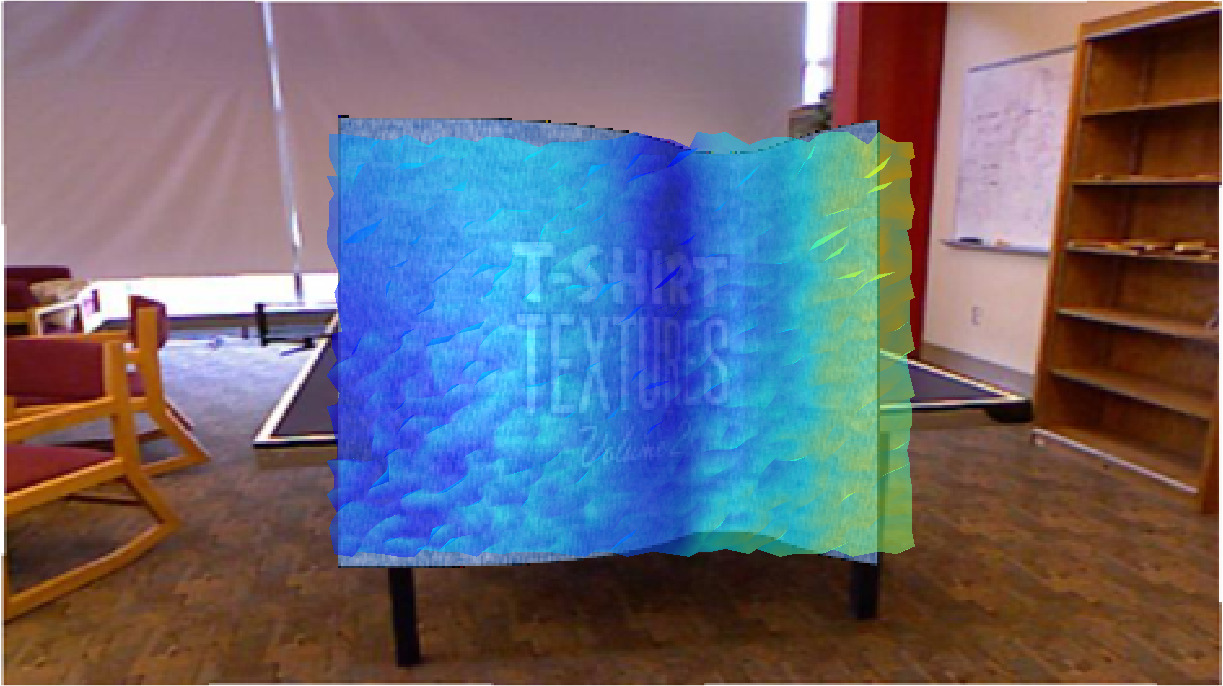}\\

			\multicolumn{1}{c}{\rule{0pt}{2ex} \centering   HDM-net}&\includegraphics[width=30mm]{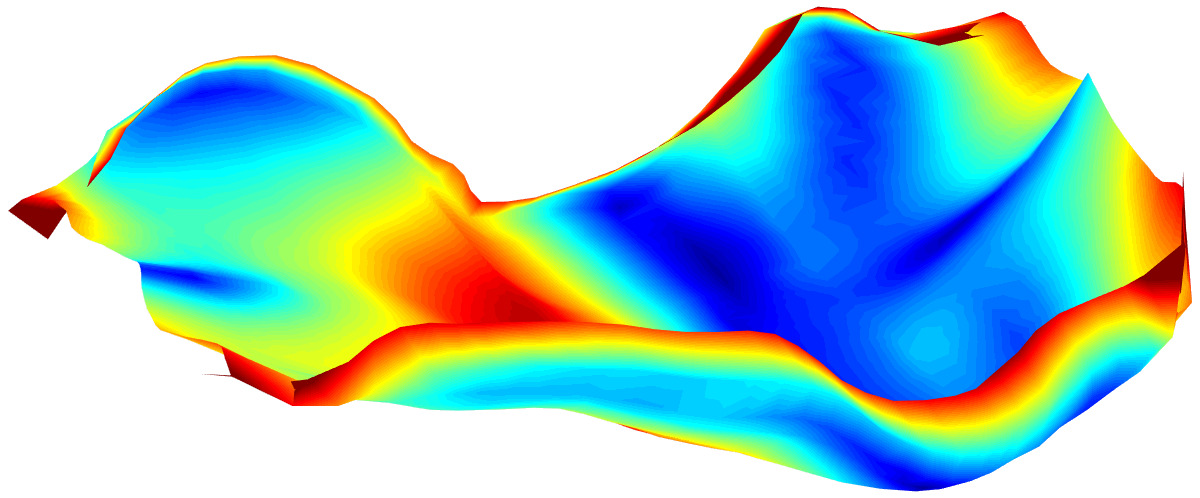}&\includegraphics[width=30mm]{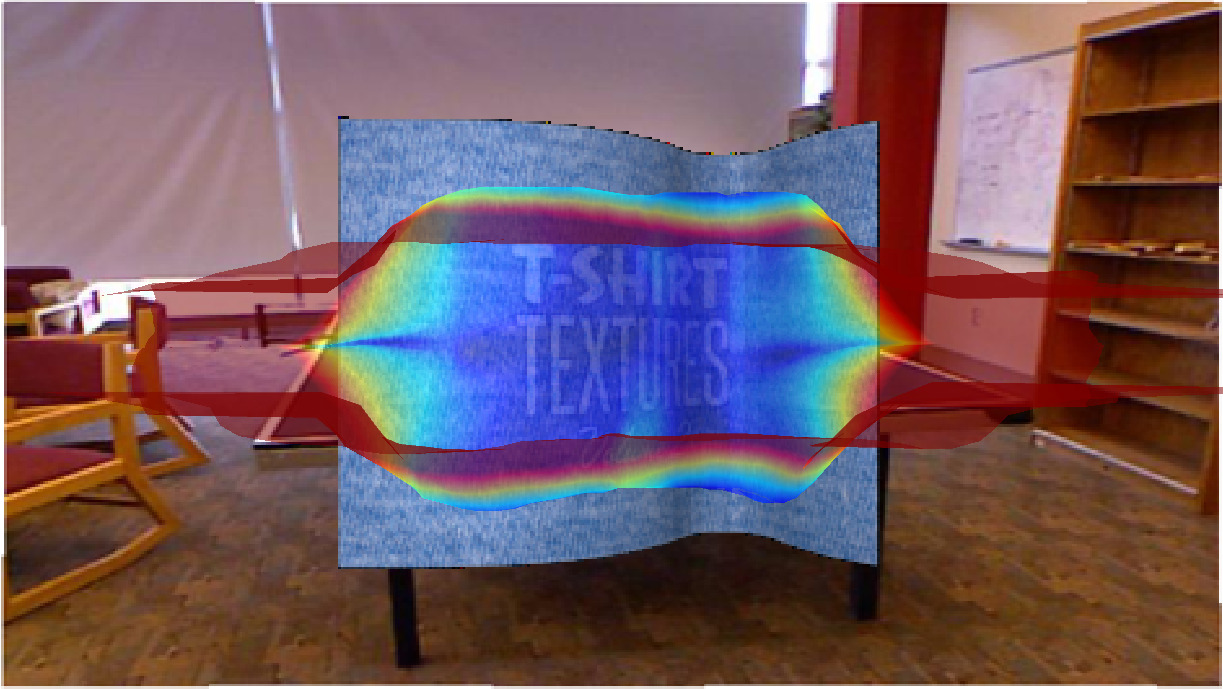}\\ 
			
			\multicolumn{1}{c}{\rule{0pt}{2ex} \centering   IsMo-GAN}&\includegraphics[width=30mm]{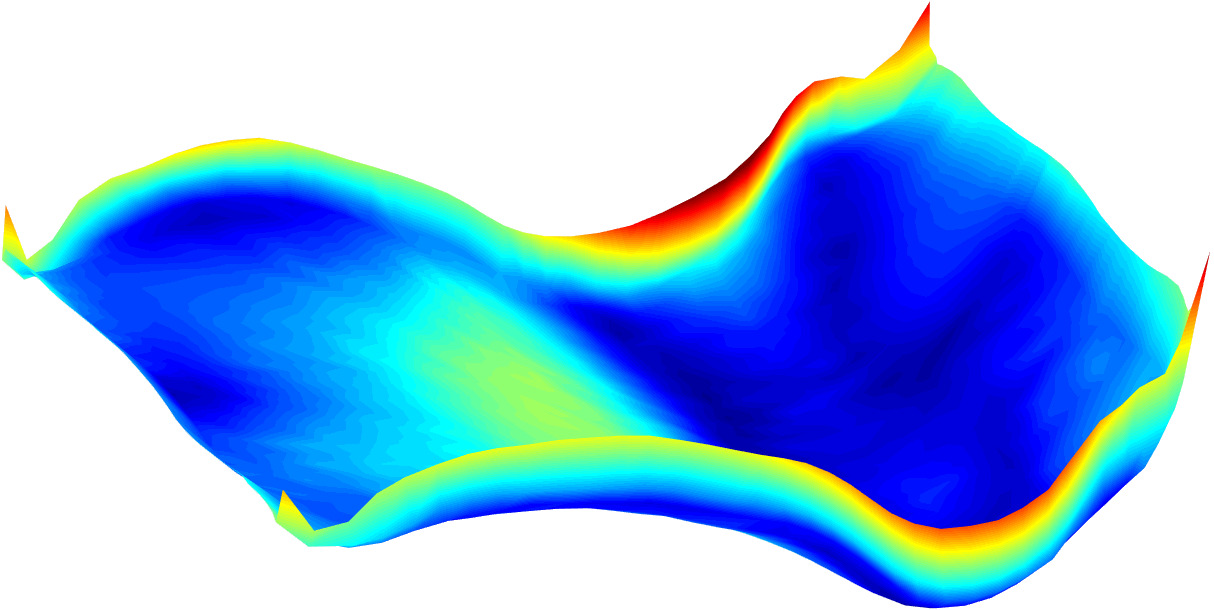}&\includegraphics[width=30mm]{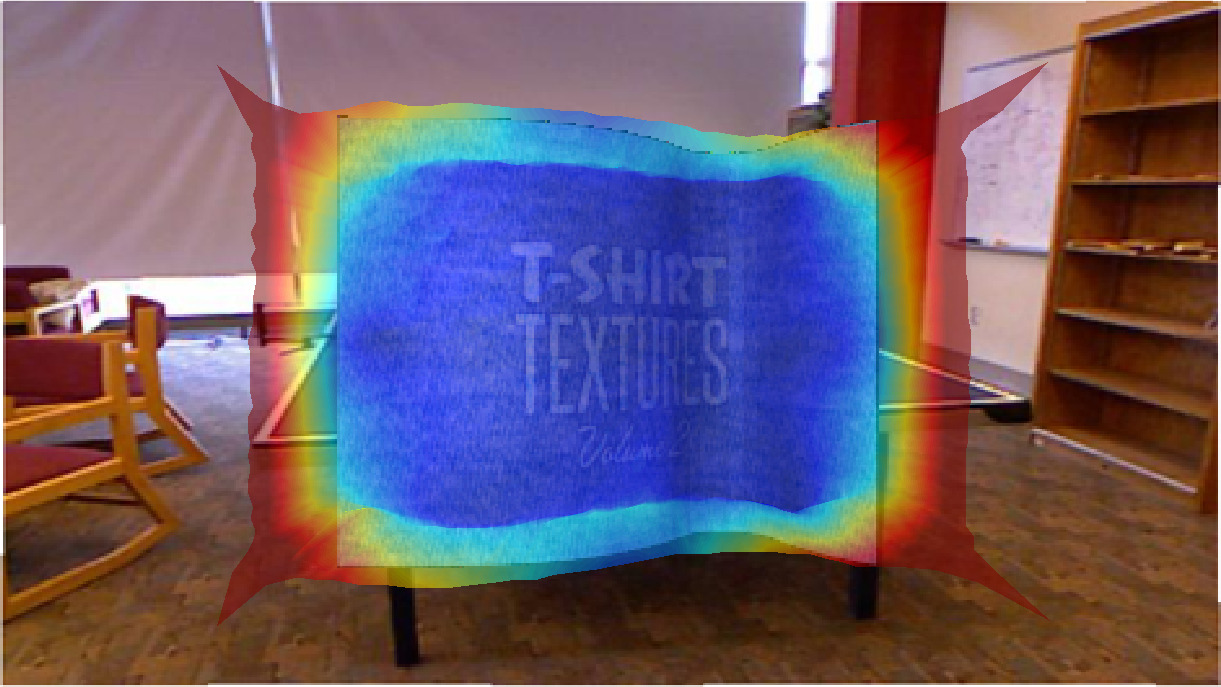}\\ 

			\multicolumn{1}{c}{\rule{0pt}{2ex} \centering   R50F}&\includegraphics[width=30mm]{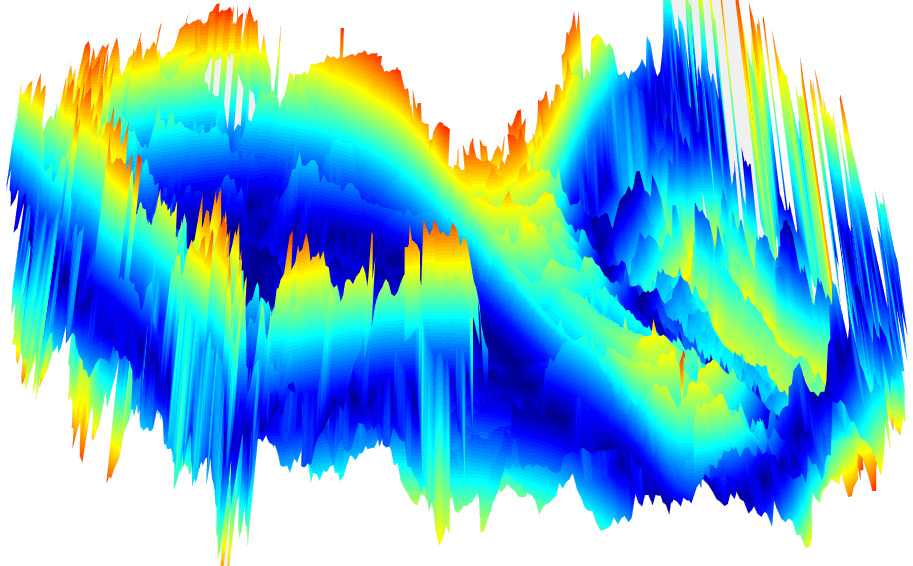}&\includegraphics[width=30mm]{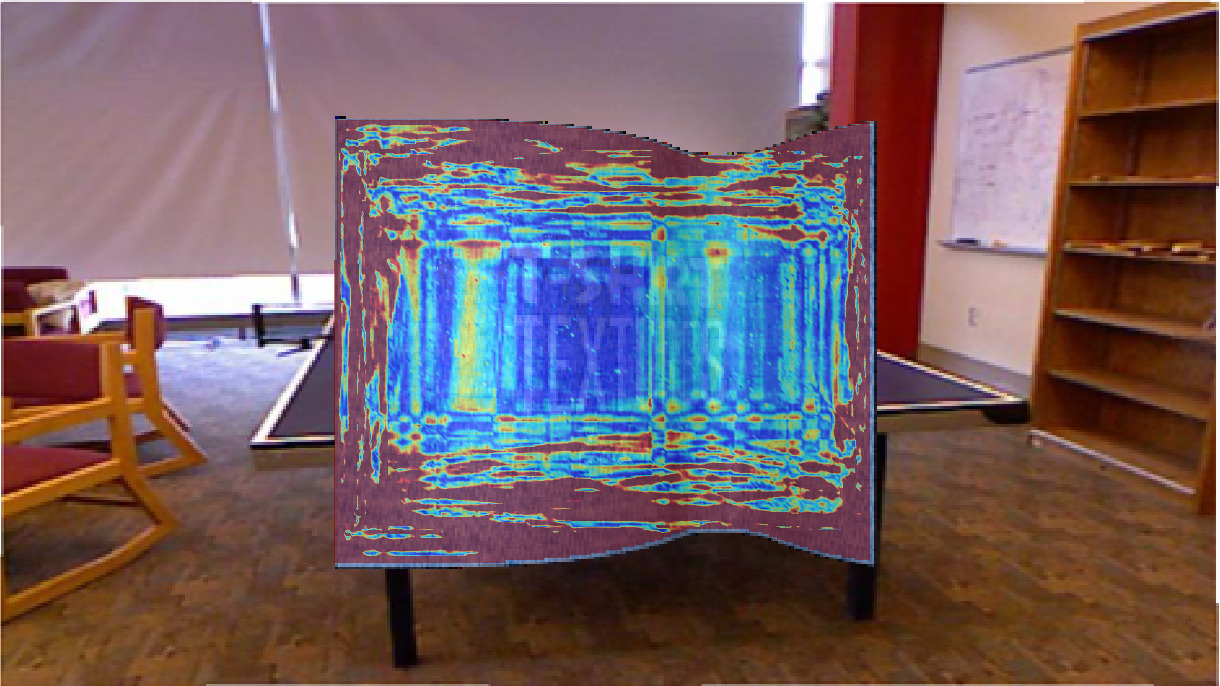}\\ 

			\multicolumn{1}{c}{\rule{0pt}{2ex} \centering   DeepSft}&\includegraphics[width=30mm]{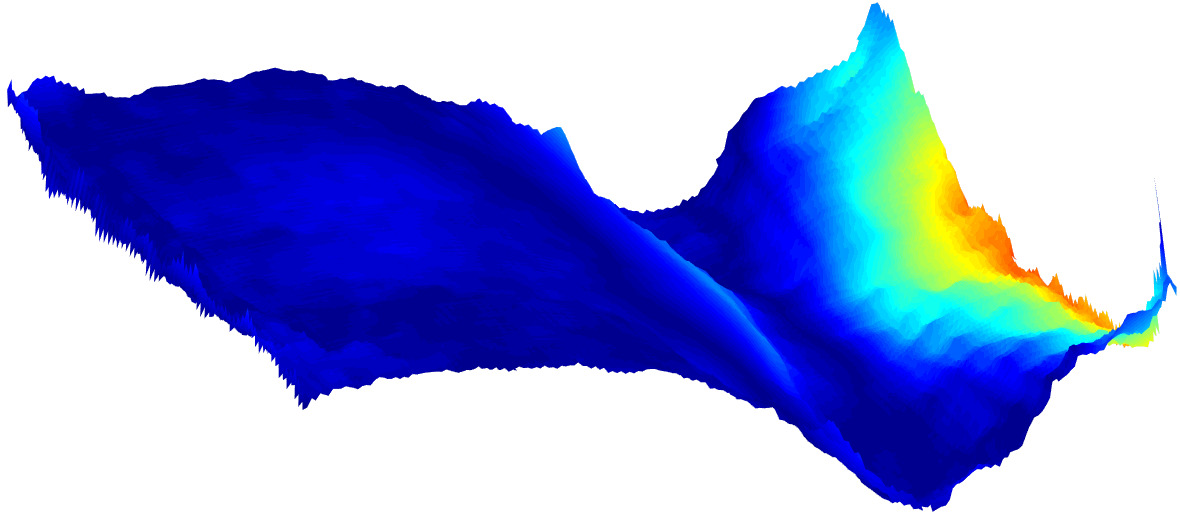}&\includegraphics[width=30mm]{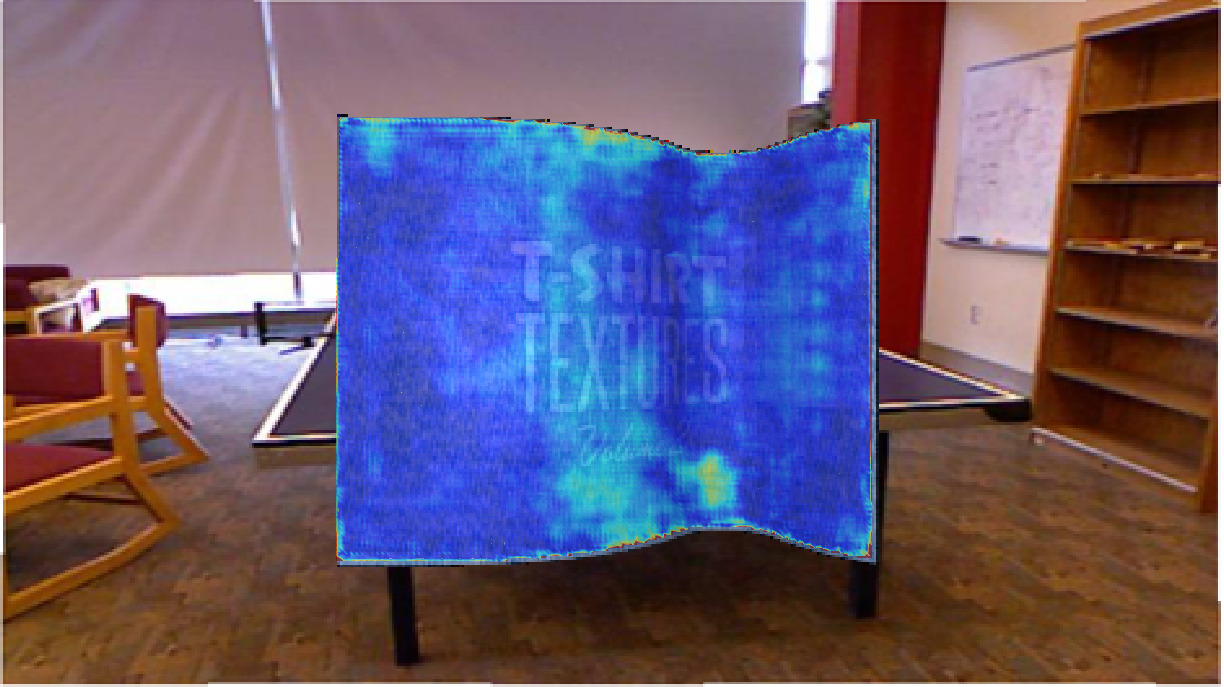}\\ \hline

			&\multicolumn{1}{c}{\centering{Ground-truth 3D surface}}&\multicolumn{1}{c}{\centering DS3 Input Image} \\ 
			
			&\includegraphics[width=30mm]{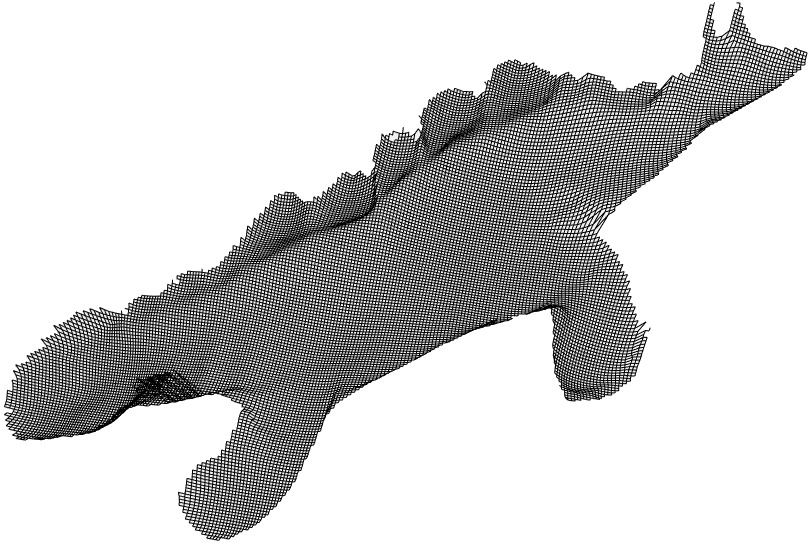}&\includegraphics[width=30mm]{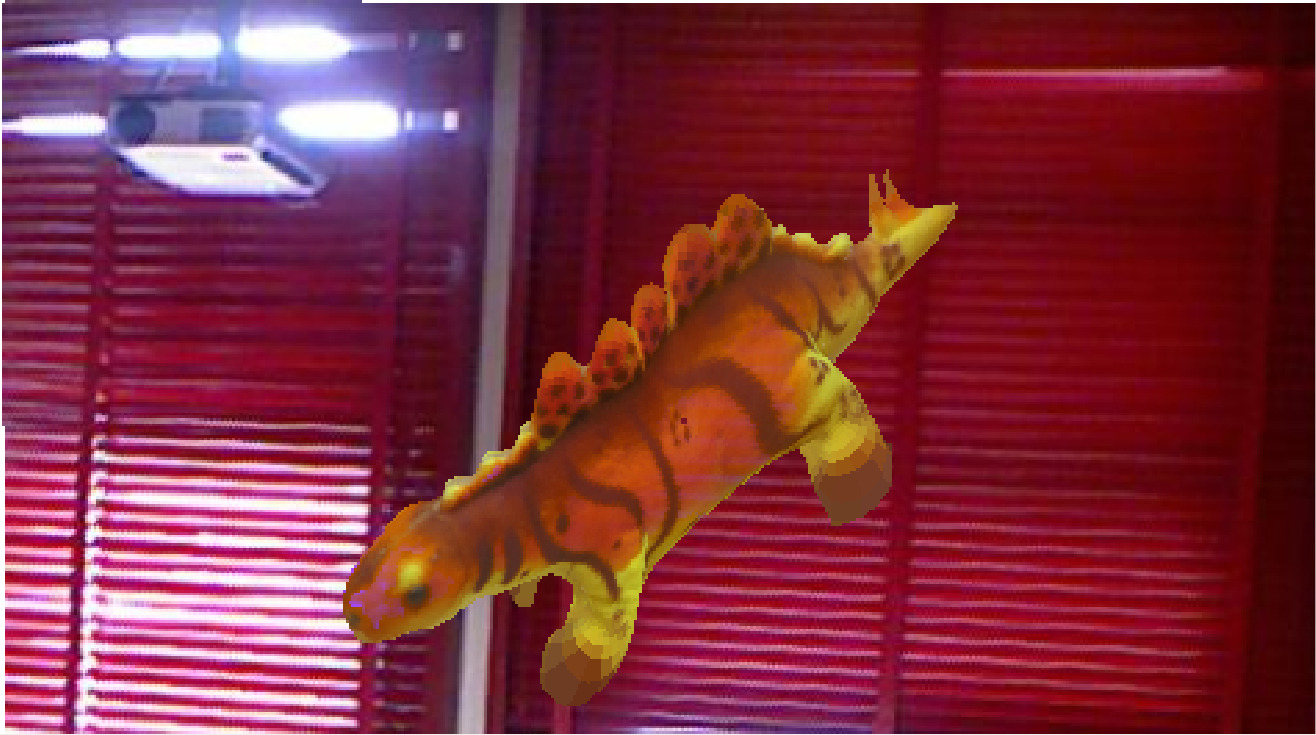}\\ 

			\multicolumn{1}{c}{Method}& \multicolumn{1}{c}{Reconstruction \& RMSE colormap}&\multicolumn{1}{c}{Registration ROI \& RMSE colormap} \\ \hline

			\multicolumn{1}{c}{\rule{0pt}{2ex} \centering   R50F}&\includegraphics[width=30mm]{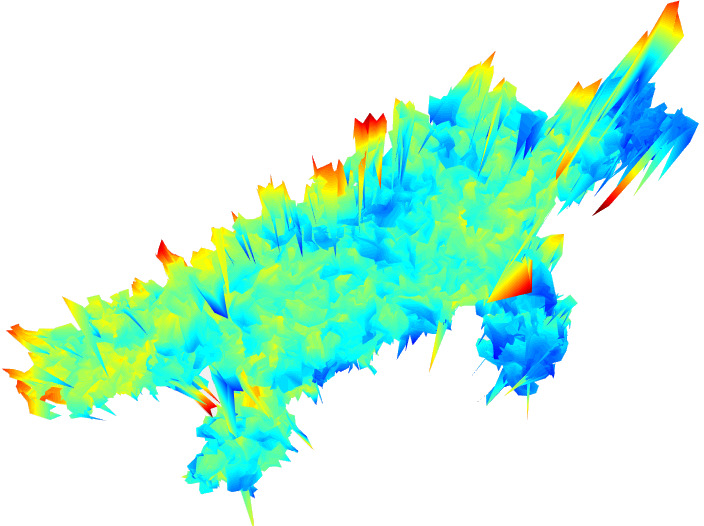}&\includegraphics[width=30mm]{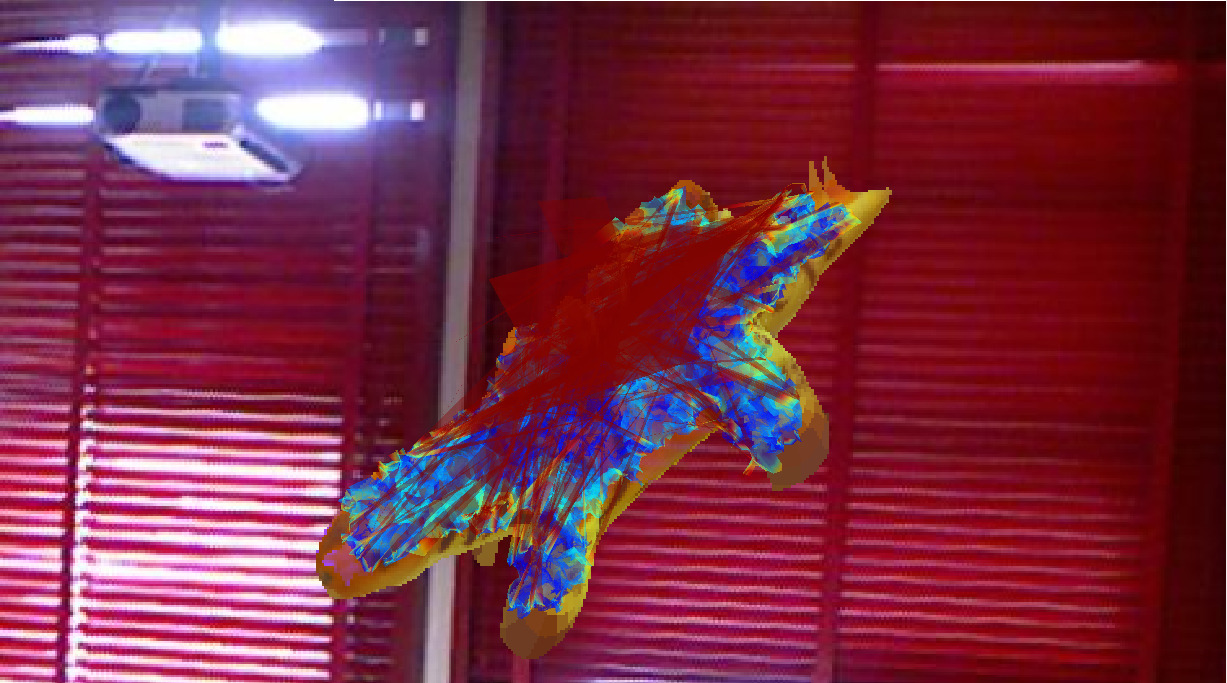}\\ 

			\multicolumn{1}{c}{\rule{0pt}{2ex} \centering   DeepSft}&\includegraphics[width=30mm]{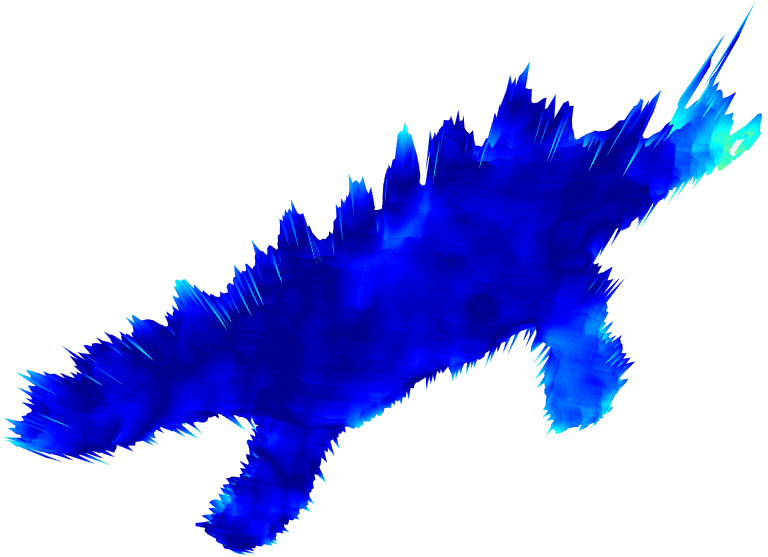}&\includegraphics[width=30mm]{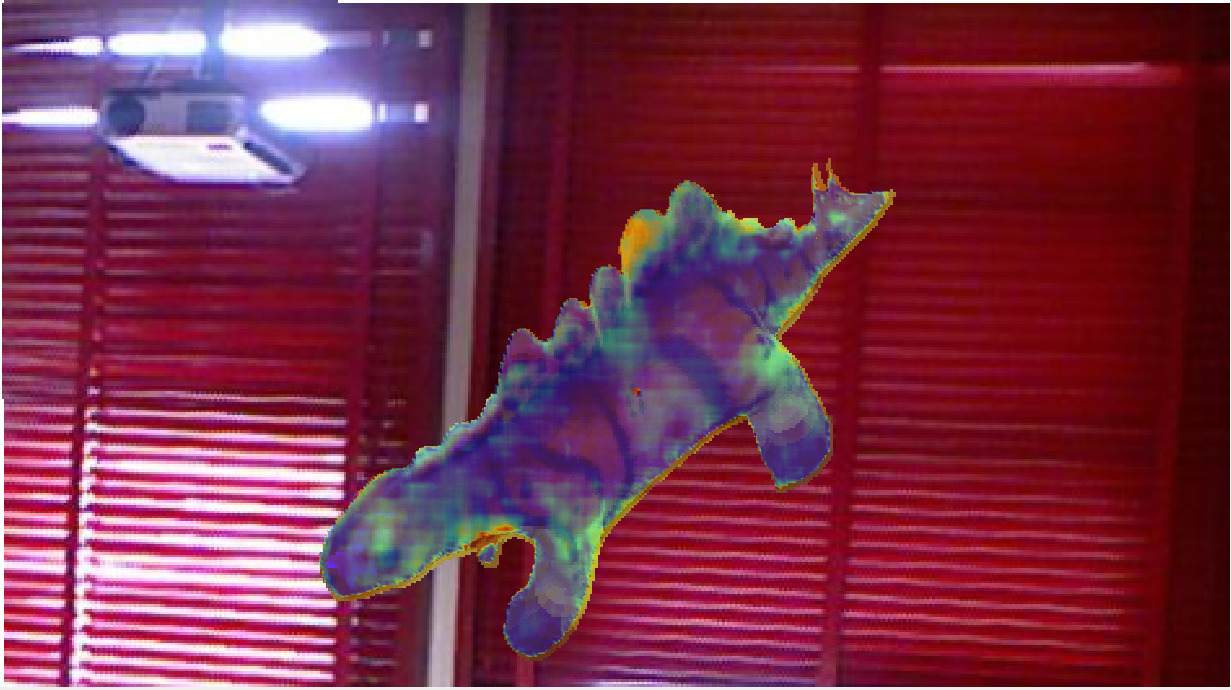}\\ 

			RMSE colormap & \multicolumn{1}{l}{\rule{0pt}{2ex} \textbf{0} \includegraphics[width=30mm,height=2mm]{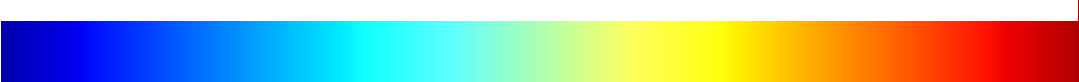} \textbf{30} mm} & \multicolumn{1}{l}{\rule{0pt}{2ex} \textbf{0} \includegraphics[width=30mm,height=2mm]{colormap_jet} \textbf{1} n.u.}
			\\ 
			
		\end{tabular} 
	\end{adjustbox}
	\centering \caption {Visual comparison of results computed from DeepSfT and other classical and DNN SfT methods with two test objects. The reconstructions are colored according to RMSE with heatmaps (middle column). The registration results are visualised with an overlay of the predicted template shape projected onto the input image. Registration errors are visualised with heatmaps (right column). n.u. stands for normalised texture map units.}
	\label{tb:qualitative}
\end{table}

The results show a similar trend as with the rectangular template datasets: DeepSfT outperforms the other methods in terms of reconstruction error, with an RMSE of the order of millimeters, and in registration with an RMSE close to 2 px. The second best method is R50F, although its results are significantly worse than DeepSfT is. The results of CH17 and its variants are very poor. This may be because CH17 is not well adapted for volumic objects with stronger non-isometric deformation.

\begin{table}[!htbp]
	\begin{adjustbox}{max width=\linewidth}
		\begin{tabular}{m{2cm}m{4cm}m{4cm}m{4cm}m{4cm}m{2cm}}

			\rule{0pt}{2.5ex} \centering \large{Dataset}& \rule{0pt}{2.5ex} \centering \large{Input Image} &\rule{0pt}{2.5ex} \centering \large{Depth Output}&\rule{0pt}{2.5ex} \centering \large{Registration Output\qquad(v)}&\rule{0pt}{2.5ex} \centering \large{Registration Output\qquad(v)} &\multicolumn{1}{m{2cm}}{\rule{0pt}{3ex}  \centering \large{Depth RMSE (mm)}}\\ \hline
			
			\centering \large{DS2}&\vspace{1.52mm}\includegraphics[width=40mm]{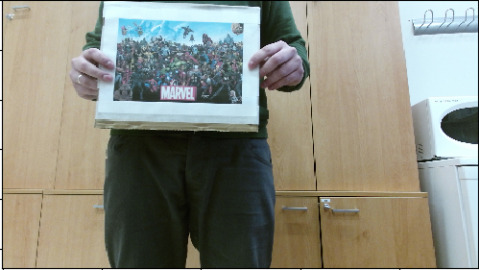} & \vspace{1.52mm}\includegraphics[width=40mm]{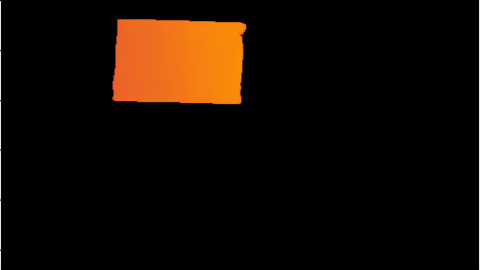}
			& \vspace{1.52mm}\includegraphics[width=40mm]{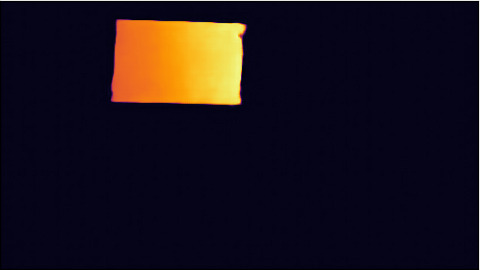}  & \vspace{1.52mm}\includegraphics[width=40mm]{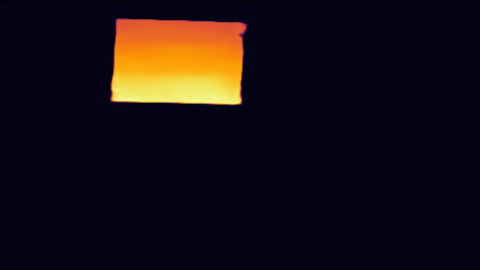}& \multicolumn{1}{c}{\large{2.82}}  \\ 
			
			\centering \large{DS5}&\vspace{1.52mm}\includegraphics[width=40mm]{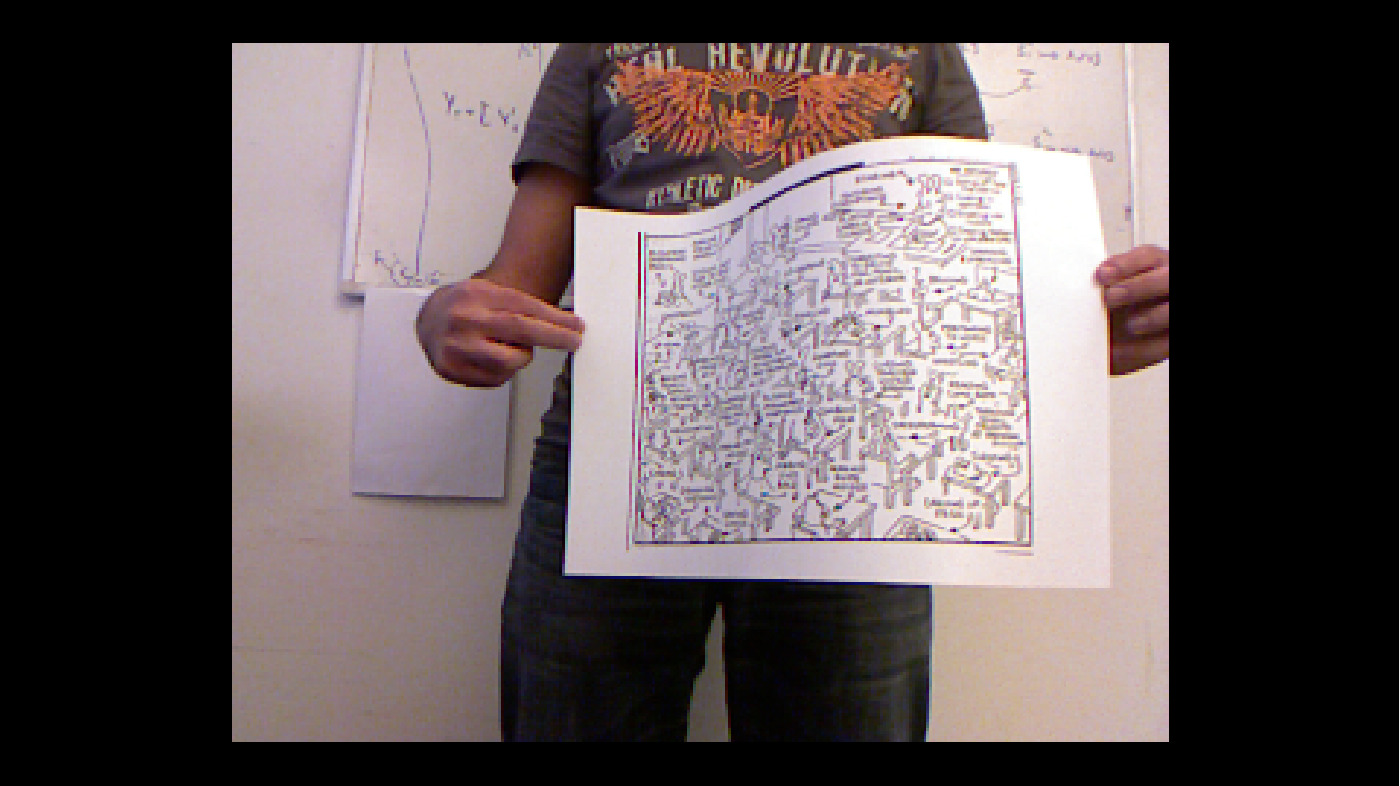} & \vspace{1.52mm}\includegraphics[width=40mm]{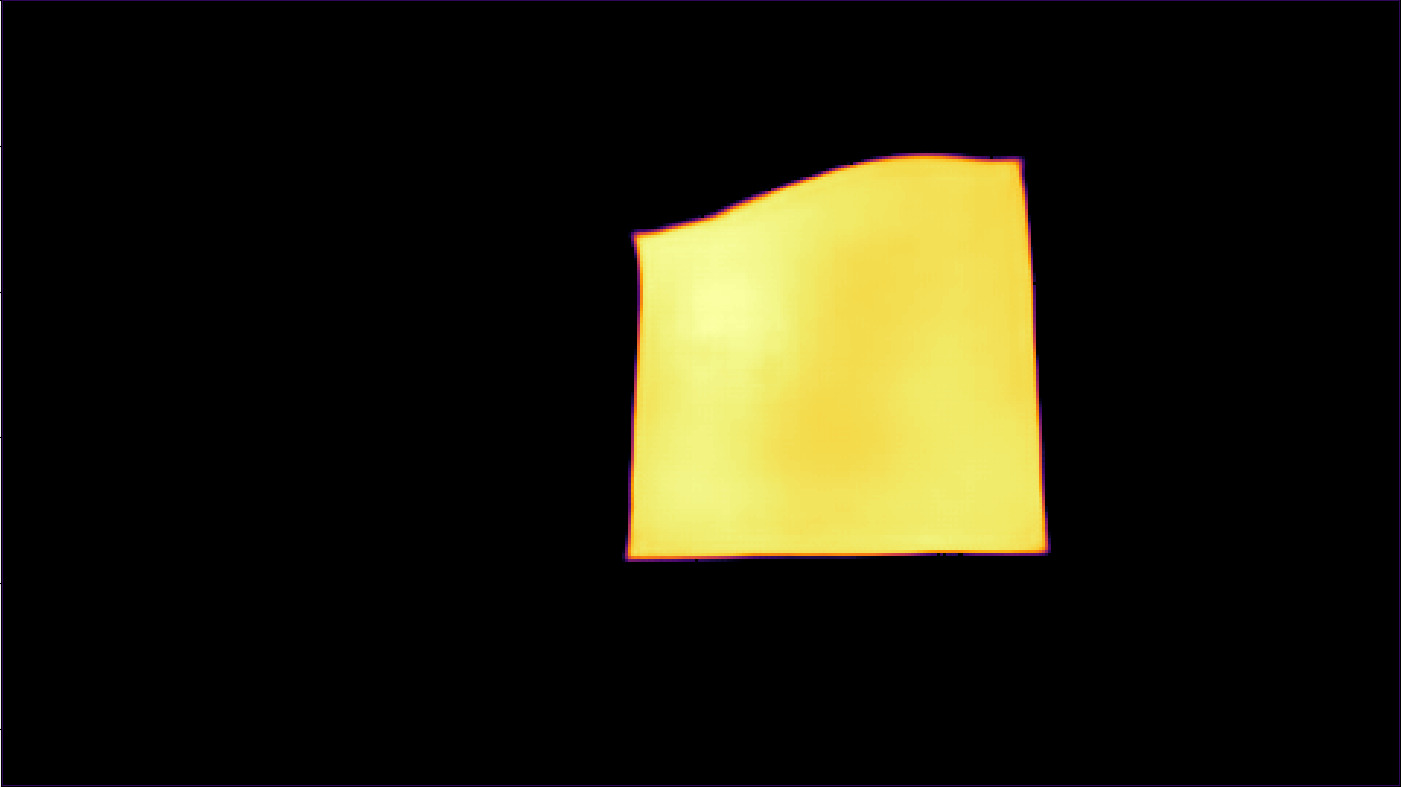}
			& \vspace{1.52mm}\includegraphics[width=40mm]{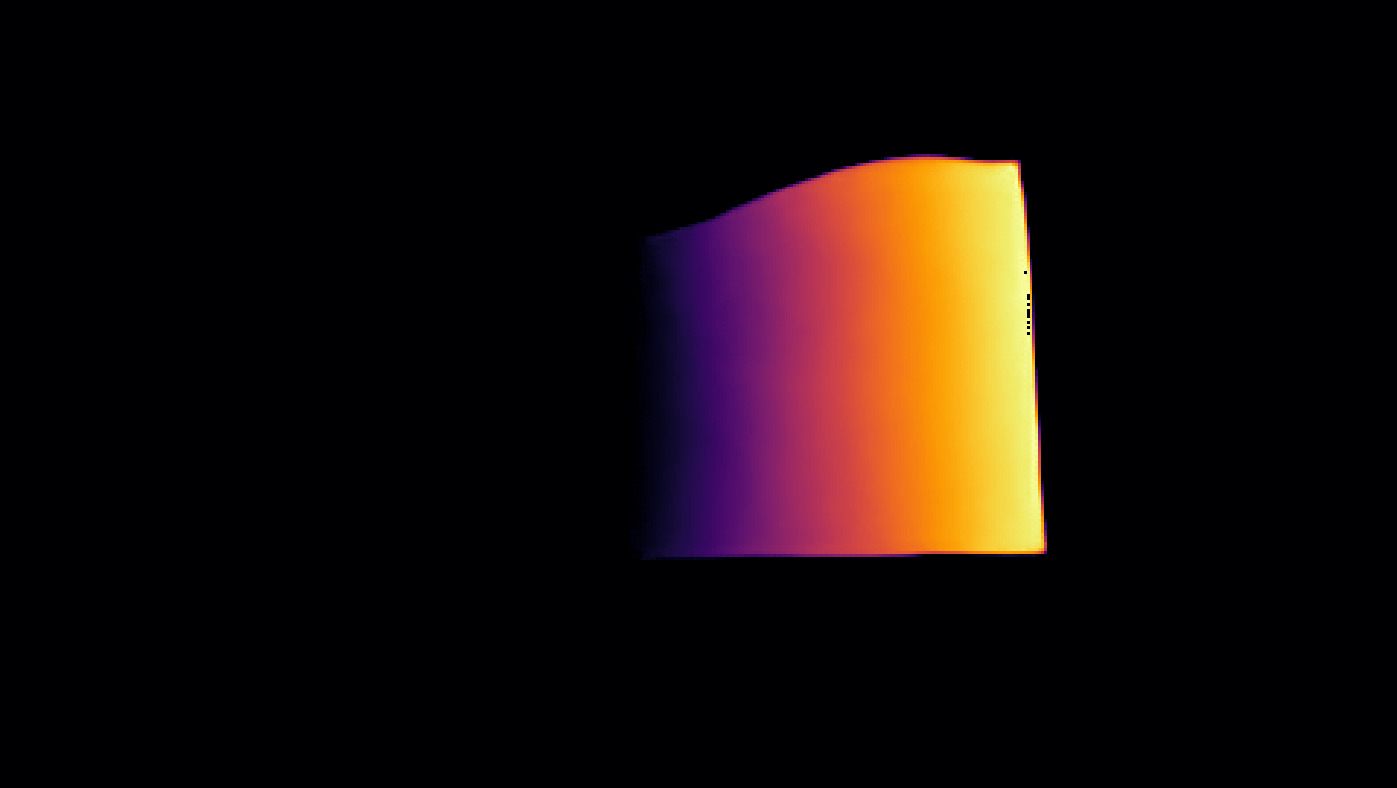}  & \vspace{1.52mm}\includegraphics[width=40mm]{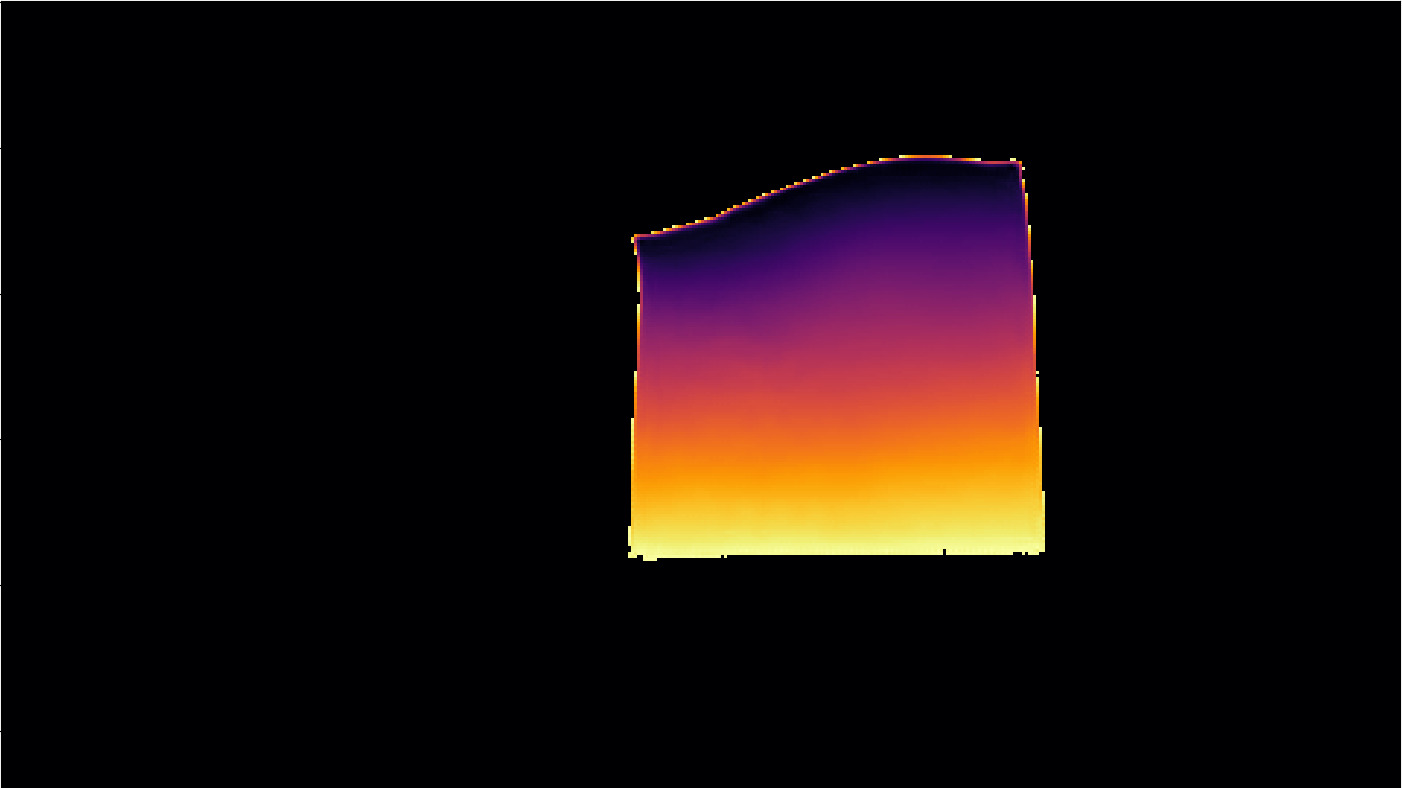}& \multicolumn{1}{c}{\large{6.01}}  \\ 
			
			\centering \large{DS1}&\vspace{1.52mm}\includegraphics[width=40mm]{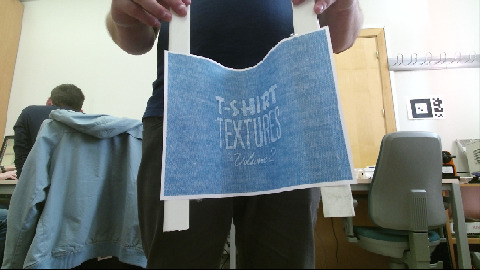} & \vspace{1.52mm}\includegraphics[width=40mm]{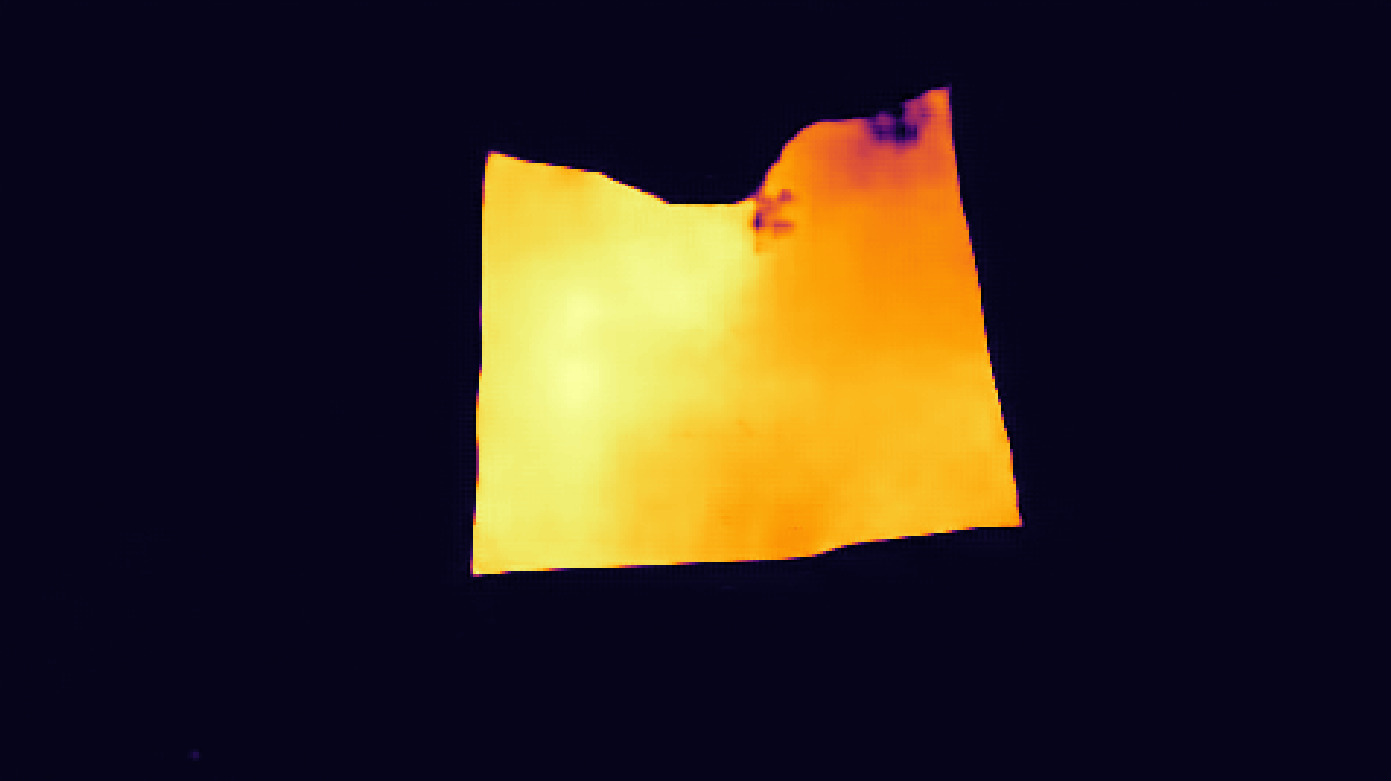}
			& \vspace{1.52mm}\includegraphics[width=40mm]{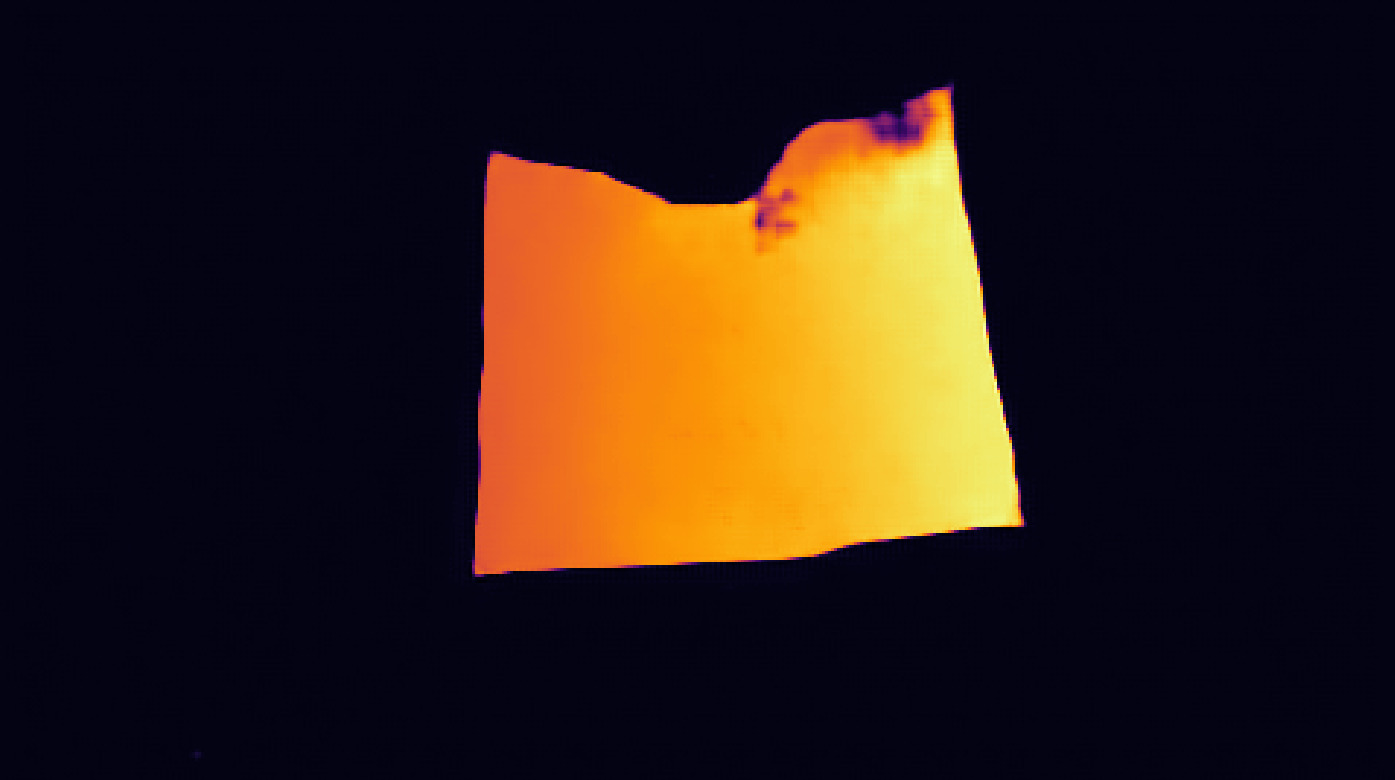}  & \vspace{1.52mm}\includegraphics[width=40mm]{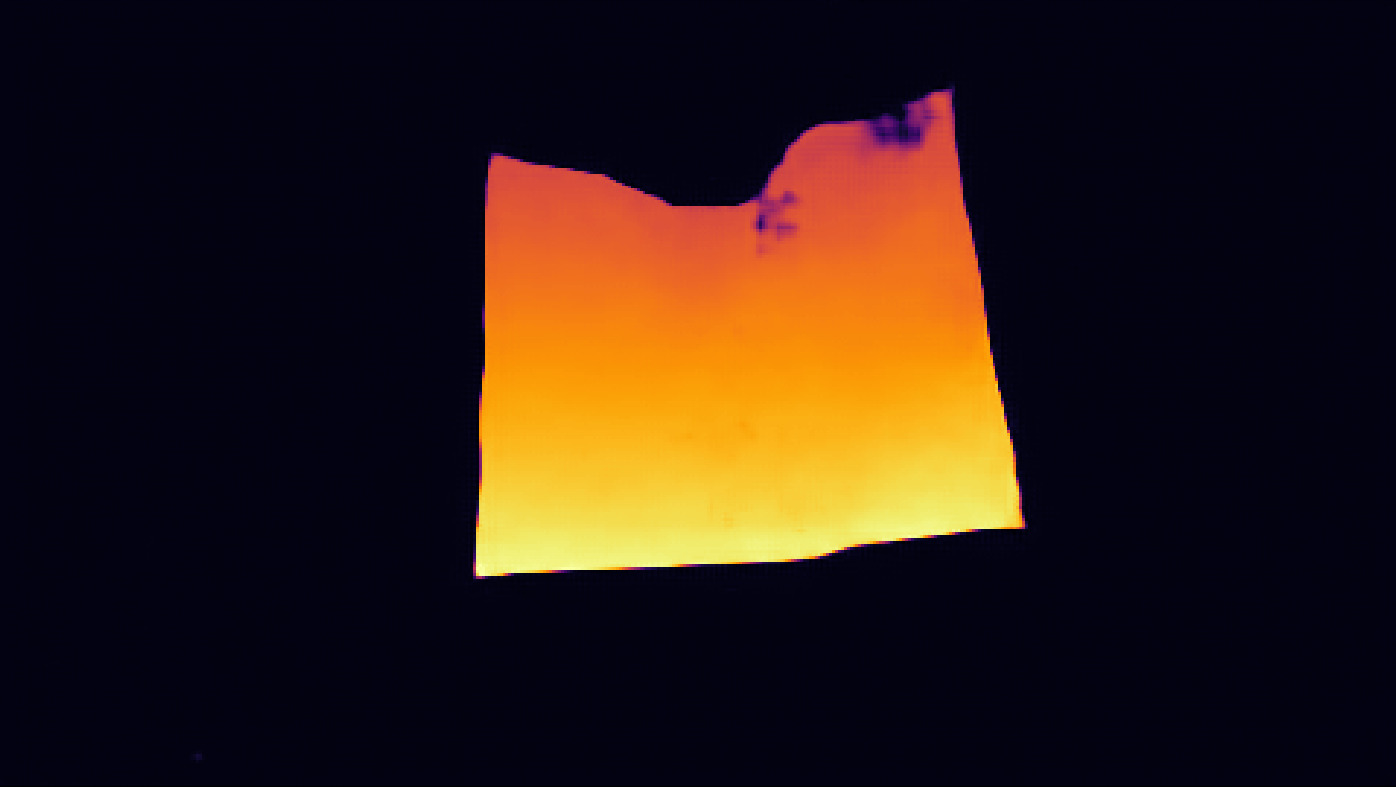}& \multicolumn{1}{c}{\large{4.69}}  \\ 
			
			\centering \large{DS3}&\vspace{1.52mm}\includegraphics[width=40mm]{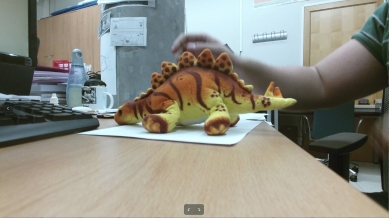} & \vspace{1.52mm}\includegraphics[width=40mm]{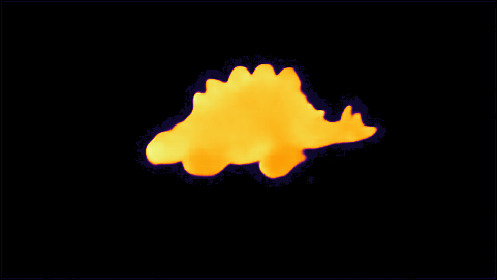}
			& \vspace{1.52mm}\includegraphics[width=40mm]{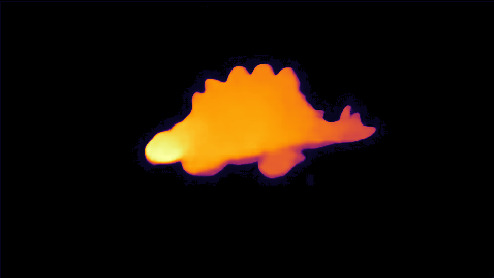} &  \vspace{1.52mm}\includegraphics[width=40mm]{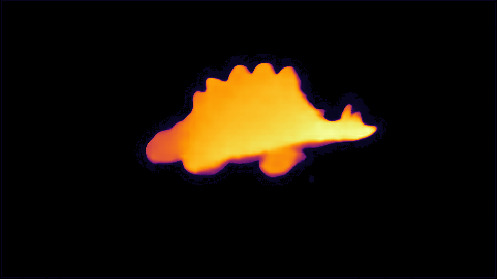}& \multicolumn{1}{c}{\large{11.26}} \\ 
			
			\centering \large{DS3}&\vspace{1.52mm}\includegraphics[width=40mm]{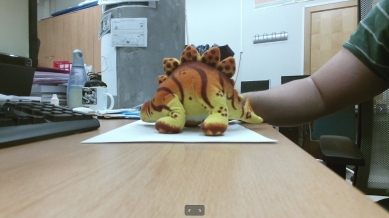}  &
			\vspace{1.52mm}\includegraphics[width=40mm]{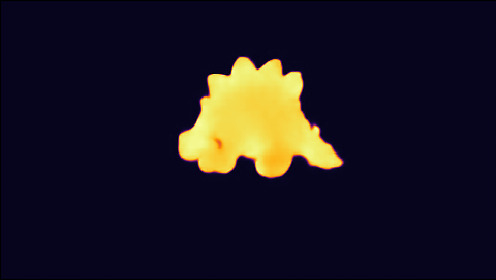} & \vspace{1.52mm}\includegraphics[width=40mm]{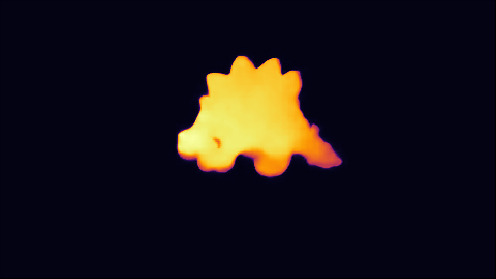} & \vspace{1.52mm}\includegraphics[width=40mm]{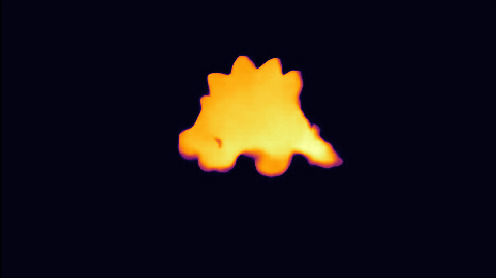}&\multicolumn{1}{c}{\large{8.96}} \\ 
			
			\centering \large{DS4}&\vspace{1.52mm}\includegraphics[width=40mm]{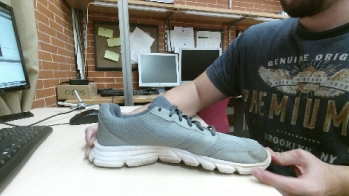}  &
			\vspace{1.52mm}\includegraphics[width=40mm]{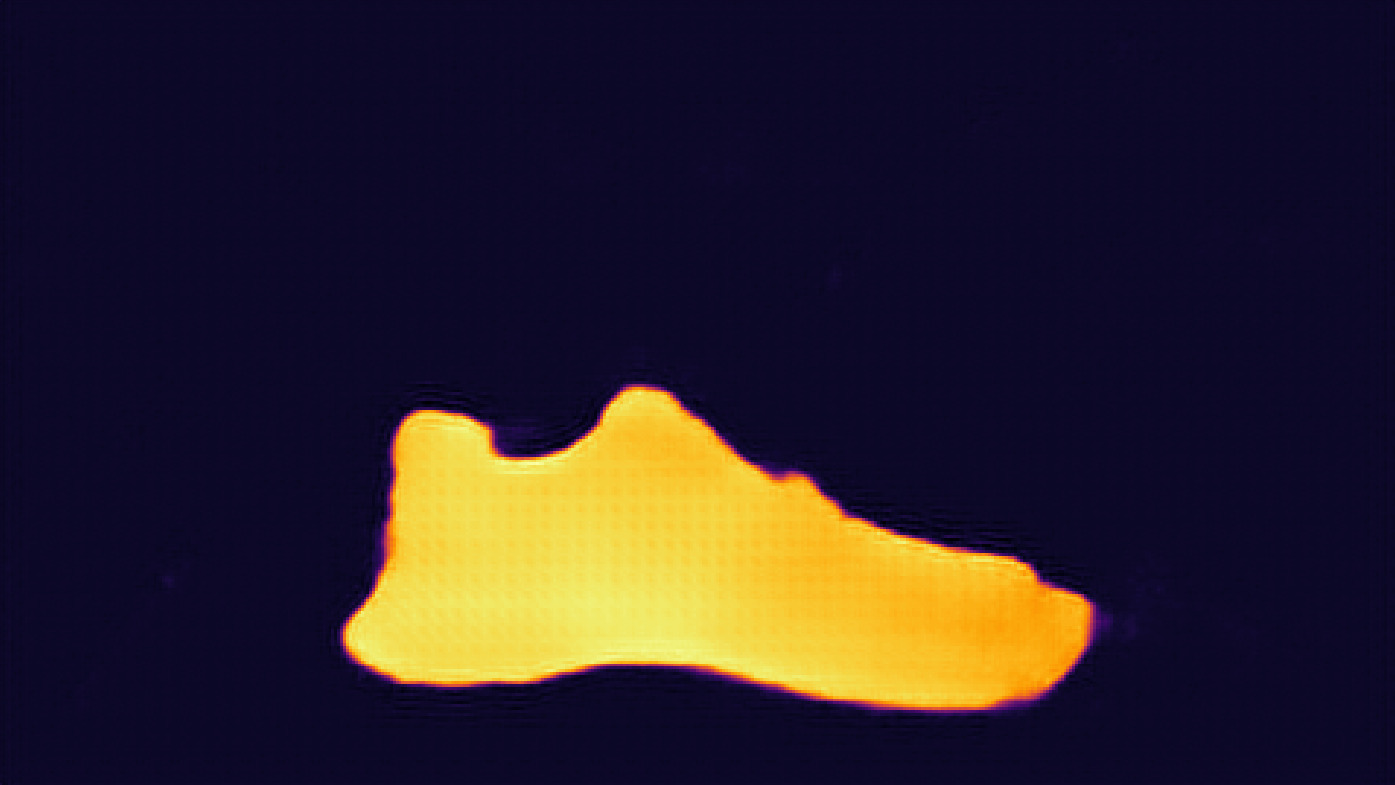} & \vspace{1.52mm}\includegraphics[width=40mm]{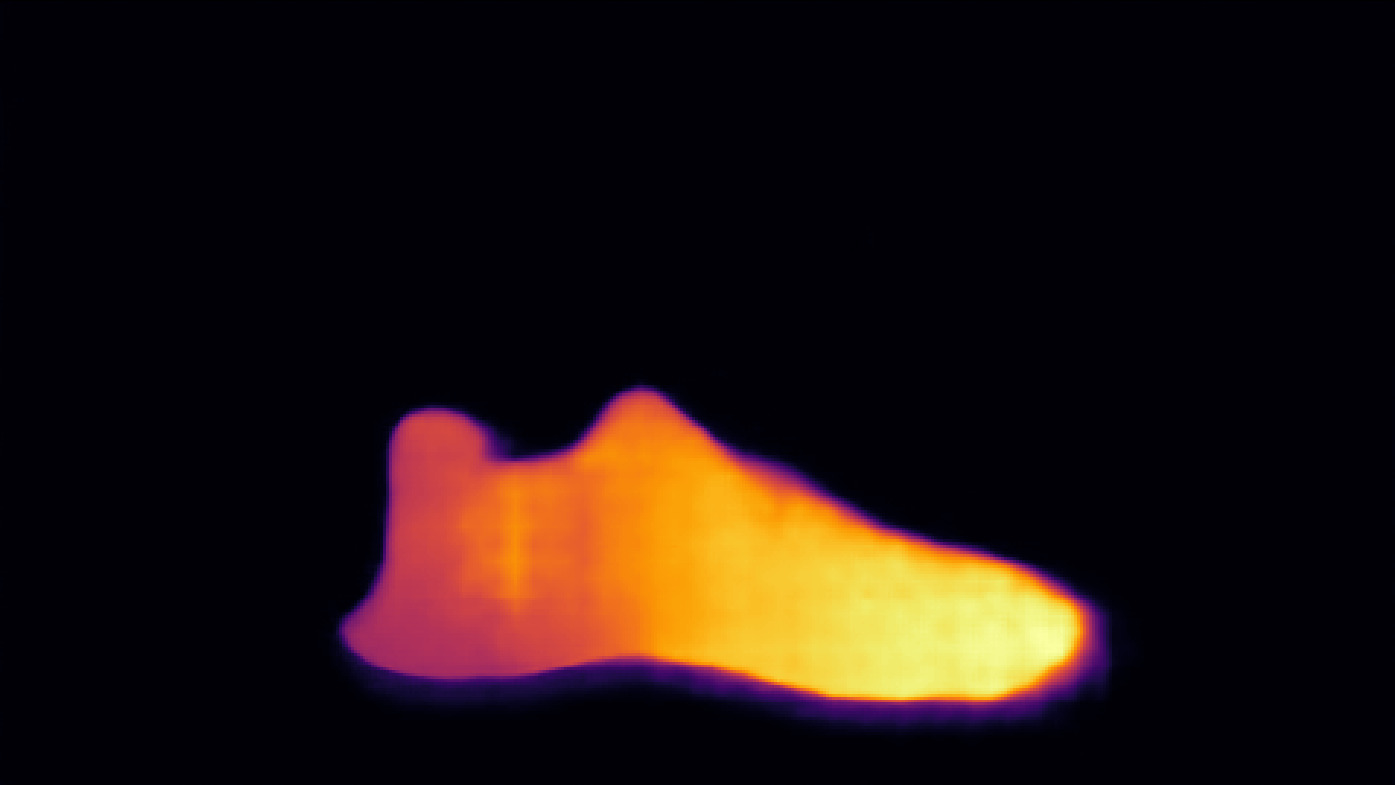} & \vspace{1.52mm}\includegraphics[width=40mm]{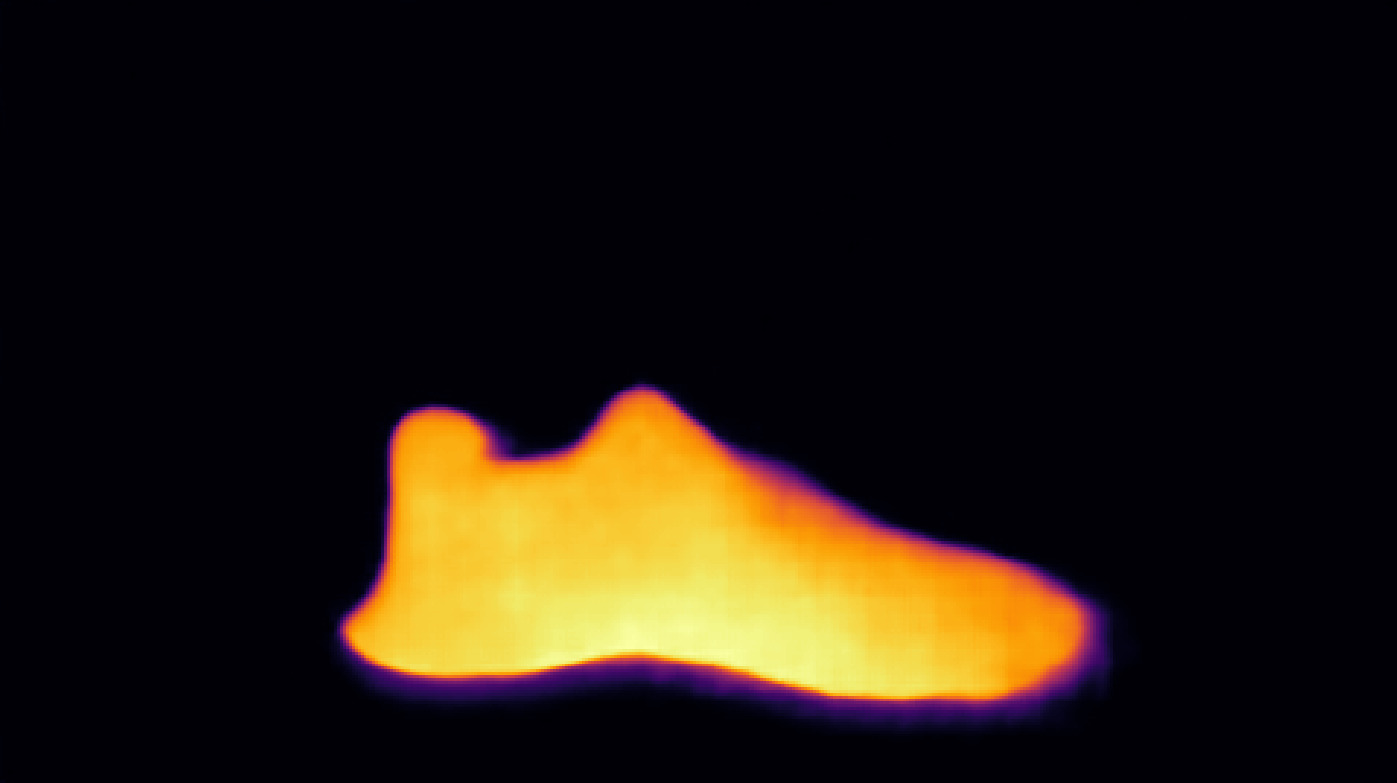}&\multicolumn{1}{c}{\large{9.08}}  \\ 
			
			\centering \large{DS4}&\vspace{1.52mm}\includegraphics[width=40mm]{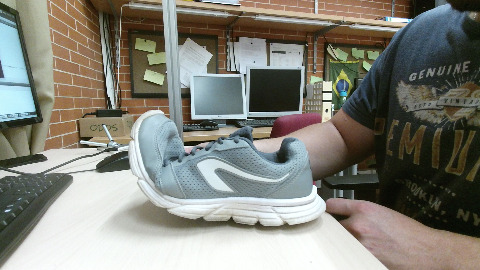}  &
			\vspace{1.52mm}\includegraphics[width=40mm]{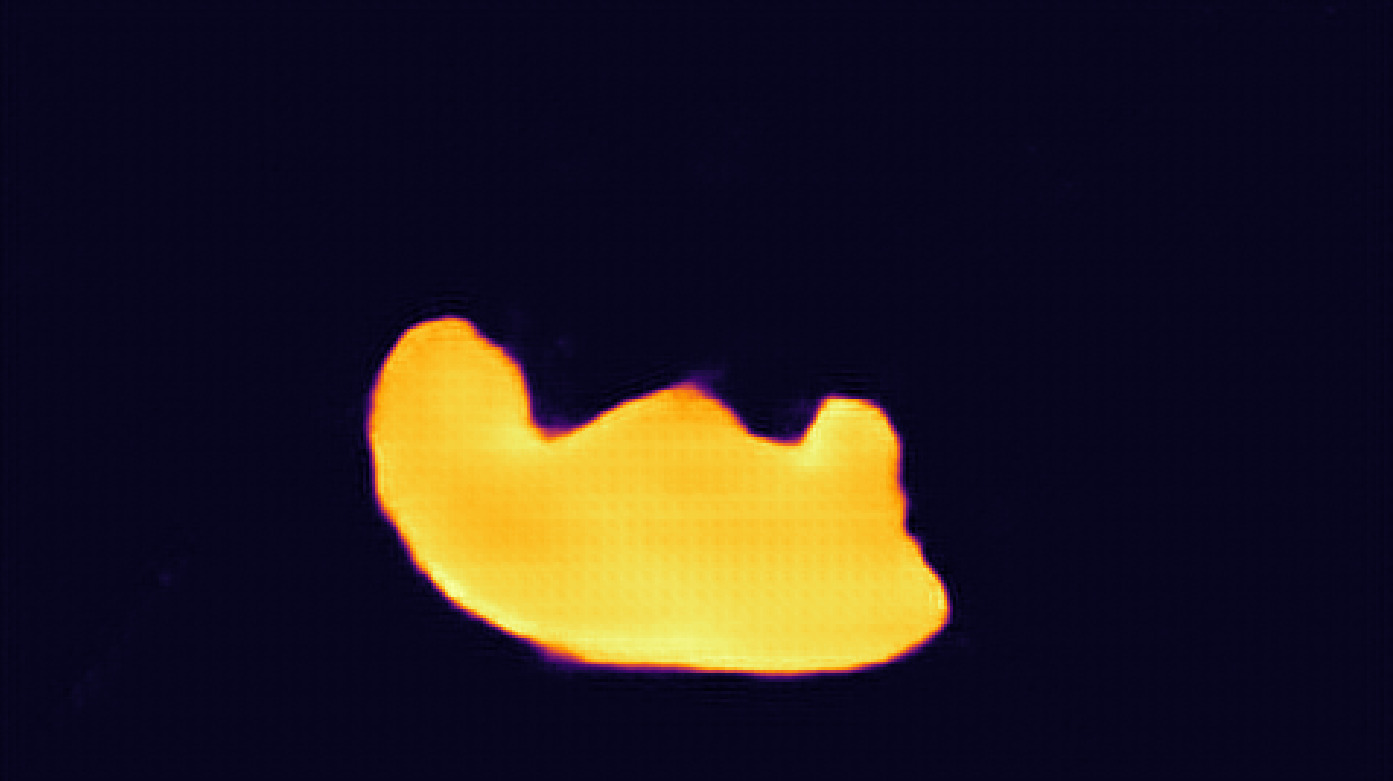} & \vspace{1.52mm}\includegraphics[width=40mm]{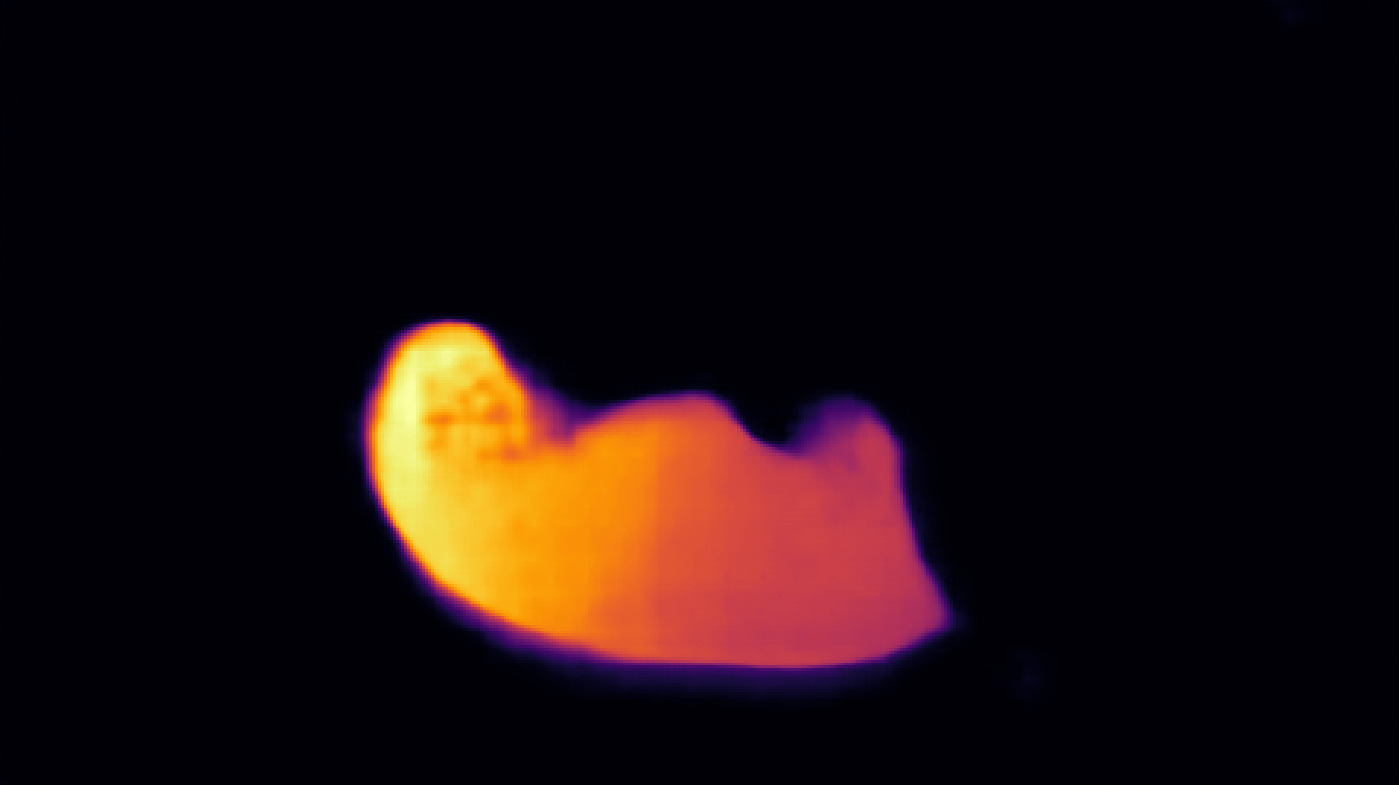} & \vspace{1.52mm}\includegraphics[width=40mm]{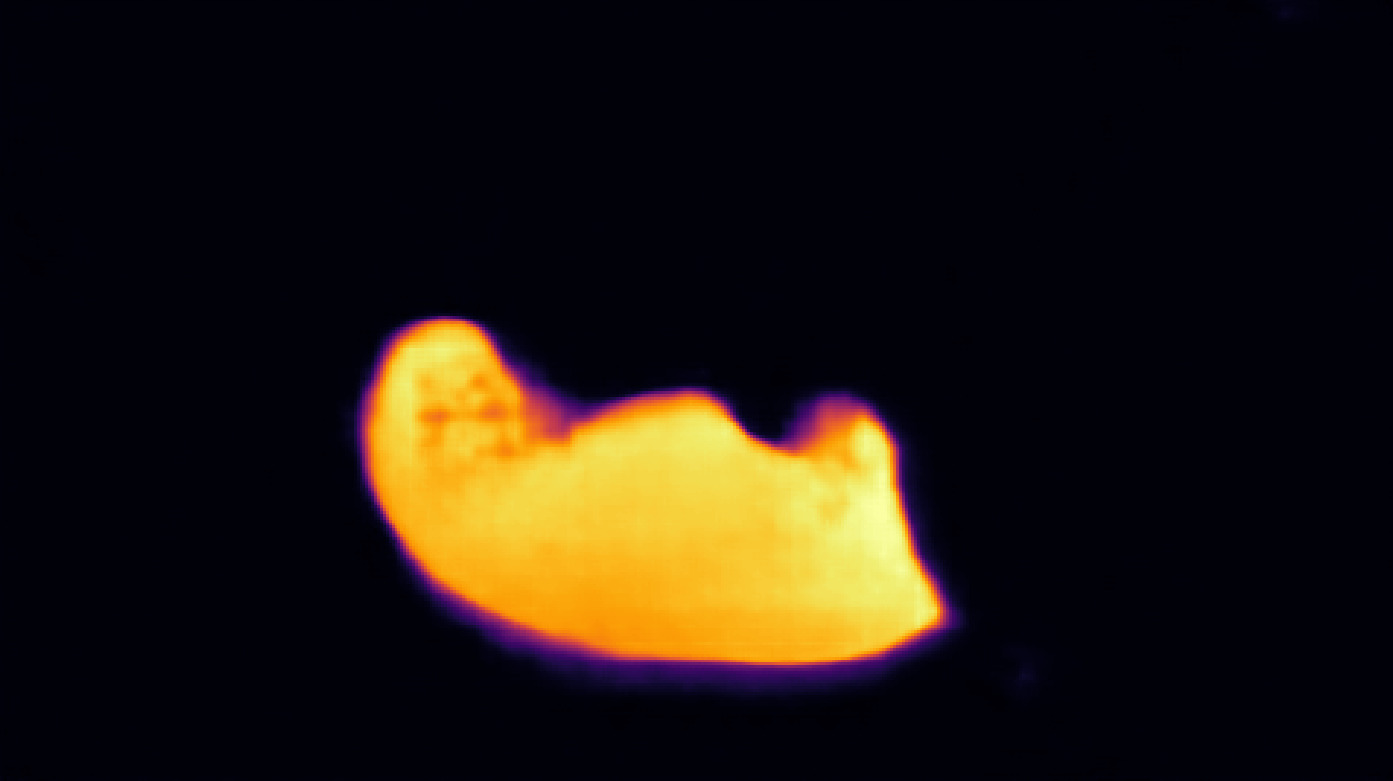}& \multicolumn{1}{c}{\large{7.49}}  \\ 
			
			\multicolumn{6}{c}{\rule{0pt}{2.5ex} \large{Colormaps}}
			\\ 
			
			\multicolumn{2}{c}{\rule{0pt}{2.5ex} \large{Registration Output (u)}}&\multicolumn{4}{l}{\large{\textbf{0}} \includegraphics[width=80mm,height=2mm]{colormap_inferno2} \large{\textbf{1}} \large{n.u.}}
			\\ 
			
			\multicolumn{2}{c}{\rule{0pt}{2.5ex}\large{Registration Output (v)}}&\multicolumn{4}{l}{\large{\textbf{0}} \includegraphics[width=80mm,height=2mm]{colormap_inferno2} \large{\textbf{1}} \large{n.u.}}
			\\ 
			
			\multicolumn{2}{c}{\rule{0pt}{2.5ex} \large{Depth}}&\multicolumn{4}{l}{\large{\textbf{0}} \includegraphics[width=80mm,height=2mm]{colormap_inferno2} \large{\textbf{500}} \large{millimeters}}
			\\ 

		\end{tabular} 
	\end{adjustbox}
	\caption {Example outputs for the five objects used to test DeepSfT. n.u. stands for normalised units in the template texture map.}
	\label{tb:examples}
\end{table}

We show in table \ref{tb:texture_mapped} qualitative reconstruction results obtained with DS1R, DS3R and DS4R with real images. We observe that shapes recovered with DeepSfT are similar to ground-truth obtained with the RGB-D camera and have no 'outliers' in their boundaries, in contrast to the RGB-D camera ground-truth. We observe that the error is larger near self-occlusion boundaries.

\begin{table*}[!htbp]
	\begin{adjustbox}{max width=\linewidth}
		\begin{tabular}{lm{3cm}m{3cm}m{3cm}m{3cm}m{3cm}}

			\rule{0pt}{2.5ex} Dataset&\rule{0pt}{2.5ex}  Input Image &\rule{0pt}{2.5ex} Ground-truth &\rule{0pt}{2.5ex} DNN reconstruction &\rule{0pt}{2.5ex}  Textured DNN&\rule{0pt}{2.5ex} 3D Shape completion \\ 
			
			&&&\rule{0pt}{2.5ex} output (blue) vs GT (red)&\rule{0pt}{2.5ex}  reconstruction output&\\ 
			
			\rule{0pt}{2.5ex}  DS1 &\vspace{1.52mm}\includegraphics[width=30mm]{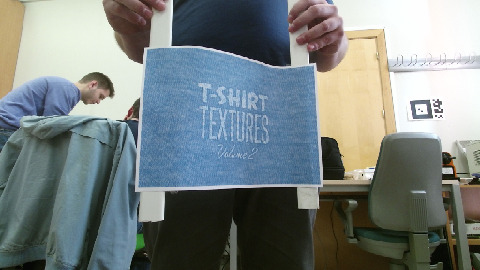} &\vspace{1.52mm}\includegraphics[width=30mm]{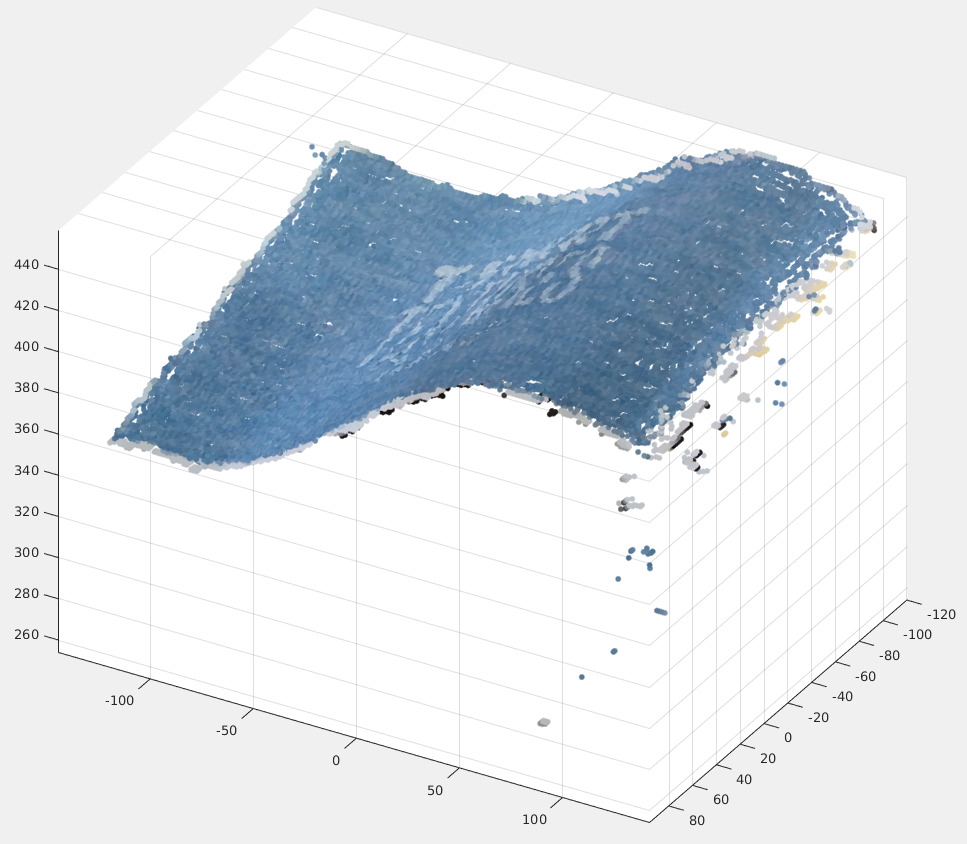}& \vspace{1.52mm}\includegraphics[width=30mm]{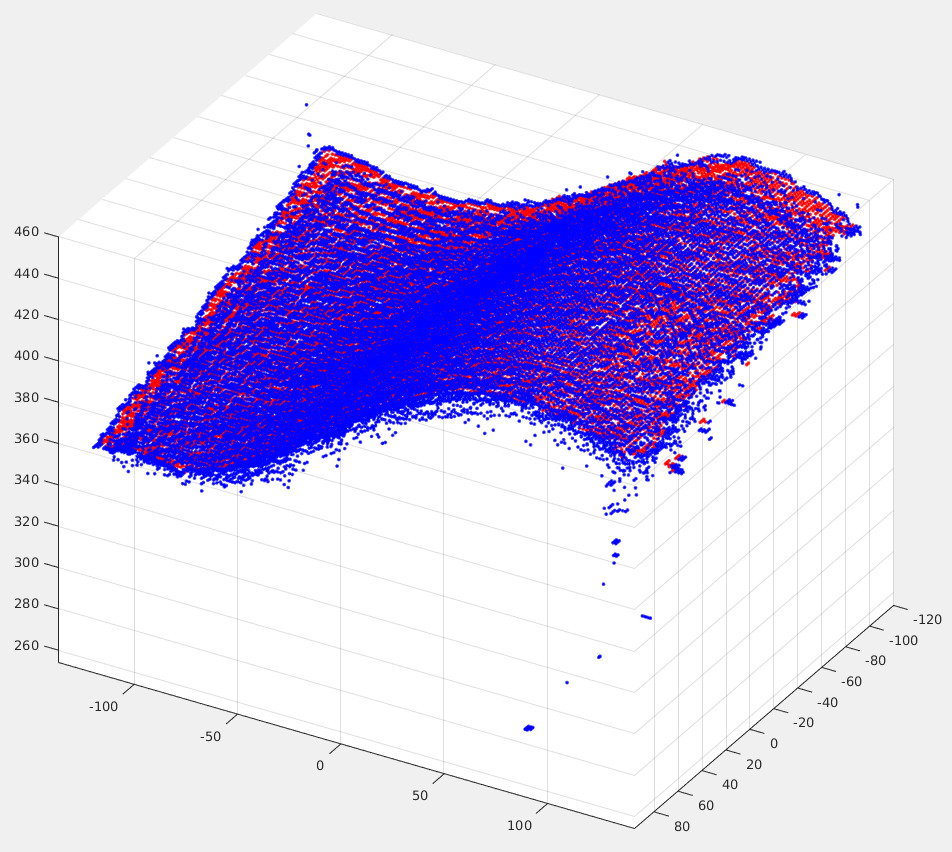}
			& \vspace{1.52mm}\includegraphics[width=30mm]{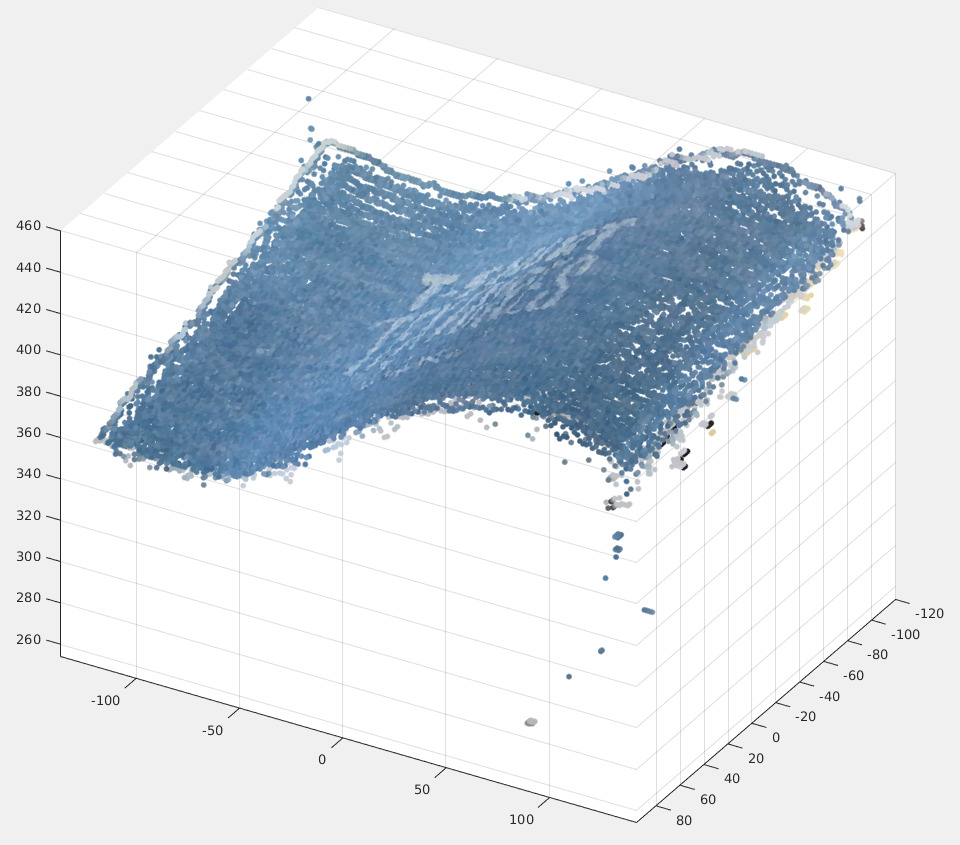}  & DNN RMSE (mm)  \vspace{1.52mm}\includegraphics[width=30mm]{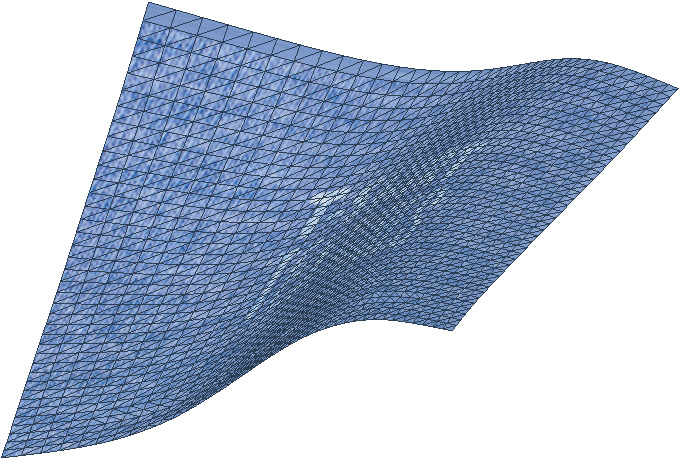} \\ 
			
			&&&&DNN RMSE (mm) 3.26&ARAP RMSE (mm) 3.41\\ 
			
			\rule{0pt}{2.5ex}  DS3&\vspace{1.52mm}\includegraphics[width=30mm]{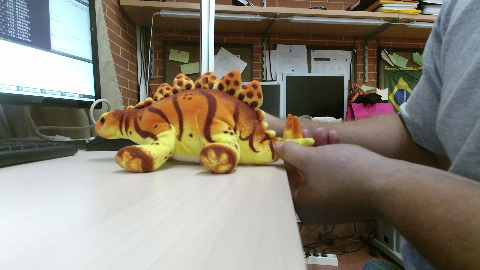}&\vspace{1.52mm}\includegraphics[width=30mm]{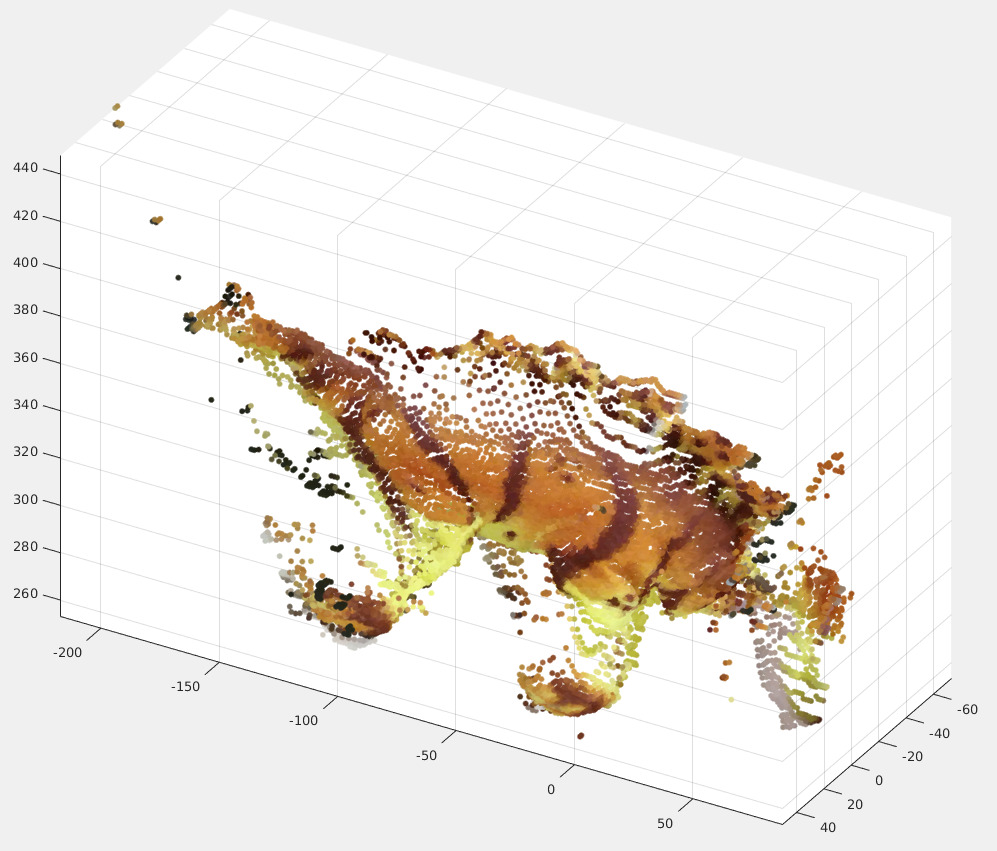} & \vspace{1.52mm}\includegraphics[width=30mm]{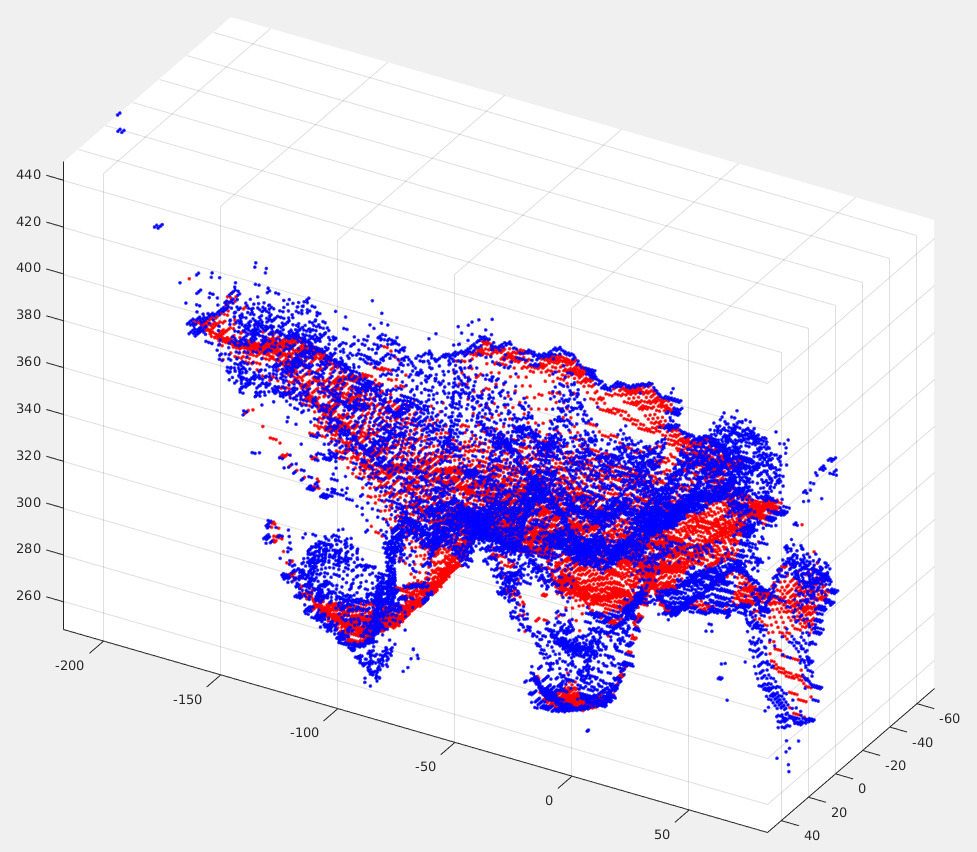}
			& \vspace{1.52mm}\includegraphics[width=30mm]{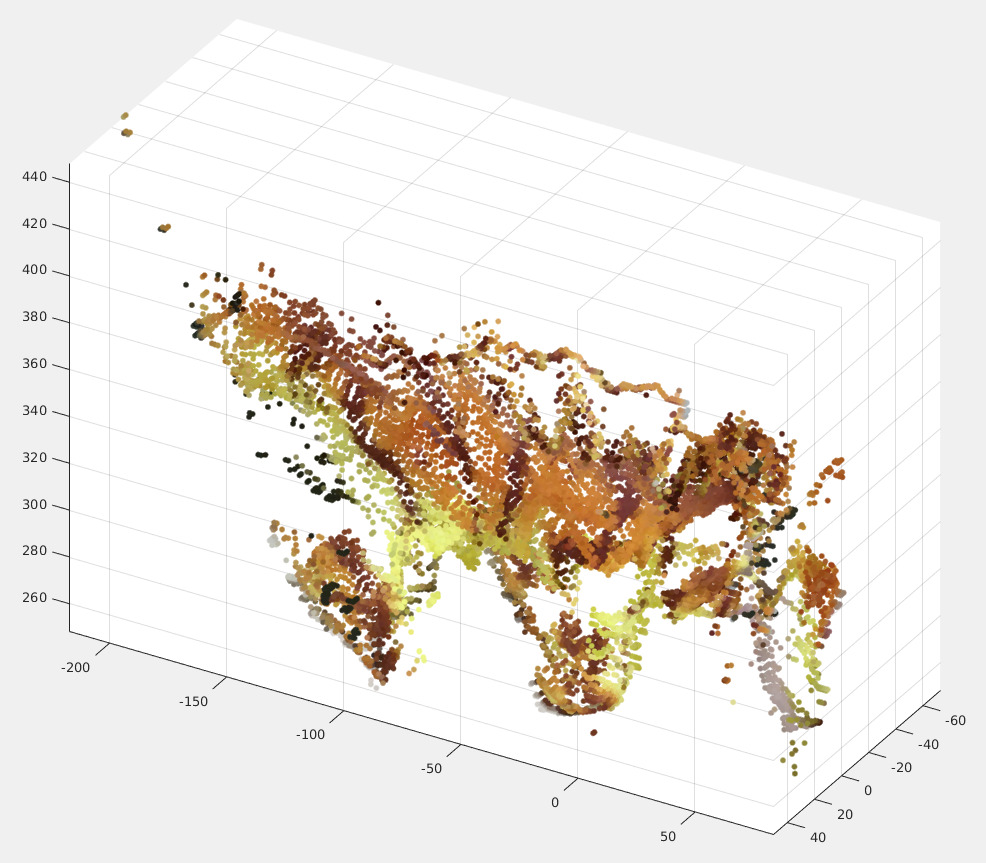}  & \vspace{1.52mm}\includegraphics[width=30mm]{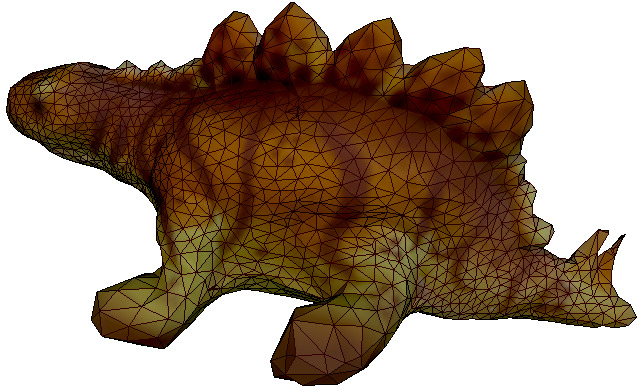}\\ 
			
			&&&&DNN RMSE (mm) 9.51&ARAP RMSE (mm) 9.84\\ 
			
			\rule{0pt}{2.5ex}  DS4&\vspace{1.52mm}\includegraphics[width=30mm]{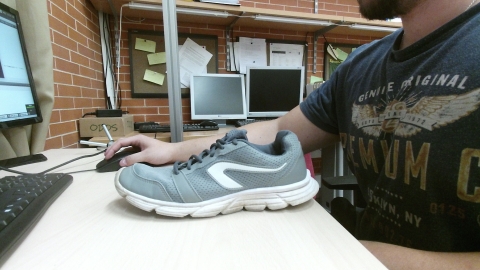} & \vspace{1.52mm}\includegraphics[width=30mm]{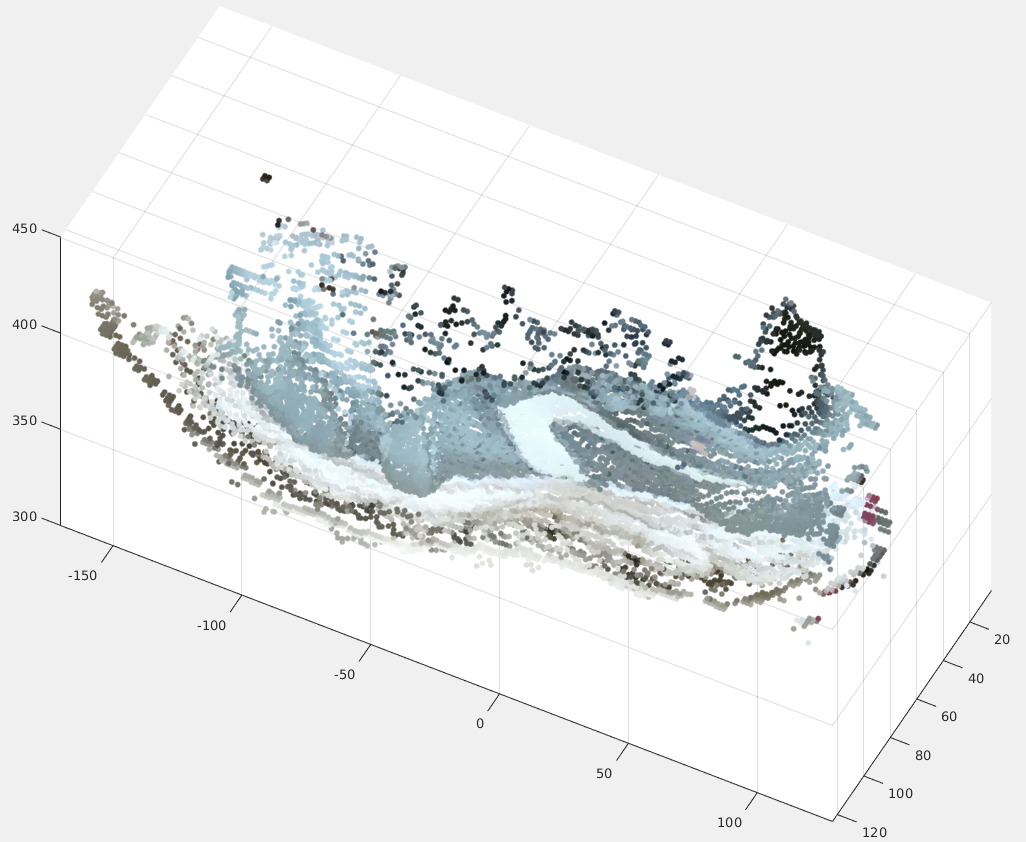}& \vspace{1.52mm}\includegraphics[width=30mm]{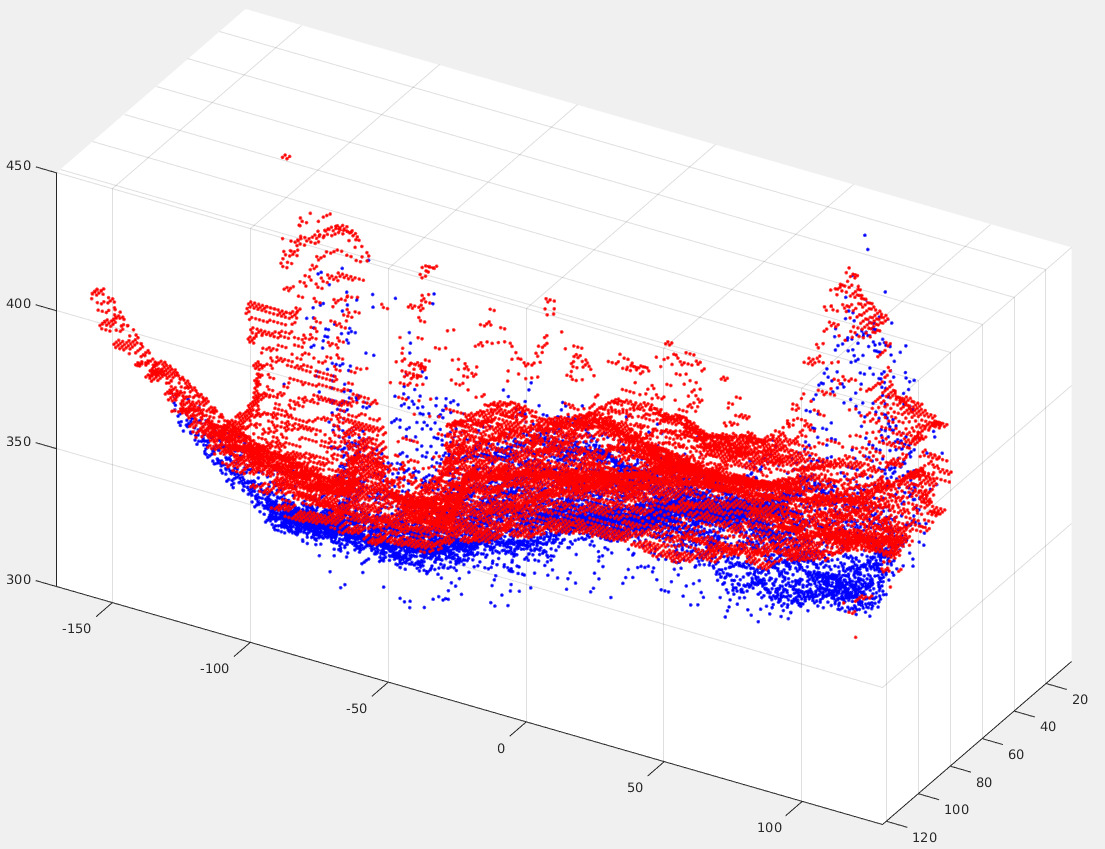}
			& \vspace{1.52mm}\includegraphics[width=30mm]{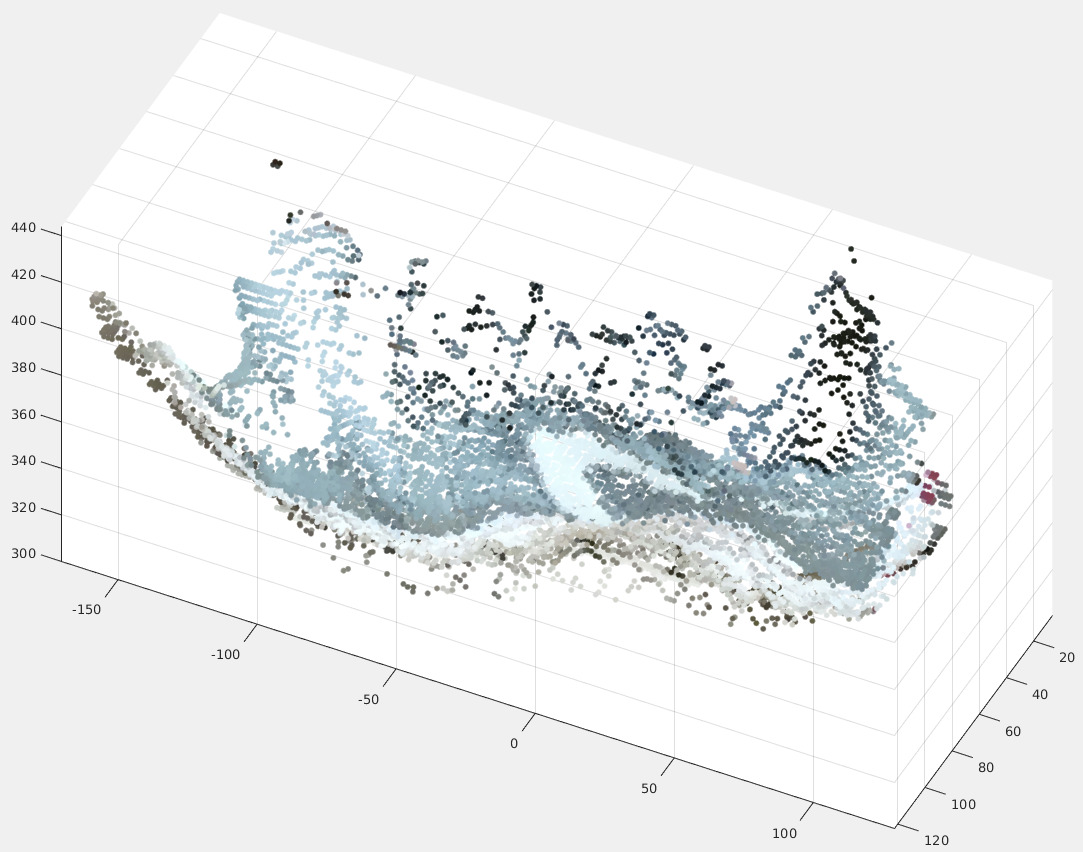} & \vspace{1.52mm}\includegraphics[width=30mm]{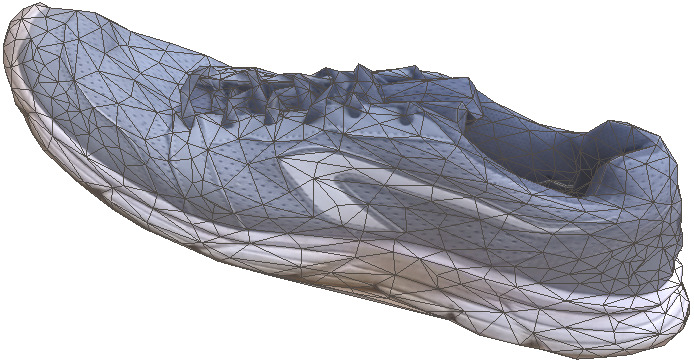} \\ 
			
			&&&&DNN RMSE (mm) 7.42&ARAP RMSE (mm) 7.47\\ 
		\end{tabular} 
	\end{adjustbox}
	\caption {Examples of DeepSfT results before and after ARAP 3D shape completion as described in \S\ref{subsec:arap} using real data.}
	\label{tb:texture_mapped}
\end{table*}

\subsection{Evaluation of ARAP shape completion}

We show in table \ref{tb:texture_mapped} example results before and after ARAP shape completion using DS1, DS3 and DS4 arranged in three rows. The table shows from left to right a representative input image, ground-truth provided by Kinect V2, registration and reconstruction outputs from DeepSfT as point clouds, outputs as coloured point clouds, and lastly the 3D shape completion results. The reconstruction errors are evaluated across the visible surface regions before and after shape completion and denoted by DNN RMSE and ARAP RMSE respectively. We can see that these errors are very similar, which implies that the benefit of shape completion is only to recover the occluded regions. It does not improve significantly the reconstruction of the visible regions compared to the DNN output. Quantitatively the completed 3D shapes look compelling and representative of the true object deformations.

\subsection{Evaluation of test camera generalisation}

\begin{table}[!htbp]
	
	\begin{adjustbox}{max width=\linewidth}
		\begin{tabular}{cm{3cm}m{3cm}m{3cm}}
			& \multicolumn{1}{r}{Kinect V2} & \multicolumn{1}{r}{Realsense D435} & \multicolumn{1}{r}{ GoproHeroV3}\\
			\vspace{1.5mm} \makecell{Corrected\\ Image} & \vspace{1.5mm} \includegraphics[width=30mm]{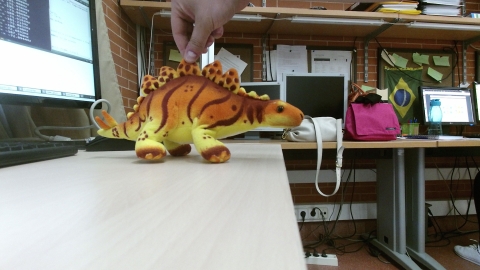} &\vspace{1.5mm} \includegraphics[width=30mm]{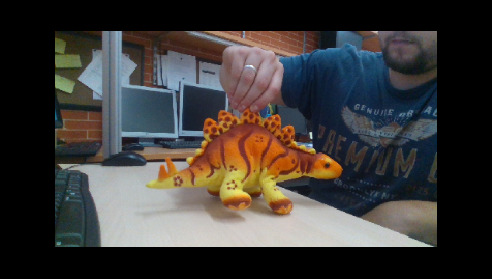} & \vspace{1.5mm} \includegraphics[width=30mm]{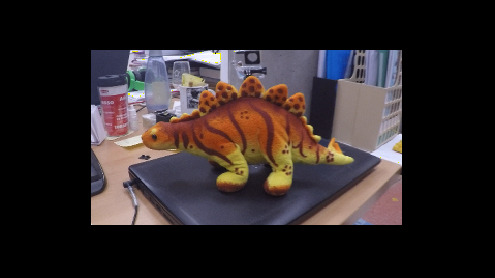}\\
			Reconstruction & \includegraphics[width=30mm]{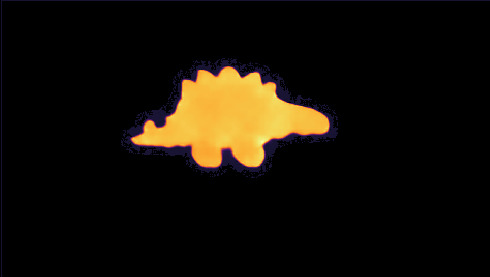} & \includegraphics[width=30mm]{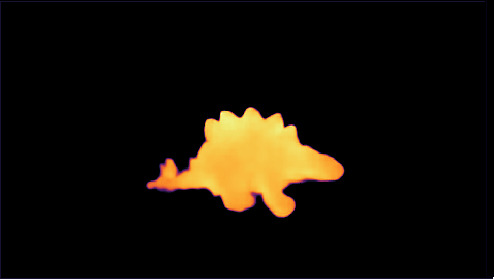} & \includegraphics[width=30mm]{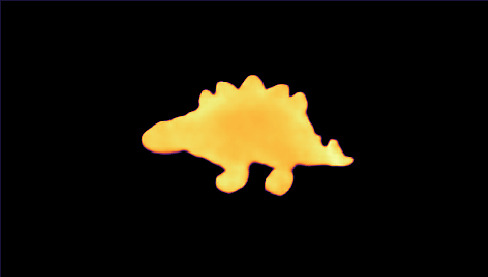}\\ 
			&\multicolumn{1}{r}{ RMSE 7.12 mm} &\multicolumn{1}{r}{ RMSE 12.34 mm} & \multicolumn{1}{r}{-} \\
			&\multicolumn{3}{l}{\textbf{0} \includegraphics[width=80mm,height=2mm]{colormap_inferno2} \textbf{500} mm}\\
			Registration (u) & \includegraphics[width=30mm]{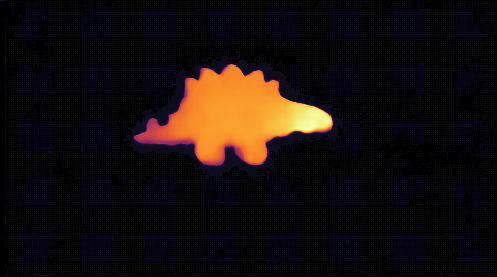} & \includegraphics[width=30mm]{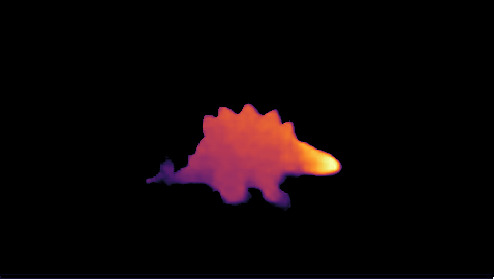} & \includegraphics[width=30mm]{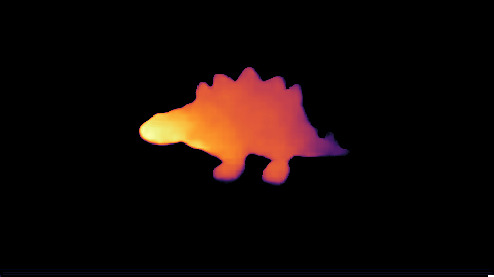}\\
			Registration (v) & \includegraphics[width=30mm]{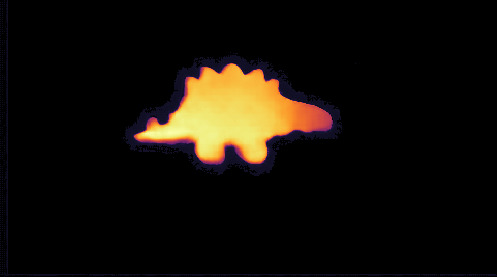} & \includegraphics[width=30mm]{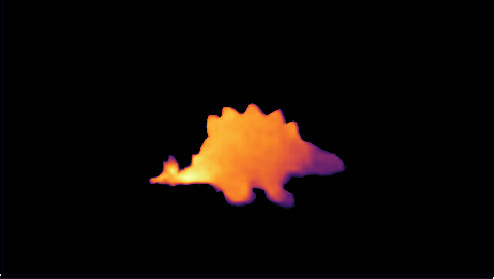} & \includegraphics[width=30mm]{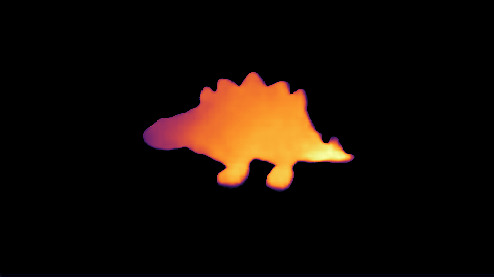}\\			
			&\multicolumn{3}{l}{\textbf{0} \includegraphics[width=80mm,height=2mm]{colormap_inferno2} \textbf{1} n.u.}\\

		\end{tabular} 
	\end{adjustbox}
	\centering \caption {Experimental results with different camera models. n.u. stands for normalised units in the template texture map.}
	\label{tb:diferent_cameras}
\end{table}

Using the technique described in \S\ref{sec:otherCams}, we test performance with three different real test cameras (Microsoft Kinect V2: same as for training, Intel Realsense D435 and Gopro Hero V3). Table \ref{tb:diferent_cameras} gives reconstruction errors with the Kinect and Realsense cameras. For the Gopro Hero V3 (an RGB camera) we show qualitative results. Quantitatively, the reconstruction errors with the Kinect and Realsense cameras are quite similar. This is an important point and clearly demonstrates the ability of DeepSfT to generalise well to images taken with a different test camera. Furthermore, DeepSfT copes with images from another camera even if the focal lengths are significantly different, as indicated qualitatively with the GoPro camera. We emphasize that this is the first time SfT has been solved with different train/test cameras with a DNN. This has a big practical benefit, because we are not limited to using the same camera at test time. 

\subsection{Evaluation of light and occlusion resistance}

We show that DeepSfT is resistant to light changes and significant occlusions in table \ref{tb:resistance}. The first two rows of the table show representative examples of scenes with external and self occlusions. DeepSfT is able to cope with them, accurately detecting the occlusion boundaries.
\begin{table}[!htbp]
	\begin{adjustbox}{max width=\linewidth}
		\begin{tabular}{m{1.8cm}m{2cm}m{2cm}m{2cm}m{2cm}m{2cm}}
			
			\multicolumn{6}{c}{\rule{0pt}{2.5ex} Occlusions} \\ \hline
			
			Dataset & \centering \vspace{1.52mm}DS1&\centering \vspace{1.52mm}DS4
			&\centering \vspace{1.52mm}DS4 &\centering \vspace{1.52mm}DS3 & \multicolumn{1}{m{2cm}}{\centering \vspace{1.52mm}DS3}\\ 
			
			Image & \centering \vspace{1.52mm}\includegraphics[width=20mm]{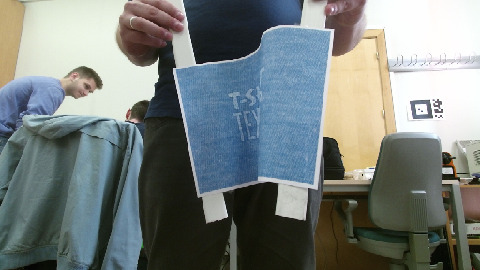} & \vspace{1.52mm}\includegraphics[width=20mm]{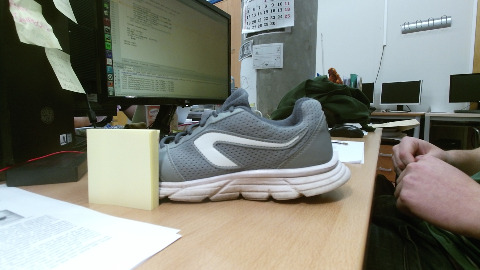}
			& \vspace{1.52mm}\includegraphics[width=20mm]{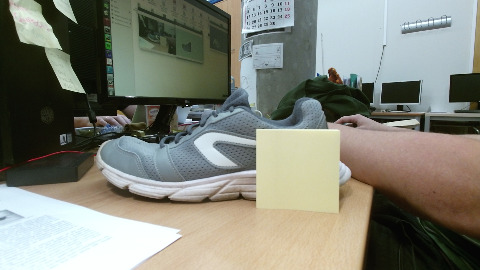}
			& \vspace{1.52mm}\includegraphics[width=20mm]{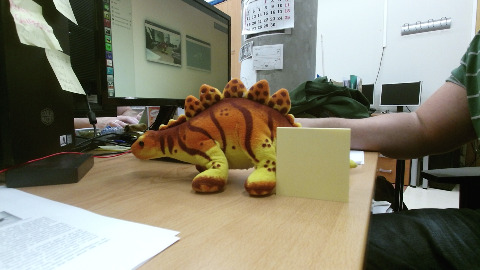}
			& \vspace{1.52mm}\includegraphics[width=20mm]{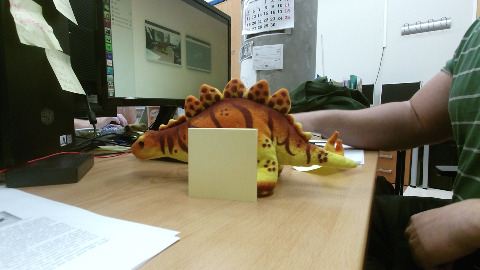}\\ 
			
			Reconstruction & \vspace{1.52mm}\includegraphics[width=20mm]{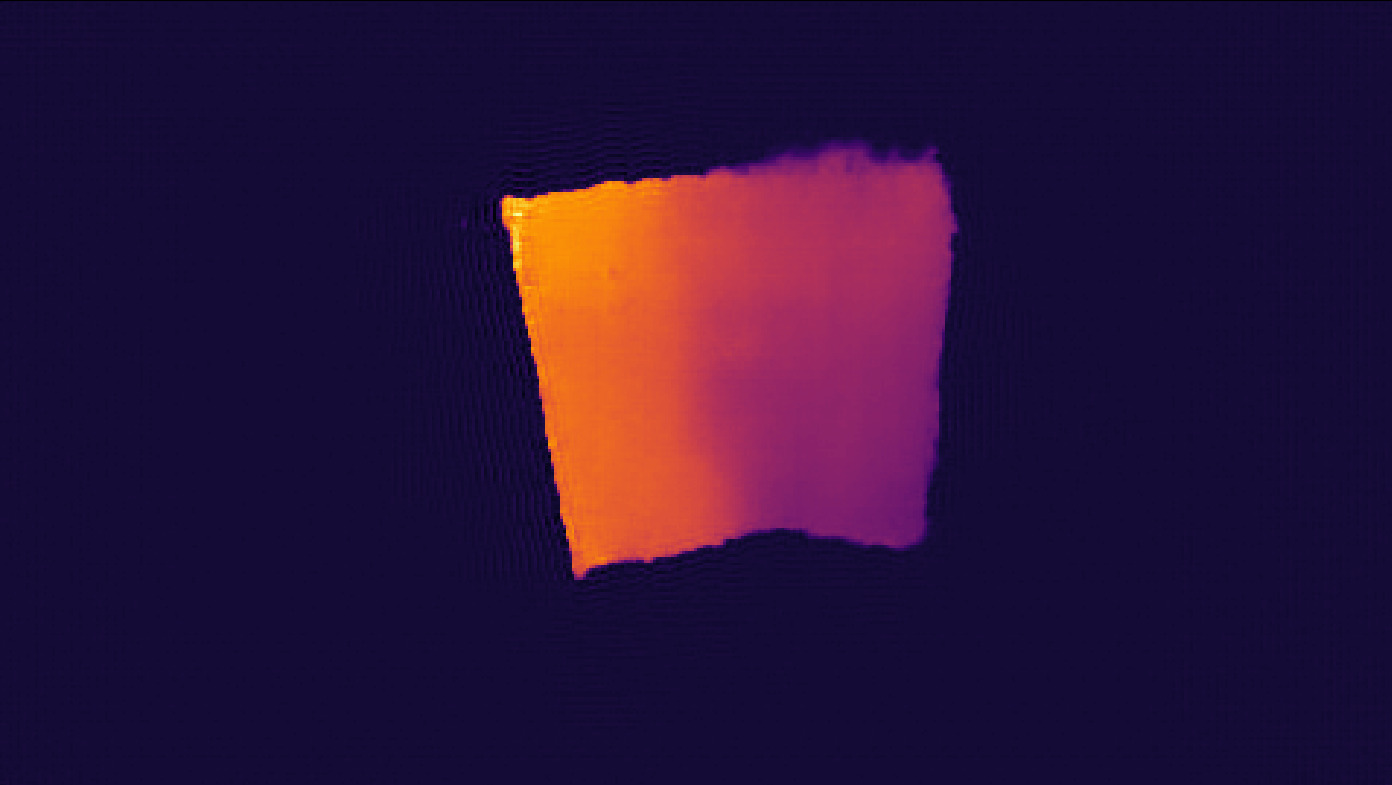} & \vspace{1.52mm}\includegraphics[width=20mm]{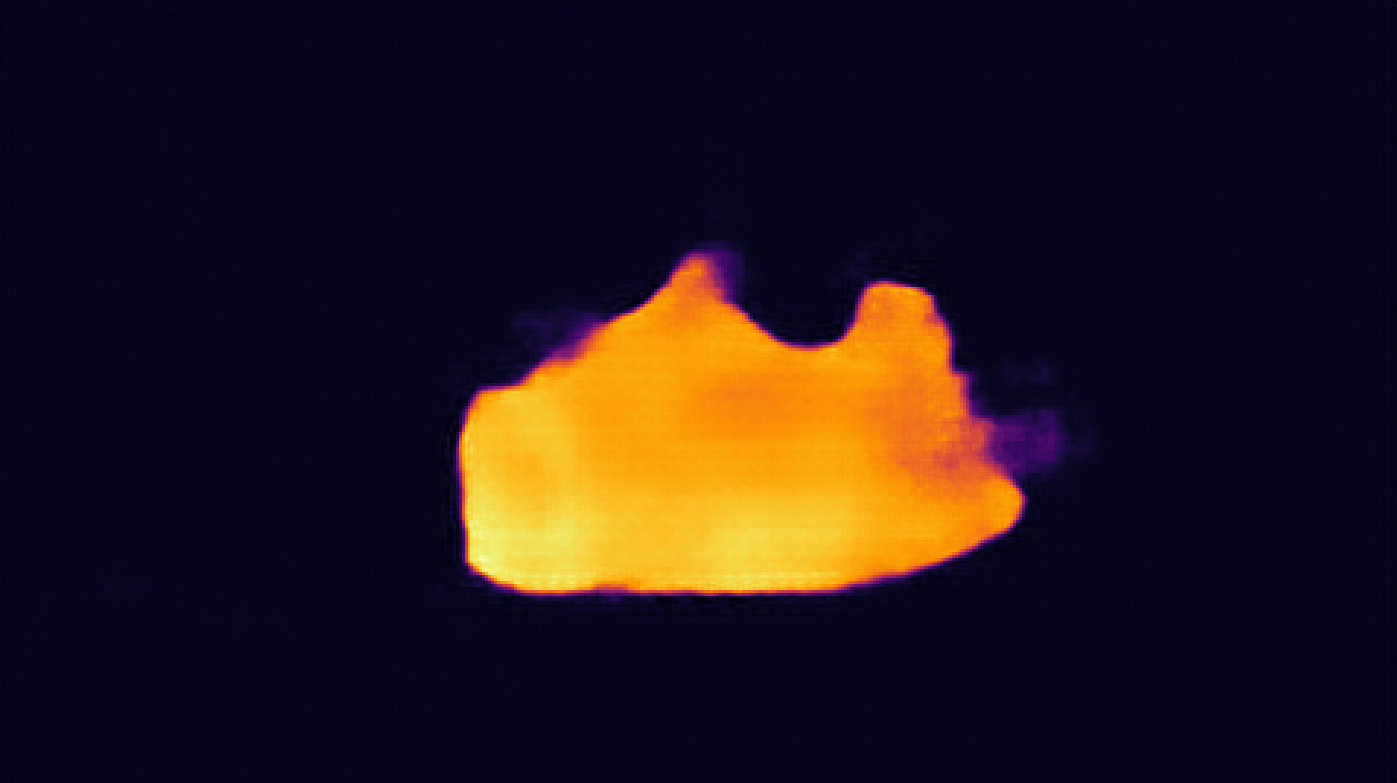}
			& \vspace{1.52mm}\includegraphics[width=20mm]{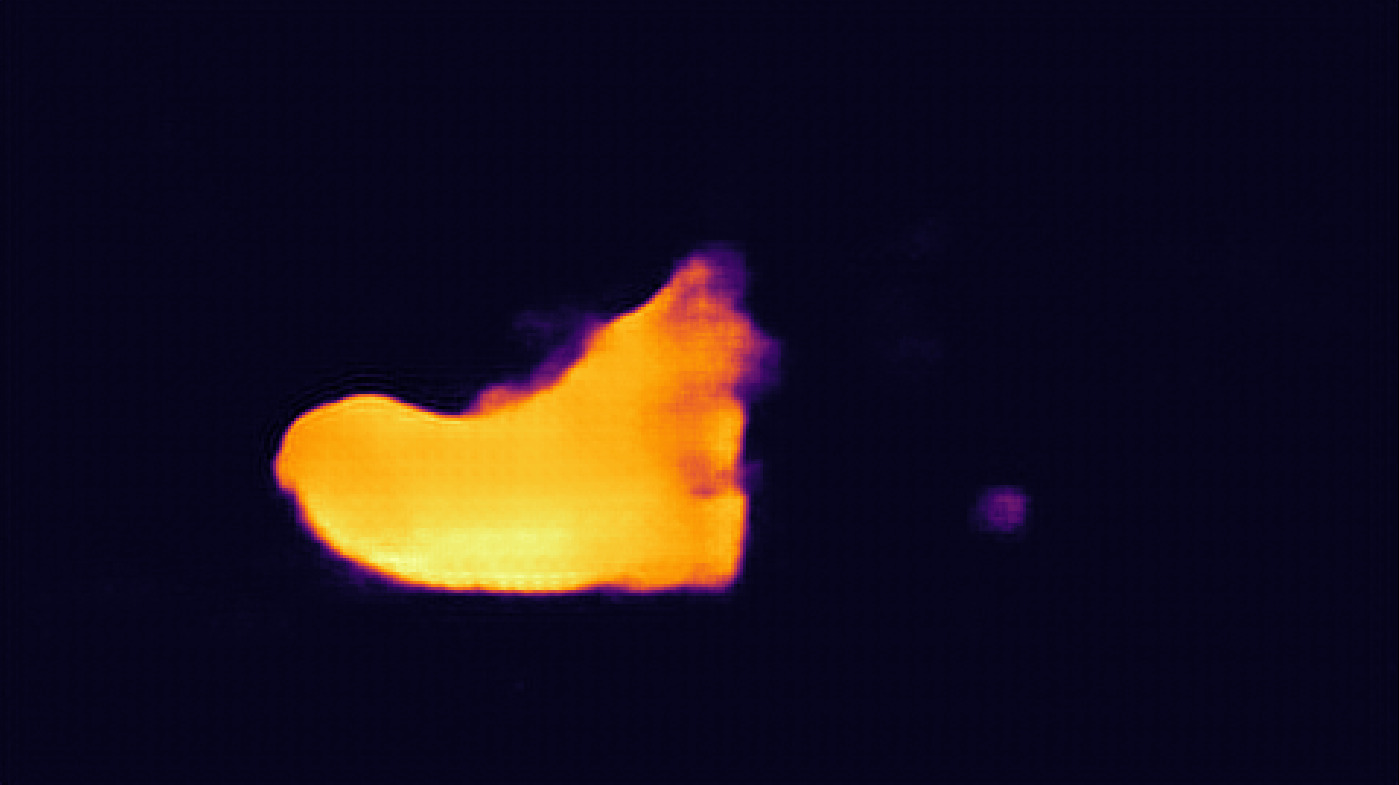} 
			& \vspace{1.52mm}\includegraphics[width=20mm]{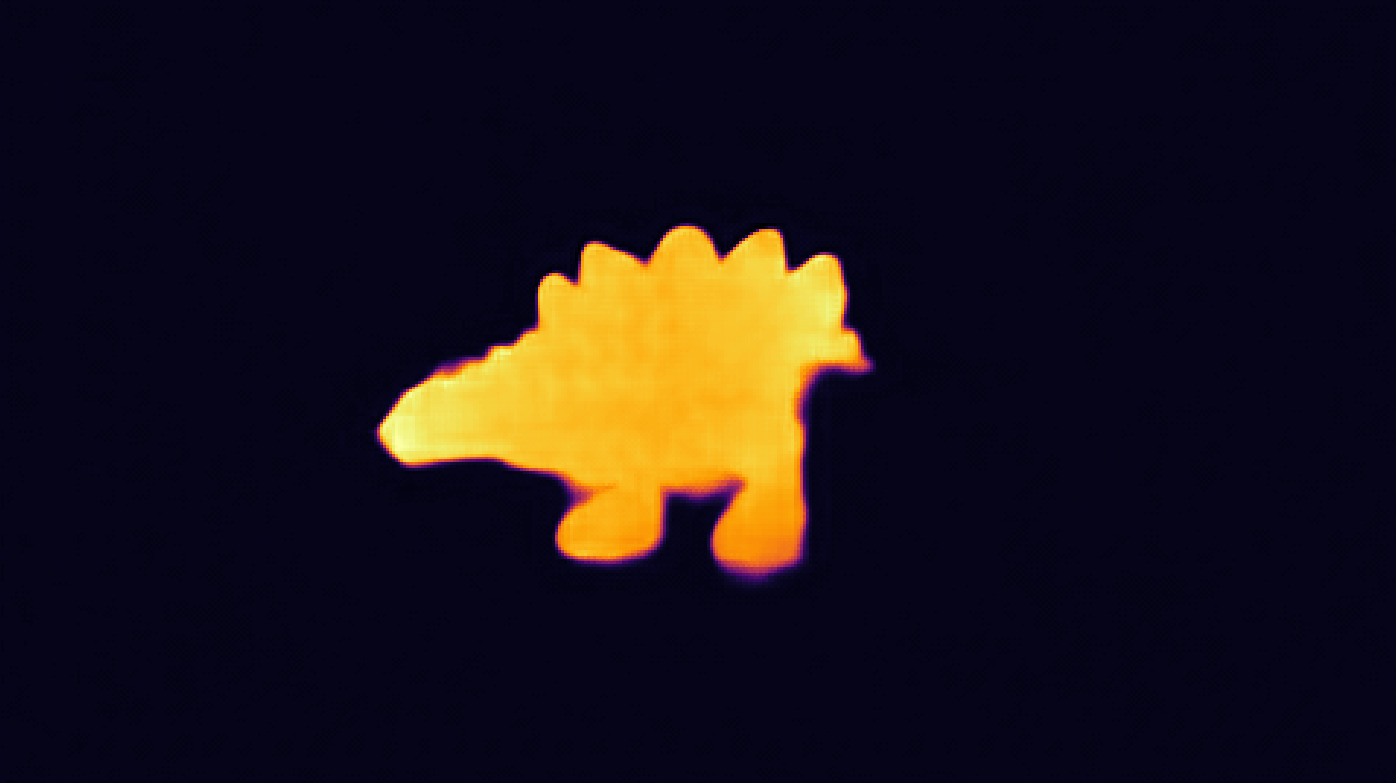}
			& \vspace{1.52mm}\includegraphics[width=20mm]{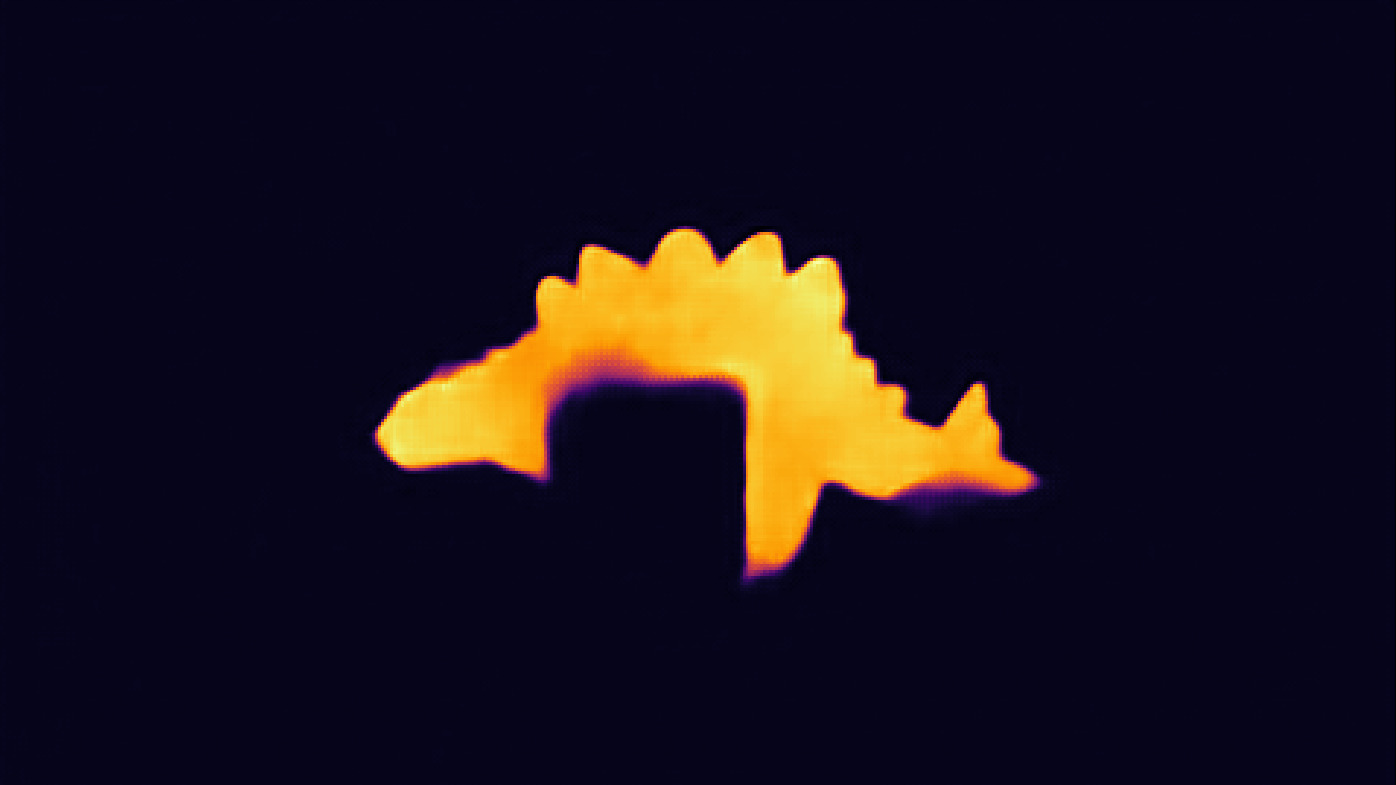}\\ 
			
			\multicolumn{6}{c}{\rule{0pt}{2.5ex} Illumination Changes} \\ \hline
			
			Dataset & \centering \vspace{1.52mm}DS1&\centering \vspace{1.52mm}DS4
			&\centering \vspace{1.52mm}DS4 &\centering \vspace{1.52mm}DS3 & \multicolumn{1}{m{2cm}}{\centering \vspace{1.52mm}DS3}\\ 
			
			Input & \centering \vspace{1.52mm}\includegraphics[width=20mm]{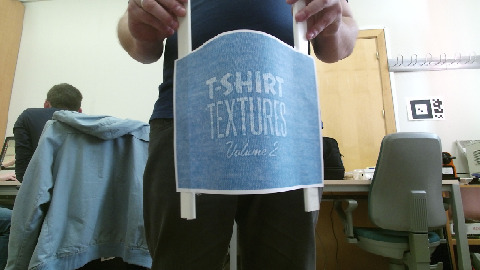} & \vspace{1.52mm}\includegraphics[width=20mm]{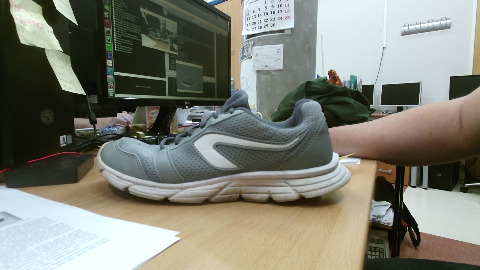}
			& \vspace{1.52mm}\includegraphics[width=20mm]{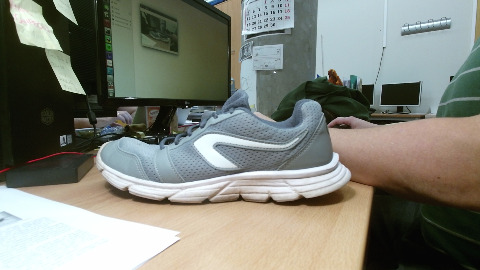}
			& \vspace{1.52mm}\includegraphics[width=20mm]{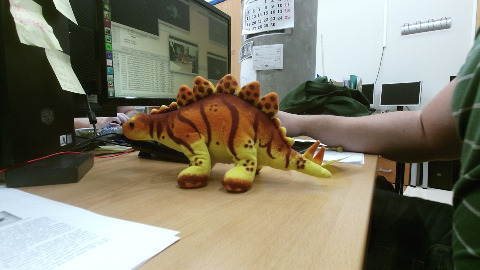}
			& \vspace{1.52mm}\includegraphics[width=20mm]{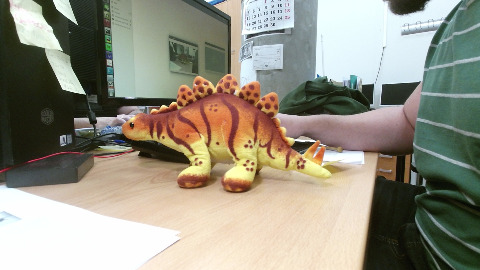}\\ 
			
			Reconstruction & \vspace{1.52mm}\includegraphics[width=20mm]{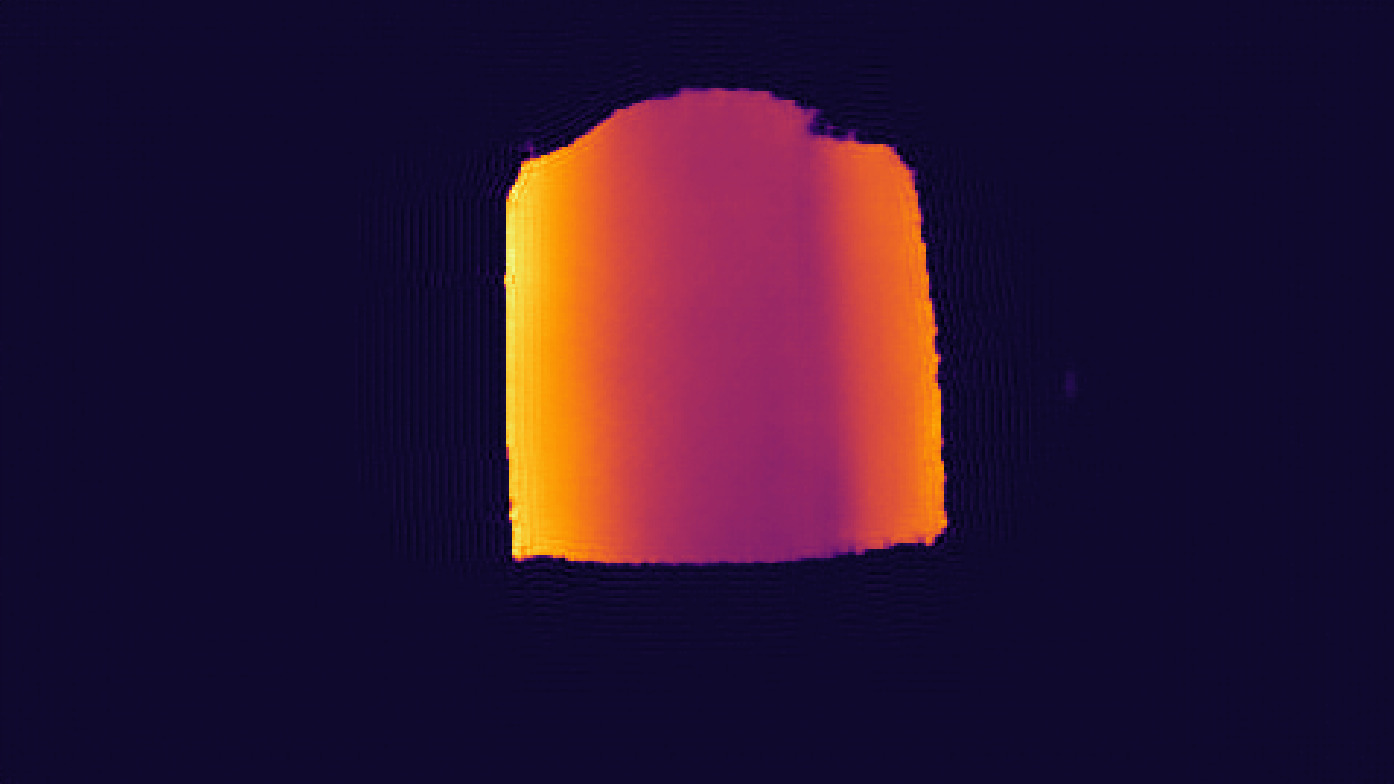} & \vspace{1.52mm}\includegraphics[width=20mm]{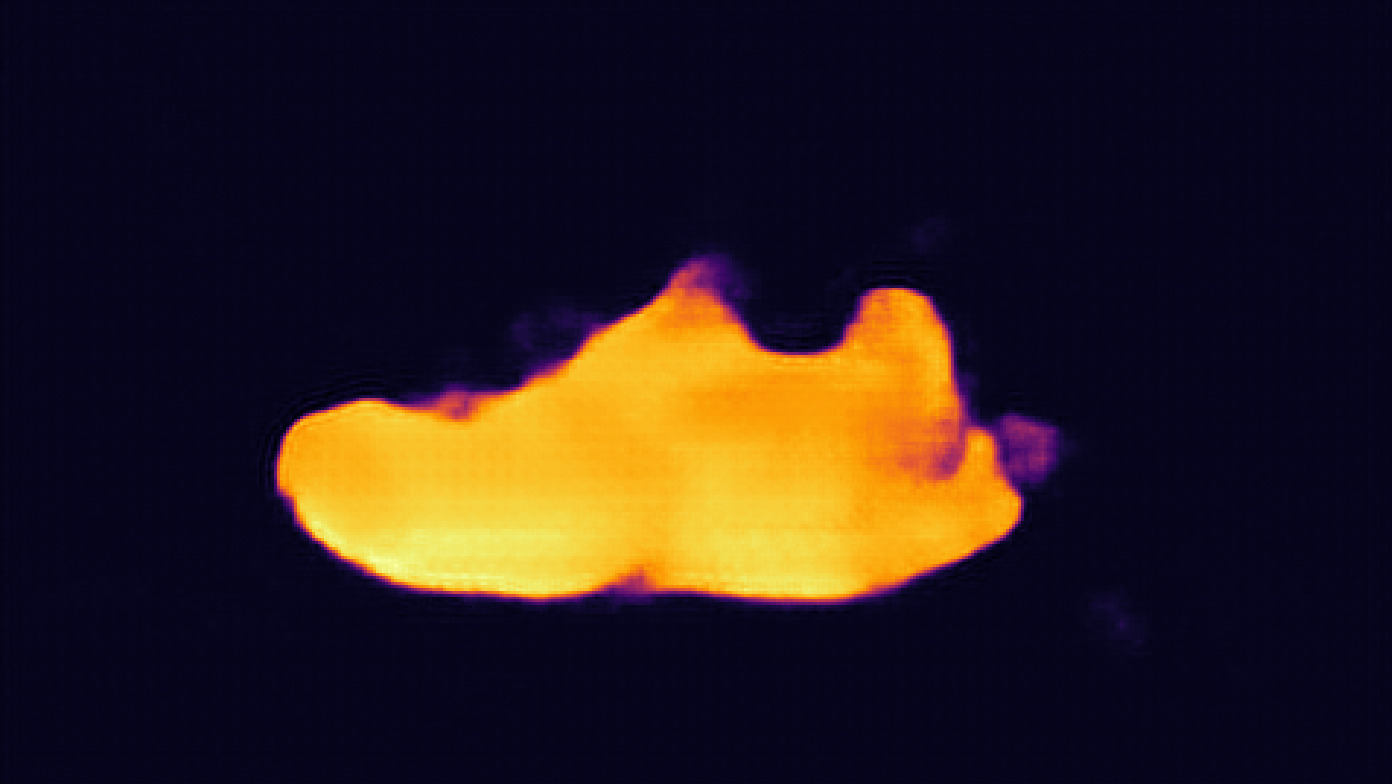}
			& \vspace{1.52mm}\includegraphics[width=20mm]{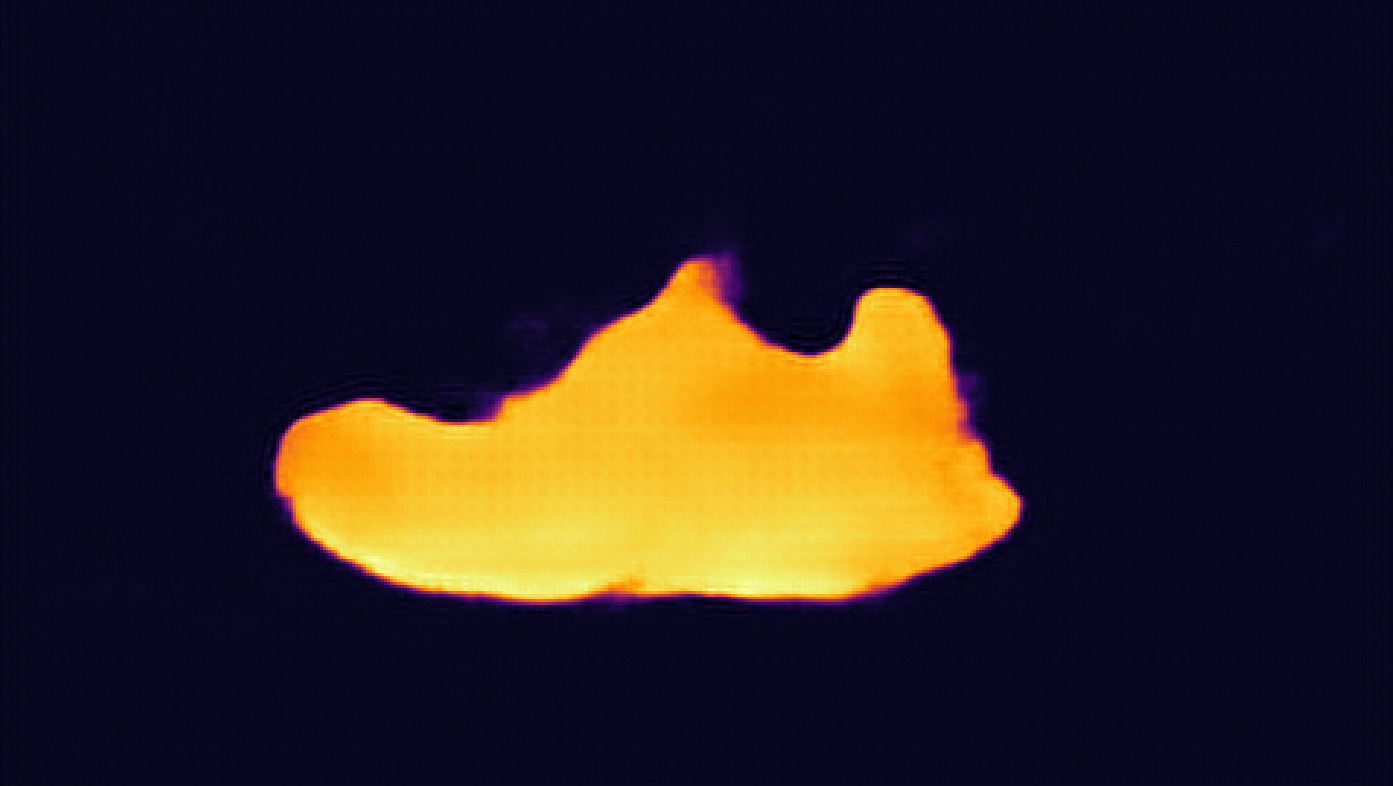} 
			& \vspace{1.52mm}\includegraphics[width=20mm]{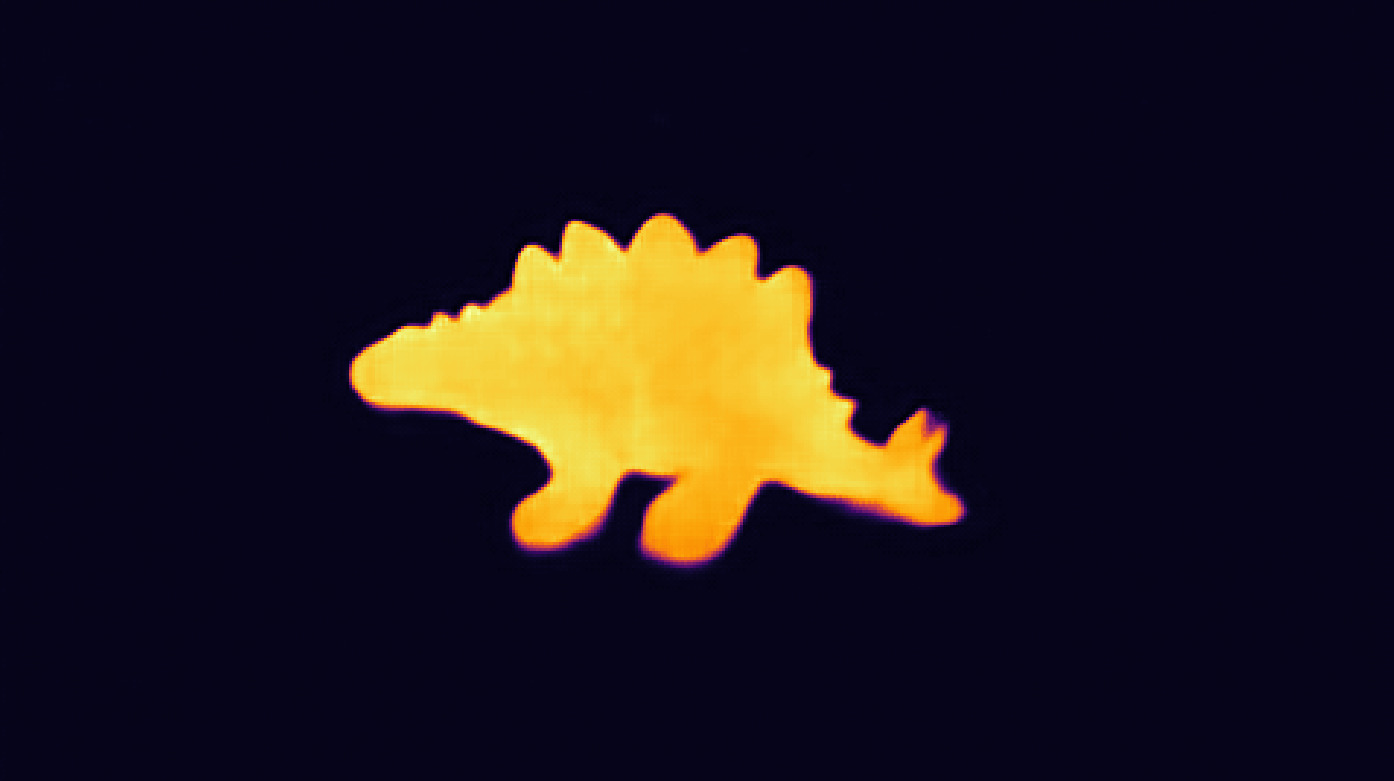}
			& \vspace{1.52mm}\includegraphics[width=20mm]{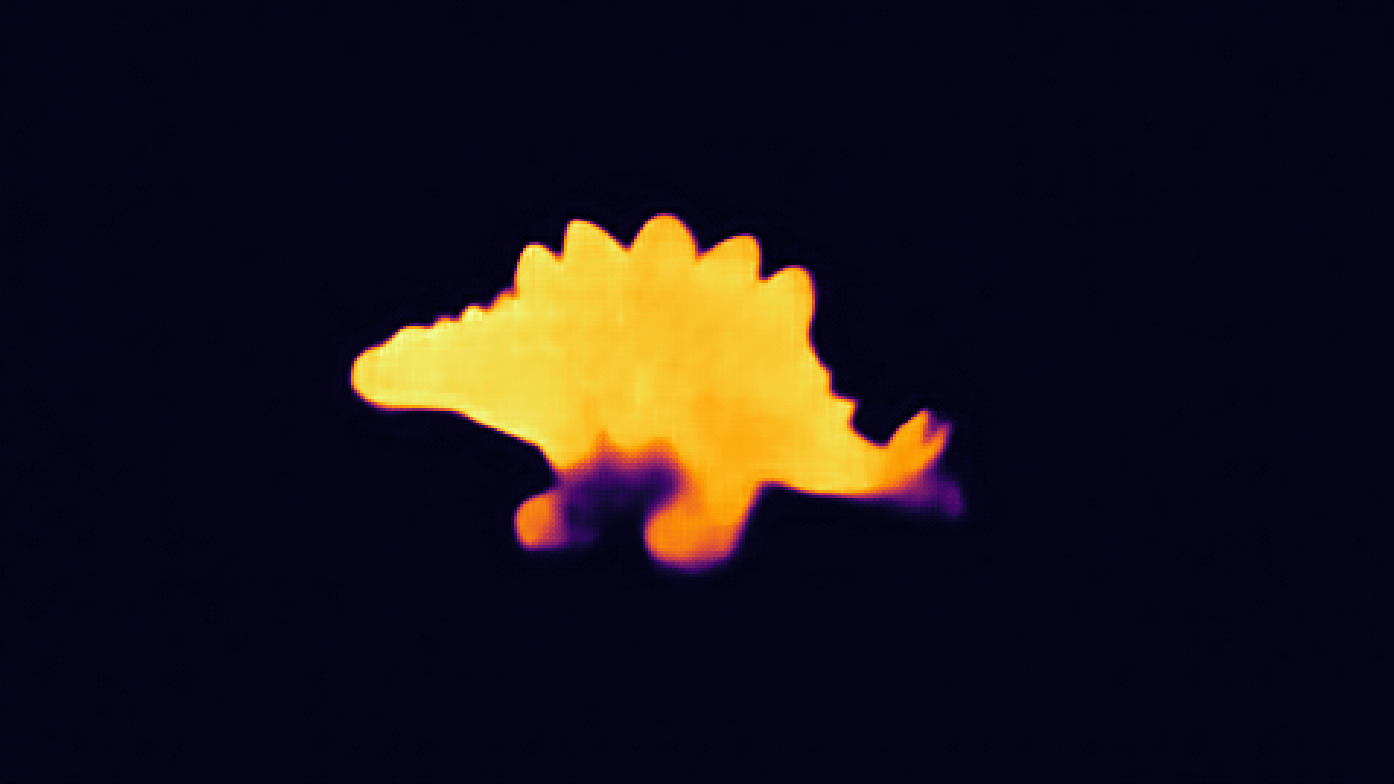}\\ 
			
			\multicolumn{6}{c}{\rule{0pt}{2.5ex} Failure Cases} \\ \hline
			
			Dataset & \centering \vspace{1.52mm}DS1&\centering \vspace{1.52mm}DS4
			&\centering \vspace{1.52mm}DS3 &\centering \vspace{1.52mm}DS3 & \multicolumn{1}{m{2cm}}{\centering \vspace{1.52mm}DS4}\\ 
			
			Input & \centering \vspace{1.52mm}\includegraphics[width=20mm]{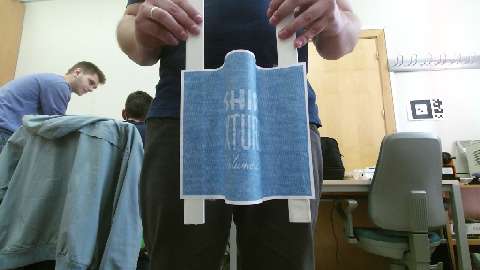} & \vspace{1.52mm}\includegraphics[width=20mm]{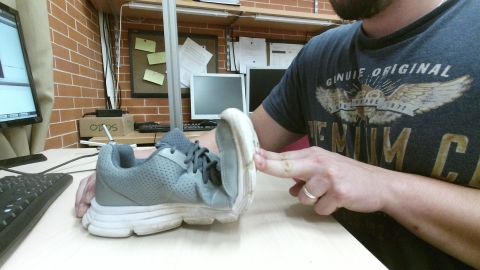}
			& \vspace{1.52mm}\includegraphics[width=20mm]{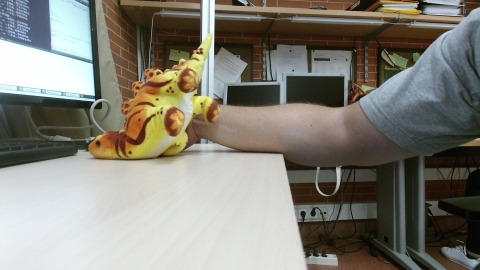}
			& \vspace{1.52mm}\includegraphics[width=20mm]{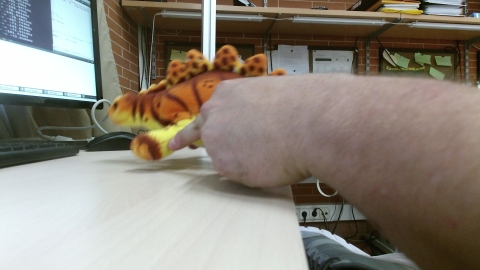}
			& \vspace{1.52mm}\includegraphics[width=20mm]{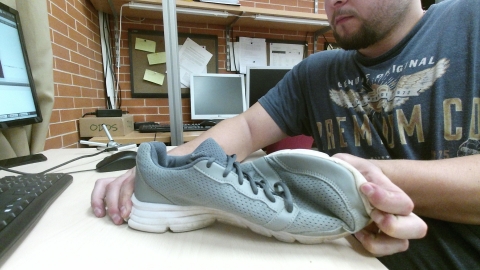}\\ 
			
			Region with information & \vspace{1.52mm}\includegraphics[width=20mm]{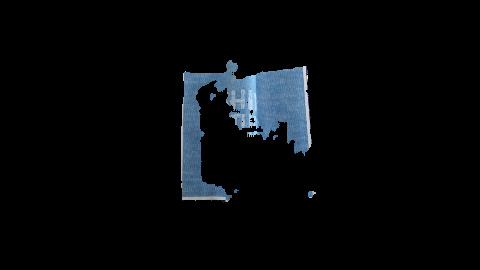} & \vspace{1.52mm}\includegraphics[width=20mm]{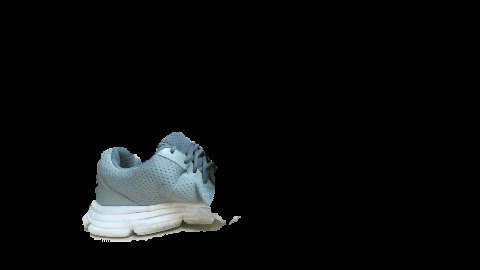}
			& \vspace{1.52mm}\includegraphics[width=20mm]{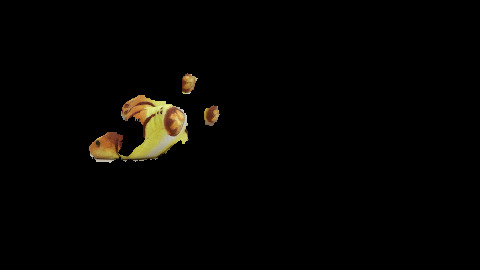} 
			& \vspace{1.52mm}\includegraphics[width=20mm]{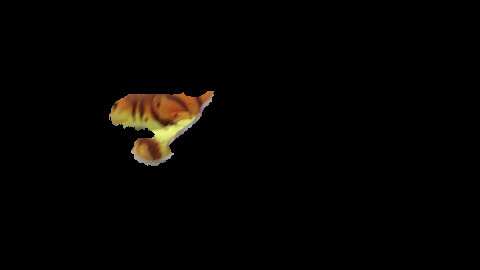} 
			& \vspace{1.52mm}\includegraphics[width=20mm]{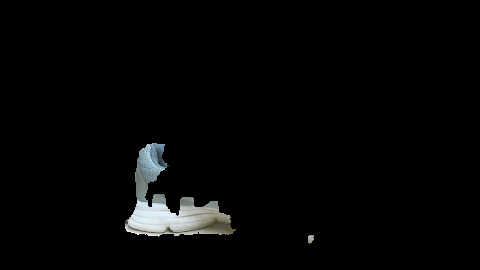}
			\\ 
			\multicolumn{1}{c}{\rule{0pt}{2.5ex}Colormap}&\multicolumn{5}{c}{\textbf{0} \includegraphics[width=80mm,height=2mm]{colormap_inferno2} \textbf{500} millimeters}
			\\ 
		\end{tabular}
	\end{adjustbox}
	\centering \caption {Representative occlusion resistance, light change resistance and failure cases.}
	\label{tb:resistance}
\end{table}
The third and fourth rows show examples of scenes with illumination that produce significant shading variations. DeepSfT shows good resistance to these variations.

\subsection{Failure modes}
There are some instances where DeepSfT fails, shown in the final two rows of table~\ref{tb:resistance}. There are general failure modes of SfT (very strong occlusions and illumination changes), for which all methods will fail at some point. There are also failure modes specific to learning-based approaches (excessive deformations that are not represented in the training set). However, recall that wide-baseline methods like DeepSft can recover easily from failures with video inputs because they process each image independently, unlike short-baseline methods. Therefore failure for some frames in a video does not prevent successful reconstruction and registration in the later frames.

\subsection{Ablation studies}

\subsubsection{The benefit of Total Variation regularisation}

We included total variation smoothness during fine tunning of the Depth Refinement Block and the Registration Refinement Block. In the Depth Refinement Block, the main objective of this term is to alleviate the effect of noise and outliers in depth data used as ground-truth in equation~\eqref{eq:depthloss}. In the Registration Refinement Block, it is used to improve convergence of the self-supervised algorithm, based on minimising the photometric error in equation~\eqref{eq:photoref}. We investigate the effect of this term in both depth and registration accuracy, when testing with real data. We show in table \ref{tb:total_variation} the quantitative results obtained with all the templates DS1, DS2, DS3, DS4 and DS5. As can be seen, the Total Variation regularisation improves the reconstruction and registration errors in all the cases, especially for the DS2 template. 

\subsubsection{The benefit of depth refinement}

We evaluate the influence of the Depth Refinement Block and its results in terms of depth RMSE. We show these errors in table \ref{tb:depthref} where the RMSE obtained using only the Main Block of DeepSfT is compared to the RMSE obtained by the Depth Refinement Block. Recall that the Main Block has been trained exclusively using synthetic data whereas the Depth Refinement Block has been fine-tuned with real data. It can be clearly seen that the Depth Refinement Block RMSE is much lower compared to the Main Block RMSE. Recall that the Main Block provides a first approximation of the depth map but, due to the render gap between synthetic and real data, this approximation is not highly accurate. The Refinement Block refines this approximation. This agrees with the widely held view that refining a network with real data can significantly reduce the render gap, and improve generalisation \cite{8851791}. 

It is important to highlight that the results indicate an error increase according to the template complexity. For the rectangular templates, like DS1 and DS2, there is less of an error gap with and without the Depth Refinement Block. This is likely because these objects are the less difficult to represent and easier for the network to generalise, because their intrinsic deformation space is smaller compared with the volumic objects that deform in more complex ways. The RMSE gap for the volumic templates is large, and the benefit of the Depth Refinement Block is very evident in these cases.

\begin{table}[!htbp]
	\setcellgapes{3pt}\makegapedcells
	\begin{adjustbox}{max width=\linewidth}
		\begin{tabular}{|c|c|c|}
			\hline
			
			Sequence &DeepSfT (Main Block only) RMSE  &DeepSfT RMSE \\ \hline
			
			DS3R & 70.55 & 8.12 \\ \hline
			
			DS4R & 85.58 & 6.86  \\ \hline
			
			DS2R & 14.43 & 1.53 \\ \hline
			
			DS1R & 17.20 & 2.32  \\ \hline

		\end{tabular} 
	\end{adjustbox}
	\centering \caption {Results and comparison of DeepSfT with and without Depth Refinement Block. All the errors are expressed in mm.}
	\label{tb:depthref}
\end{table}
\subsubsection{The benefit of registration refinement}

We evaluate the impact of the Registration Refinement Block in terms of registration accuracy. Given that we lack registration ground-truth with real data we use photometric error as a proxy, computed as follows. We compute the Mean Square Error between the rendered images $I'_i$ and the input image $I_i$ in the visible region $\mathcal{I}_{i}$:
\begin{equation}
\mathcal{E}_{pr} = \left(\frac{1}{|\mathcal{I}_i|}\sum_{(u,v)\in \mathcal{I}_i} \left( I'_i(u,v) - I_i(u,v) \right)^2 \right)^\frac{1}{2}.
\label{eq:photoerrors}
\end{equation}
We show in table \ref{tb:photo_error} the photometric error and qualitative results when using the output of the Main Block, and when using the Registration Refinement Block. In terms of photometric error, the Registration Refinement Block output has less error than the Main Block output, which we recall was trained using only synthetic data. The templates with more texture features like DS5, DS2, and DS3 show qualitatively more improvement than DS1 and DS4, which have less texture. tables \ref{tb:photo_error} and \ref{tb:total_variation} show quantitative and qualitative results before and after registration refinement. 

\begin{figure}[!htbp]
	\centering
	\includegraphics[width=\linewidth]{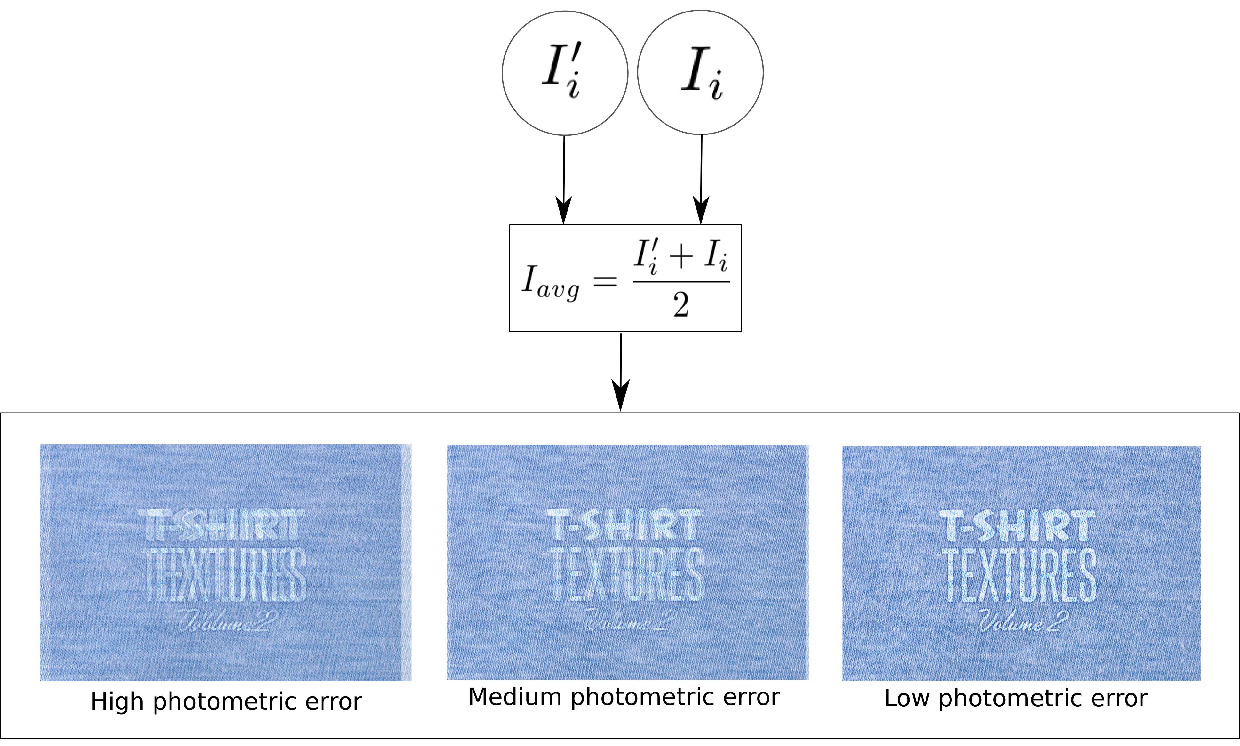}
	\caption{ Visualization of registration accuracy using image blending.}
	\label{fig:avg_image}
\end{figure}

We also give a qualitative visualization of the registration error, computed by blending the input image $I$ and the rendered image $I'$,
computed from the DeepSfT registration:
\begin{equation}
\begin{split}
I_{avg} =\frac{I'_i + I_i}{2}.
\end{split}
\label{eq:avgimg}
\end{equation}
A sharper $I_{avg}$ indicates a better registration. We show this visualization in figure \ref{fig:avg_image}, where the greater the photometric error, the worse the accuracy of the registration, and the more blurred the average image. In table \ref{tb:photo_error} we show the registration error visualization zoomed in the region of interest to provide a better visualization. We can clearly see a strong registration improvement with DS5, DS2 and DS4, with a smaller improvement for DS1 and no clear improvement with DS3. 

\begin{table*}[!htbp]
	\begin{adjustbox}{max width=\linewidth}
		\begin{tabular}{c m{5cm} m{5cm} m{5cm} m{5cm} m{5cm}}
			
			\rule{0pt}{2ex}   \large{Dataset}&\rule{0pt}{2ex} \large{Input Image}&\rule{0pt}{2ex}\large Average Image&\rule{0pt}{2ex}\large  Average Image&\rule{0pt}{2ex}\large Zoomed average Image&\rule{0pt}{2ex}\large Zoomed average Image\\
			
			&&\rule{0pt}{2ex}\large without registration&\rule{0pt}{2ex}  \large with registration refinement&\rule{0pt}{2ex}\large  without registration &\rule{0pt}{2ex}\large with registration refinement\\
			
			&&\rule{0pt}{2ex}\large refinement&&\rule{0pt}{2ex}\large  refinement&\\ \hline
			
			\rule{0pt}{2ex}   \centering {\large DS5}&\vspace{1.52mm}\includegraphics[width=50mm]{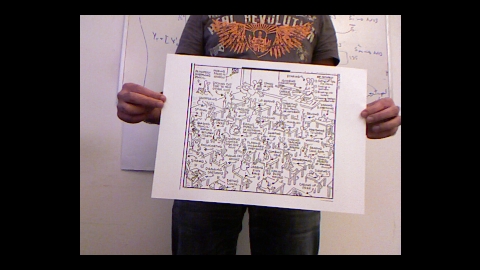}
			& \vspace{1.52mm}\includegraphics[width=50mm]{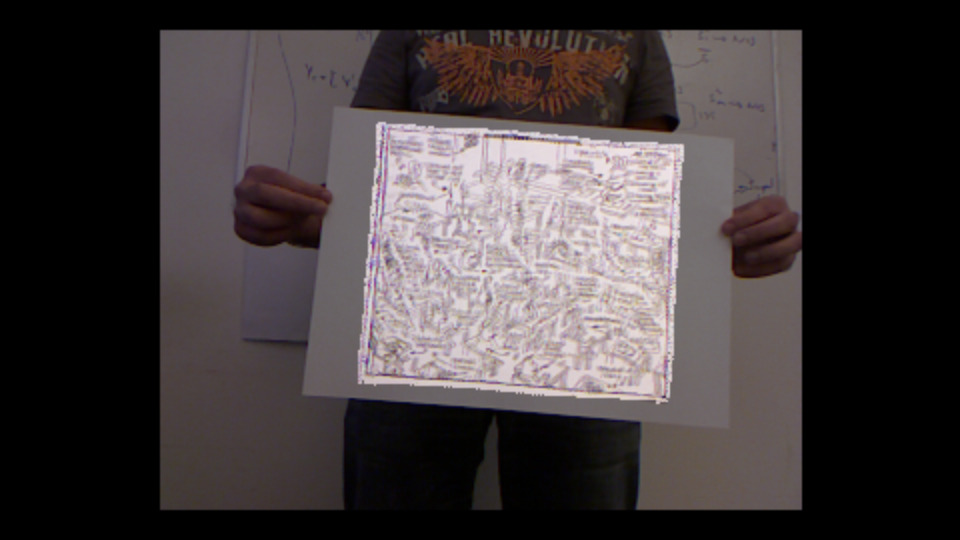}  & \vspace{1.52mm}\includegraphics[width=50mm]{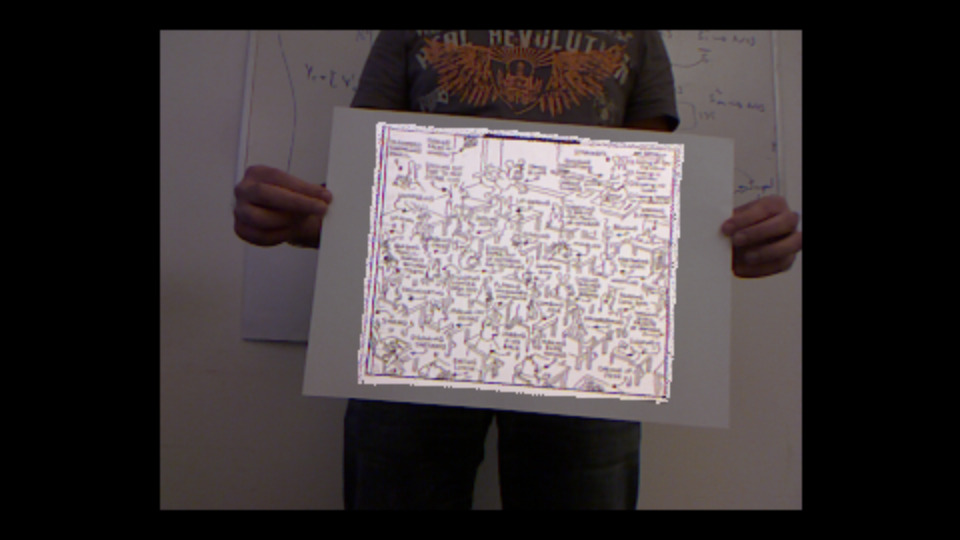}& \vspace{1.52mm}\includegraphics[width=50mm]{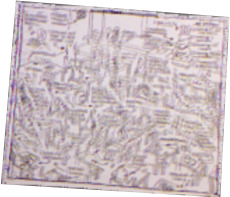}& \vspace{1.52mm}\includegraphics[width=50mm]{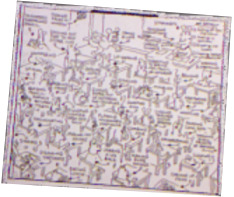}  \\ 
			
			\rule{0pt}{2ex}   \centering $\mathcal{E}_{pr}$&
			& \multicolumn{1}{r}{ \large{0.380}}& \multicolumn{1}{r}{\textbf{\large{0.345}}}&&   \\ 
			
			\rule{0pt}{2ex}   \centering{\large DS1}&\vspace{1.52mm}\includegraphics[width=50mm]{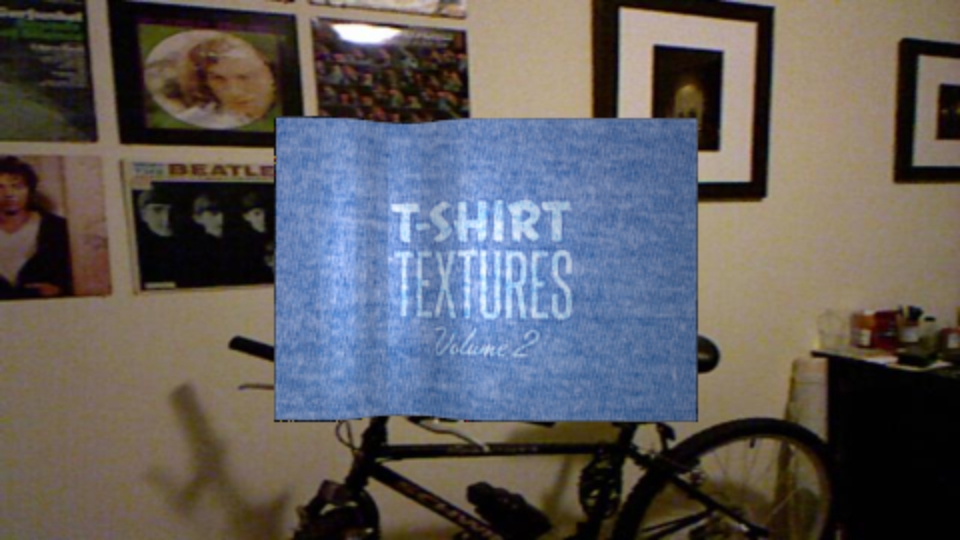}
			& \vspace{1.52mm}\includegraphics[width=50mm]{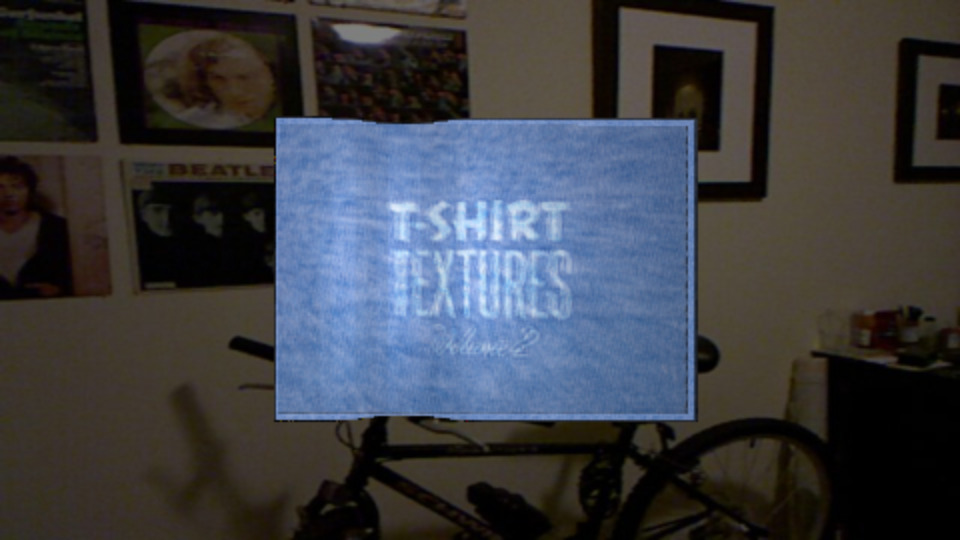}  & \vspace{1.52mm}\includegraphics[width=50mm]{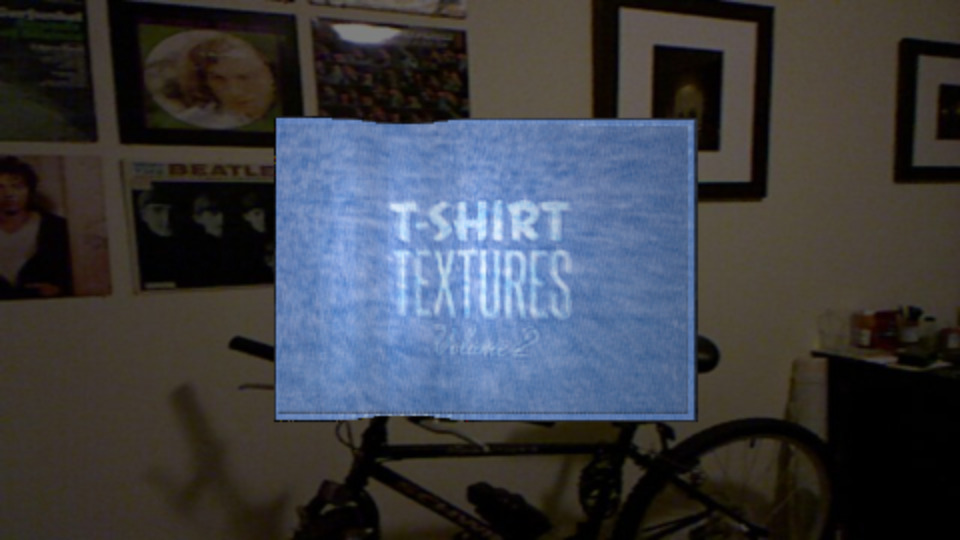}& \vspace{1.52mm}\includegraphics[width=50mm]{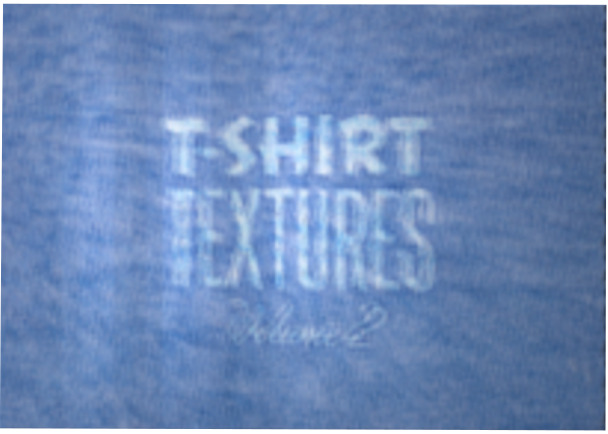}& \vspace{1.52mm}\includegraphics[width=50mm]{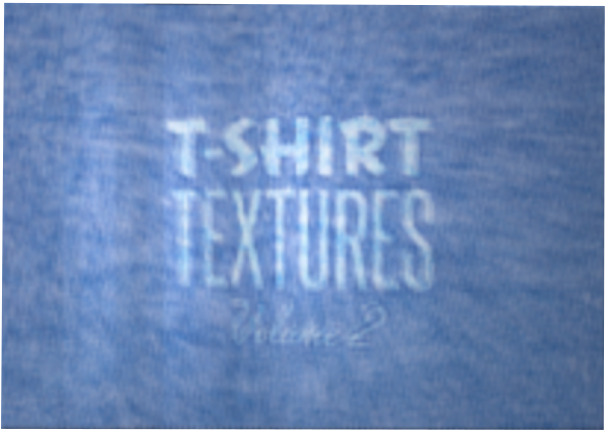}  \\ 
			
			\rule{0pt}{2ex}   \centering $\mathcal{E}_{pr}$&
			& \multicolumn{1}{r}{ \large{0.309}}& \multicolumn{1}{r}{\textbf{\large{0.250}}}&&  \\ 
			
			{\large DS3}&\vspace{1.52mm}\includegraphics[width=50mm]{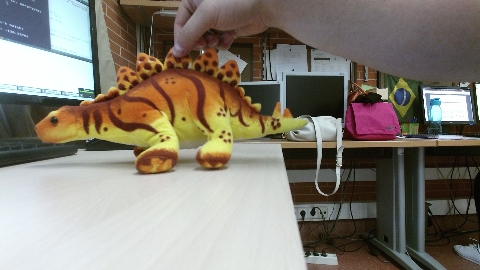}
			& \vspace{1.52mm}\includegraphics[width=50mm]{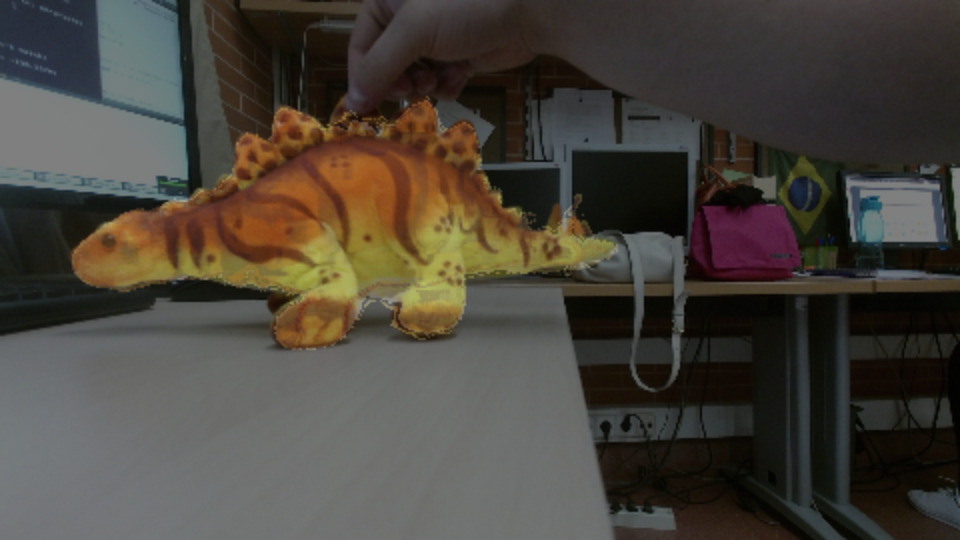} &  \vspace{1.52mm}\includegraphics[width=50mm]{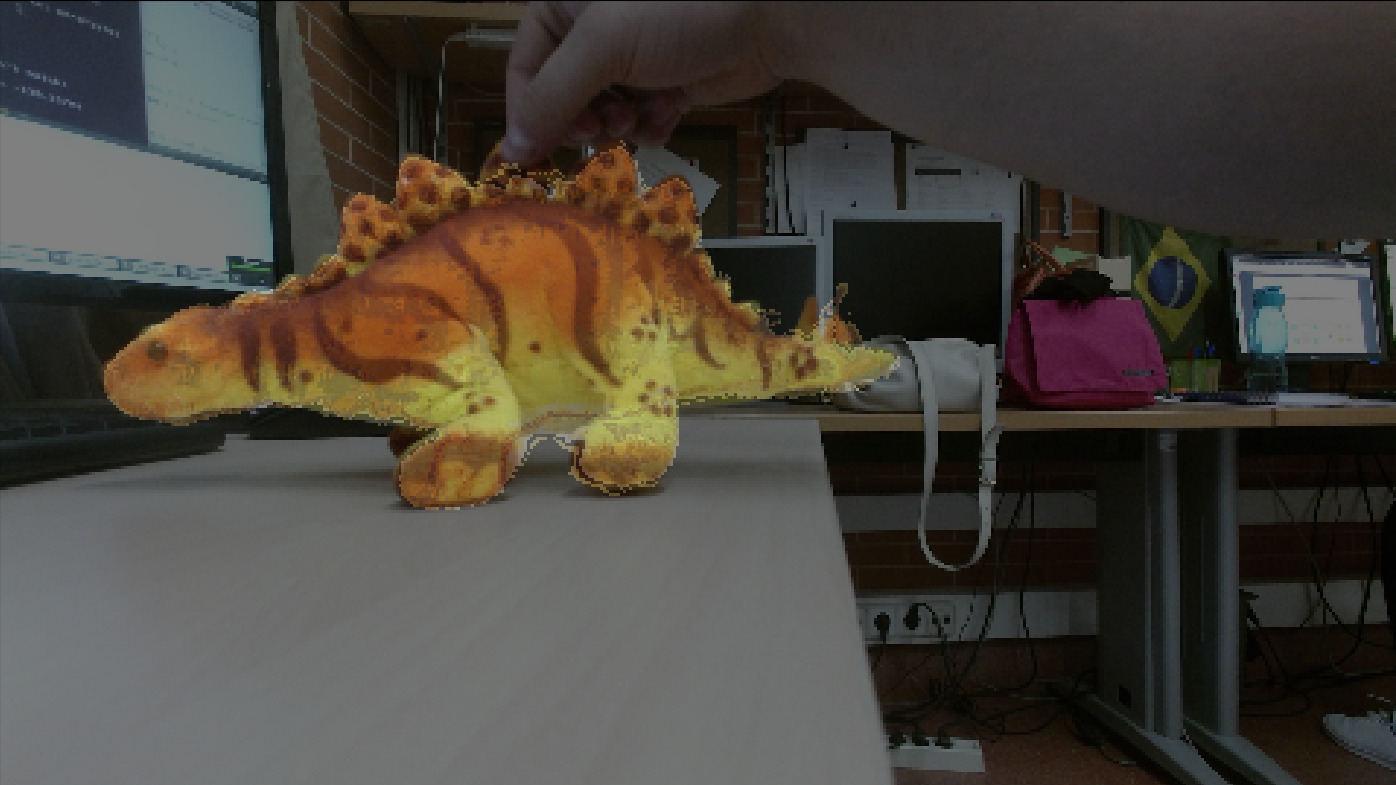}&  \vspace{1.52mm}\includegraphics[width=50mm]{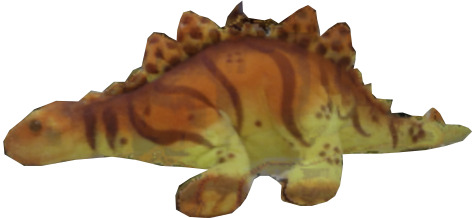}& \vspace{1.52mm}\includegraphics[width=50mm]{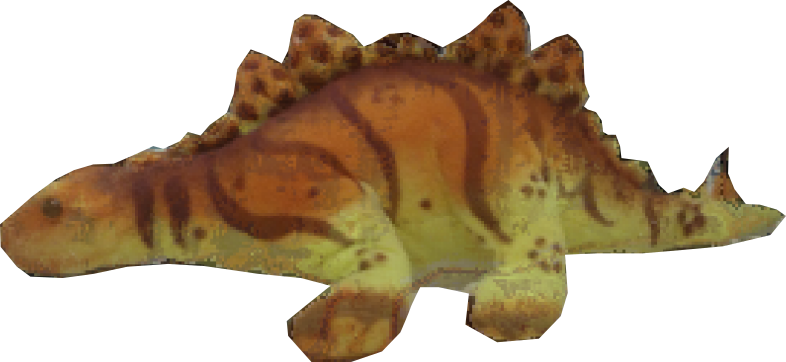} \\ 
			
			\rule{0pt}{2ex}   \centering $\mathcal{E}_{pr}$&
			& \multicolumn{1}{r}{ \large{0.133}}& \multicolumn{1}{r}{\textbf{\large{0.101}}}&&   \\ 
			
			\rule{0pt}{2ex}   \centering  {\large DS2}&\vspace{1.52mm}\includegraphics[width=50mm]{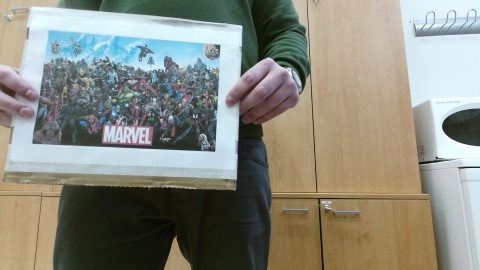}
			& \vspace{1.52mm}\includegraphics[width=50mm]{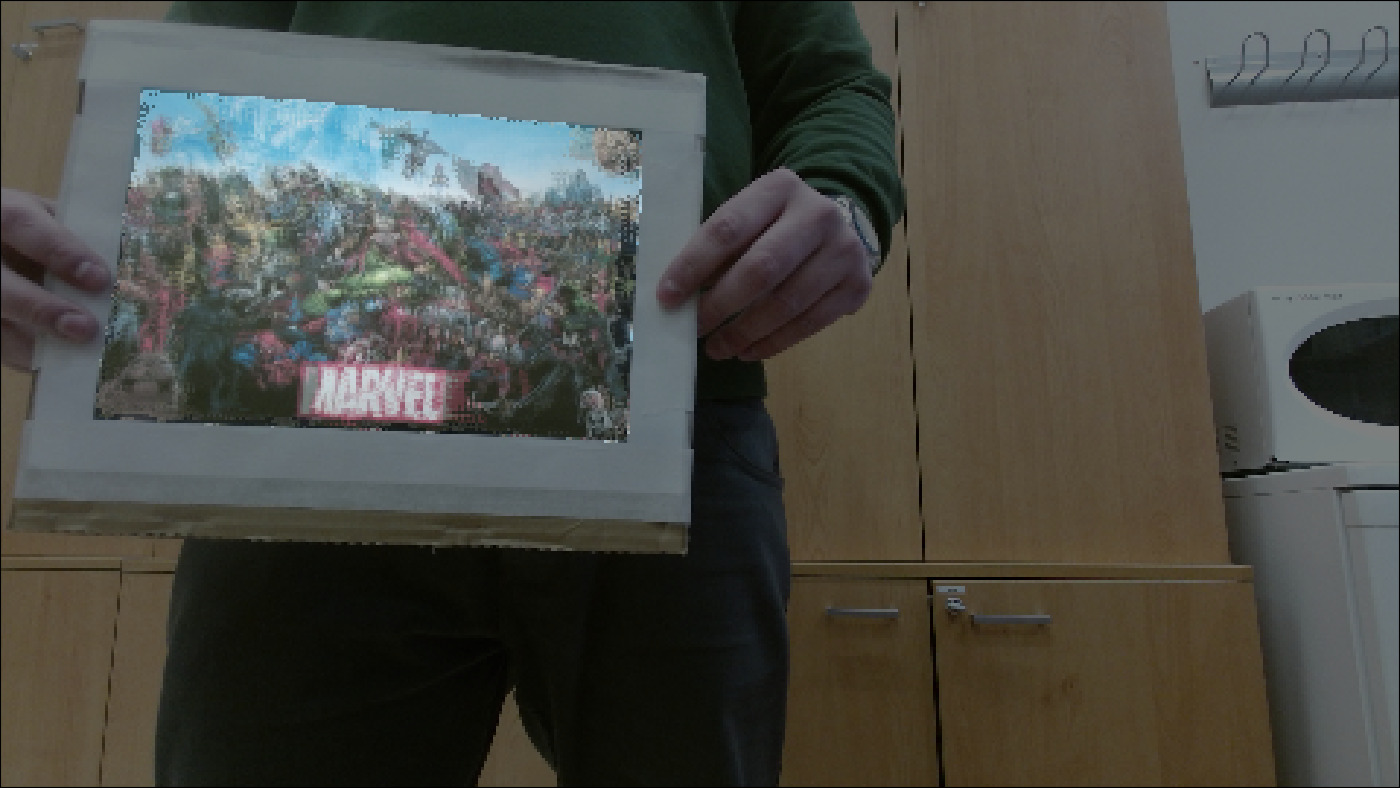} &  \vspace{1.52mm}\includegraphics[width=50mm]{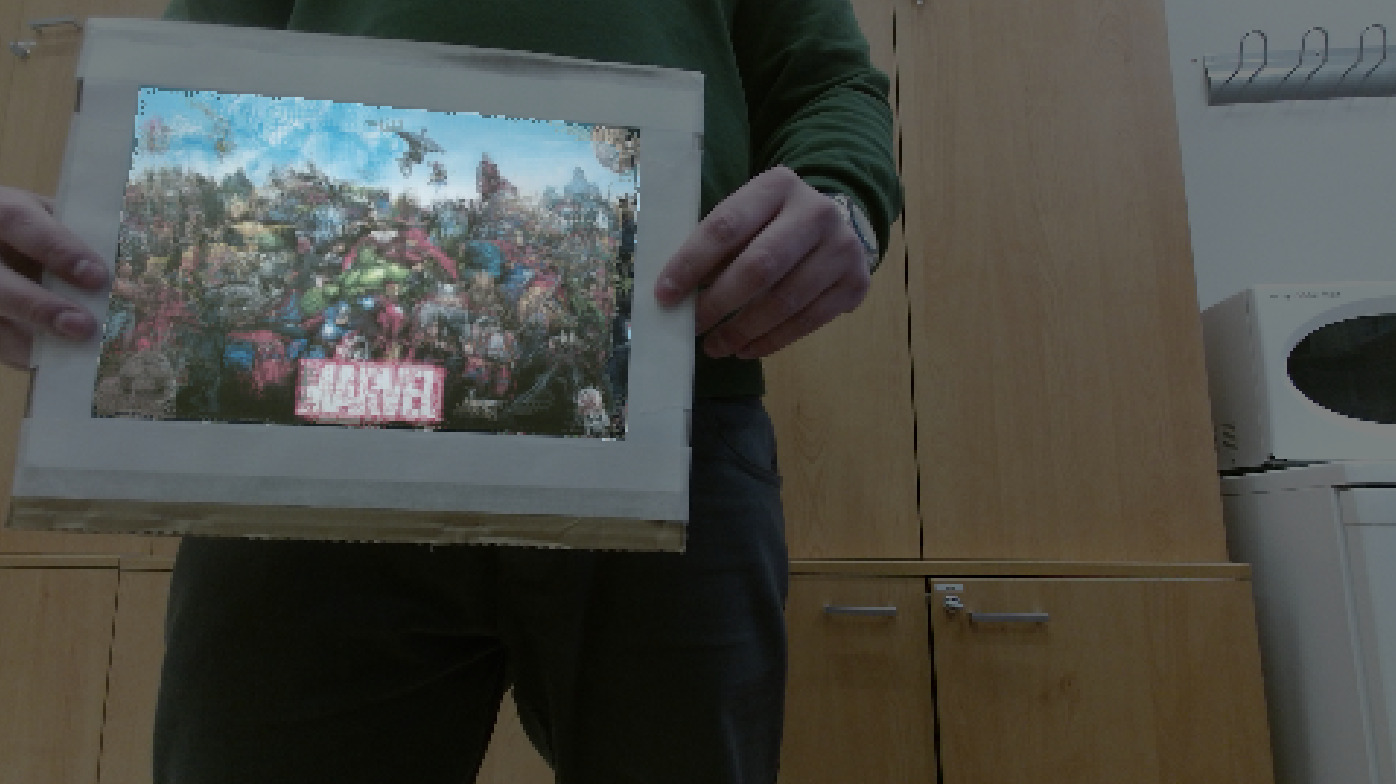}& \vspace{1.52mm}\includegraphics[width=50mm]{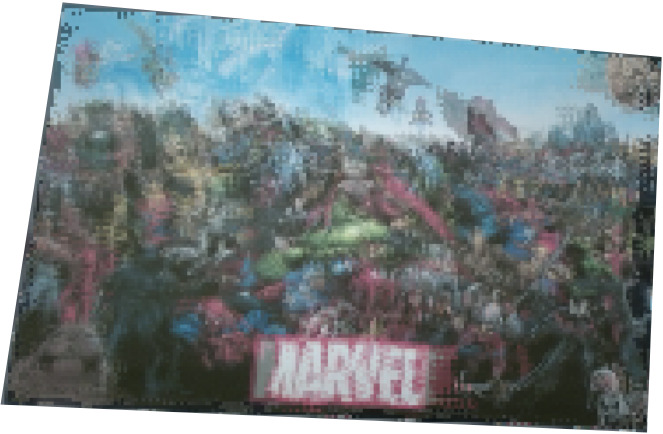}& \vspace{1.52mm}\includegraphics[width=50mm]{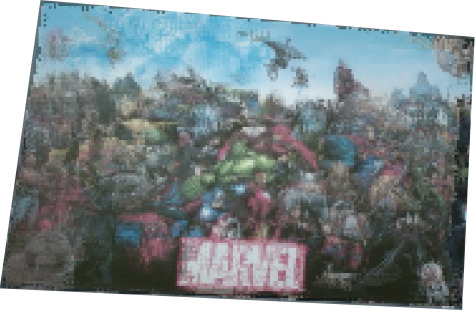}  \\ 
			
			\rule{0pt}{2ex}   \centering $\mathcal{E}_{pr}$&
			& \multicolumn{1}{r}{ \large{0.090}}& \multicolumn{1}{r}{\textbf{\large{0.015}}}&&   \\ 
			
			\rule{0pt}{2ex}   \centering {\large DS4}&\vspace{1.52mm}\includegraphics[width=50mm]{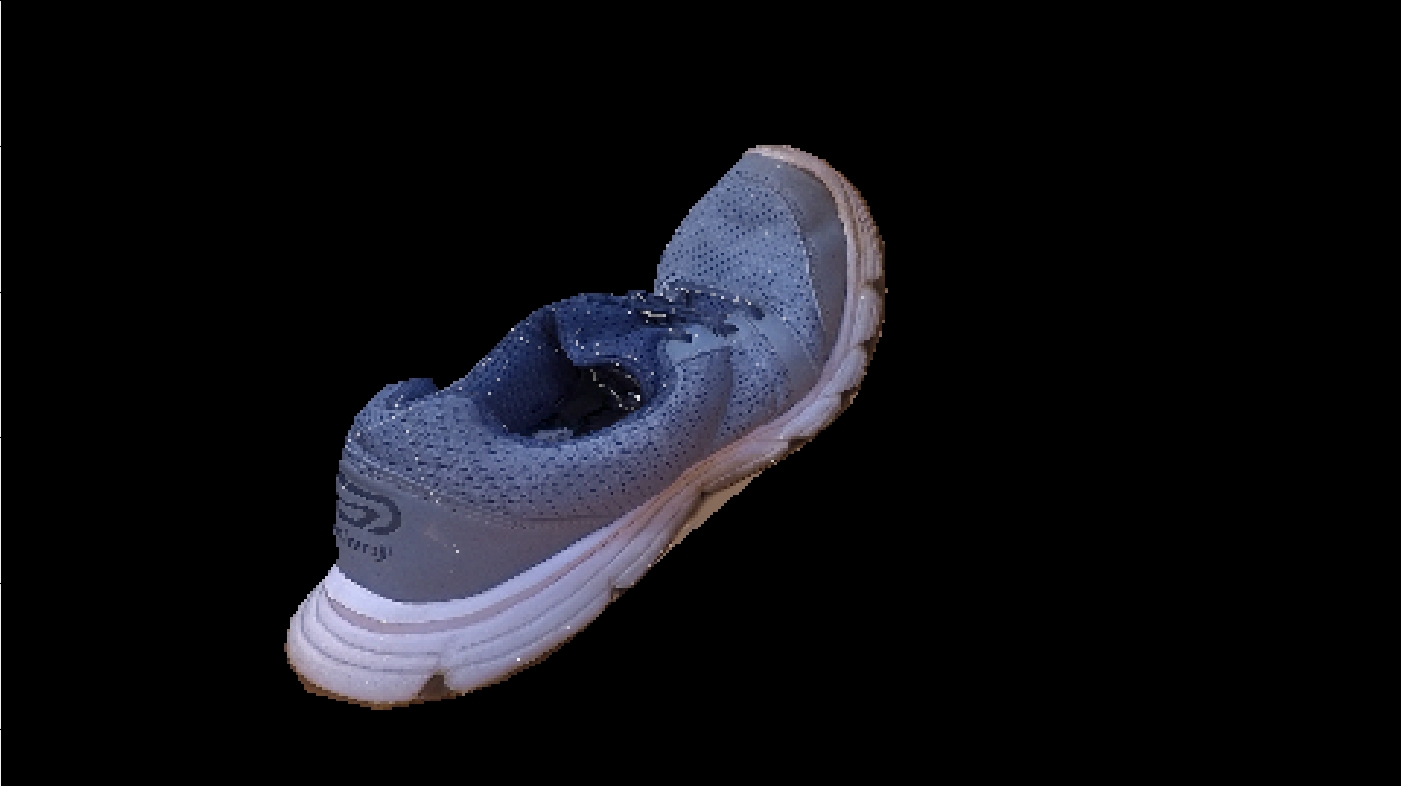}
			& \vspace{1.52mm}\includegraphics[width=50mm]{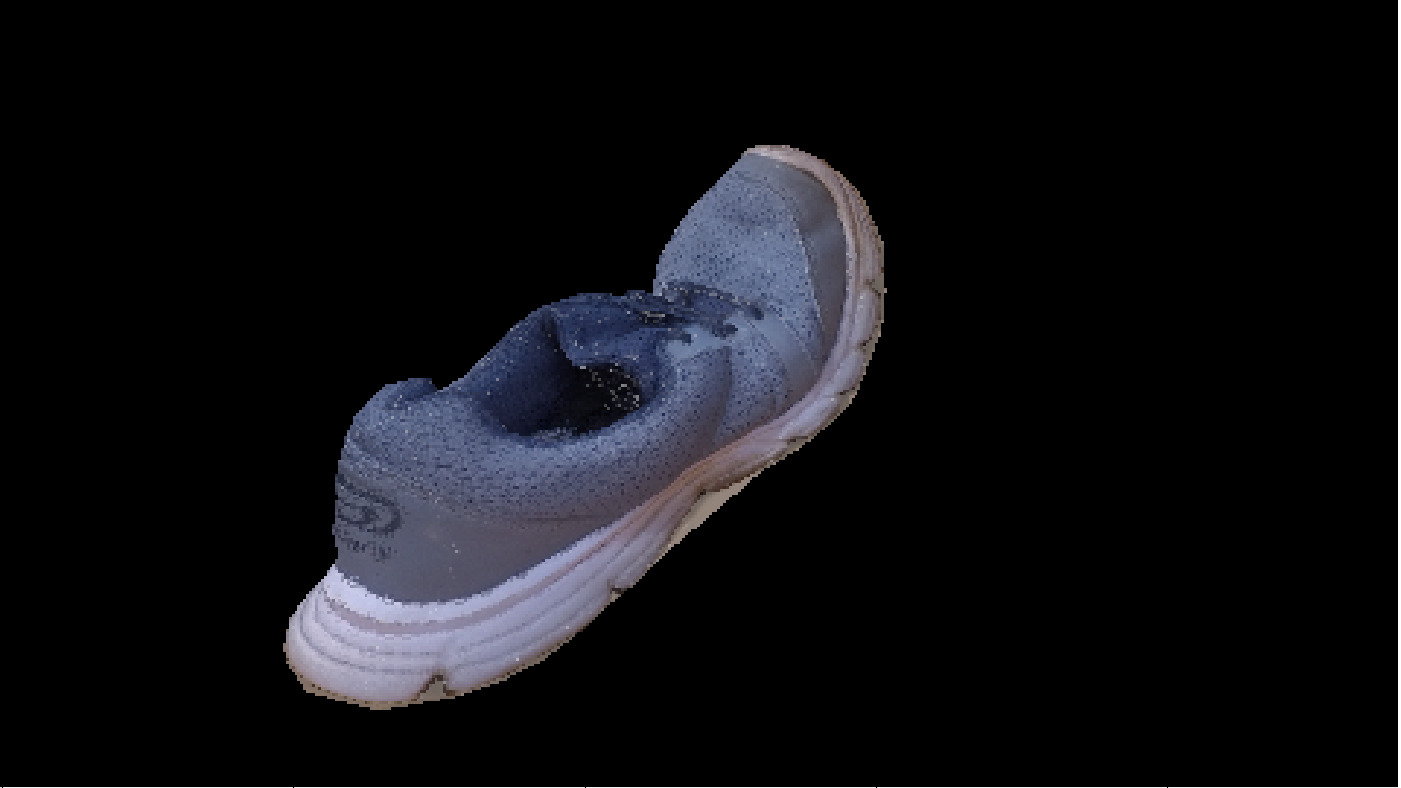} &  \vspace{1.52mm}\includegraphics[width=50mm]{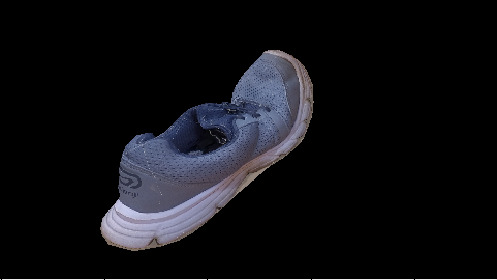}& \vspace{1.52mm}\includegraphics[width=45mm]{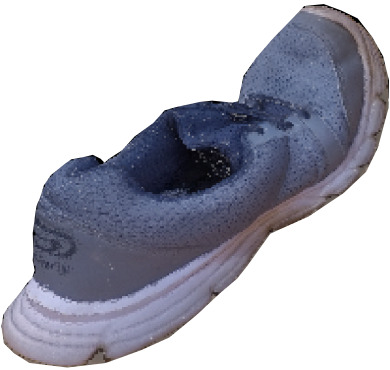}& \vspace{1.52mm}\includegraphics[width=45mm]{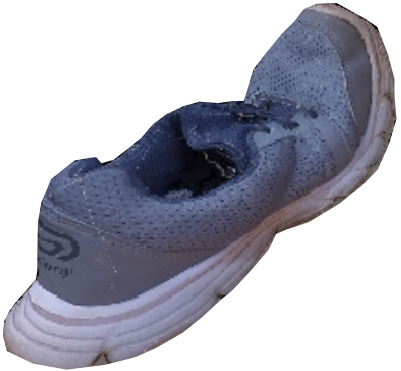}  \\ 
			
			\rule{0pt}{2ex}{\centering $\mathcal{E}_{pr}$ }&
			& \multicolumn{1}{r}{ \large{0.192}}& \multicolumn{1}{r}{\textbf{\large{0.172}}}&&   \\ 
			
		\end{tabular} 
	\end{adjustbox}
	\caption {Examples from DS1, DS2, DS3, DS4 and DS5 templates showing photometric error with and without registration refinement. The first column shows the input images and the third, fourth, fifth and sixth columns show the Main Block results, the Registration Refinement Block results and the corresponding photometric errors from equation (\ref{eq:photoerrors}).}
	\label{tb:photo_error}
	
\end{table*}

\subsection{Timing experiments}

Table \ref{tb:speed} shows the average frame rate of the compared methods, benchmarked on a conventional Linux desktop PC with a single NVIDIA GTX-1080 GPU. 
\begin{table}[!htbp]
	\setcellgapes{3pt}\makegapedcells
	\begin{adjustbox}{max width=\linewidth}
		\begin{tabular}{|c|c|c|c|c|c|c|c|c|}
			\hline
			& DeepSfT & R50F &  CH17 & CH17R & DOF & NGO15 & HDM-net & IsMo-GAN\\ \hline
			
			Time (fps) & 20.40 & 37.00 & 0.75 & 0.19 & 8.84&0.03&25.12& 10.47 \\ \hline
			
		\end{tabular} 
	\end{adjustbox}
	\centering \caption {Average framerate of the evaluated methods.}
	\label{tb:speed}
\end{table}
The DNN methods are considerably faster than the other methods, with frame rates close to real time for DeepSfT. Solutions based on CH17 are far from real-time.  

\subsection{Monocular depth estimation comparison}
We have compared object-generic monocular reconstruction method with DeepSfT. We use DenseDepth~\cite{densedepth} and BTS~\cite{BTS}, two state-of-the-art DNN monocular reconstruction methods. The former is based on DenseNet~\cite{densenet}. We evaluated their ability to recover the object's depth with two experiments. \emph{1)} We test DenseDepth and BTS depth accuracy on the real datasets when they are pre-trained with the NYUDepth dataset~\cite{nyudepth}, which contains RGB-D images from indoor scenes with different types of common objects. To make a fair comparison we have adapted our images to match the intrinsics of the NYUDepth dataset. We compute depth error only in the visible region of each image. We see that DenseDepth and BTS depth RMSE are several orders of magnitude higher compared to DeepSfT. \emph{2)} We fine tune DenseDepth and BTS with all training examples from DS1R, DS3R, DS4R and DS5R datasets at the same time. This restricts DenseDepth and BTS to detect four different objects. The results are shown in tables \ref{tb:densedepth_WE} and \ref{tb:qualitative_monocular} where DenseDepth and BTS fine tuned versions are named as DenseDepth+FT and BTS+FT respectively. In this case the errors have considerably reduced but they are still much larger than the error achieved by DeepSfT. With this experiment we show that instance-level monocular reconstruction solutions such as DeepSfT are able to achieve much more accurate reconstruction results compared to object-generic methods such as \cite{densedepth,BTS}. DenseDepth and BTS obtain an approximately correct average shape, the latter being best. However, they are not able to achieve comparable results to DeepSfT, even when training them with only a reduced set of objects.

\begin{table*}[!htbp]
	\setcellgapes{3pt}\makegapedcells
	\begin{adjustbox}{max width=\linewidth}
		\begin{tabular}{|c|c|c|c|c|c|}
			\hline
			
			\rule{0pt}{2ex} \centering Sequence &\rule{0pt}{2ex} \centering  DenseDepth RMSE  &\rule{0pt}{2ex} \centering  DenseDepth+FT RMSE &\rule{0pt}{2ex} \centering  BTS RMSE  &\rule{0pt}{2ex} \centering  BTS+FT RMSE &\rule{0pt}{2ex}   DeepSft RMSE   \\ \hline
			
			DS3 & 184.52 & 32.53& 122.06 & 23.45 &\textbf{5.52} \\ \hline
			
			DS4 & 221.84 & 18.96& 96.80 & 17.83 &\textbf{4.76}  \\ \hline
			
			DS5 & 395.53 & 22.47& 84.45 & 12.34& \textbf{6.47}  \\ \hline
			
			DS1 & 93.87 & 51.25& 89.22 &14.76 &\textbf{2.33}  \\ \hline

		\end{tabular} 
	\end{adjustbox}
	\centering \caption {Reconstruction, results comparison of \cite{densedepth,BTS}, with and without fine tuning and DeepSfT. All the errors are in mm.}
	\label{tb:densedepth_WE}
\end{table*}

\section{Conclusions}
We have presented DeepSfT, the first dense, real-time solution for wide-baseline SfT with general templates. This has been an open computer vision problem for over a decade. No previous method is able to accurately solve SfT for weakly-textured, non-flat object templates, such as the dinosaur or shoe examples. DeepSfT will enable many real-world applications that require dense registration and 3D reconstruction of deformable objects, in particular augmented reality with deforming objects. In future work we aim to generalise DeepSfT to multiple templates, using them as explicit inputs to our network, or by using object detectors to select an object-specific SfT network. We also will investigate how to train DeepSfT with self-supervised learning approaches, which may require incorporating other priors, such as temporal and spatial smoothness, and other deformation models. Our DeepSfT architecture and results may also contribute to develop future DNN NRSfM solutions. In particular, the use of dense maps as the network output followed by post-processing steps for mesh completion significantly reduces the complexity of the learning process, as opposed to inferring the entire surface or volume. Semi-supervised learning approaches, similar to the one implemented in DeepSfT, can also boost the goal.

{\small
\bibliographystyle{ieee}
\bibliography{egbib.bib}

\begin{thebibliography}{10}\itemsep=-1pt

\bibitem{photoscan}
{\em Agisoft Photoscan}.
\newblock \url{https://www.agisoft.com}.

\bibitem{tensorflow}
M.~Abadi, A.~Agarwal, P.~Barham, E.~Brevdo, Z.~Chen, C.~Citro, G.~S. Corrado,
  and et~al.
\newblock {TensorFlow}: Large-scale machine learning on heterogeneous systems,
  2015.
\newblock Software available from tensorflow.org.

\bibitem{Agudo2015}
A.~Agudo and F.~Moreno-Noguer.
\newblock Simultaneous pose and non-rigid shape with particle dynamics.
\newblock In {\em IEEE Conference on Computer Vision and Pattern Recognition},
  pages 2179--2187, 2015.

\bibitem{Agudo2016}
A.~Agudo, F.~Moreno-Noguer, B.~Calvo, and J.~M.~M. Montiel.
\newblock Sequential non-rigid structure from motion using physical priors.
\newblock {\em IEEE Transactions on Pattern Analysis and Machine Intelligence},
  38(5):979--994, 2016.

\bibitem{densedepth}
I.~Alhashim and P.~Wonka.
\newblock High quality monocular depth estimation via transfer learning.
\newblock {\em arXiv e-prints}, abs/1812.11941, 2018.

\bibitem{alp2018densepose}
R.~Alp~G{\"u}ler, N.~Neverova, and I.~Kokkinos.
\newblock Densepose: Dense human pose estimation in the wild.
\newblock In {\em IEEE Conference on Computer Vision and Pattern Recognition},
  pages 7297--7306, 2018.

\bibitem{segnet}
V.~Badrinarayanan, A.~Kendall, and R.~Cipolla.
\newblock Segnet: A deep convolutional encoder-decoder architecture for image
  segmentation.
\newblock {\em CoRR}, abs/1511.00561, 2015.

\bibitem{Bartoli2015}
A.~Bartoli, Y.~G{\'e}rard, F.~Chadebecq, T.~Collins, and D.~Pizarro.
\newblock Shape-from-template.
\newblock {\em IEEE Transactions on Pattern Analysis and Machine Intelligence},
  37(10):2099--2118, 2015.

\bibitem{bednarik}
J.~Bednarik, P.~Fua, and M.~Salzmann.
\newblock Learning to reconstruct texture-less deformable surfaces from a
  single view.
\newblock In {\em International Conference on 3D Vision (3DV)}, pages 606--615,
  09 2018.

\bibitem{Blender}
{Blender Online Community}.
\newblock {\em Blender - a 3D modelling and rendering package}.
\newblock Blender Foundation, Blender Institute, Amsterdam.

\bibitem{Bregler2000}
C.~Bregler, A.~Hertzmann, and H.~Biermann.
\newblock Recovering non-rigid {3D} shape from image streams.
\newblock In {\em IEEE Conference on Computer Vision and Pattern Recognition}.
  IEEE, 2000.

\bibitem{brickwedde2019monosf}
F.~Brickwedde, S.~Abraham, and R.~Mester.
\newblock Mono-sf: Multi-view geometry meets single-view depth for monocular
  scene flow estimation of dynamic traffic scenes, 2019.

\bibitem{Brunet2014}
F.~Brunet, A.~Bartoli, and R.~I. Hartley.
\newblock Monocular template-based 3d surface reconstruction: Convex
  inextensible and nonconvex isometric methods.
\newblock {\em Computer Vision and Image Understanding}, 125:138--154, 2014.

\bibitem{equiareal}
D.~Casillas-Perez, D.~Pizarro, D.~Fuentes-Jimenez, M.~Mazo, and A.~Bartoli.
\newblock Equiareal shape-from-template.
\newblock {\em J. Math. Imaging Vis.}, 61(5):607–626, June 2019.

\bibitem{Chhatkuli2017}
A.~Chhatkuli, D.~Pizarro, A.~Bartoli, and T.~Collins.
\newblock A stable analytical framework for isometric shape-from-template by
  surface integration.
\newblock {\em IEEE Transactions on Pattern Analysis and Machine Intelligence},
  39(5):833--850, 2017.

\bibitem{Chhatkuli2017a}
A.~Chhatkuli, D.~Pizarro, T.~Collins, and A.~Bartoli.
\newblock Inextensible non-rigid structure-from-motion by second-order cone
  programming.
\newblock {\em IEEE Transactions on Pattern Analysis and Machine Intelligence},
  pages 1--1, 2017.

\bibitem{collins14b}
T.~Collins and A.~Bartoli.
\newblock Using isometry to classify correct/incorrect {3D-2D} correspondences.
\newblock In {\em European Conference on Computer Vision}, 2014.

\bibitem{Collins2015}
T.~Collins and A.~Bartoli.
\newblock Realtime shape-from-template: System and applications.
\newblock In {\em International Symposium on Mixed and Augmented Reality,
  ISMAR}, pages 116--119, 2015.

\bibitem{Collins2016}
T.~Collins, A.~Bartoli, N.~Bourdel, and M.~Canis.
\newblock Robust, real-time, dense and deformable 3d organ tracking in
  laparoscopic videos.
\newblock In {\em International Conference on Medical Image Computing and
  Computer-Assisted Intervention}, pages 404--412. Springer, 2016.

\bibitem{Collins2014}
T.~Collins, P.~Mesejo, and A.~Bartoli.
\newblock An analysis of errors in graph-based keypoint matching and proposed
  solutions.
\newblock In {\em European Conference on Computer Vision}, pages 138--153.
  Springer, 2014.

\bibitem{cvlab}
{Computer Vision Laboratory}.
\newblock {\em Deformable Surface ReconstructionDatabase}.
\newblock Computer Vision Laboratory, Ecole Polytechnique Fédérale de
  Lausanne – EPFL.

\bibitem{adam}
J.~B. Diederik P.~Kingma.
\newblock Adam: A method for stochastic optimization.
\newblock {\em Arxiv}, arXiv:1412.6980(6), December 2014.

\bibitem{FlowNet}
A.~Dosovitskiy, P.~Fischery, E.~Ilg, P.~Hausser, C.~Hazirbas, V.~Golkov,
  P.~van~der Smagt, D.~Cremers, and T.~Brox.
\newblock Flownet: Learning optical flow with convolutional networks.
\newblock In {\em IEEE International Conference on Computer Vision}, pages
  2758--2766. IEEE Computer Society, 2015.

\bibitem{Eigen2015}
D.~Eigen and R.~Fergus.
\newblock Predicting depth, surface normals and semantic labels with a common
  multi-scale convolutional architecture.
\newblock In {\em IEEE International Conference on Computer Vision}, pages
  2650--2658, 2015.

\bibitem{Hughes}
J.~D. Foley, A.~van Dam, S.~K. Feiner, and J.~F. Hughes.
\newblock {\em Computer Graphics: Principles and Practice (2nd Ed.)}.
\newblock Addison-Wesley Longman Publishing Co., Inc., USA, 1990.

\bibitem{Garg2016}
R.~Garg, V.~K. B.G., G.~Carneiro, and I.~Reid.
\newblock Unsupervised cnn for single view depth estimation: Geometry to the
  rescue.
\newblock In {\em European Conference on Computer Vision}, pages 740--756.
  Springer International Publishing, 2016.

\bibitem{Gay-Bellile2010}
V.~Gay-Bellile, A.~Bartoli, and P.~Sayd.
\newblock Direct estimation of nonrigid registrations with image-based
  self-occlusion reasoning.
\newblock {\em IEEE Transactions on Pattern Analysis and Machine Intelligence},
  32(1):87--104, Jan 2010.

\bibitem{hdm_net}
V.~Golyanik, S.~Shimada, K.~Varanasi, and D.~Stricker.
\newblock Hdm-net: Monocular non-rigid 3d reconstruction with learned
  deformation model.
\newblock {\em CoRR}, abs/1803.10193, 2018.

\bibitem{gopro}
GoPro.
\newblock Gopro hero silver v3 rgb camera.
\newblock \url{https://es.gopro.com/update/hero3}.

\bibitem{rendergap}
N.~Hafez, V.~Dietrich, and S.~Roehrl.
\newblock Analysis of different methods to close the reality gap for instance
  segmentation in a flexible assembly cell.
\newblock In S.~Zeghloul, M.~A. Laribi, and J.~S. Sandoval~Arevalo, editors,
  {\em Advances in Service and Industrial Robotics}, pages 505--515, Cham,
  2020. Springer International Publishing.

\bibitem{Haouchine2017}
N.~Haouchine and S.~Cotin.
\newblock Template-based monocular {3D} recovery of elastic shapes using
  lagrangian multipliers.
\newblock In {\em IEEE Conference on Computer Vision and Pattern Recognition}.
  {IEEE}, July 2017.

\bibitem{Haouchine2014}
N.~Haouchine, J.~Dequidt, M.-O. Berger, and S.~Cotin.
\newblock Single view augmentation of {3D} elastic objects.
\newblock In {\em International Symposium on Mixed and Augmented Reality},
  pages 229--236. IEEE, 2014.

\bibitem{Hartley2003}
R.~Hartley and A.~Zisserman.
\newblock {\em Multiple view geometry in computer vision}.
\newblock Cambridge university press, 2003.

\bibitem{he2016deep}
K.~He, X.~Zhang, S.~Ren, and J.~Sun.
\newblock Deep residual learning for image recognition.
\newblock In {\em IEEE conference on computer vision and pattern recognition},
  pages 770--778, 2016.

\bibitem{hornacek2014sphereflow}
M.~Hornacek, A.~Fitzgibbon, and C.~Rother.
\newblock Sphereflow: 6 dof scene flow from rgb-d pairs.
\newblock In {\em IEEE Conference on Computer Vision and Pattern Recognition},
  pages 3526--3533, 2014.

\bibitem{densenet}
G.~Huang, Z.~Liu, and K.~Q. Weinberger.
\newblock Densely connected convolutional networks.
\newblock {\em CoRR}, abs/1608.06993, 2016.

\bibitem{hur2020}
J.~Hur and S.~Roth.
\newblock Self-supervised monocular scene flow estimation.
\newblock In {\em IEEE/CVF Conference on Computer Vision and Pattern
  Recognition}, pages 7396--7405, 2020.

\bibitem{FlowNet20}
E.~Ilg, N.~Mayer, T.~Saikia, M.~Keuper, A.~Dosovitskiy, and T.~Brox.
\newblock Flownet 2.0: Evolution of optical flow estimation with deep networks.
\newblock In {\em IEEE conference on computer vision and pattern recognition},
  pages 2462--2470, 2017.

\bibitem{realsense}
Intel.
\newblock Intel realsense d435 stereo depth camera.
\newblock \url{http://realsense.intel.com}.

\bibitem{posenet}
A.~Kendall, M.~Grimes, and R.~Cipolla.
\newblock Posenet: A convolutional network for real-time 6-dof camera
  relocalization.
\newblock In {\em IEEE International Conference on Computer Vision (ICCV)},
  pages 2938--2946, 12 2015.

\bibitem{swats}
N.~S. Keskar and R.~Socher.
\newblock Improving generalization performance by switching from adam to {SGD}.
\newblock {\em CoRR}, abs/1712.07628, 2017.

\bibitem{Torresani2008}
{L. Torresani, A. Hertzmann and C. Bregler.}
\newblock Nonrigid structure-from-motion: Estimating shape and motion with
  hierarchical priors.
\newblock {\em IEEE Transactions on Pattern Analysis and Machine Intelligence},
  30(5):878--892, 2008.

\bibitem{BTS}
J.~H. Lee, M.~Han, D.~W. Ko, and I.~H. Suh.
\newblock From big to small: Multi-scale local planar guidance for monocular
  depth estimation.
\newblock {\em CoRR}, abs/1907.10326, 2019.

\bibitem{Liu2016}
F.~Liu, C.~Shen, G.~Lin, and I.~D. Reid.
\newblock Learning depth from single monocular images using deep convolutional
  neural fields.
\newblock {\em IEEE Transactions on Pattern Analysis and Machine Intelligence},
  38(10):2024--2039, 2016.

\bibitem{Liu-Yin2017}
Q.~Liu-Yin, R.~Yu, L.~Agapito, A.~Fitzgibbon, and C.~Russell.
\newblock Better together: Joint reasoning for non-rigid 3d reconstruction with
  specularities and shading.
\newblock {\em British Machine Vision Conference (BMVC)}, pages 42.1--42.12,
  2016.

\bibitem{sift}
D.~G. Lowe.
\newblock Distinctive image features from scale-invariant keypoints.
\newblock {\em International Journal of Computer Vision}, 60:91--110, 2004.

\bibitem{lv2018learning}
Z.~Lv, K.~Kim, A.~Troccoli, D.~Sun, J.~M. Rehg, and J.~Kautz.
\newblock Learning rigidity in dynamic scenes with a moving camera for 3d
  motion field estimation.
\newblock In {\em European Conference on Computer Vision (ECCV)}, pages
  468--484, 2018.

\bibitem{Malti2015}
A.~Malti, A.~Bartoli, and R.~Hartley.
\newblock A linear least-squares solution to elastic shape-from-template.
\newblock In {\em IEEE Conference on Computer Vision and Pattern Recognition},
  pages 1629--1637, 2015.

\bibitem{Malti2013}
A.~Malti, R.~Hartley, A.~Bartoli, and J.-H. Kim.
\newblock Monocular template-based 3d reconstruction of extensible surfaces
  with local linear elasticity.
\newblock In {\em IEEE Conference on Computer Vision and Pattern Recognition},
  pages 1522--1529, 2013.

\bibitem{Martinez2017}
J.~Martinez, R.~Hossain, J.~Romero, and J.~J. Little.
\newblock A simple yet effective baseline for 3d human pose estimation.
\newblock In {\em IEEE International Conference on Computer Vision}, volume~1,
  page~5, 2017.

\bibitem{nyudepth}
P.~K. Nathan~Silberman, Derek~Hoiem and R.~Fergus.
\newblock Indoor segmentation and support inference from rgbd images.
\newblock In {\em European Conference on Computer Vision}, 2012.

\bibitem{Ngo2016}
D.~T. Ngo, J.~{\"O}stlund, and P.~Fua.
\newblock Template-based monocular 3d shape recovery using laplacian meshes.
\newblock {\em IEEE Transactions on Pattern Analysis and Machine Intelligence},
  38(1):172--187, 2016.

\bibitem{7410619}
D.~T. {Ngo}, S.~{Park}, A.~{Jorstad}, A.~{Crivellaro}, C.~D. {Yoo}, and
  P.~{Fua}.
\newblock Dense image registration and deformable surface reconstruction in
  presence of occlusions and minimal texture.
\newblock In {\em IEEE International Conference on Computer Vision (ICCV)},
  pages 2273--2281, 2015.

\bibitem{Oezguer2017}
E.~{\"O}zg{\"u}r and A.~Bartoli.
\newblock Particle-sft: A provably-convergent, fast shape-from-template
  algorithm.
\newblock {\em International Journal of Computer Vision}, 123(2):184--205,
  2017.

\bibitem{asrigidasposible}
S.~Parashar, D.~Pizarro, A.~Bartoli, and T.~Collins.
\newblock As-rigid-as-possible volumetric shape-from-template.
\newblock In {\em IEEE International Conference on Computer Vision}, December
  2015.

\bibitem{Perriollat2011}
M.~Perriollat, R.~Hartley, and A.~Bartoli.
\newblock Monocular template-based reconstruction of inextensible surfaces.
\newblock {\em International Journal of Computer Vision}, 95(2):124--137, 2011.

\bibitem{pilet08a}
J.~Pilet, V.~Lepetit, and P.~Fua.
\newblock Fast non-rigid surface detection, registration and realistic
  augmentation.
\newblock {\em International Journal on Computer Vision}, 76(2):109--122,
  February 2008.

\bibitem{Pizarro2012}
D.~Pizarro and A.~Bartoli.
\newblock Feature-based deformable surface detection with self-occlusion
  reasoning.
\newblock {\em International Journal of Computer Vision}, 97(1):54--70, 2012.

\bibitem{Pumarola2018}
A.~Pumarola, A.~Agudo, L.~Porzi, A.~Sanfeliu, V.~Lepetit, and F.~Moreno-Noguer.
\newblock Geometry-aware network for non-rigid shape prediction from a single
  view.
\newblock In {\em IEEE Conference on Computer Vision and Pattern Recognition},
  pages 4681--4690, 2018.

\bibitem{Ren2017UnsupervisedDL}
Z.~Ren, J.~Yan, B.~Ni, B.~Liu, X.~Yang, and H.~Zha.
\newblock Unsupervised deep learning for optical flow estimation.
\newblock In {\em AAAI Conference on Artificial Intelligence}, 2017.

\bibitem{rudin}
L.~I. Rudin, S.~Osher, and E.~Fatemi.
\newblock Nonlinear total variation based noise removal algorithms.
\newblock {\em Physica D: Nonlinear Phenomena}, 60(1):259--268, 1992.

\bibitem{Salzmann2009}
M.~Salzmann and P.~Fua.
\newblock Reconstructing sharply folding surfaces: A convex formulation.
\newblock In {\em IEEE Conference on Computer Vision and Pattern Recognition},
  pages 1054--1061. IEEE, 2009.

\bibitem{Salzmann2008}
M.~Salzmann, F.~Moreno-Noguer, V.~Lepetit, and P.~Fua.
\newblock Closed-form solution to non-rigid 3d surface registration.
\newblock {\em European Conference on Computer Vision}, pages 581--594, 2008.

\bibitem{schuster2018sceneflowfields}
R.~Schuster, O.~Wasenmuller, G.~Kuschk, C.~Bailer, and D.~Stricker.
\newblock Sceneflowfields: Dense interpolation of sparse scene flow
  correspondences.
\newblock In {\em IEEE Winter Conference on Applications of Computer Vision
  (WACV)}, pages 1056--1065. IEEE, 2018.

\bibitem{Shimada2019}
S.~Shimada, V.~Golyanik, C.~Theobalt, and D.~Stricker.
\newblock {IsMo-GAN}: Adversarial learning for monocular non-rigid 3d
  reconstruction.
\newblock In {\em IEEE Conference on Computer Vision and Pattern Recognition
  Workshops (CVPRW)}, 2019.

\bibitem{ARAP}
O.~Sorkine and M.~Alexa.
\newblock As-rigid-as-possible surface modeling.
\newblock In {\em Fifth Eurographics symposium on Geometry processing}, pages
  109--116, 2007.

\bibitem{strong}
D.~Strong and T.~Chan.
\newblock Edge-preserving and scale-dependent properties of total variation
  regularization.
\newblock {\em Inverse Problems}, 19(6):S165--S187, nov 2003.

\bibitem{optical_flow}
N.~Sundaram, T.~Brox, and K.~Keutzer.
\newblock Dense point trajectories by gpu-accelerated large displacement
  optical flow.
\newblock {\em European Conference on Computer Vision}, 2010.

\bibitem{patchbased}
A.~Tsoli and A.~Argyros.
\newblock Patch-based reconstruction of a textureless deformable 3d surface
  from a single rgb image.
\newblock In {\em IEEE International Conference on Computer Vision (ICCV)},
  pages 4034--4043, 10 2019.

\bibitem{wedel2011stereoscopic}
A.~Wedel, T.~Brox, T.~Vaudrey, C.~Rabe, U.~Franke, and D.~Cremers.
\newblock Stereoscopic scene flow computation for 3d motion understanding.
\newblock {\em International Journal of Computer Vision}, 95(1):29--51, 2011.

\bibitem{xavier_glorot}
Y.~B. Xavier~Glorot, Antoine~Bordes.
\newblock Understanding the difficulty of training deep feedforward neural
  networks.
\newblock {\em Machine Learning Research}, 2010.

\bibitem{Dai2012}
{Y. Dai, H. Li, and M. He.}
\newblock A simple prior-free method for non-rigid structure-from-motion
  factorization.
\newblock In {\em IEEE Conference on Computer Vision and Pattern Recognition},
  2012.

\bibitem{8851791}
H.~{Zhang}, Y.~{Tian}, K.~{Wang}, H.~{He}, and F.~{Wang}.
\newblock Synthetic-to-real domain adaptation for object instance segmentation.
\newblock In {\em International Joint Conference on Neural Networks (IJCNN)},
  pages 1--7, 2019.

\bibitem{adamsgd}
Z.~Zhang, L.~Ma, Z.~Li, and C.~Wu.
\newblock Normalized direction-preserving adam.
\newblock {\em CoRR}, abs/1709.04546, 2017.

\bibitem{zou2018df}
Y.~Zou, Z.~Luo, and J.-B. Huang.
\newblock Df-net: Unsupervised joint learning of depth and flow using
  cross-task consistency.
\newblock In {\em European conference on computer vision (ECCV)}, pages 36--53,
  2018.

\end{thebibliography}
}

\label{app:appendixA}
\section{Appendix: Warping with Bilinear Interpolation}

We describe the process to create the image $I'(u,v)$ from the registration $\hat{\eta}(u,v)=(\hat{\eta}_U(u,v),\hat{\eta}_V(u,v))$ and the texture map $A(U,V)$. We recall that $U,V$ are normalised coordinates drawn from the unit square. We define the texture map image $\bar{A}(\bar{U},\bar{V})$ of size $H\times W$ with $(\bar{U},\bar{V}) \in [1,W]\times[1,H]$ being pixel coordinates, obtained by de-normalising $(U,V)$ with $W$ and $H$. Image coordinates $(u,v)\in [1,w]\times[1,h]$ are already in pixels and $I'(u,v)$ is of size $h\times w$ pixels. In addition, we assume that both $\bar{A}$ and $I'$ are single channel images. The generalisation to 3-channel images is straightforward. We have that $\hat{\eta}(u,v)$ is a differentiable function of the Photometric Refinement Block weights $\theta_{\mathcal{R}}$ and $\dfrac{\partial \hat{\eta}_U(u,v)}{\partial \theta_{\mathcal{R}}}$ and $\dfrac{\partial \hat{\eta}_V(u,v)}{\partial \theta_{\mathcal{R}}}$ are the first derivatives of the registration with respect to $\theta_{\mathcal{R}}$. We recall that  $\mathcal{I}$ is a subset of coordinates $(u,v)$ where the object is visible, and thus $I'(u,v)$ is set to a constant value outside $\mathcal{I}$. By using billinear interpolation we obtain $I'$ as follows:

\begin{equation}
\label{eq:bl1}
I'(u,v) =\begin{cases}\sum_{i=1}^4 w_i \bar{A}(\bar{U}_i,\bar{V}_i)  & (u,v) \in \mathcal{I}\\ 0 & \text{otherwise}\end{cases}
\end{equation}
where:
\begin{eqnarray}
\label{eq:bl2}
\bar{U}_1 = \bar{U}_2 =\zeta_U\left(\floor{\hat{\eta}_U(u,v)}\right)& \bar{U}_3 = \bar{U}_4 =\zeta_U\left( \ceil{\hat{\eta}_U(u,v)}\right )\\\nonumber
\bar{V}_1 = \bar{V}_3 = \zeta_V\left(\floor{\hat{\eta}_V(u,v)}\right)& \bar{V}_2 = \bar{V}_4 =\zeta_V\left( \ceil{\hat{\eta}_V(u,v)}\right)
\end{eqnarray}
and
\begin{eqnarray}
\label{eq:bl3}
w_1 &=& (1+\bar{U}_1 - \hat{\eta}_U(u,v)) (1+\bar{V}_1 - \hat{\eta}_V(u,v))\\\nonumber
w_2 &=& (1+\bar{U}_1 - \hat{\eta}_U(u,v)) (1-\bar{V}_2 + \hat{\eta}_V(u,v))\\\nonumber
w_3 &=& (1-\bar{U}_3 +\hat{\eta}_U(u,v))(1+\bar{V}_3 - \hat{\eta}_V(u,v))\\\nonumber
w_4 &=& (1-\bar{U}_4 +\hat{\eta}_U(u,v))(1-\bar{V}_4 + \hat{\eta}_V(u,v))
\end{eqnarray}
with $\floor{ . }$ and $\ceil{ . }$ being the `floor' and `ceil' operators respectively. We assume that their derivatives with respect to $\theta_{\mathcal{R}}$ are zero. Also, $\zeta_U(x)=\text{max}(\min(x,W),1)$ and $\zeta_V(x)=\text{max}(\min(x,H),1)$ are functions ensuring that coordinates remain inside the domain of $\bar{A}$.   
Therefore, all terms in equation~(\ref{eq:bl3}) are bilinear in $\eta(u,v)$ with:
\begin{equation}
\label{eq:bl1}
\dfrac{\partial I'(u,v)}{\partial \theta_{\mathcal{R}}} = \begin{cases}\sum_{i=1}^4 \dfrac{\partial w_i}{\partial \theta_{\mathcal{R}}} \bar{A}(\bar{U}_i,\bar{V}_i)  & (u,v) \in \mathcal{I}\\ 0 & \text{otherwise.}\end{cases}
\end{equation}

\end{document}